%% file: arxiv_main.tex
\documentclass[11pt,twoside]{article} 

\input{math_commands.tex}

\usepackage[utf8]{inputenc} 
\usepackage[T1]{fontenc}    
\usepackage{booktabs}       
\usepackage{amsfonts}       
\usepackage{nicefrac}       
\usepackage{microtype}      

\usepackage{subfigure}
\usepackage{epsf}
\usepackage{epsfig}
\usepackage{fancyhdr}
\usepackage{graphics}
\usepackage{graphicx}
\usepackage{psfrag}
\usepackage{fullpage}
\usepackage{pdfpages}
\usepackage{url}
\usepackage[colorlinks=True,linkcolor=magenta,citecolor=blue,urlcolor=blue,pagebackref=true,backref=true]
{hyperref}
\renewcommand*{\backref}[1]{\ifx#1\relax \else Page #1 \fi}
\renewcommand*{\backrefalt}[4]{%
    \ifcase #1 \footnotesize{(Not cited.)}%
    \or        \footnotesize{(Cited on page~#2.)}%
    \else      \footnotesize{(Cited on pages~#2.)}%
    \fi}
\usepackage{color}
\usepackage{hyperref}
\usepackage{url}
\usepackage{multicol}
\usepackage{multirow}
\usepackage{diagbox}
\input{import}

\newcommand{\vMF}{\text{vMF}}
\title{Improving relational regularized \\ Autoencoders with spherical sliced fused Gromov Wasserstein}

\begin{document}
\begin{center}

{\bf{\LARGE{Improving Relational Regularized  Autoencoders with Spherical Sliced Fused Gromov Wasserstein}}}
  
\vspace*{.2in}
{\large{
\begin{tabular}{ccccc}
Khai Nguyen$^{\diamond}$ & Son Nguyen$^{\diamond}$ & Nhat Ho$^{\dagger}$ & Tung Pham$^{\diamond}$ & Hung Bui$^{\diamond}$
\end{tabular}
}}

\vspace*{.2in}

\begin{tabular}{c}
VinAI Research$^\diamond$, University of Texas, Austin$^{\dagger}$\\
\end{tabular}

\today

\vspace*{.2in}

\begin{abstract}
Relational regularized autoencoder (RAE) is a framework to learn the distribution of data by minimizing a reconstruction loss together with a relational regularization on the latent space. A recent attempt to reduce the inner discrepancy between the prior and aggregated posterior distributions is to incorporate  sliced fused Gromov-Wasserstein (SFG) between these distributions. That approach has a weakness since it treats every slicing direction similarly, meanwhile several directions are not useful for the discriminative task. To improve the discrepancy and consequently the relational regularization, we propose a new relational discrepancy, named \emph{spherical sliced fused Gromov Wasserstein} (SSFG), that can find an important area of projections characterized by a von Mises-Fisher distribution. Then, we introduce two variants of SSFG to improve its performance. The first variant, named \emph{mixture spherical sliced fused Gromov Wasserstein} (MSSFG), replaces the vMF distribution by a mixture of von Mises-Fisher distributions to capture multiple important areas of directions that are far from each other. The second variant, named \emph{power spherical sliced fused Gromov Wasserstein} (PSSFG), replaces the vMF distribution by a power spherical distribution to improve the sampling time in high dimension settings. We then apply the new discrepancies to the RAE framework to achieve its new variants. Finally, we conduct extensive experiments to show that the new proposed autoencoders have favorable performance in learning latent manifold structure, image generation ,and reconstruction.  
\end{abstract}
\end{center}
\section{Introduction}
\label{sec:introduction}
In recent years, autoencoders have been used widely as important frameworks in several machine learning and deep learning models, such as generative models \cite{kingma2013auto,tolstikhin2018wasserstein,kolouri2018sliced} and representation learning models~\cite{tschannen2018recent}. Formally, autoencoders consist of two components, namely, an encoder and a decoder. The encoder denoted by $E_{\phi}$ maps the data, which is presumably in a low dimensional manifold, to a latent space. Then the data could be generated by sampling points  from the latent space via a prior distribution $p$, then decoding those points by  the decoder $G_\theta$. The decoder is formally a function from latent space to the data space and it induces a distribution $p_{G_{\theta}}$ on the data space.  In generative modeling, the major task is to obtain a decoder $G_{\theta^*}$ such that its induced distribution $p_{G_{\theta^*}}$ and the data distribution are very close under some discrepancies. 
Two popular instances of autoencoders are the variational autoencoder (VAE)~\cite{kingma2013auto}, which uses KL divergence, and the Wasserstein autoencoder (WAE)~\cite{tolstikhin2018wasserstein}, which chooses the Wasserstein distance~\cite{Villani-09-Optimal} as the discrepancy between the induced distribution and the data distribution. 

In order to implement the WAE, a relaxed version was introduced by removing the constraint on the prior and the aggregated posterior (latent code distribution). In particular, a chosen discrepancy between these distributions is added to the objective function and plays a role as a regularization term.  With that relaxation approach, the WAE becomes a flexible framework for customized choices of the discrepancies~\cite{patrini2020sinkhorn,kolouri2018sliced}. However, the WAE suffers either from the over-regularization problem when the prior distribution is too simple~\cite{dai2018diagnosing,ghosh2019variational}, which is usually chosen to be isotropic Gaussian, or from the under-regularization problem when learning an expressive prior distribution jointly with the autoencoder without additional regularization, e.g., structural regularization~\cite{xu2020learning}. In order to circumvent these issues of WAE, relational regularized autoencoder (RAE) was proposed in~\cite{xu2020learning} with two major changes. The first change is to use a mixture of Gaussian distributions as the prior while the second change is to set a regularization on the structural difference between the prior and the aggregated posterior distribution, which is called the relational regularization. The state-of-the-art version of RAE, deterministic relational regularized autoencoder (DRAE), utilizes the sliced fused Gromov Wasserstein (SFG)~\cite{xu2020learning} as the relational regularization. Although DRAE performs well in practice and has good computational complexity~\cite{xu2020learning}, the SFG does not fully 
exploit the benefits of relational regularization due to its slicing drawbacks. Similar to sliced Wasserstein (SW) ~\cite{bonnotte2013unidimensional, bonneel2015sliced} and sliced Gromov Wasserstein (SG)~\cite{vayer2019sliced}, SFG uses the uniform distribution over the unit sphere to sample projecting directions. However, that leads to the underestimation of the discrepancy between two target distributions ~\cite{deshpande2019max,kolouri2019generalized} since many unimportant directions are included in that estimation. A potential solution is by using only the best Dirac measure over the unit sphere to sample projecting directions in SFG, which was employed in max-sliced Wasserstein distance~\cite{deshpande2019max}. However, this approach focuses on the discrepancy of the target probability measures based on only one important direction while other important directions are not considered.  

\vspace{0.5 em}
\noindent
\textbf{Our contributions.} To improve the effectiveness of the relational regularization in the autoencoder framework, we propose novel sliced relational discrepancies between the prior and the aggregated posterior.  The new sliced discrepancies utilize von Mises-Fisher distribution and its variants instead of the uniform distribution as the distributions over slices. An advantage of the vMF distribution and its variants is that they could interpolate between the Dirac measure and uniform measure, thereby improving the quality of the projections sampled from these measures and overcoming the weaknesses of both the SFG and its max version---max-SFG. In summary, our major contributions are as follows:

\textbf{1.} First, we propose a novel discrepancy, named \textit{spherical sliced fused Gromov Wassersetein} (SSFG). This discrepancy utilizes vMF distribution as the slicing distribution to focus on the area of directions that can separate the target probability measures on the projected space. Moreover, we show that SSFG is a well-defined pseudo-metric on the probability space and does not suffer from the curse of dimensionality for the inference purpose. With favorable theoretical properties of SSFG, we apply it to the RAE framework and obtain a variant of RAE, named \textit{spherical deterministic RAE} (s-DRAE).

\textbf{2.} Second, we propose an extension of SSFG to \textit{mixture SSFG} (MSSFG)  where we utilize a mixture of vMF distributions as the slicing distribution (see Appendix~\ref{Sec:MSFG} for the details). Comparing to the SSFG, the MSSFG is able to simultaneously search for multiple areas of important directions, thereby capturing more important directions that could be far from each other. Based on the MSSFG, we then propose another variant of RAE, named \textit{mixture spherical deterministic RAE} (ms-DRAE).

\textbf{3.} Third, to improve the sampling time and stability of vMF distribution in high dimension settings, we introduce another variant of SSFG, named \emph{power SSFG} (PSSFG), which uses power spherical distribution instead of the vMF distribution as the slicing distribution. Then, we apply the PSSFG to the RAE framework to obtain the \textit{power spherical deterministic RAE} (ps-DRAE).  

\textbf{4.} Finally, we carry out extensive experiments on standard datasets to show that proposed autoencoders achieve the best generative quality among well-known autoencoders, including the state-of-the-art RAE---DRAE. Furthermore, the experiments indicate that the s-DRAE, ms-DRAE, and ps-DRAE can learn  a nice latent manifold structure, a good mixture of Gaussian prior which can cover well the latent manifold, and provide more stable results in both generation and reconstruction than DRAE.

\vspace{0.5 em}
\noindent
\textbf{Organization.} The remainder of the paper is organized as follows. In Section~\ref{Sec:background}, we provide backgrounds for DRAE and vMF distribution. In Section~\ref{sec:general_SSFG}, we propose the spherical sliced fused Gromov Wasserstein and its extension. We then apply these spherical discrepancies to the relational regularized autoencoder. Extensive experiment results are presented in Section~\ref{Sec:experiments} followed by conclusion in Section~\ref{Sec:conclusion}. Proofs of key results and extra materials are in the supplementary material.

\vspace{0.5 em}
\noindent
\textbf{Notation:} Let $\mathbb{S}^{d-1}$ be the $d$-dimensional hypersphere and $\mathcal{U}(\mathbb{S}^{d-1})$ be the uniform distribution on $\mathbb{S}^{d-1}$. For a metric space $(\mathcal{X},d_{1})$, we denote by $\mathcal{P}(\mathcal{X})$ the space of probability distributions over $\mathcal{X}$ with finite moments. We say that $d_{1}$ is a pseudo-metric in space $\mathcal{X}$ if it is non-negative, symmetric, and satisfies the inequality:  $d_{1}(x, z) \leq C\big[ d_{1}(x,y ) + d_{1}(y, z)\big]$ for a universal constant $C > 0$ and for all $x, y, z \in \mathcal{X}$. For any distribution $\mu$ and $\nu$, $\Pi(\mu,\nu)$ is the set of all transport plans between $\mu$ and $\nu$. For $x\in \mathbb{R}^d$, denote $\delta_x$ to be the Dirac measure at $x$. For any $\theta \in \mathbb{S}^{d - 1}$ and any measure $\mu$, $\theta \sharp \mu$ denotes the pushforward measure of $\mu$ through the mapping $\mathcal{R}_{\theta}$ where $\mathcal{R}_{\theta}(x) = \theta^{\top} x$ for all $x$. 
\section{Background}
\label{Sec:background}
In this section,  we provide backgrounds for the sliced fused Gromov Wasserstein and the relational regularized autoencoders. Then, we give backgrounds for the von Mises-Fisher distribution.
\subsection{Sliced Fused Gromov Wasserstein and deterministic relational regularized autoencoder}
First, we review the WAE framework~\cite{tolstikhin2018wasserstein}, which is used to  learn a generative model by minimizing a relaxed version of Wasserstein distance~\cite{Villani-09-Optimal} between data distribution $p_d(x)$ and model distribution $p_\theta(x) := G_\theta \sharp p(z)$, where $p(z)$ is a noise distribution. The model aims to find 
the autoencoder which solves the following objective function:
\begin{align}
    \min_{\theta,\phi} \mathbb{E}_{p_d(x)} \mathbb{E}_{q_\phi (z|x)}[d(x,G_\theta(z))]+ \lambda D(q_\phi(z)||p(z)), \label{eq:autoencod_def}
\end{align}
where $d$ is the ground metric of Wasserstein distance, $D$ is a discrepancy between distributions, and $q_\phi (z|x)$ is a distribution for encoder $E_\phi: X \to Z$, parameterized by $\phi$. Due to the efficiency in training generative models, several autoencoder models are derived from this framework. For example, WAE uses standard Gaussian distribution for $p(z)$ and chooses $D$ to be either maximum mean discrepancy (MMD) or GAN. Later, by using sliced Wasserstein distance~\cite{bonneel2015sliced} for $D$,~\cite{kolouri2018sliced} achieved another type of autoencoder, which is called SWAE.

\vspace{0.5 em}
\noindent
\textbf{Deterministic relational regularized autoencoder (DRAE):} In DRAE,~\cite{xu2020learning} parametrizes the prior as a mixture of Gaussians $(p_{\mu_{1:k},\Sigma_{1:k}}(z))$ and makes it learnable. Additionally, they introduce the sliced fused Gromov Wasserstein as the discrepancy between the posterior and the prior distributions. 

\begin{definition}
\label{def:sliced_fused_Gromov}
(SFG) Let  $\mu, \nu \in \mathcal{P}(\mathbb{R}^d)$ be two probability distributions, $\beta$ be a constant in  $[0,1]$, and $d_{1}: \mathbb{R} \times \mathbb{R} \to \mathbb{R}_{+}$ be a pseudo-metric on $\mathbb{R}$. The \textbf{sliced fused Gromov Wasserstein} (SFG) between $\mu$ and $\nu$ is defined as:
\begin{align}
    \text{SFG}(\mu,\nu; \beta) : = \mathbb{E}_{\theta \sim \mathcal{U}(\mathbb{S}^{d-1})}[D_{fgw}(\theta \sharp \mu,\theta \sharp \nu;\beta,d_1)], \label{eq:sfg}
\end{align}
where the \textbf{fused Gromov Wasserstein} $D_{fgw}$ is given by:
\begin{align}
   D_{fgw}(\theta \sharp \mu,\theta \sharp \nu;\beta,d_1) : = & \min_{\pi \in \Pi (\theta \sharp \mu,\theta \sharp \nu)} \biggr\{ (1-\beta) \int_{\mathbb{R}^d \times \mathbb{R}^d}  d_{1}(\theta^\top x, \theta^\top y) d \pi(x,y) \nonumber \\ & \hspace{-2 em} + \beta  \int_{(\mathbb{R}^d)^4}\big[d_{1}(\theta^\top x, \theta^\top x^{\prime}) - d_{1}(\theta^\top y, \theta^\top y^\prime)\big]^2 d\pi(x,y) d \pi (x^\prime, y^\prime) \biggr\}. 
  \label{def:fg}
\end{align}
\end{definition}
Given the definition of SFG, the objective function of the deterministic relational regularized autoencoder (DRAE) takes the following form:
\begin{align}
    \min_{\theta,\phi,\mu_{1:k},\Sigma_{1:k}} \mathbb{E}_{p_d(x)} \mathbb{E}_{q_\phi (z|x)}\big[d\big(x,G_\theta(z)\big)\big]+ \lambda  \mathbb{E}_{q_\phi(z),p_{\mu_{1:k},\Sigma_{1:k}}(z)} SFG \big[\big(\hat{q}_{N}(z)||\hat{p}_{N}(z)\big)\big], \label{eq:drae_sfg}
\end{align}
where $\hat{q}_N (z)$ and  $\hat{p}_{N}(z)$ are the empirical distributions of $q_\phi(z)$ and $p_{\mu_{1:k},\Sigma_{1:k}}(z)$ respectively.

\vspace{0.5 em}
\noindent
\textbf{Properties of SFG:} From equation~(\ref{eq:sfg}), SFG is a linear  combination of sliced Wasserstein (SW) and sliced Gromov Wasserstein (SG). In particular,  SFG becomes SW and SG when $\beta=0$ and $\beta=1$, respectively. Hence SFG is able to take advantages of both of them. If $\mu$ and $\nu$ have $n$ supports and uniform weights and $d_{1}(x, y) = (x - y)^2$, SFG has computational complexity of the order $\mathcal{O} (n \log n)$. It is because under $d_1$, both SW and SG have closed-form expressions~\cite{vayer2019sliced, bonnotte2013unidimensional} 
 where the optimal transport map $\pi$ in $D_{fgw}$ can be obtained by sorting the projected supports of $\mu$ and $\nu$.

\vspace{0.5 em}
\noindent
\textbf{Limitation of SFG:} The major limitation of SFG is that the outer expectation with respect to $\theta \sim \mathcal{U}(\mathbb{S}^{d - 1})$ in SFG is generally intractable. In practice,  projections from the unit sphere are uniformly sampled and we then apply the Monte Carlo method to obtain an approximate of that expectation. However, the difference between two distributions is certainly not distributed uniformly, meaning that informative directions are mixed up with many non-informative ones. Hence, sampling blindly slices in high dimensional space not only is ineffective but also underestimates the discrepancy between the two distributions. The von Mises-Fisher (vMF) distribution provides a way to have concentrated weight on the most important directions and assigns less weight to further directions. Therefore, we gain a better representation of the discrepancy between probability measures. 
\begin{figure}[!t]
\begin{center}

  \begin{tabular}{ccc}
 \includegraphics[scale=0.45]{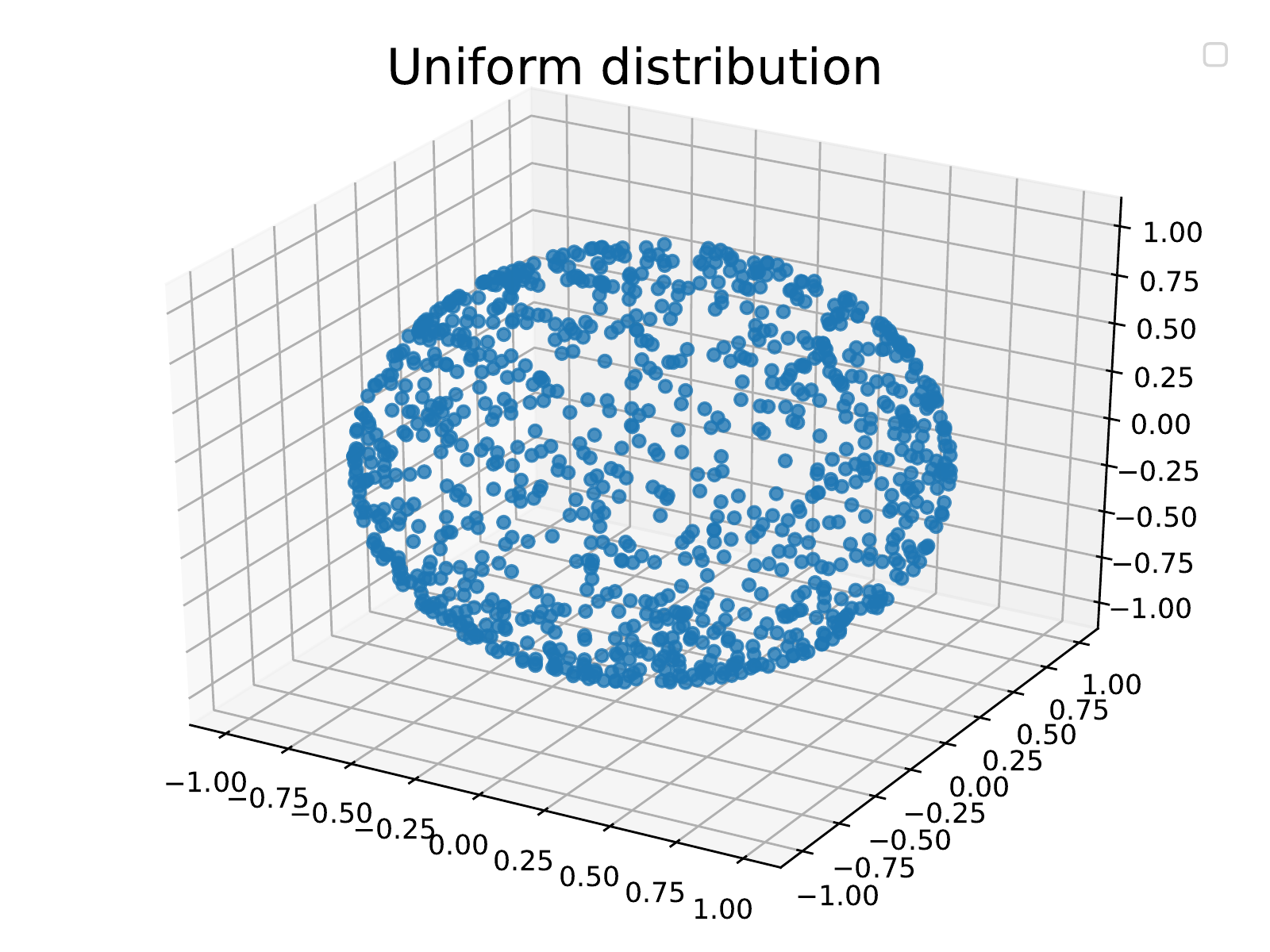} 
    &
    \includegraphics[scale=0.45]{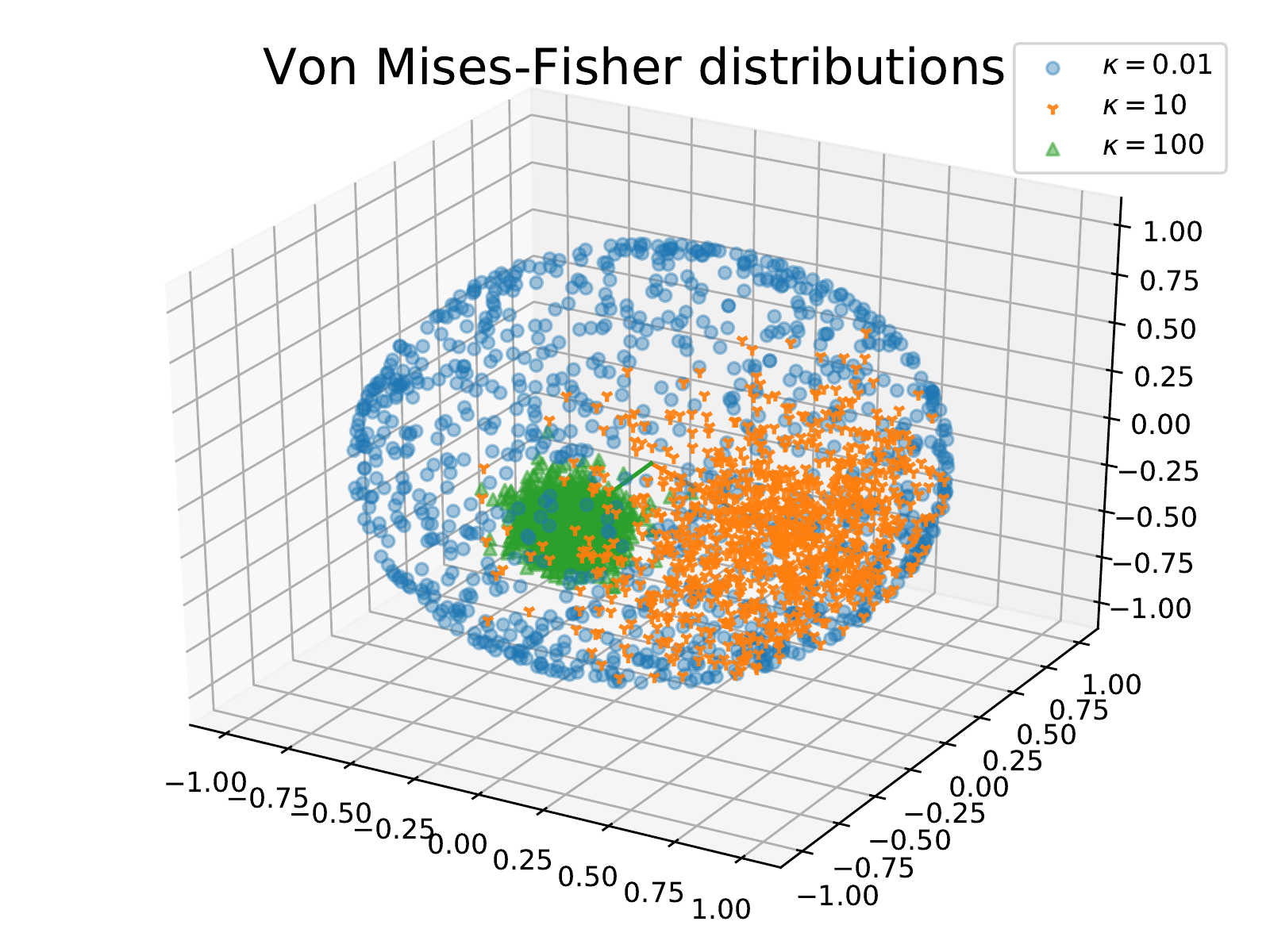}
   & 
   
  \end{tabular}
 \end{center}
  \caption{
  \footnotesize{Illustrations of uniform distribution and von Mises-Fisher distribution.}}
  \label{fig:sphere}
  \vspace{-0.2 em}
\end{figure}


\subsection{Von Mises-Fisher distribution}
Now, we review the definition of the von Mises-Fisher distribution.
\begin{definition}
\label{def:vMF}
The von Mises–Fisher distribution (\emph{vMF}) is a probability distribution on the unit sphere $\mathbb{S}^{d-1}$  where its density function is given by \cite{jupp1979maximum}:
\begin{align}
    f(x| \epsilon, \kappa ) : = C_d (\kappa) \exp(\kappa \epsilon^\top x),
\end{align}
where $\kappa \geq 0$ is the concentration parameter,  $\epsilon\in \mathbb{S}^{d-1}$ is  the location vector, and $C_d(\kappa) : = \frac{\kappa^{d/2 -1}}{(2 \pi)^{d/2} I_{d/2 -1 }(\kappa) }$ is the normalization constant. Here, $I_v$ is the modified Bessel function of the first kind at order $v$ \cite{temme2011special}. 
\end{definition} 
 The vMF concentrates around mode $\epsilon$ and its density decreases when $x$ goes away from $\epsilon$.  When $\kappa \to 0$, vMF converges to the uniform distribution, and when $\kappa \to \infty$, vMF approaches to the Dirac distribution centered at $\epsilon$~\cite{Suvrit_directional}. These properties are illustrated by a toy example in Figure \ref{fig:sphere}.
\section{Spherical sliced fused Gromov Wasserstein and its relational regularized  Autoencoder}
\label{sec:general_SSFG}

In this section, we introduce a novel discrepancy, named \textit{spherical sliced fused Gromov Wasserstein} (SSFG), that searches for the best vMF distribution which distributes more masses to the most important area of projections on the unit sphere $\mathbb{S}^{d - 1}$. Then, we discuss an application of SSFG to the relational regularized autoencoder framework.
\subsection{Spherical sliced fused Gromov Wasserstein}
\label{sec:SSFG}
We first start with a definition of spherical sliced fused Gromov Wasserstein.
\begin{definition}
\label{def:SSFG}
(SSFG) Let  $\mu, \nu \in \mathcal{P}(\mathbb{R}^d)$ be two probability distributions, $\kappa > 0$, $\beta \in [0, 1]$, $d_{1}: \mathbb{R} \times \mathbb{R} \to \mathbb{R}_{+}$ be a pseudo-metric on $\mathbb{R}$. The \textbf{spherical sliced fused Gromov Wasserstein} (SSFG) between $\mu$ and $\nu$ is defined as follows:
\begin{align}
    \text{SSFG}(\mu,\nu; \beta, \kappa) : = \max_{\epsilon  \in \mathbb{S}^{d-1}}   \mathbb{E}_{\theta \sim \emph{vMF} ( \cdot | \epsilon,\kappa)}\big[D_{fgw}(\theta \sharp \mu,\theta \sharp \nu;\beta,d_1)\big] , \label{eq:def_SSFG}
\end{align}
where the fused Gromov Wasserstein $D_{fgw}$ is defined at equation~(\ref{def:fg}).
\end{definition}
A few comments on the SSFG are in order.  The family of vMF distributions is controlled by two parameters $\epsilon$ and $\kappa$, where $\epsilon$ is the mode of the vMF distribution while $\kappa$ controls its concentration. By changing $\kappa$ from $0$ to infinity, the vMF family could interpolate from the uniform distribution to any Dirac distribution on the sphere. In other words, it allows us to control distributing weight to the most important direction and other directions based on the geodesic distance on the sphere. Optimizing over the family of vMF distributions helps us to identify where the best direction is as well as how much weight we need to put there in comparison with other less important directions. 


\vspace{0.5 em}
\noindent
\textbf{Sampling procedure and reparameterization trick with the vMF:} To generate samples from vMF, we follow the procedure in~\cite{ulrich1984computer}, which is described in Algorithm \ref{Alg:vMF_sampling} in Appendix \ref{Sec:computational_detail_SSFG}. Note that this procedure does not suffer from the curse of dimensionality. Furthermore, to compute integral with respect to the vMF distribution, we use the reparametrization scheme in~\cite{naesseth2017reparameterization}, which was extended for vMF in Lemma 2 in~\cite{davidsonhyperspherical}. Finally,~\cite{davidsonhyperspherical} proved that samples from Algorithm~\ref{Alg:vMF_sampling} (cf. Appendix~\ref{subsec:vMF_sampling}) can provide a differentiable estimator for the parameters of vMF distribution. More details of this scheme are given in Appendix \ref{subsec:gradient_estimator}.
\begin{figure}[t]
\begin{center}

  \begin{tabular}{c}
 \includegraphics[scale=0.8]{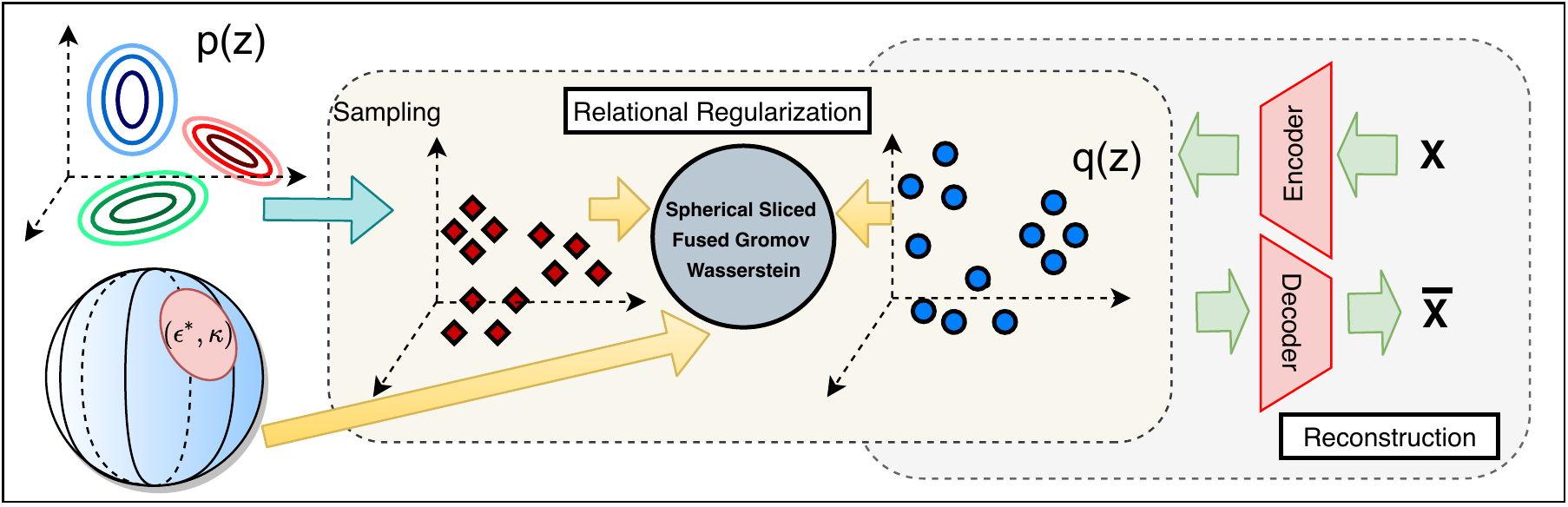} 
   
  \end{tabular}
  \caption{
  \footnotesize{ Visualization of the spherical deterministic relational regularized autoencoder (s-DRAE).}}
  \label{fig:drae}
  \vspace{-0.2 em}
  \end{center}
\end{figure}

\vspace{0.5 em}
\noindent
\textbf{Complexity of computing SSFG:} Let $\mu$ and $\nu$ be two discrete distributions that have $n$ supports with uniform weights. For the general case of $d_{1}$, similar to SFG, the complexity of computing SSFG can be expensive (at least of the order $\mathcal{O}(n^4)$ as the fused Gromov Wasserstein $D_{fgw}$ is a quadratic programming problem). However, the complexity of SSFG can be greatly improved under specific choices of $d_{1}$. For example, when $d_{1}(x, y) = (x - y)^2$, with a similar argument to the SFG case, the SSFG has computational complexity of the order $\mathcal{O}(n \log n)$.


\vspace{0.5 em}
\noindent
\textbf{Key properties of SSFG:} We first prove that SSFG is a pseudo-metric on the probability space.
\begin{theorem}
\label{theorem:SSFG-distance}
For any $\beta \in [0,1]$ and $\kappa > 0$, $\text{SSFG}(.,.;\beta, \kappa)$ is a pseudo-metric in the space of probability measures, namely, it is non-negative, symmetric, and satisfies the weak triangle inequality.
\end{theorem}
The proof of Theorem~\ref{theorem:SSFG-distance} is in Appendix~\ref{sec:proof:theorem:SSFG-distance}. Our next result establishes relations between SSFG and SFG and the max version of SFG, named as max-SFG.
\begin{theorem} 
\label{theorem:asymptotic_vMFsliced}
For any probability measures $\mu, \nu \in \mathcal{P}(\mathbb{R}^d)$, the following holds:
\begin{align*}
\text{(a)} \quad \quad  \quad \quad \quad \quad \quad \quad \quad \lim \limits_{\kappa \to 0} \text{SSFG} (\mu, \nu; \beta, \kappa) & = \text{SFG} (\mu,\nu; \beta), \\
    \lim \limits_{\kappa \to \infty} \text{SSFG} (\mu, \nu; \beta, \kappa) & = \max_{\theta \in \mathbb{S}^{d - 1}} D_{fgw}(\theta \sharp \mu,\theta \sharp \nu;\beta) : = \text{max-SFG} (\mu,\nu; \beta).
\end{align*}
(b) For any $\kappa > 0$, we find that
\begin{align*}
    \exp(-\kappa) C_{d}(\kappa) \text{SFG} (\mu, \nu; \beta) \leq \text{SSFG} (\mu, \nu; \beta, \kappa) & \leq \exp(\kappa) C_{d}(\kappa) \text{SFG} (\mu, \nu; \beta), \\
    \text{SSFG} (\mu, \nu; \beta, \kappa) & \leq \text{max-SFG}(\mu,\nu; \beta).
\end{align*}
\end{theorem}
The proof of Theorem~\ref{theorem:asymptotic_vMFsliced} is in Appendix~\ref{sec:proof:theorem:asymptotic_vMFsliced}.  Theorem~\ref{theorem:asymptotic_vMFsliced} shows that SSFG is an interpolation between SFG and max-SFG, namely, it combines the properties of both SFG and max-SFG. Furthermore, the result of part (b) of Theorem~\ref{theorem:asymptotic_vMFsliced} indicates that SSFG is strongly equivalent to SFG. 

Our next result shows that SSFG does not suffer from the curse of dimensionality for the inference purpose under certain choices of $d_{1}$. Therefore, it will be a statistically efficient discrepancy to compare the prior distribution to the encoder distribution in the DRAE framework.
\begin{theorem}
\label{theorem:stats_SSFG}
Assume that $\mu$ is a probability measure supported on a compact subset $\Theta \subset \mathbb{R}^{d}$. Let $X_{1}, \ldots, X_{n}$ be i.i.d. data from $P$ and $d_{1}(x, y) = |x - y|^{r}$ for a positive integer $r$. We denote by $\mu_{n} = \frac{1}{n} \sum_{i = 1}^{n} \delta_{X_{i}}$ the empirical measure of the data points $X_{1}, \ldots, X_{n}$. Then, for any $\beta \in [0,1]$ and $\kappa > 0$, there exists a constant $c$ depending only on $r$ and diameter of $\Theta$ such 
 that
\begin{align*}
     \mathbb{E} \Big[\text{SSFG}(\mu_{n},\mu; \beta, \kappa)\Big] 
\leq \frac{c}{n}.
\end{align*}
\end{theorem}
Theorem~\ref{theorem:stats_SSFG} together with the earlier argument about the computational complexity of SSFG suggests that the choice of $d_{1}(x, y) = (x - y)^2$ is not only convenient for the computation but also statistically efficient. Therefore, we will specifically use this choice of $d_{1}$ in our experiments in Section~\ref{Sec:experiments}. 

\vspace{0.5 em}
\noindent
\textbf{Spherical deterministic relational regularized autoencoder:}
We replace SFG by SSFG in the deterministic relational regularized autoencoder framework in equation~(\ref{eq:drae_sfg}) to obtain a new variant of DRAE with a stronger relational regularization. The new autoencoder is named as \textit{spherical deterministic relational regularized autoencoder} (s-DRAE). Intuitive visualization of s-DRAE is presented in Figure~\ref{fig:drae}. The detailed training procedure for s-DRAE is left in Appendix~\ref{subsec:training_SDRAE}.

\subsection{Extensions and Variants of SSFG}
\label{sub:extension_SSFG}
We first propose an extension of SSFG to its mixture variant. 
\begin{definition}
\label{def:MSFG}
(MSSFG) Let $\mu,\nu \in \mathcal{P}(\mathbb{R}^d)$ be two probability distributions, $\beta \in [0,1]$ be a constant, $\{\alpha_i\}_{i=1}^k$ be given mixture weights, and $\{\kappa_{i}\}_{i = 1}^{k}$ be given mixture concentration parameters where $k \geq 1$. Furthermore, let $d_1:\mathbb{R} \times \mathbb{R} \to \mathbb{R}_+$ be a pseudo-metric on $\mathbb{R}$. Then, the \textbf{mixture spherical sliced fused Gromov Wasserstein} (MSSFG) between $\mu$ and $\nu$ is defined as follows:
\begin{align}
    \text{MSSFG}(\mu,\nu;\beta,\{\kappa_i\}_{i=1}^k, \{\alpha_i\}_{i=1}^k) & \nonumber \\
    & \hspace{- 6 em} : = \max_{\epsilon_{1:k}  \in \mathbb{S}^{d-1}}  \mathbb{E}_{\theta \sim \text{MovMF} ( \cdot | \epsilon_{1:k},\{\kappa_i\}_{i=1}^k, \{\alpha_i\}_{i=1}^k)}\big[D_{fgw}(\theta \sharp \mu,\theta \sharp \nu;\beta,d_1)\big] ,
\end{align}
where $D_{fgw}$ is defined in equation~(\ref{def:fg}) and the mixture of vMF distributions is defined as \\ $\text{MovMF} ( \cdot | \epsilon_{1:k},\{\kappa_i\}_{i=1}^k, \{\alpha_i\}_{i=1}^k) : = \sum_{i=1}^k \alpha_i \emph{vMF}(\cdot|\epsilon_i,\kappa_i)$.
\end{definition}
\paragraph{Comparison between MSSFG and SSFG:} When $k = 1$, the MSSFG becomes SSFG. Recall that SSFG tries to search for the best location parameter in the unit sphere $\mathbb{S}^{d - 1}$ that maximizes the expected value of the fused Gromov Wasserstein between the projected probability measures. Intuitively, it places a large weight on the best projection and some weights on other important projections. However, if these important projections are far from the best projection, i.e., the center of the best von Mises-Fisher distribution, their weights will be very small, which can be undesirable. To account for this issue, the mixture of von Mises-Fisher distributions aims to find $k$ best location parameters whose weights are guaranteed to be large enough. Furthermore, when $k$ is chosen to be sufficiently large, mixture of von Mises-Fisher distributions will give a good coverage of the unit sphere; therefore, the important directions that MSSFG can find will be able to reflect more accurate differences between the target probability distributions than those from SSFG. 

\vspace{0.5 em}
\noindent
\textbf{Properties of MSSFG and its DRAE version:} As SSFG, MSSFG is a pseudo-metric in the probability space and does not suffer from the curse of dimensionality. Its computational complexity is of the order $\mathcal{O}(n \log n)$ when $\mu$ and $\nu$ are discrete measures with $n$ atoms and uniform weights and $d_{1}(x, y) = (x - y)^2$. The detailed discussion of the properties of MSSFG is in Appendix~\ref{Sec:MSFG}. An application of MSSFG to the DRAE framework leads to the \emph{mixture spherical DRAE} (ms-DRAE). 

\vspace{0.5 em}
\noindent
\textbf{Improving computational time of (M)SSFG:} Drawing the samples from the vMF distribution and its mixtures can be slow in high dimension settings, which affects the computation of (M)SSFG. To account for this issue, we propose using \emph{power spherical distribution}~\cite{de2020power} instead of vMF and its mixtures as the slicing distribution to improve the computational time of (M)SSFG. It leads to a new discrepancy, named \emph{power SSFG} (PSSFG), between the probability distributions (see Appendix~\ref{sec:power_SFFG} for the definition). In Section~\ref{Sec:experiments}, we show that PSSFG has better computational time than (M)SSFG while its DRAE version, named \emph{power spherical DRAE} (ps-DRAE), has comparable performance to SSFG.  

\section{Experiments}
\label{Sec:experiments}
In this section,  we conduct extensive experiments on MNIST \cite{lecun1998gradient} and CelebA datasets \cite{liu2015faceattributes} to evaluate the performance of s-DRAE, ms-DRAE, and ps-DRAE with various autoencoders, including DRAE (trained by SFG), PRAE~\cite{xu2020learning}, m-DRAE (trained by max-SFG---see its definition in Theorem~\ref{theorem:asymptotic_vMFsliced}), VAE~\cite{kingma2013auto}, WAE~\cite{tolstikhin2018wasserstein}, SWAE~\cite{kolouri2018sliced}, GMVAE~\cite{dilokthanakul2016deep}, and the VampPrior~\cite{tomczak2018vae}. We use two standard scores as evaluation metrics: (i) the Frechet Inception distance (FID) score~\cite{heusel2017gans} is used to measure the generative ability; (ii) the reconstruction score is used to evaluate the reconstruction performance computed on the test set. For the computational details of the FID  score, we compute the score between 10000 randomly generated samples and all samples from the test set of each dataset. To guarantee the fairness of the comparison, we use the same autoencoder architecture, Adam optimizer with learning rate = 0.001, $\beta_1=0.5$ and $\beta_2=0.999$; batch size = 100; latent size = 8 on MNIST and  64 on CelebA; coefficient $\lambda$=1; fused parameter $\beta$ = 0.1.  We set the number of components $K=10$ for autoencoder with a mixture of Gaussian distribution as the prior. More detailed descriptions of these settings are in Appendix~\ref{Sec:exp_setting}. 

\vspace{0.5 em}
\noindent
\textbf{Comparing with other autoencoders:}
We first report the performances of autoencoders on MNIST~\cite{lecun1998gradient} and CelebA datasets~\cite{liu2015faceattributes}. Table~\ref{tab:FIDtable} presents the FID scores and reconstruction losses of trained autoencoders. All results are obtained from five different runs and reported with empirical mean and standard deviation. On the MNIST dataset, ms-DRAE achieves the lowest scores in both FID score and reconstruction loss among all the autoencoders. In addition, s-DRAE and ps-DRAE also have better scores than DRAE. On the CelebA dataset, we cannot reproduce results from VAE, PRAE, and WAE; therefore, we use the results with these autoencoders from DRAE paper~\cite{xu2020learning}. Table~\ref{tab:FIDtable} suggests that ms-DRAE also obtains the lowest mean and standard deviation in FID score than other autoencoders, meanwhile its reconstruction loss is almost the same as other DRAEs. The FID scores of s-DRAE and ps-DRAE are also better than those of DRAE. These results suggest that the proposed spherical discrepancies truly improve the performances of the DRAE framework. In these experiments, we set the number of projections  $L=50$ for every sliced-discrepancy. For s-DRAE, ps-DRAE and ms-DRAE (10 vMF components with uniform weights and same concentration parameters), we search for $\kappa \in \{ 1,5,10,50,100\}$ which gives the best FID score on the validation set of the corresponding dataset. By tuning $\kappa$, we find that the performance of both  s-DRAE and ps-DRAE is close to that of DRAE when $\kappa= 1$, namely, the reconstruction loss and FID score are nearly equal to the scores of DRAE. On the other extreme, when $\kappa=100$, s-DRAE and ps-DRAE behave like m-DRAE in both evaluation metrics. Further details are given in Figures~\ref{fig:kappa} and~\ref{fig:psdrae_kappa} in Appendices~\ref{subsec:more_result_sdrae} and~\ref{subsec:exp_ps-DRAE} respectively.

Detailed results including generated images, reconstruction images and visualizing latent spaces  are in Appendix \ref{subsec:more_result_sdrae}. These results indicate that s-DRAE, ps-DRAE, and ms-DRAE can learn nice latent structures and mixture Gaussian priors which can cover well these latent spaces. As a consequence, the spherical DRAEs can produce good generated images and reconstruct images correctly.
\begin{table}[t]
 \caption{FID scores and reconstruction losses of different autoencoders. (*) denotes the results that are taken from~\cite{xu2020learning} due to the reproducing failure. The results are taken from 5 different runs.}
    \centering
    \begin{tabular}{c|cc|cc}
    \toprule
         \multirow{2}{*}{Method}& \multicolumn{2}{c}{MNIST} & \multicolumn{2}{c}{CelebA}   \\
         
         & FID&Reconstruction&FID&Reconstruction\\
         \midrule
         VAE & 71.55 $\pm$ 26.65&18.59 $\pm$ 2.22  &59.99(*)&96.36(*)\\
         GMVAE& 75.68 $\pm$11.95&18.19 $\pm$ 0.14&212.59 $\pm$ 18.15&97.77$\pm$ 0.19\\
         Vampprior &138.03 $\pm$ 34.09  &29.98$\pm$ 4.09&- & - \\
         PRAE  & 100.25 $\pm$41.72&16.20 $\pm$3.14 &52.20 (*)&  \textbf{63.21(*)}\\
         \midrule
         WAE & 80.77 $\pm$ 11&11.53 $\pm$0.33 &52.07 (*)&63.83(*) \\
         SWAE & 80.28 $\pm$19.22&14.12 $\pm$ 2.06&86.53 $\pm$ 2.49 &89.71$\pm$2.15\\
         DRAE& 58.04 $\pm$ 20.74& 14.07 $\pm$ 4.31 & 50.09 $\pm$1.33 &66.05 $\pm$ 2.56\\ 
         m-DRAE (ours) & 52.92 $\pm$13.81& 13.13 $\pm$ 0.33& 49.05 $\pm$ 0.93&66.30 $\pm$ 0.22\\
         s-DRAE  (ours)& 47.97 $\pm$ 13.83 & 11.17 $\pm$1.73 & 46.63 $\pm$0.83&66.62 $\pm$ 0.51\\
          ps-DRAE  (ours)& 49.15 $\pm$ 12.93 & 11.71 $\pm$1.21 & 48.21 $\pm$1.02 &66.31 $\pm$ 0.43\\
           ms-DRAE  (ours)& \textbf{43.57 $\pm$ 10.98} & \textbf{11.12 $\pm$0.91} & \textbf{46.01 $\pm$0.91}&65.91$\pm$ 0.4\\
         
    \bottomrule
    \end{tabular}
   
    \label{tab:FIDtable}
\end{table}

\begin{figure}[t]
\begin{center}

  \begin{tabular}{ccc}
 \includegraphics[scale=0.5]{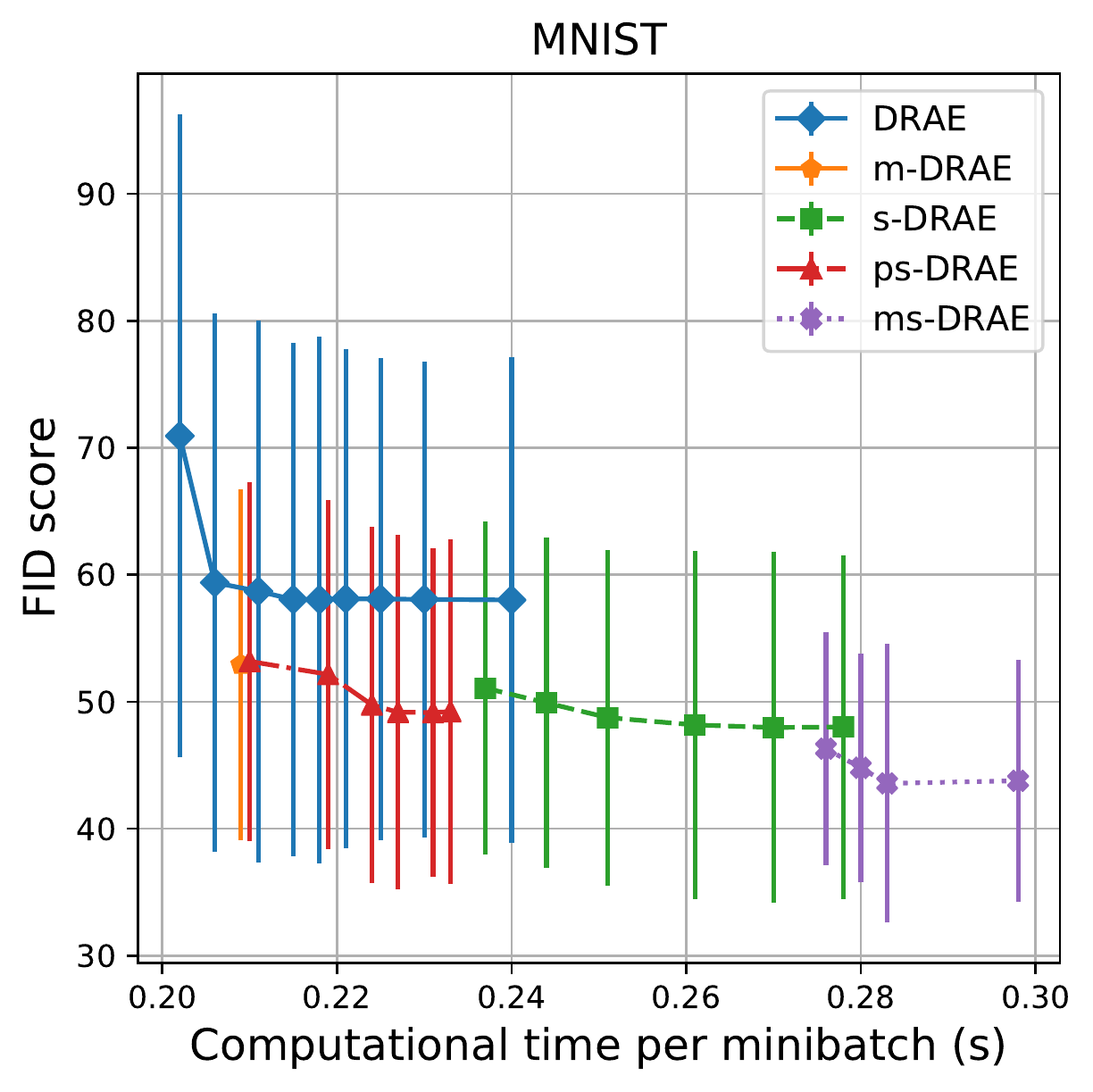} 
    \quad \quad \quad \quad \quad &
    \includegraphics[scale=0.5]{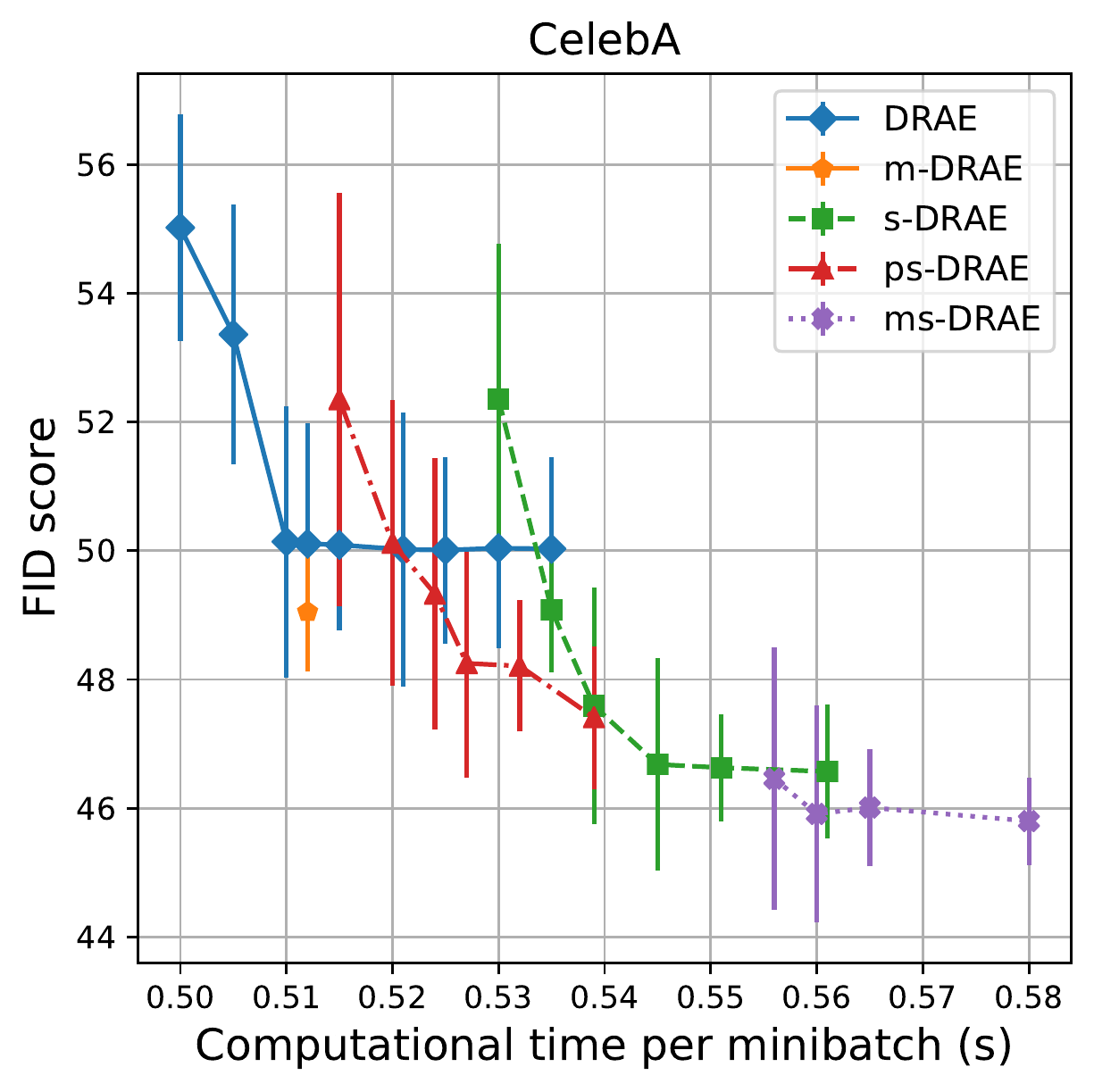}
   & 
   
  \end{tabular}
 \end{center}
  \caption{
  \footnotesize{Each dot  represents the computational time per minibatch and FID score. For DRAE, we vary the number of projections $L \in \{1,5,10,20,50,100,200,500,1000\}$}; for s-DRAE we set $\kappa=10$, $L \in \{1,5,10,20,50,100\}$; for ps-DRAE we set $\kappa=50, L \in \{1,5,10,20,50,100\}$; and for ms-DRAE we set $L=50$, the number of vMF components $k \in \{2,5,10,50\}$ (for each $k$, we find the best $\kappa \in \{1,5,10,50,100\}$).}
  \label{fig:comaring_MNIST}
  \vspace{-0.2 em}
\end{figure}

\vspace{0.5 em}
\noindent
\textbf{Detailed comparisons among deterministic RAEs:} 
\label{subsec:compare_DRAE}
It is well-known that the quality and computational time of sliced-discrepancies depend on the number of projections~\cite{kolouri2018sliced,kolouri2019generalized,deshpande2019max}. Therefore, we carry out experiments on MNIST and CelebA datasets to compare  ps-DRAE, ms-DRAE, s-DRAE to DRAE in a wide range the of number of projections. In detail, we set the number of projections $L \in \{1,5,10,20,50,100,200,500,1000\}$ in DRAE ; and $L \in \{1,5,10,20,50,100\}$ for s-DRAE ($\kappa=10$), and ps-DRAE ($\kappa=50$). We then report the  (minibatch) computation time and FID score
of these autoencoders in Figure~\ref{fig:comaring_MNIST}. In this figure, we also plot the time and FID scores of m-DRAE and ms-DRAE (using $L=50$ and the number of vMF components $k \in \{2,5,10,50\}$.  On the MNIST dataset, with only 1 projection, s-DRAE achieves lower FID score than all settings of DRAE; however, s-DRAE requires more time to train due to the sampling of vMF and its optimization problem. Having faster sampling procedure of PS distribution, ps-DRAE has better computational time than s-DRAE while it still has a comparable performance to s-DRAE. With the computational time greater than about 0.21(s), ps-DRAE always produces lower FID score than DRAE. We also observe the same phenomenon on CelebA dataset, namely, ps-DRAE and s-DRAE have lower FID score than DRAE with any value of $L$ but $L = 1$. Between s-DRAE and ps-DRAE, s-DRAE gives better results but ps-DRAE is faster in training. On both the datasets, m-DRAE has  faster speed than (p)s-DRAE but its FID score is higher. In terms of the FID score, ms-DRAE is the best autoencoder though it can be more expensive in training. Finally, we observe that increasing the number of vMF components can enhance the FID score but also worsen the speed.

\vspace{0.5 em}
\noindent
\textbf{Additional experiments:} We provide further comparisons between MSSFG, PSSFG, SSFG and SFG in ex-post density estimation of autoencoder~\cite{ghosh2019variational} and GAN~\cite{goodfellow2014generative} applications in Appendices~\ref{subsec:pde} and~\ref{subsec:GAN}. In the ex-post density estimation framework, we find that MSSFG, PSSFG and SSFG give better FID score than SFG. Like traditional training procedures, MSSFG achieves the best performance in this task. In the GAN application, we use a toy example, which is to learn a generator to produce 4 Gaussian modes. We observe that MSSFG, PSSFG and SSFG help the model distribution converge faster than SFG does.

\section{Conclusion}
In the paper, we first introduced a new spherical relational discrepancy, named spherical sliced fused Gromov Wasserstein (SFFG), between the probability measures. This discrepancy is obtained by replacing the uniform distribution over slicing direction in sliced fused Gromov Wasserstein by a von Mises-Fisher distribution that can cover the most informative area of directions. To improve the performance and stability of SFFG, we then propose two variants of SSFG: (i) the first variant is mixture SSFG (MSSFG), obtained by using a mixture of vMF distributions instead of a single vMF distribution to capture more informative areas of directions; (ii) the second variant is power SSFG (PSSFG), obtained by replacing the vMF distribution by the power spherical distribution to improve the sampling time of vMF distribution in high dimension settings. An application of these discrepancies to the DRAE framework leads to several new variants of DRAE. Extensive experiments show that these new autoencoders are more stable and achieve better generative performance than the previous autoencoders, including DRAE, in comparable computational time. 
\label{Sec:conclusion}



\bibliographystyle{abbrv}
\bibliography{iclr2021_conference}
\clearpage
\appendix
\begin{center}
\textbf{\Large{Supplement to ``Improving Relational Regularized Autoencoders \\ with Spherical Sliced Fused Gromov Wasserstein''}}
\end{center}
In this supplementary material, we collect several proofs and remaining materials that were deferred from the main paper. In Appendix~\ref{sec:proof}, we provide the proofs of the main results in the paper. In Appendix~\ref{Sec:computational_detail_SSFG}, further computational details of spherical sliced fused Gromov Wasserstein and its corresponding relational regularized autoencoder are given. We discuss two key extensions of spherical sliced fused Gromov Wasserstein (SSFG) in Appendices~\ref{Sec:MSFG} and~\ref{sec:power_SFFG}. Additional experiments for the generative models are presented in Appendix~\ref{sec:additional_experiments}. Finally, detailed experimental settings are in Appendix~\ref{Sec:exp_setting}.  
\section{Proofs}
\label{sec:proof}
In this appendix, we give the detailed proof for all the results in the main text.
\subsection{Proof of Theorem~\ref{theorem:SSFG-distance}}
\label{sec:proof:theorem:SSFG-distance}
In order to facilitate the ensuing presentation, for any probability measures $\mu$ and $\nu$ on $\mathbb{P}(\mathbb{R}^{d})$ and joint probability measure $\gamma$ of $\mu$ and $\nu$, we denote $\text{proj}^{1}(\gamma) = \mu$ and $\text{proj}^{2}(\gamma) = \nu$. From the definition of spherical sliced fused Gromov Wasserstein, it is clear that SSFG is symmetric and non-negative. Therefore, we will only need to prove that it satisfies the weak triangle inequality. Recall that, since $d_{1}$ is pseudo-metric, there exists a constant $C > 0$ such that for any $x, y, z \in \mathbb{R}$, we have
\begin{align*}
    d_{1}(x, z) \leq C \big[d_{1}(x, y) + d_{1}(y, z)\big].
\end{align*}
Now, assume that $\mu, \nu, \xi$ are probability measures on $\mathbb{R}^{d}$. For any $\theta \in \mathbb{S}^{d- 1}$, we respectively denote $\pi_{1}$ and $\pi_{2}$ as the optimal couplings that minimize the fused Gromov Wasserstein between $\theta \sharp \mu$ and $\theta \sharp \nu$ and $\pi_{2}$ and the fused Gromov Wasserstein between $\theta \sharp \nu$ and $\theta \sharp \xi$. Then, by the gluing lemma~\cite{Villani-03}, there exists a probability measure $\gamma \in \mathbb{P}(\mathbb{R} \times \mathbb{R} \times \mathbb{R})$ such that $\text{proj}^{1,2}(\gamma) = \pi_{1}$ and $\text{proj}^{2,3}(\gamma) = \pi_{2}$. Since $\text{proj}^{1}(\gamma) = \theta \sharp \mu$ and $\text{proj}^{3}(\gamma) = \theta \sharp \xi$, we obtain that $\text{proj}^{1,3}(\gamma) \in \Pi(\theta \sharp \mu, \theta \sharp \xi)$. From the definition of fused Gromov Wasserstein, we find that
\begin{align}
    D_{fgw}(\theta \sharp \mu,\theta \sharp \xi;\beta,d_1) & = \min_{\pi \in \Pi (\theta \sharp \mu,\theta \sharp \xi;\beta)} \biggr\{ (1-\beta) \int_{\mathbb{R}^d \times \mathbb{R}^d}  d_{1}(\theta^\top x, \theta^\top z) d \pi(x,z) \nonumber \\ & + \beta  \int_{(\mathbb{R}^d)^4}\big[d_{1}(\theta^\top x, \theta^\top x^{\prime}) - d_{1}(\theta^\top z, \theta^\top z^\prime)\big]^2 d\pi(x,z) d \pi (x^\prime, z^\prime) \biggr\} \nonumber \\
    & \leq (1-\beta) \int_{\mathbb{R}^d \times \mathbb{R}^d}  d_{1}(\theta^\top x, \theta^\top z) d \text{proj}^{1,3}(\gamma)(x, z) \nonumber \\ & + \beta  \int_{(\mathbb{R}^d)^4}\big[d_{1}(\theta^\top x, \theta^\top x^{\prime}) - d_{1}(\theta^\top z, \theta^\top z^\prime)\big]^2 d \text{proj}^{1,3}(\gamma)(x, z) d \text{proj}^{1,3}(\gamma)(x', z') \nonumber \\
    & = (1-\beta) \int_{\mathbb{R}^d \times \mathbb{R}^d \times \mathbb{R}^{d}}  d_{1}(\theta^\top x, \theta^\top z) d \gamma(x, y, z) \nonumber \\ & + \beta  \int_{(\mathbb{R}^d)^6}\big[d_{1}(\theta^\top x, \theta^\top x^{\prime}) - d_{1}(\theta^\top z, \theta^\top z^\prime)\big]^2 d \gamma(x, y, z) d \gamma(x', y', z'). \label{eq:triangle_zero}
\end{align}
Since $d_{1}(\theta^{\top}x, \theta^{\top}z) \leq C\big[ d_{1}(\theta^{\top}x, \theta^{\top}y) + d_{1}(\theta^{\top}y, \theta^{\top}z)\big]$, we obtain that
\begin{align}
    \int_{\mathbb{R}^d \times \mathbb{R}^d}  d_{1}(\theta^\top x, \theta^\top z) d \gamma(x, y, z) & \leq C \int_{\mathbb{R}^d \times \mathbb{R}^d \times \mathbb{R}^{d}} \big[ d_{1}(\theta^{\top}x, \theta^{\top}y) + d_{1}(\theta^{\top}y, \theta^{\top}z)\big] d \gamma(x, y, z) \nonumber \\
    & = C \biggr\{ \int_{\mathbb{R}^d \times \mathbb{R}^d} d_{1}(\theta^{\top}x, \theta^{\top}y) d \pi_{1}(x, y) \nonumber \\
    & \hspace{8 em} + \int_{\mathbb{R}^d \times \mathbb{R}^d} d_{1}(\theta^{\top}y, \theta^{\top}z) d \pi_{1}(y, z)\biggr\}. \label{eq:triangle_first}
\end{align}
Furthermore, an application of Cauchy-Schwarz inequality leads to
\begin{align}
    & \hspace{- 5 em} \int_{(\mathbb{R}^d)^6}\big[d_{1}(\theta^\top x, \theta^\top x^{\prime}) - d_{1}(\theta^\top z, \theta^\top z^\prime)\big]^2 d \gamma(x, y, z) d \gamma(x', y', z')  \nonumber \\
    & \leq 2 \biggr\{ \int_{(\mathbb{R}^d)^6}\big[d_{1}(\theta^\top x, \theta^\top x^{\prime}) - d_{1}(\theta^\top y, \theta^\top y^\prime)\big]^2 d \gamma(x, y, z) d \gamma(x', y', z') \nonumber \\
    & \quad + \int_{(\mathbb{R}^d)^6}\big[d_{1}(\theta^\top y, \theta^\top y^{\prime}) - d_{1}(\theta^\top z, \theta^\top z^\prime)\big]^2 d \gamma(x, y, z) d \gamma(x', y', z') \biggr\} \nonumber \\
    & = 2 \biggr\{ \int_{(\mathbb{R}^d)^4}\big[d_{1}(\theta^\top x, \theta^\top x^{\prime}) - d_{1}(\theta^\top y, \theta^\top y^\prime)\big]^2 d \pi_{1}(x, y) d \pi_{1}(x', y') \nonumber \\
    & \quad + \int_{(\mathbb{R}^d)^4}\big[d_{1}(\theta^\top y, \theta^\top y^{\prime}) - d_{1}(\theta^\top z, \theta^\top z^\prime)\big]^2 d \pi_{2}(y, z) d \pi_{2}(y', z') \biggr\}. \label{eq:triangle_second}
\end{align}
Plugging the results in equations~(\ref{eq:triangle_first}) and~(\ref{eq:triangle_second}) into the upper bound of $D_{fgw}(\theta \sharp \mu,\theta \sharp \xi;\beta,d_1)$ in equation~(\ref{eq:triangle_zero}), for any $\theta \in \mathbb{S}^{d - 1}$ we have
\begin{align*}
    D_{fgw}(\theta \sharp \mu,\theta \sharp \xi;\beta,d_1) \leq \max \{C, 2\} \biggr\{D_{fgw}(\theta \sharp \mu,\theta \sharp \nu;\beta,d_1) + D_{fgw}(\theta \sharp \nu,\theta \sharp \xi;\beta,d_1) \biggr\}.
\end{align*}
Based on that inequality, we obtain that
\begin{align*}
    \text{SSFG}(\mu,\xi; \beta, \kappa) \leq \max \{C, 2\} \big[\text{SSFG}(\mu,\nu; \beta, \kappa) + \text{SSFG}(\nu,\xi; \beta, \kappa)\big].
\end{align*}
As a consequence, the spherical sliced fused Gromov Wasserstein satisfies the weak triangle inequality, which concludes the theorem. 
\subsection{Proof of Theorem~\ref{theorem:asymptotic_vMFsliced}}
\label{sec:proof:theorem:asymptotic_vMFsliced}
(a) We first prove that $\text{SSFG} (\mu,\nu; \beta, \kappa)$ is continuous in terms of $\kappa$. Indeed, the function \\$\mathbb{E}_{\theta \sim \vMF ( \cdot | \epsilon,\kappa)}\big[D_{fgw}^2(\theta \sharp \mu,\theta \sharp \nu;\beta)\big]$ is continuous in terms of both $\epsilon$ and $\kappa$. Therefore, based on maximum theorem, $\text{SSFG} (\mu,\nu; \beta, \kappa)$ is continuous in terms of $\kappa$. 

When $\kappa$ goes to 0, the density of $\vMF (.|\epsilon,\kappa)$ converges to that of $\mathcal{U}(\mathbb{S}^{d-1})$~\cite{Suvrit_directional}. Therefore, given the continuity of SSFG in terms of $\kappa$ we have
\begin{align*}
    \lim \limits_{\kappa \to 0} \text{SSFG} (\mu, \nu; \beta, \kappa) & = \max_{\epsilon \in \mathbb{S}^{d - 1}} \biggr( \lim \limits_{\kappa \to 0} \mathbb{E}_{\theta \sim \text{vMF}(.|\epsilon,\kappa)} \big[D_{fgw}(\theta \sharp \mu,\theta \sharp \nu;\beta)\big] \biggr) \\
    & = \max_{\epsilon \in \mathbb{S}^{d - 1}} \biggr( \mathbb{E}_{\theta \sim \mathcal{U}(\mathbb{S}^{d-1})} \big[D_{fgw}(\theta \sharp \mu,\theta \sharp \nu;\beta)\big] \biggr) = \text{SFG}(\mu,\nu; \beta),
\end{align*}
which confirms our conclusion that the spherical sliced fused Gromov Wasserstein becomes the sliced fused Gromov Wasserstein when $\kappa \to 0$. 

When $\kappa \to \infty$, the density of $\text{vMF}(.|\epsilon,\kappa)$ converges to the density of $\delta_{\epsilon}$~\cite{Suvrit_directional}. Therefore, based on Scheffe's lemma~\cite{Feller_introduction}, the density of $\text{vMF}(.|\epsilon,\kappa)$ converges in L1-norm to that of $\delta_{\epsilon}$. Given that result, as $D_{fgw}(\theta \sharp \mu,\theta \sharp \nu;\beta)$ is uniformly bounded for all $\theta \in \mathbb{S}^{d- 1}$, we arrive at
\begin{align*}
    \lim \limits_{\kappa \to \infty} \text{SSFG} (\mu, \nu; \beta, \kappa) & = \max_{\epsilon \in \mathbb{S}^{d - 1}} \biggr( \lim \limits_{\kappa \to \infty} \mathbb{E}_{\theta \sim \text{vMF}(.|\epsilon,\kappa)} \big[D_{fgw}(\theta \sharp \mu,\theta \sharp \nu;\beta)\big] \biggr) \\
    & = \max_{\epsilon \in \mathbb{S}^{d - 1}} \biggr( \mathbb{E}_{\theta \sim \delta_{\epsilon}} \big[D_{fgw}(\theta \sharp \mu,\theta \sharp \nu;\beta)\big] \biggr) = \text{max-SFG}(\mu,\nu; \beta).
\end{align*}
As a consequence, we obtain the conclusion of part (a) of the theorem.

(b) For any $\epsilon, \theta \in \mathbb{S}^{d - 1}$, using Cauchy-Schwarz inequality, we get $-1\leq \epsilon^{\top} x\leq 1$. Hence, for $\kappa > 0$, it is clear that $\exp(-\kappa) \leq \exp(\kappa \epsilon^\top x) \leq \exp(\kappa)$. Consequently, we obtain that
\begin{align*}
    \text{SSFG} (\mu, \nu; \beta, \kappa) & = \max_{\epsilon  \in \mathbb{S}^{d-1}} \biggr( \mathbb{E}_{\theta \sim \vMF ( \cdot | \epsilon,\kappa)}\big[D_{fgw}(\theta \sharp \mu,\theta \sharp \nu;\beta)\big] \biggr) \\
    & \leq \max_{\epsilon  \in \mathbb{S}^{d-1}} \biggr( \exp(\kappa) \mathbb{E}_{\theta \sim \vMF ( \cdot | \epsilon,\kappa)}\big[D_{fgw}^2(\theta \sharp \mu,\theta \sharp \nu;\beta)\big] \biggr) \\
    & = \exp(\kappa) C_{d}(\kappa) \text{SFG} (\mu, \nu; \beta).
\end{align*}
Similarly, by using the bound $\exp(-\kappa) \leq \exp(\kappa \epsilon^\top x)$, we also arrive at the bound $\exp(-\kappa) C_{d}(\kappa) \text{SFG} (\mu, \nu; \beta) \leq \text{SSFG} (\mu, \nu; \beta, \kappa)$. Therefore, we obtain the bounds of SSFG based on SFG. 

Regarding the upper bound of SSFG based on max-SFG, it is straight-forward from the inequality $D_{fgw}(\theta \sharp \mu,\theta \sharp \nu;\beta) \leq \max_{\theta' \in \mathbb{S}^{d - 1}} D_{fgw}(\theta' \sharp \mu,\theta' \sharp \nu;\beta)$. As a consequence, we obtain the conclusion of part (b) of the theorem.
\subsection{Proof of Theorem~\ref{theorem:stats_SSFG}}
For any positive integer  $r$, we first prove the following simple inequality
\begin{align}
    \biggr\{|\theta^{\top}x - \theta^{\top}x'|^{r} - |\theta^{\top}y - \theta^{\top}y'|^{r} \biggr\}^2 \leq C \biggr \{|\theta^{\top} x - \theta^{\top}y|^{2} + |\theta^{\top} x' - \theta^{\top} y'|^2 \biggr\}, \label{eq:simple_ineq}
\end{align}
for any $\theta \in \mathbb{S}^{d - 1}$ and $x, y, x', y' \in \Theta$. Here, $C$ is some universal constant depending only on the diameter of $\Theta$. For simplicity, denote 
\begin{align*}
    \theta^{\top} x =a ; \quad \theta^{\top} x^{\prime} = b; \quad \theta^{\top} y = c; \quad \theta^{\top} y^{\prime} = d;\\
    C(\Omega) = \text{diameter of $\Omega$}.
\end{align*}
By triangle's inequality, we find that
\begin{align*}
    \Big||a-b| - |c-d| \Big| \leq |a-c| + |b-d|.
\end{align*}
 Moreover, we have the identity
\begin{align*}
    |a-b|^r - |c-d|^r = \big[|a-b|-|c-d| \big] \sum_{i=0}^{r-1} |a-b|^i |c-d|^{r-i-1}.   
\end{align*}
Note that, the absolute values of $a,b,c,d$ are not greater than $C(\Omega)$. It follows that
\begin{align*}
    \Big||a-b|^r - |c-d|^r \Big| \leq \big[|a-c| + |b-d| \big] r 2^{r-1}C(\Omega)^{r-1}.
\end{align*}
By using Cauchy-Schwarz inequality, we obtain
\begin{align*}
    \Big\{|a-b|^r - |c-d|^r\Big\}^2 &\leq 2^{2r-2}r^2 C(\Omega)^{2r-2} \Big[|a-c| + |b-d| \Big]^2 \\
    &\leq 2^{2r-1} r^2 C(\Omega)^{2r-2} \Big[|a-c|^2 + b-d|^2 \Big].
\end{align*}
Given the claim~(\ref{eq:simple_ineq}), we obtain that
\begin{align*}
    \text{SFG}(\mu_{n},\mu; \beta) \leq C_{1} \mathbb{E}_{\theta \sim \mathcal{U}(\mathbb{S}^{d-1})} \Big[\min_{\pi \in \Pi (\theta \sharp \mu_{n},\theta \sharp \mu)} (\theta^\top x - \theta^\top y)^2 d \pi(x,y)\Big] : = SW(\mu_{n}, \mu),
\end{align*}
where $C_{1}$ is some universal constant. The RHS of the above inequality is known as second order sliced Wasserstein distance between $\mu_{n}$ and $\mu$. Since $\Theta$ is compact, the result of~\cite{Bobkov-2019} leads to
\begin{align*}
    \mathbb{E} \big[ SW(\mu_{n}, \mu)\big] \leq \frac{C_{2}}{n}.
\end{align*}
Hence, it demonstrates that 
\begin{align*}
    \mathbb{E} \big[\text{SFG}(\mu_{n},\mu; \beta)\big] \leq \frac{C_{1} C_{2}}{n},
\end{align*}
for all $\beta \in [0, 1]$. 
Combining the above result with the result of part (b) in Theorem~\ref{sec:proof:theorem:asymptotic_vMFsliced}, we obtain that
\begin{align*}
    \mathbb{E} \biggr[\text{SSFG}(\mu_{n},\mu; \beta, \kappa)\biggr] 
\leq \mathbb{E} \big[\text{SFG}(\mu_{n},\mu; \beta) \big] \leq \frac{c}{n},
\end{align*}
where $c = C_{1} C_{2}$. As a consequence, we obtain the conclusion of the theorem.

\section{Computational details of spherical sliced fused Gromov Wasserstein and its relational regularized autoencoder}
\label{Sec:computational_detail_SSFG}
In this section, we present the sampling algorithm (Algorithm~\ref{Alg:vMF_sampling}) of von Mises-Fisher distribution, which follows the scheme in \cite{ulrich1984computer,davidsonhyperspherical}, and how to derive the gradient estimator of the SSFG. Finally, we give the detail of the training procedure of s-DRAE (Algorithm \ref{Alg:ssfg_drae}).

\subsection{Sampling from von Mises-Fisher distribution}
\label{subsec:vMF_sampling}
As discussed in \cite{davidsonhyperspherical} to sample  $\theta \sim \vMF(.|\epsilon,\kappa)$, we first need to sample  $h_1 \sim \vMF(\cdot|e_1, \kappa)$ where $e_1 = (1,0,..,0)$. To do this step, we need to sample $\omega$ from the  univariate density $g(\omega|\kappa,d) \propto \exp{(\kappa \omega)} (1-\omega^2)^{\frac{d-3}{2}}$  ($\omega\in [-1,1]$) using an acceptance-rejection scheme, then compute $h_1= (\omega,\sqrt{1-\omega^2} v^\top)^\top$ where $v\sim \mathcal{U}(\mathbb{S}^{d-2})$. Next, we find a orthogonal transformation $U$ such that $U(\epsilon) e_1= \epsilon $ by using Householder reflection. Finally, we compute $\theta = U h_1$, then $\theta$ is a sample from $\vMF(.|\epsilon,\kappa)$ as proved in \cite{ulrich1984computer}. Detailed process is provided in pseudo-code in Algorithm \ref{Alg:vMF_sampling}.
\begin{algorithm}
  \caption{Sampling from vMF distribution}
  \label{Alg:vMF_sampling}
\begin{algorithmic}
  \STATE {\bfseries Input:} location $\epsilon$, concentration $\kappa$, dimension $d$, unit vector $e_1= (1,0,..,0)$
  \STATE Sample $v \sim \mathcal{U}(\mathbb{S}^{d-2})$ 
\STATE $b \leftarrow \frac{-2 \kappa+\sqrt{4 \kappa^{2}+(d-1)^{2}}}{d-1}$, $a \leftarrow \frac{(d-1)+2 \kappa+\sqrt{4 \kappa^{2}+(d-1)^{2}}}{4}$, $m \leftarrow \frac{4 a b}{(1+b)}-(d-1) \ln (d-1)$
  \REPEAT 
    \STATE Sample $\psi \sim \operatorname{Beta}\left(\frac{1}{2}(d-1), \frac{1}{2}(d-1)\right)$
    \STATE $\omega \leftarrow h(\psi, \kappa)=\frac{1-(1+b) \psi}{1-(1-b) \psi}$
    \STATE $t \leftarrow \frac{2 a b}{1-(1-b) \psi}$
    \STATE Sample $u \sim \mathcal{U}(0,1)$
  \UNTIL{{$(d-1) \ln (t)-t+m \geq \ln (u)$}}
 \STATE $h_1\leftarrow (\omega, \sqrt{1-\omega^2} v^\top)^\top$
\STATE $\epsilon^\prime \leftarrow e_1 - \epsilon$
\STATE $u = \frac{\epsilon^\prime}{\norm{\epsilon^\prime}}$
\STATE $U = \mathbb{I} - 2uu^\top$
  \STATE {\bfseries Output:} $Uh_1$
\end{algorithmic}
\end{algorithm}

\subsection{Gradient estimator}
\label{subsec:gradient_estimator}
Recall that  $d$ is the dimension of latent space, $(\epsilon, \kappa)$ be the parameters of vMF distribution, $b=\frac{-2 \kappa+\sqrt{4 \kappa^{2}+(d-1)^{2}}}{d-1}$, two distributions:
\begin{align}
    &g(\omega \mid \kappa)=\frac{2\left(\pi^{d / 2}\right)}{\Gamma(d / 2)} \mathcal{C}_{d}(\kappa) \frac{\exp (\omega \kappa)\left(1-\omega^{2}\right)^{\frac{1}{2}(d-3)}}{\text{Beta}\left(\frac{1}{2}, \frac{1}{2}(d-1)\right)}, \nonumber\\
    &r(\omega|\kappa)= \frac{2 b^{1 / 2(d-1)}}{\text{Beta}\left(\frac{1}{2}(d-1), \frac{1}{2}(d-1)\right)} \frac{\left(1-\omega^{2}\right)^{1 / 2(d-3)}}{[(1+b)-(1-b) \omega]^{d-1}}, \nonumber
\end{align}
distribution $s(\psi) := \operatorname{Beta}\left(\frac{1}{2}(d-1), \frac{1}{2}(d-1)\right)$, function $h(\psi, \kappa)=\frac{1-(1+b) \psi}{1-(1-b) \psi}$, distributions  $\pi_1(\psi|\kappa)=s(\psi)\frac{g(h(\psi,\kappa)|\kappa)}{r(h(\psi,\kappa)|\kappa)}$, $\pi_2(v):= \mathcal{U}(\mathbb{S}^{d-2})$, and function  
\begin{align}
    T(\omega,v,\epsilon)= \Big(\mathbb{I} - 2\frac{e_1-\epsilon}{\norm{e_1-\epsilon}}\frac{e_1-\epsilon}{\norm{e_1-\epsilon}}^\top\Big) \big(\omega, \sqrt{1-\omega^2} v^\top\big)^\top:=\theta. \nonumber
\end{align}

From  Lemma 2 in \cite{davidsonhyperspherical}, we have:
\begin{align}
    \mathbb{E}_{\vMF(\theta|\epsilon,\kappa)}\big[f(\theta)\big]= \mathbb{E}_{(\psi,v ) \sim \pi_1(\psi|\kappa) \pi_2(v)}\Big[f\big(T(h(\psi,\kappa),v,\epsilon)\big)\Big],
\end{align}

 Note that, in SSFG we only need $\nabla_\epsilon \mathbb{E}_{\vMF(\theta|\epsilon,\kappa)}[f(\theta)]$ which can be obtained directly from the previous lemma:
 \begin{align}
     \nabla_\epsilon \mathbb{E}_{\vMF(\theta|\epsilon,\kappa)}\big[f(\theta)\big]= \mathbb{E}_{(\psi,v ) \sim \pi_1(\psi|\kappa) \pi_2(v)}\Big[ \nabla_\epsilon f\big(T(h(\psi,\kappa),v,\epsilon)\big)\Big],
 \end{align}

Then we can get a gradient estimator by using Monte-Carlo estimation scheme:
\begin{align}
    \nabla_\epsilon \mathbb{E}_{\vMF(\theta|\epsilon,\kappa)}\big[f(\theta)\big] \approx \frac{1}{L}\sum_{i=1}^L \Big[ \nabla_\epsilon f\big(T(h(\psi_i,\kappa),v_i,\epsilon)\big)\Big],
\end{align}
where $\{\psi_i\}_{i=1}^L \sim \pi_{1}(\psi|\kappa)$ and $\{v_i\}_{i=1}^L \sim \pi_2(v)$ and $L$ is the number of projections.
Sampling from $\pi_1(\psi|\kappa)$ is equivalent to the acceptance-rejection scheme in vMF sampling procedure, sampling $\pi_2(v)$  is directly  from $\mathcal{U}(\mathbb{S}^{d-2})$. Moreover, we also can derive a gradient estimator for $\nabla_\kappa \mathbb{E}_{\vMF(\theta|\epsilon,\kappa)}\big[f(\theta)\big]$ by using the log-derivative trick, however, this is unnecessary in our SSFG.

From this estimator we can estimate the gradient $\nabla_{\epsilon} \mathbb{E}_{\vMF(\theta|\epsilon,\kappa)} \big[D_{fgw}(\theta \sharp \mu,\theta \sharp \nu;\beta,d_1)\big]$, then use stochastic gradient ascent to find the optimal location $\epsilon^*$, which induces the optimal vMF distribution.

\subsection{Computational detail of spherical sliced fused Gromov Wasserstein}
\label{subsec:computation_SSFG}
By using the estimator in section \ref{subsec:gradient_estimator}, we present the algorithm to compute SSFG (Algorithm \ref{Alg:ssfg}).

\begin{algorithm}
  \caption{Computation of spherical sliced fused Gromov Wasserstein}
  \label{Alg:ssfg}
\begin{algorithmic}
  \STATE {\bfseries Input:}  Empirical distribution $\{x_i\}_{i=1}^N$,$\{y_i\}_{i=1}^N$, concentration $\kappa$, fused parameter $\beta$, the number of projections $L$, $max\_iter$
\REPEAT
\STATE Initialize SSFG $\leftarrow$ 0 
\FOR{$l=1$ \TO $L$}
\STATE Sample $\theta_l \sim \vMF(\theta|\epsilon,\kappa)$ based on Algorithm \ref{Alg:vMF_sampling}
\STATE Sort $\{\theta_l^\top x_i\}_{i=1}^N$ and $\{\theta_l^\top y_i\}_{i=1}^N$ respectively
\STATE Caculate $D_{fgw}(\{\theta_l^\top x_i\}_{i=1}^N,\{ \theta_l^\top y_i\}_{i=1}^N;\beta)$  based on its closed-form and sorted samples
\STATE SSFG $\leftarrow$ SSFG + $D_{fgw}(\{\theta_l^\top z_i\}_{i=1}^N,\{ \theta_l^\top z^\prime_i\}_{i=1}^N;\beta)$
\ENDFOR
\STATE SSFG $\leftarrow$ $\frac{\text{SSFG}}{L}$
\STATE Estimate $\nabla_\epsilon \text{SSFG}$
\STATE Update $\epsilon \leftarrow \text{Proj}_{\mathbb{S}^{d-1}}(\text{Adam}(\nabla_\epsilon \text{SSFG}))$ 
\UNTIL{$\epsilon$ converges \OR reach $max\_iter$}

  \STATE {\bfseries Output:} SSFG
\end{algorithmic}
\end{algorithm}
\subsection{Training procedure of s-DRAE}
\label{subsec:training_SDRAE}

In this section, we give a detailed procedure to train the spherical deterministic relational autoencoder in Algorithm \ref{Alg:ssfg_drae}. For each minibatch, the training procedure of s-DRAE includes computation of the reconstruction loss and the SSFG between the empirical distribution of the prior and the encoded latent distribution. The reconstruction loss is easy to compute while the computation of SSFG is harder. To obtain the best vMF distribution from the SSFG, we use the stochastic gradient ascent algorithm where the update depends on the gradient estimator of the location parameter of vMF, which is derived in Lemma 2 in ~\cite{davidsonhyperspherical}. Its details are in Appendix~\ref{subsec:gradient_estimator}. Then we sample $L$ unit vectors from the obtained vMF and apply 1-d projections to obtain two sliced distributions. The SSFG distance can be computed easily after sorting the supports of the two distributions.
\begin{figure}[t]
\begin{center}

  \begin{tabular}{ccc}
 \includegraphics[scale=0.45]{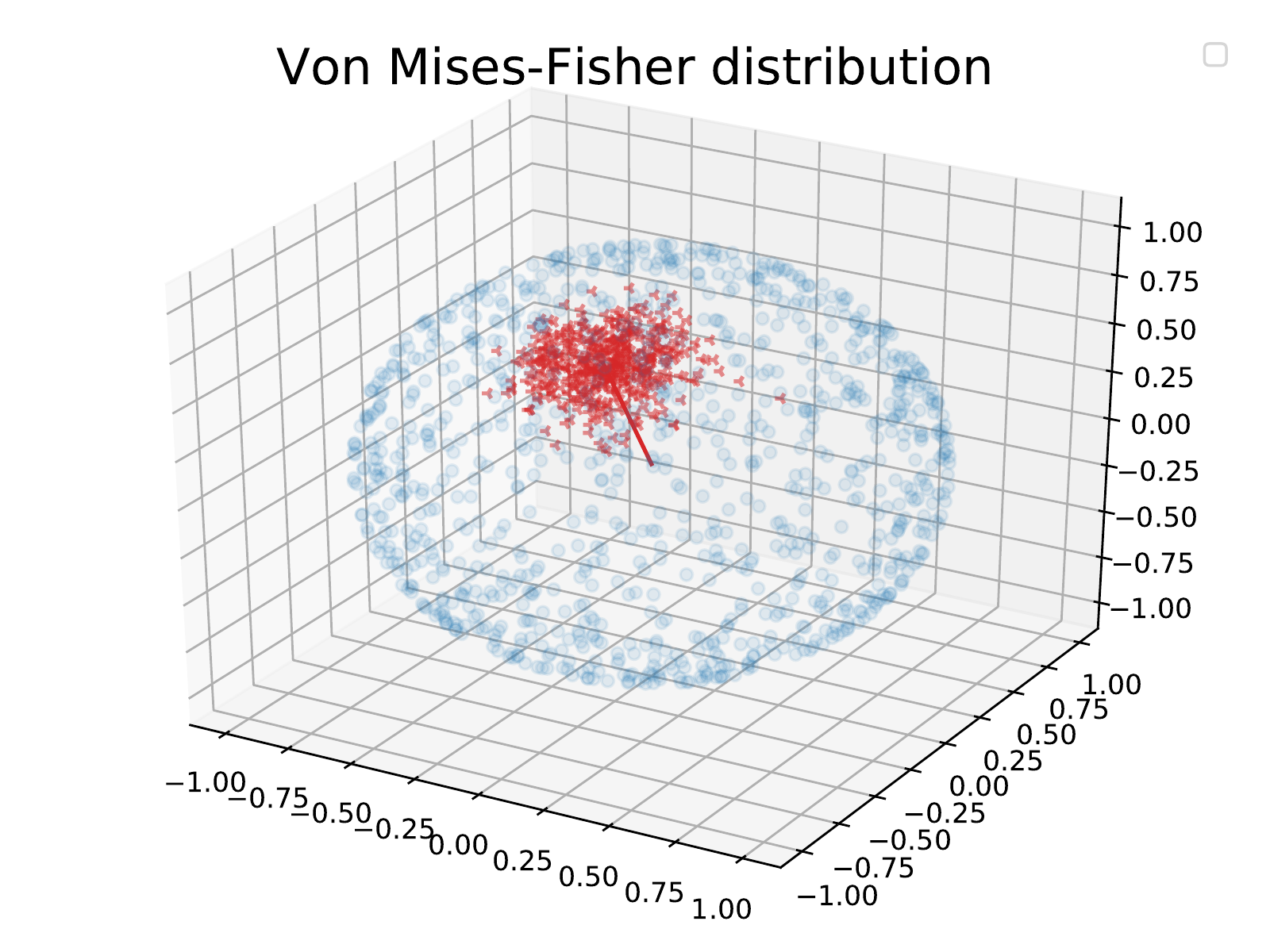} 
    &
    \includegraphics[scale=0.45]{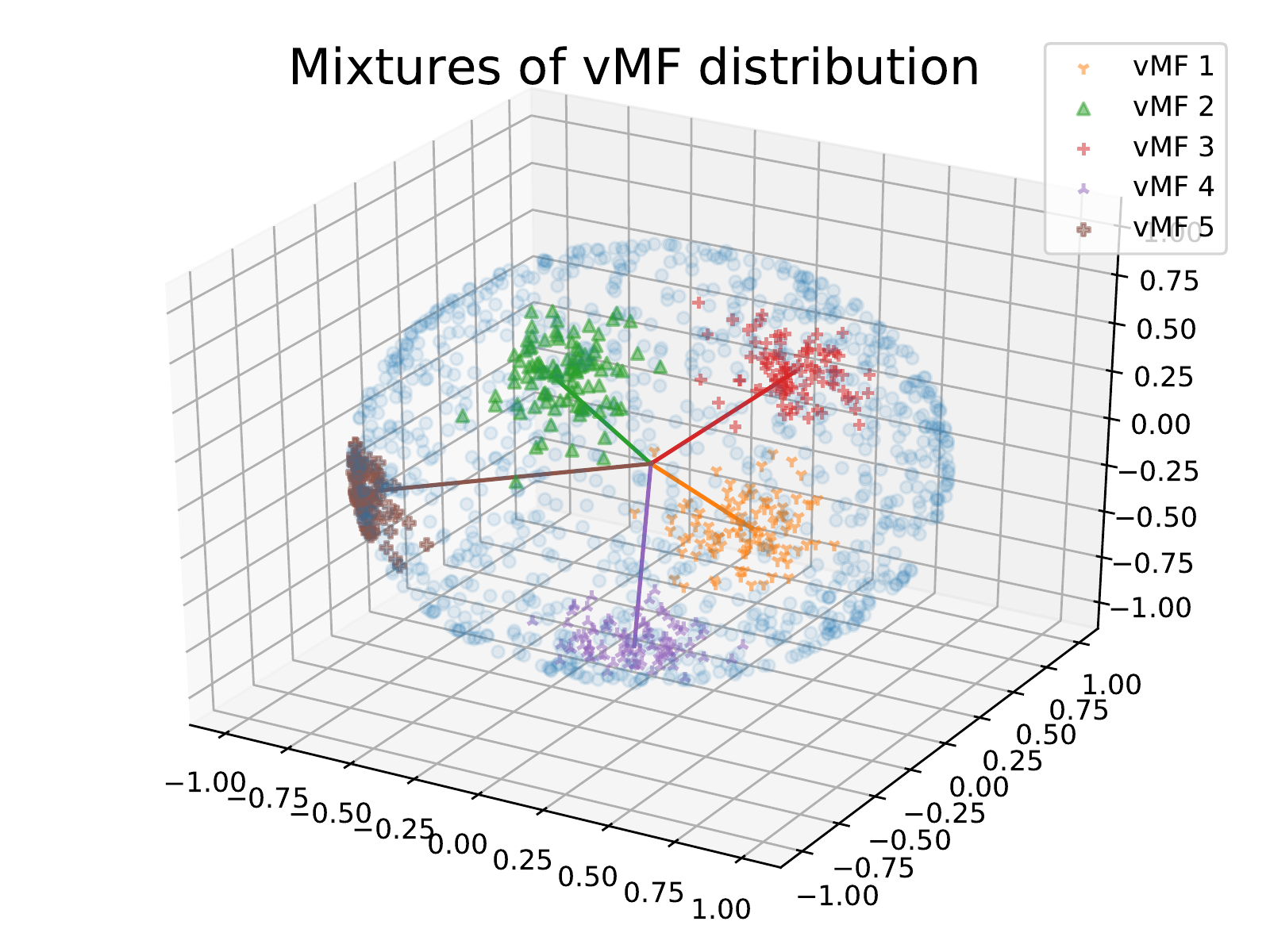}
   & 
   
  \end{tabular}
 \end{center}
  \caption{
  \footnotesize{Illustrations of  von Mises-Fisher distribution and mixture of von Mises-Fisher distributions.}}
  \label{fig:visual_MovMF}
  \vspace{-0.2 em}
\end{figure}
\begin{algorithm}
  \caption{Training DRAE with spherical sliced fused Gromov Wasserstein}
  \label{Alg:ssfg_drae}
\begin{algorithmic}
  \STATE {\bfseries Input:} Concentration $\kappa$, fused parameter $\beta$, coefficient $\lambda$, the number of mixtures $K$, $max\_iter$ minibatch's size $N$, the number of projections $L$, empirical data distribution $\hat{p}(x)$
  \STATE Initialize $G_\theta , E_\phi, \mu_{1:K},\Sigma_{1:K},\epsilon$
\FOR{each epoch}
\FOR{each minibatch $\{x_i\}_{i=1}^N \sim \hat{p}(x)$ }
\STATE $\{z_i\}_{i=1}^N \leftarrow E_\phi (\{x_i\}_{i=1}^N)$
\STATE Sample $\{z^\prime_i\}_{i=1}^N \sim p(z):= \frac{1}{K}\sum_{i=1}^K \mathcal{N}(\mu_k,\Sigma_k)$
\STATE Compute SSFG between $\{z_i\}_{i=1}^N$ and $\{z^\prime_i\}_{i=1}^N$ with $\beta,\kappa,L,max\_iter$  via the Algorithm \ref{Alg:ssfg}.
\STATE Reconstruction $\leftarrow$ $\frac{1}{N} \sum_{i=1}^N d(x_i,G_\theta(z_i))$
\STATE Update $\theta,\phi,\mu_{1:K},\Sigma_{1:K} \leftarrow \text{Adam}(\nabla_{\theta,\phi,\mu_{1:K},\Sigma_{1:K}} (\text{Reconstruction} + \lambda \text{SSFG}))$ 
\ENDFOR
\ENDFOR

  \STATE {\bfseries Output:} $G_{\theta^*}, E_{\phi^*}, \mu_{1:K}^*, \Sigma_{1:K}^*$
\end{algorithmic}
\end{algorithm}
\section{Mixture spherical sliced fused Gromov Wasserstein}
\label{Sec:MSFG}
In this appendix, we consider an extension of the spherical sliced fused Gromov Wasserstein to the mixture spherical sliced fused Gromov Wasserstein (MSSFG). Then, we discuss an application of MSSFG to the deterministic relational regularized autoencoder framework.

In order to facilitate the ensuing discussion, we recall the definitions of mixture of vMF distributions and mixture spherical sliced fused Gromov Wasserstein (MSSFG). We first define mixture of von Mises-Fisher distributions~\cite{banerjee2005clustering}, which plays an important role in the definition of mixture spherical sliced fused Gromov Wasserstein.
\begin{definition}
\label{def:mixture_vMF}
Given $k \geq 1$ distinct pairs $(\epsilon_{1}, \kappa_{1}), \ldots, (\epsilon_{k}, \kappa_{k})$ and the mixture weights $\{\alpha_i\}_{i=1}^k$, i.e., $\alpha_i \geq 0$ and $\sum_{i=1}^k \alpha_i=1$, the \textbf{mixture of vMF distributions} is defined as:
\begin{align*}
    \emph{MovMF}(\cdot|\epsilon_{1:k},\kappa_{1:k},\alpha_{1:k}) : = \sum_{i=1}^k \alpha_i \emph{vMF}(\cdot|\epsilon_i,\kappa_i).
\end{align*}
\end{definition}
When $k = 1$, the mixture of von Mises-Fisher distributions becomes the standard von Mises-Fisher distribution. When $k \geq 2$, we provide an illustration of mixture of vMF distributions in Figure~\ref{fig:visual_MovMF}. In order to sample from mixture of vMF distribution, we first sample the mixture index by categorical distribution parametrized by $\{\alpha_i\}_{i=1}^k$. Then, we sample the corresponding vMF component. Details of the sampling procedure with mixture of vMF distributions in Algorithm~\ref{Alg:MovMF_sampling}.

\begin{algorithm}
  \caption{Sampling from mixture of vMF distributions}
  \label{Alg:MovMF_sampling}
\begin{algorithmic}
  \STATE {\bfseries Input:}  The number of vMF components $k$, location $\{\epsilon_i\}_{i=1}^k$, concentration $\{\kappa_i\}_{i=1}^k$, mixture weights $\{\alpha_i\}_{i=1}^k$ , dimension $d$, unit vector $e_1= (1,0,..,0)$.
  \STATE Sample index $i \sim \text{Categorical}(\alpha_1,...,\alpha_k)$
  \STATE Sample $v \sim \mathcal{U}(\mathbb{S}^{d-2})$ 
\STATE $b \leftarrow \frac{-2 \kappa_i+\sqrt{4 \kappa_i^{2}+(d-1)^{2}}}{d-1}$, $a \leftarrow \frac{(d-1)+2 \kappa_i+\sqrt{4 \kappa_i^{2}+(d-1)^{2}}}{4}$, $m \leftarrow \frac{4 a b}{(1+b)}-(d-1) \ln (d-1)$
  \REPEAT 
    \STATE Sample $\psi \sim \operatorname{Beta}\left(\frac{1}{2}(d-1), \frac{1}{2}(d-1)\right)$
    \STATE $\omega \leftarrow h(\psi, \kappa_i)=\frac{1-(1+b) \psi}{1-(1-b) \psi}$
    \STATE $t \leftarrow \frac{2 a b}{1-(1-b) \psi}$
    \STATE Sample $u \sim \mathcal{U}(0,1)$
  \UNTIL{{$(d-1) \ln (t)-t+m \geq \ln (u)$}}
 \STATE $h_1 \leftarrow (\omega, \sqrt{1-\omega^2} v^\top)^\top$
\STATE $\epsilon^\prime \leftarrow e_1 - \epsilon_i$
\STATE $u = \frac{\epsilon^\prime}{\norm{\epsilon^\prime}}$
\STATE $U = \mathbb{I} - 2uu^\top$
  \STATE {\bfseries Output:} $Uh_1$
\end{algorithmic}
\end{algorithm}

With the definition of the mixture of vMF distributions in hand, we now define the mixture spherical sliced fused Gromov Wasserstein between the probability distributions.

\begin{definition}
(MSSFG) Let $\mu,\nu \in \mathcal{P}(\mathbb{R}^d)$ be two probability distributions, $\beta \in [0,1]$ be a constant, $\{\alpha_i\}_{i=1}^k$ be given mixture weights, and $\{\kappa_{i}\}_{i = 1}^{k}$ be given mixture concentration parameters. Furthermore, let $d_1:\mathbb{R} \times \mathbb{R} \to \mathbb{R}_+$ be a pseudo-metric on $\mathbb{R}$. Then, the \textbf{mixture spherical sliced fused Gromov Wasserstein} (MSSFG) between $\mu$ and $\nu$ is defined as follows:
\begin{align}
    \text{MSSFG}(\mu,\nu;\beta,\{\kappa_i\}_{i=1}^k, \{\alpha_i\}_{i=1}^k) & \nonumber \\
    & \hspace{- 6 em} : = \max_{\epsilon_{1:k}  \in \mathbb{S}^{d-1}}  \mathbb{E}_{\theta \sim \emph{MovMF} ( \cdot | \epsilon_{1:k},\{\kappa_i\}_{i=1}^k, \{\alpha_i\}_{i=1}^k)}\big[D_{fgw}(\theta \sharp \mu,\theta \sharp \nu;\beta,d_1)\big] ,
\end{align}
where the fused Gromov Wasserstein $D_{fgw}$ is defined in equation~(\ref{def:fg}).
\end{definition}
\paragraph{Computational complexity of MSSFG:}
Let $\mu$ and $\nu$ be two discrete distributions that have $n$ supports with uniform weights. For the general case of $d_1$, computing MSSFG is costly as SFG and SSFG due to the quadratic programming problem. However, MSSFG also has  the complexity of order $\mathcal{O}(n \log n)$ with $d_1(x,y) = (x-y)^2$. To solve the optimization problem of MSSFG, we can reuse the vMF's gradient estimators in Section \ref{subsec:gradient_estimator} and find the locations of components of the mixture vMF by stochastic gradient ascent. In detail, the gradient estimator of each vMF component's location is derived as follow:
\begin{align}
    &\nabla_{\epsilon_i}  \mathbb{E}_{ \text{MovMF} ( \theta | \epsilon_{1:k},\{\kappa_i\}_{i=1}^k, \{\alpha_i\}_{i=1}^k)}\big[D_{fgw}(\theta \sharp \mu,\theta \sharp \nu;\beta,d_1)\big] \nonumber \\
    &= \nabla_{\epsilon_i}  \mathbb{E}_{\frac{1}{k}\sum_{i=1}^k \alpha_i \vMF(\theta|\epsilon_i,\kappa_i)}\big[D_{fgw}(\theta \sharp \mu,\theta \sharp \nu;\beta,d_1)\big] \nonumber\\
    &= \frac{\alpha_i}{k} \nabla_{\epsilon_i} \mathbb{E}_{\vMF(\theta|\epsilon_i,\kappa_i)}\big[D_{fgw}(\theta \sharp \mu,\theta \sharp \nu;\beta,d_1)\big]
\end{align}
Here, we can reuse the result from Section \ref{subsec:gradient_estimator} to get an estimator. 
\paragraph{Key properties of MSSFG:} Similar to SSFG, MSSFG is also a pseudo-metric on the probability space.
\begin{theorem}
\label{theorem:metric_MSSFG}
For any $\beta \in [0,1]$, mixture concentration parameters $\{\kappa_{i}\}_{i = 1}^{k}$, and mixture weights $\{\alpha_{i}\}_{i = 1}^{k}$, $\text{MSSFG}(.,.;\beta,\{\kappa_i\}_{i=1}^k, \{\alpha_i\}_{i=1}^k)$ is a well-defined pseudo-metric in the space of probability measures, namely, it is non-negative, symmetric, and satisfies the weak triangle inequality.
\end{theorem}
The proof of Theorem~\ref{theorem:metric_MSSFG} is similar to the proof of Theorem~\ref{theorem:SSFG-distance}; therefore, it is omitted. Our next result establishes some relations between MSSFG, SSFG, SFG, and max-SFG (see part (a) in Theorem~\ref{theorem:asymptotic_vMFsliced} for a definition of max-SFG).
\begin{theorem} 
\label{theorem:asymptotic_mssfg}
For any probability measures $\mu, \nu \in \mathcal{P}(\mathbb{R}^d)$, the following holds:

(a) For any mixture weights $\{\alpha_i\}_{i=1}^k$, we obtain that
\begin{align*}
\lim \limits_{\kappa_{1}, \ldots, \kappa_{k} \to 0} \text{MSSFG}(\mu,\nu;\beta,\{\kappa_i\}_{i=1}^k, \{\alpha_i\}_{i=1}^k) & = \text{SFG} (\mu,\nu; \beta), \\
    \lim \limits_{\kappa_{1}, \ldots, \kappa_{k} \to \infty} \text{MSSFG}(\mu,\nu;\beta,\{\kappa_i\}_{i=1}^k, \{\alpha_i\}_{i=1}^k) & = \text{max-SFG} (\mu,\nu; \beta).
\end{align*}
(b) For any mixture concentration parameters $\{\kappa_i\}_{i=1}^k$ and mixture weights $\{\alpha_i\}_{i=1}^k$, we find that
\begin{align*}
    \alpha_{\bar{i}} \max_{1 \leq i \leq k} \{\text{SSFG}(\mu, \nu; \beta, \kappa_{i}) \} \leq \text{MSSFG}(\mu,\nu;\beta,\{\kappa_i\}_{i=1}^k, \{\alpha_i\}_{i=1}^k) \leq \max_{1 \leq i \leq k} \{\text{SSFG}(\mu, \nu; \beta, \kappa_{i}) \},
\end{align*}
where $\bar{i} \in \mathop{\arg \max} \limits_{1 \leq i \leq k} \{\text{SSFG}(\mu, \nu; \beta, \kappa_{i}) \}$.
\end{theorem}
The proof of part (a) in Theorem~\ref{theorem:asymptotic_mssfg} is a simple extension of the proof of part (a) in Theorem~\ref{theorem:asymptotic_vMFsliced} while the proof of part (b) in Theorem~\ref{theorem:asymptotic_mssfg} is straight-forward from the definition of MSSFG. Therefore, the proof of Theorem~\ref{theorem:asymptotic_mssfg} is omitted. The result of part (a) in Theorem~\ref{theorem:asymptotic_mssfg} demonstrates that MSSFG is an interpolation between SFG and max-SFG. Furthermore, the result of part (b) in Theorem~\ref{theorem:asymptotic_mssfg} shows that MSSFG is equivalent to $\max_{1 \leq i \leq k} \{\text{SSFG}(\mu, \nu; \beta, \kappa_{i}) \}$. Based on the result of part (b) in Theorem~\ref{theorem:asymptotic_vMFsliced}, MSSFG is also equivalent to SFG. 

Our next result shows that MSSFG also does not suffer from the curse of dimensionality as SSFG.
\begin{theorem}
\label{theorem:stats_MSSFG}
Assume that $\mu$ is a probability measure supported on a compact subset $\Theta \subset \mathbb{R}^{d}$. Let $X_{1}, \ldots, X_{n}$ be i.i.d. data from $P$ and $d_{1}(x, y) = |x - y|^{r}$ for a positive integer $r$. We denote $\mu_{n} = \frac{1}{n} \sum_{i = 1}^{n} \delta_{X_{i}}$ the empirical measure of the data points $X_{1}, \ldots, X_{n}$. Then, for any $\beta \in [0,1]$, mixture concentration parameters $\{\kappa_i\}_{i=1}^k$, and mixture weights $\{\alpha_i\}_{i=1}^k$, there exists a constant $c$ depending only on $r$ and the diameter of $\Omega$ such 
 that
\begin{align*}
     \mathbb{E} \Big[\text{MSSFG}(\mu_{n},\mu; \beta, \{\kappa_i\}_{i=1}^k, \{\alpha_i\}_{i=1}^k)\Big] 
\leq \frac{c}{n}.
\end{align*}
\end{theorem}
The proof of Theorem~\ref{theorem:stats_MSSFG} is direct from the upper bound of MSSFG based on SSFG in part (b) of Theorem~\ref{theorem:asymptotic_mssfg} and the convergence rate of SSFG in expectation in Theorem~\ref{theorem:stats_SSFG}; therefore, it is omitted. 
\paragraph{Application of MSSFG to relational regularized autoencoder framework:} Similar to the SSFG, we can apply MSSFG to relational regularized autoencoder by using it as the discrepancy between prior distribution and latent code distribution. We name this autoencoder -- \textit{mixture spherical deterministic autoencoder} (ms-DRAE). The training procedure of ms-DRAE is similar to s-DRAE, it includes computation of the reconstruction loss and the MSSFG between the empirical  distribution of the prior and the encoded latent distribution. To compute MSSFG, it also requires the stochastic gradient ascent scheme to find the best MovMF distribution. Note that, in this autoencoder, we set a uniform mixture weights $\alpha_i= \frac{1}{k}$ and each vMF component uses the same value of concentration parameter $\kappa$. 
\section{Power spherical sliced fused Gromov Wasserstein}
\label{sec:power_SFFG}
\begin{figure}[t]
\begin{center}

  \begin{tabular}{ccc}
 \includegraphics[scale=0.45]{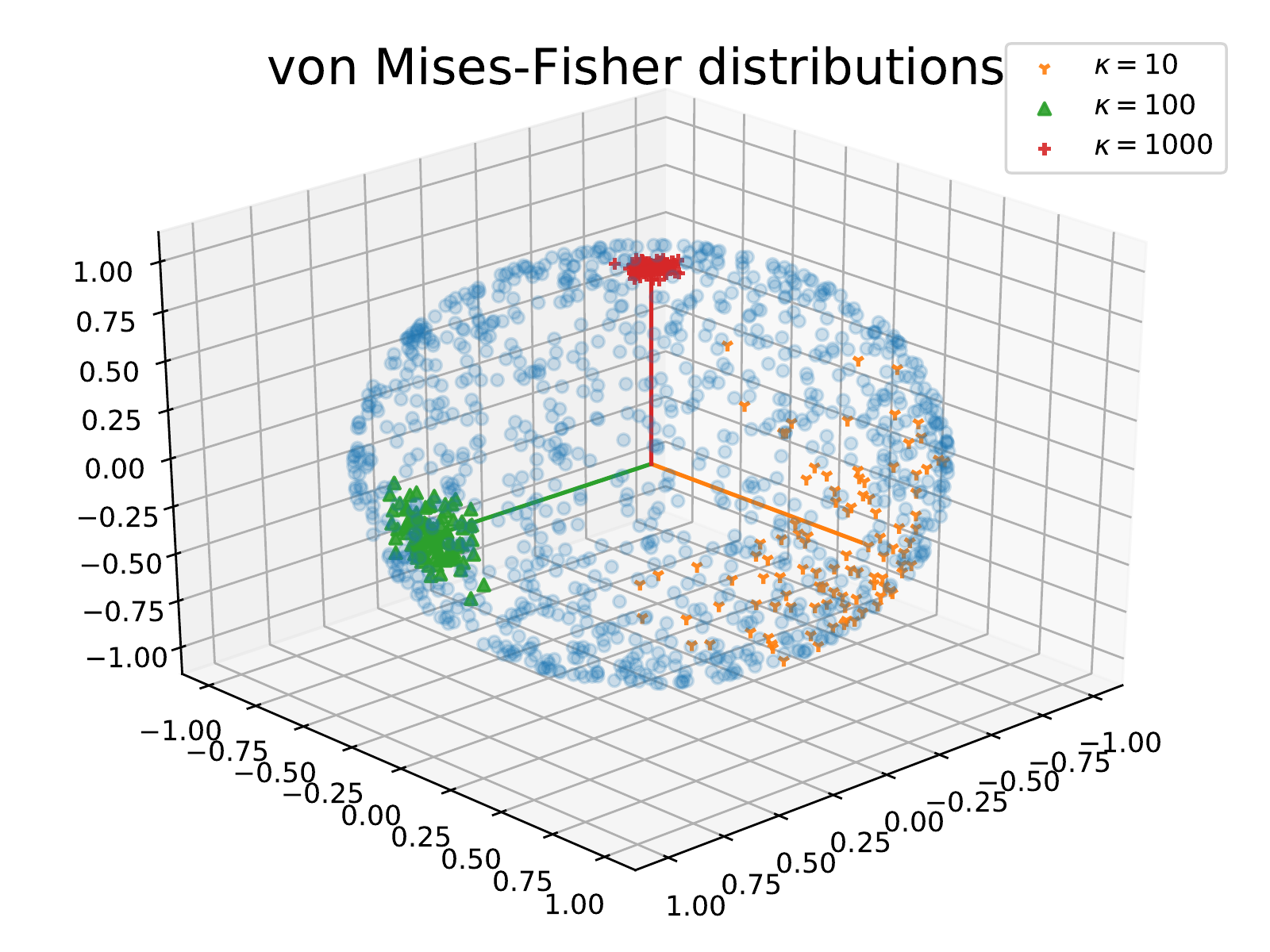} 
    &
    \includegraphics[scale=0.45]{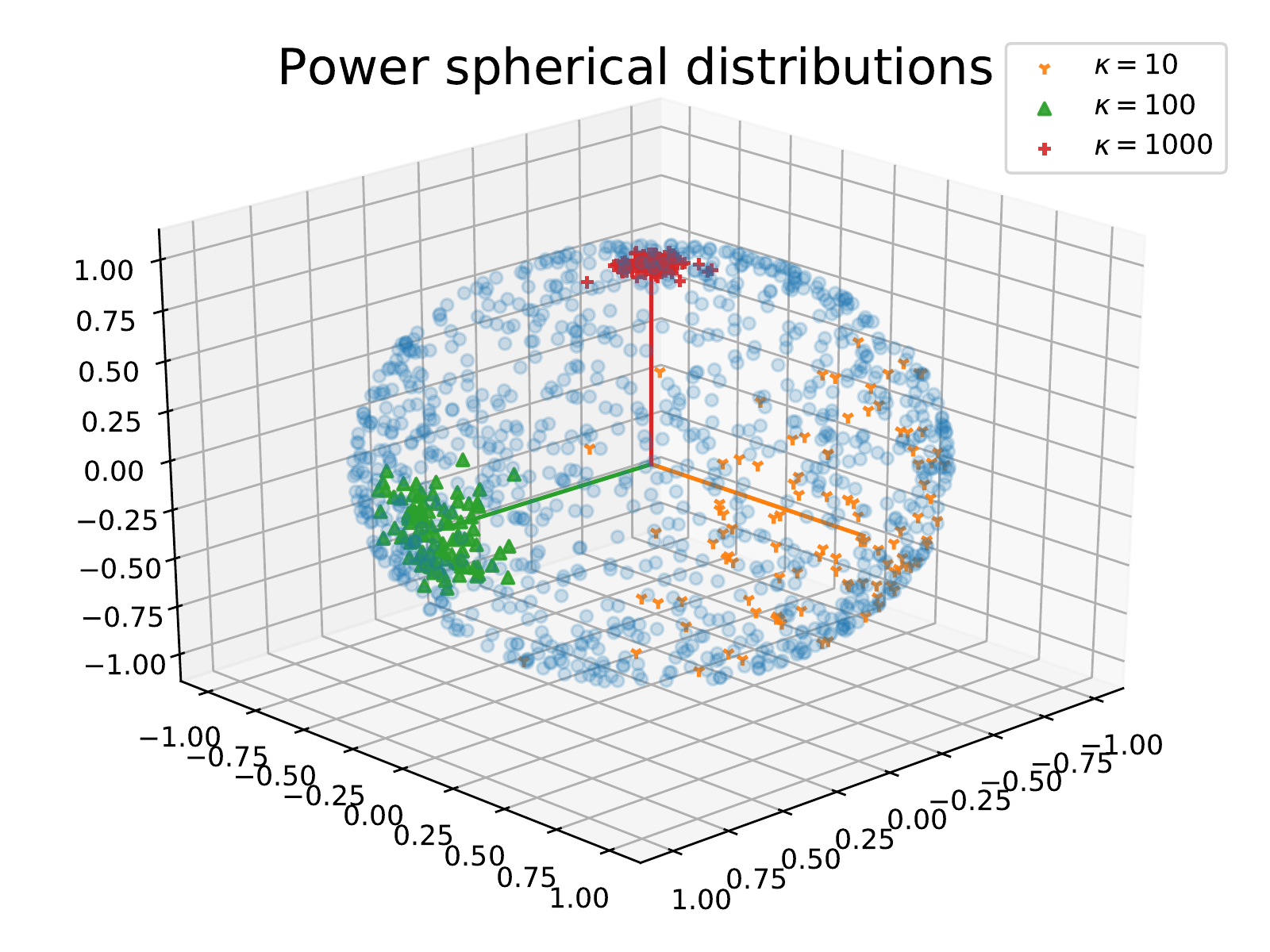}
   & 
   
  \end{tabular}
 \end{center}
  \caption{
  \footnotesize{Illustrations of von Mises-Fisher and Power spherical distribution.}}
  \label{fig:ps_sphere}
  \vspace{-0.2 em}
\end{figure}
In this appendix, we consider another variant of spherical sliced fused Gromov Wasserstein, which is power spherical sliced fused Gromov Wasserstein (PSSFG). Then, we discuss an application of PSSFG to the deterministic relational regularized autoencoder framework.

We first define power spherical distribution~\cite{de2020power}:
\begin{definition}
\label{def:power_spherical}
The power spherical distribution (PS) is a probability distribution on the unit sphere $\mathbb{S}^{d-1}$  where its density function is given by:
\begin{align}
    f(x| \epsilon, \kappa ) = C( \kappa,d) (1 + \epsilon^\top x)^{\kappa},
\end{align}
where $\kappa \geq 0$ is the concentration parameter,  $\epsilon\in \mathbb{S}^{d-1}$ is  the location vector, and $C_d(\kappa) : = \left\{ 2^{d - 1 + \kappa} \pi^{\frac{d - 1}{2}} \frac{\Gamma(\frac{d - 1}{2} + \kappa)}{\Gamma(d - 1 + \kappa)}\right\}^{-1}$. 
\end{definition}
When $\kappa \to 0$, the power spherical distribution approaches to the uniform distribution, because for $\kappa \in (0, 1]$, the density function $f(x|\epsilon,\kappa)$ is uniformly bounded. Moreover, for all $x$ is different from $- \epsilon$
\begin{align*}
\lim_{\kappa \rightarrow 0} f(x|\epsilon,\kappa) = C(0,d),
\end{align*}
which corresponds to the density of the uniform distribution on $\mathbb{S}^{d-1}$.
Hence,  by Lebesgue dominated convergence theorem, for any bounded function $g$ on $\mathbb{S}^{d-1}$, we have
\begin{align*}
    \lim_{\kappa \rightarrow 0}\int_{\mathbb{S}^{d-1}} g(x) f(x|\epsilon,\kappa) dx = \int_{\mathbb{S}^{d-1}} g(x) \mathcal{U}(\mathbb{S}^{d-1})  dx, 
\end{align*}
where $\mathcal{U}(\mathbb{S}^{d-1})$ stands for the uniform distribution on $\mathbb{S}^{d-1}$. It means the convergence of the density $f$ of the power spherical distribution to the uniform distribution when $\kappa \to 0$.  When $\kappa \rightarrow \infty$, as long as the function $f$ at the mode $\epsilon$ goes to infinity, then the following holds for any uniformly bounded function $g$ on $\mathbb{S}^{d-1}$
\begin{align*}
   \lim_{\kappa \rightarrow \infty} \int_{\mathbb{S}^{d-1}} g(x) f(x|\epsilon,\kappa) dx =  \int_{\mathbb{S}^{d-1}} g(x) \delta_{\epsilon}(x) dx.
   \end{align*}
  It is true since for $x = \epsilon$, and $\kappa \rightarrow \infty$ we have
  \begin{align*}
      f(\epsilon|\epsilon,\kappa) = 2^{\kappa} \frac{\Gamma(d-1 + \kappa)}{\Gamma(\frac{d-1}{2} + \kappa)} \pi^{\frac{1-d}{2}} 2^{1-d-\kappa} = \pi^{\frac{1-d}{2}} 2^{1-d} \frac{\Gamma(d-1 + \kappa)}{\Gamma(\frac{d-1}{2} + \kappa)} \rightarrow \infty.
      \end{align*}
As a consequence, the density of the power spherical distribution with location vector $\epsilon$ converges to the Dirac delta measure at $\epsilon$ when $\kappa \to \infty$.

\paragraph{Sampling procedure of the power spherical distribution:} We review the sampling algorithm of PS in Algorithm \ref{Alg:PS_sampling} in \cite{de2020power}. The important difference between  the sampling of the PS to   that of the vMF is that it does not require rejection sampling as in vMF's sampling algorithm (Algorithm \ref{Alg:vMF_sampling}). As a result, sampling from PS is faster than sampling from vMF. Furthermore, we can get an estimation of gradient of parameters of the density of the PS easier than that of vMF since all the operations in the sampling algorithm of the PS are differentiable (note that, it also includes sampling from Beta distribution which can use the implicit reparametrization trick~\cite{figurnov2018implicit}).

\begin{algorithm}
  \caption{Sampling from power spherical distribution}
  \label{Alg:PS_sampling}
\begin{algorithmic}
  \STATE {\bfseries Input:}   location parameter $\epsilon$, concentration $\kappa$, dimension $d$, unit vector $e_1= (1,0,..,0)$.
  \STATE Sample $z \sim \text{Beta}(\frac{(d-1)}{2}+\kappa,\frac{(d-1)}{2})$
  \STATE Sample $v \sim \mathcal{U}(\mathbb{S}^{d-2})$ 
\STATE  $w \leftarrow 2z - 1$
 \STATE $h_1\leftarrow (\omega, \sqrt{1-\omega^2} v^\top)^\top$
\STATE $\epsilon^\prime \leftarrow e_1 - \epsilon$
\STATE $u = \frac{\epsilon^\prime}{\norm{\epsilon^\prime}}$
\STATE $U = \mathbb{I} - 2uu^\top$
  \STATE {\bfseries Output:} $Uh_1$
\end{algorithmic}
\end{algorithm}

Given the definition of the power spherical distribution, we are ready to define the power spherical sliced fused Gromov Wasserstein between the probability distributions.

\begin{definition}
\label{def:PSSFG}
(PSSFG) Let  $\mu, \nu \in \mathcal{P}(\mathbb{R}^d)$ be two probability distributions, $\kappa > 0$, $\beta \in [0, 1]$, $d_{1}: \mathbb{R} \times \mathbb{R} \to \mathbb{R}_{+}$ be a pseudo-metric on $\mathbb{R}$. The \textbf{power spherical sliced fused Gromov Wasserstein} (PSSFG) between $\mu$ and $\nu$ is defined as follows:
\begin{align}
    \text{PSSFG}(\mu,\nu; \beta, \kappa) : = \max_{\epsilon  \in \mathbb{S}^{d-1}}   \mathbb{E}_{\theta \sim \emph{PS} ( \cdot | \epsilon,\kappa)}\big[D_{fgw}(\theta \sharp \mu,\theta \sharp \nu;\beta,d_1)\big] , \label{eq:def_PSSFG}
\end{align}
where the fused Gromov Wasserstein $D_{fgw}$ is defined in equation~(\ref{def:fg}). Here, $\text{PS}(.|\epsilon, \kappa)$ denotes the power spherical distribution with location vector $\epsilon$ and concentration parameter $\kappa$.
\end{definition}
\paragraph{Comparison between PSSFG and SSFG:}  The PS distribution has similar behavior as the vMF distribution on the unit sphere, suggesting that PSSFG and SSFG are similar discrepancies. However, PS shows better sampling speed and stability than vMF, which leads to the computational benefits of PSSFG over SSFG. In particular, to compute PSSFG and SSFG, we need to approximate the gradient of the location parameter $\epsilon$ with $L$ samples from the PS and vMF distributions respectively. With the faster sampling algorithm of PS, we can find the best location parameter in PSSFG faster than in SSFG. Therefore, PSSFG can be computed in lower time than SSFG. Moreover, as reported in \cite{de2020power}, the vMF distribution suffers from numerical issues on high dimension settings or on high concentration parameter settings, namely, sampling vectors from the vMF distribution can be returned as "not a number" (NaN), which greatly affects the quality of SSFG. In contrast, PSSFG does not have the similar numerical issues because the PS distribution provides more stable samples in high-dimension settings with any values of the concentration parameter.
\paragraph{Computational complexity of PSSFG:} Similar to SSFG, PSSFG between $\mu$ and $\nu$, two discrete distributions that have $n$ supports and uniform weights, has the complexity of order $\mathcal{O}(n \log n)$ when $d_1(x,y)=(x-y)^2$ for all $x, y \in \mathbb{R}$. With a faster sampling process, the gradient estimation step in PSSFG consumes a smaller amount of time than SSFG. Therefore, it leads to faster optimization to find the optimal location parameter $\epsilon$ in PSSFG than in SSFG.
\paragraph{Key properties of PSSFG:}
Similar to SSFG and MSSFG, PSSFG is also a pseudo-metric on the probability space.
\begin{theorem}
\label{theorem:metric_PSSFG}
For any $\beta \in [0,1]$ and $\kappa > 0$, $\text{PSSFG}(.,.;\beta, \kappa)$ is a pseudo-metric in the space of probability measures, namely, it is non-negative, symmetric, and satisfies the weak triangle inequality.
\end{theorem}
The proof of Theorem~\ref{theorem:metric_PSSFG} is similar to the proof of Theorem~\ref{theorem:SSFG-distance}; therefore, it is omitted. 

Our next result establishes some connections between PSSFG, SSFG, SFG, and max-SFG (see part (a) in Theorem~\ref{theorem:asymptotic_vMFsliced} for a definition of max-SFG).
\begin{theorem} 
\label{theorem:asymptotic_pssfg}
For any probability measures $\mu, \nu \in \mathcal{P}(\mathbb{R}^d)$, the following holds:
\begin{align*}
\hspace{-10 em} \text{(a)} \quad \quad \quad  \quad \quad \quad \quad \quad \quad \quad \lim \limits_{\kappa \to 0} \text{PSSFG} (\mu, \nu; \beta, \kappa) & = \text{SFG} (\mu,\nu; \beta), \\
    \lim \limits_{\kappa \to \infty} \text{PSSFG} (\mu, \nu; \beta, \kappa) & = \text{max-SFG} (\mu,\nu; \beta).
\end{align*}
(b) For any $\kappa > 0$, we find that
\begin{align*}
     \text{PSSFG} (\mu, \nu; \beta, \kappa) & \leq 2^{\kappa} C(\kappa,d) \text{SFG} (\mu, \nu; \beta), \\
    \text{PSSFG} (\mu, \nu; \beta, \kappa) & \leq \text{max-SFG}(\mu,\nu; \beta).
\end{align*}
\end{theorem}
The proof of part (a) in Theorem~\ref{theorem:asymptotic_pssfg} is a straightforward application of the proof of Theorem~\ref{theorem:asymptotic_vMFsliced} and the asymptotic properties of the power spherical distribution. The proof of part (b) in Theorem~\ref{theorem:asymptotic_pssfg} is also similar to the proof of part (b) of Theorem~\ref{theorem:asymptotic_vMFsliced}. Note that, we do not have the lower bound of $\text{PSSFG} (\mu, \nu; \beta, \kappa)$ in terms of $\text{SFG} (\mu, \nu; \beta)$. It is because the value of $(1 + \epsilon^{\top} x)^{\kappa}$ can get arbitrarily close to 0 as $\epsilon^{\top}x$ gets close to $-1$. Combining with the result of Theorem~\ref{theorem:asymptotic_pssfg}, we reach to a conclusion that the SSFG, SFG, max-SFG are stronger discrepancies than the PSSFG.

Our next result shows that PSSFG does not suffer from the curse of dimensionality as SSFG and MSSFG. Therefore, it is also a good discrepancy to use in the deterministic relational regularized autoencoder framework.
\begin{theorem}
\label{theorem:stats_PSSFG}
Assume that $\mu$ is a probability measure supported on a compact subset $\Theta \subset \mathbb{R}^{d}$. Let $X_{1}, \ldots, X_{n}$ be i.i.d. data from $P$ and $d_{1}(x, y) = |x - y|^{r}$ for a positive integer $r$. We denote $\mu_{n} = \frac{1}{n} \sum_{i = 1}^{n} \delta_{X_{i}}$ the empirical measure of the data points $X_{1}, \ldots, X_{n}$. Then, for any $\beta \in [0,1]$ and $\kappa > 0$, there exists a constant $c$ depending only on $r$ and diameter of $\Theta$ such 
 that
\begin{align*}
     \mathbb{E} \Big[\text{PSSFG}(\mu_{n},\mu; \beta, \kappa)\Big] 
\leq \frac{c}{n}.
\end{align*}
\end{theorem}
The proof of Theorem~\ref{theorem:stats_PSSFG} is straightforward from the upper bound of PSSFG in terms of SSFG and the convergence rate of SSFG in expectation in Theorem~\ref{theorem:stats_SSFG}; therefore, it is omitted. 
\paragraph{Application of PSSFG to relational regularized autoencoder framework:} Similar to the SSFG, we can apply PSSFG to relational regularized autoencoder by using it as the discrepancy between prior distribution and latent code distribution. We name this autoencoder \textit{power spherical deterministic autoencoder} (ps-DRAE). The training procedure of ps-DRAE is similar to s-DRAE, namely, it includes computation of the reconstruction loss and the PSSFG between the empirical  distribution of the prior and the encoded latent distribution. 
\section{Additional experiments}
\label{sec:additional_experiments}
In this appendix,  we provide qualitative results on MNIST and CelebA datasets, including randomly generated images, reconstruction images, and latent space visualizations  for the autoencoders in the main paper. In addition, we provide more applications where the proposed discrepancies can be applied. The first application in Section \ref{subsec:pde} trains autoencoder via ex-post density estimation, which is a new procedure to train an autoencoder and is proposed currently in~\cite{ghosh2019variational}. In this application, we compare the SSFG with SFG and max-SFG to show the benefits of the SSFG. The second application is GAN \cite{goodfellow2014generative} which is used to learn a generator only. In this second application, we compare qualitatively our spherical sliced discrepancies (SSFG and MSSFG) with SFG and max-SFG. 
\label{Sec:addtional_experiments}
\begin{figure}[!h]
\begin{center}

  \begin{tabular}{ccc}
 \includegraphics[scale=0.34]{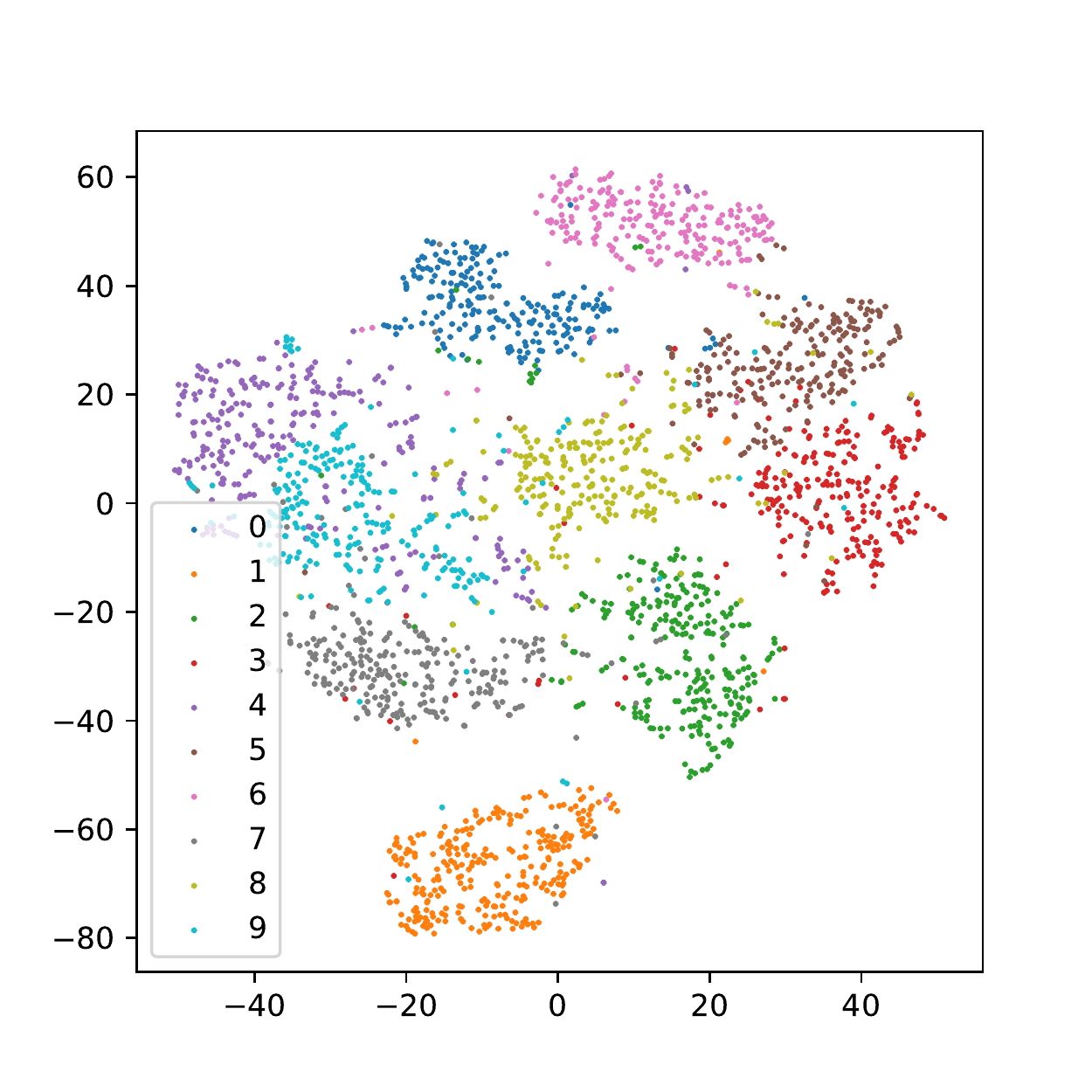} 
    
    &\includegraphics[scale=0.34]{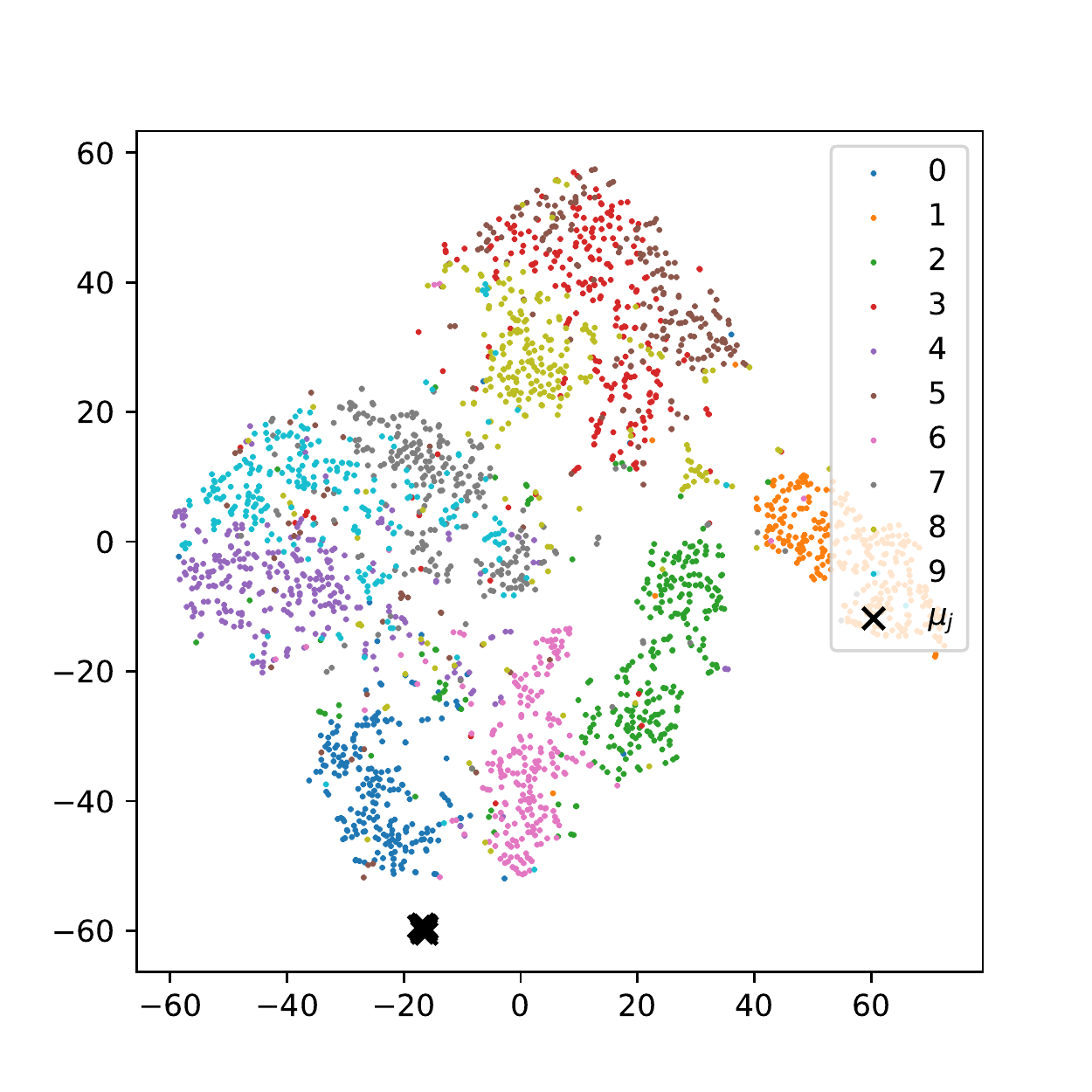}
     &
 \includegraphics[scale=0.34]{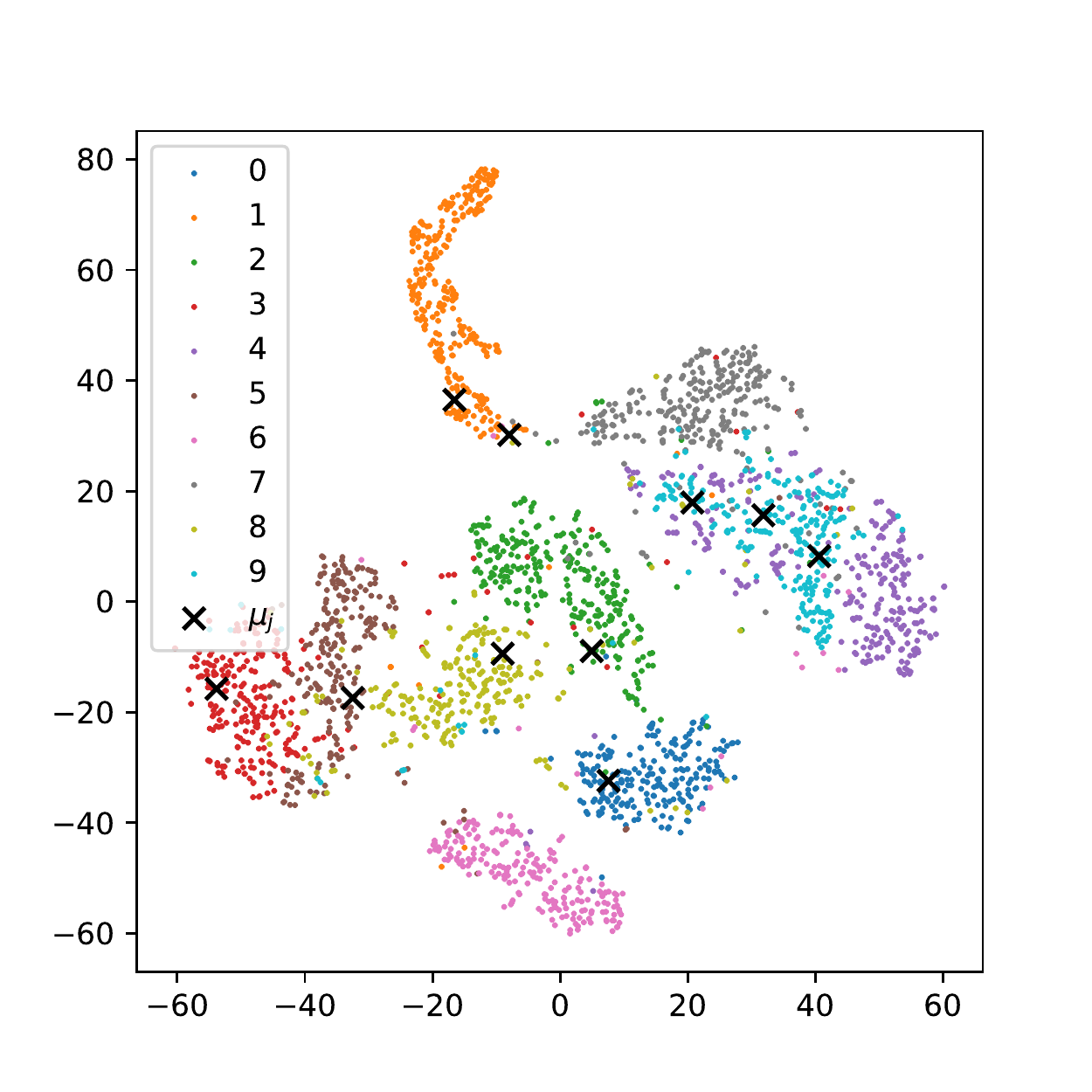} 
 \\
     VAE&Vampprior&GMVAE
     
 \\
    \includegraphics[scale=0.34]{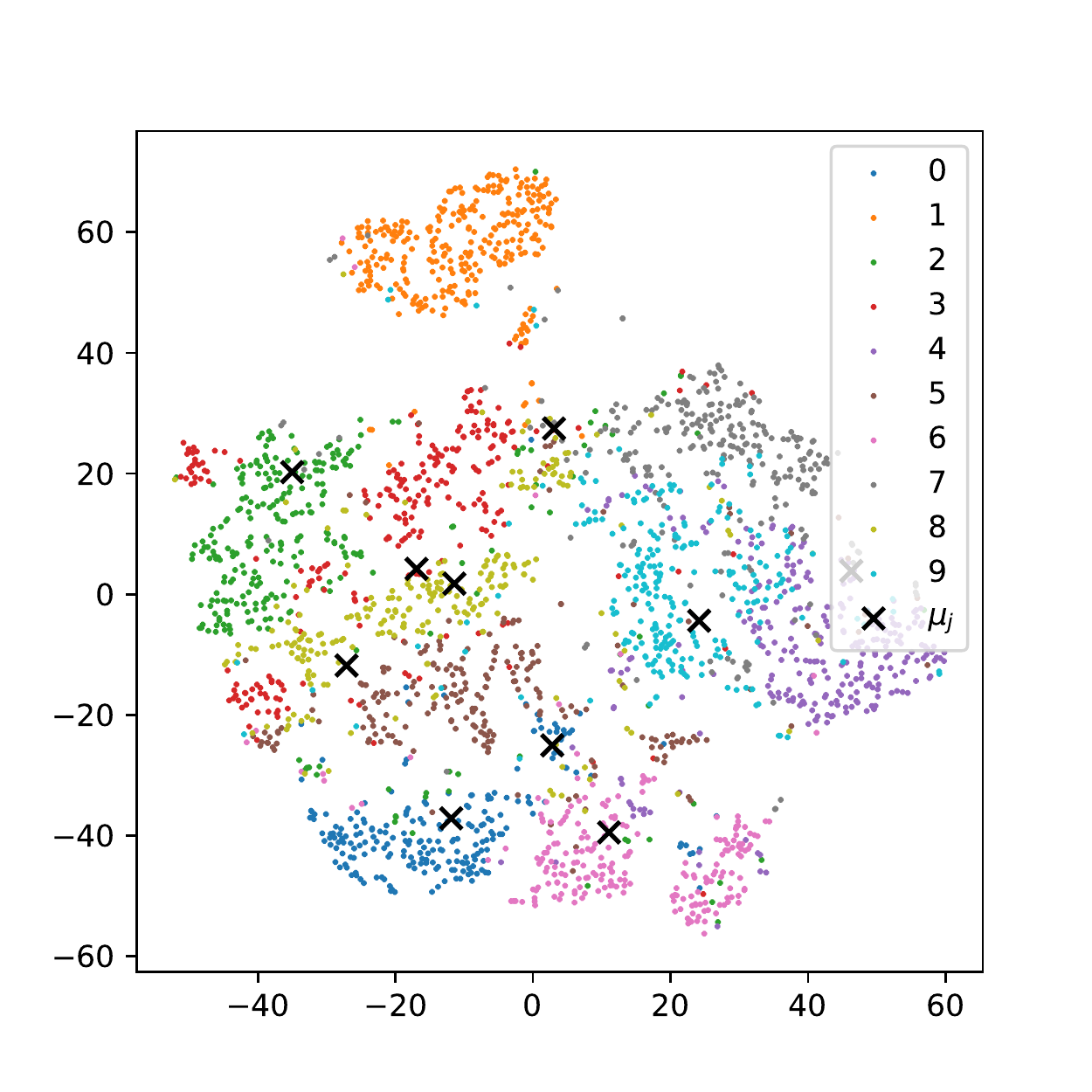} 
    &\includegraphics[scale=0.34]{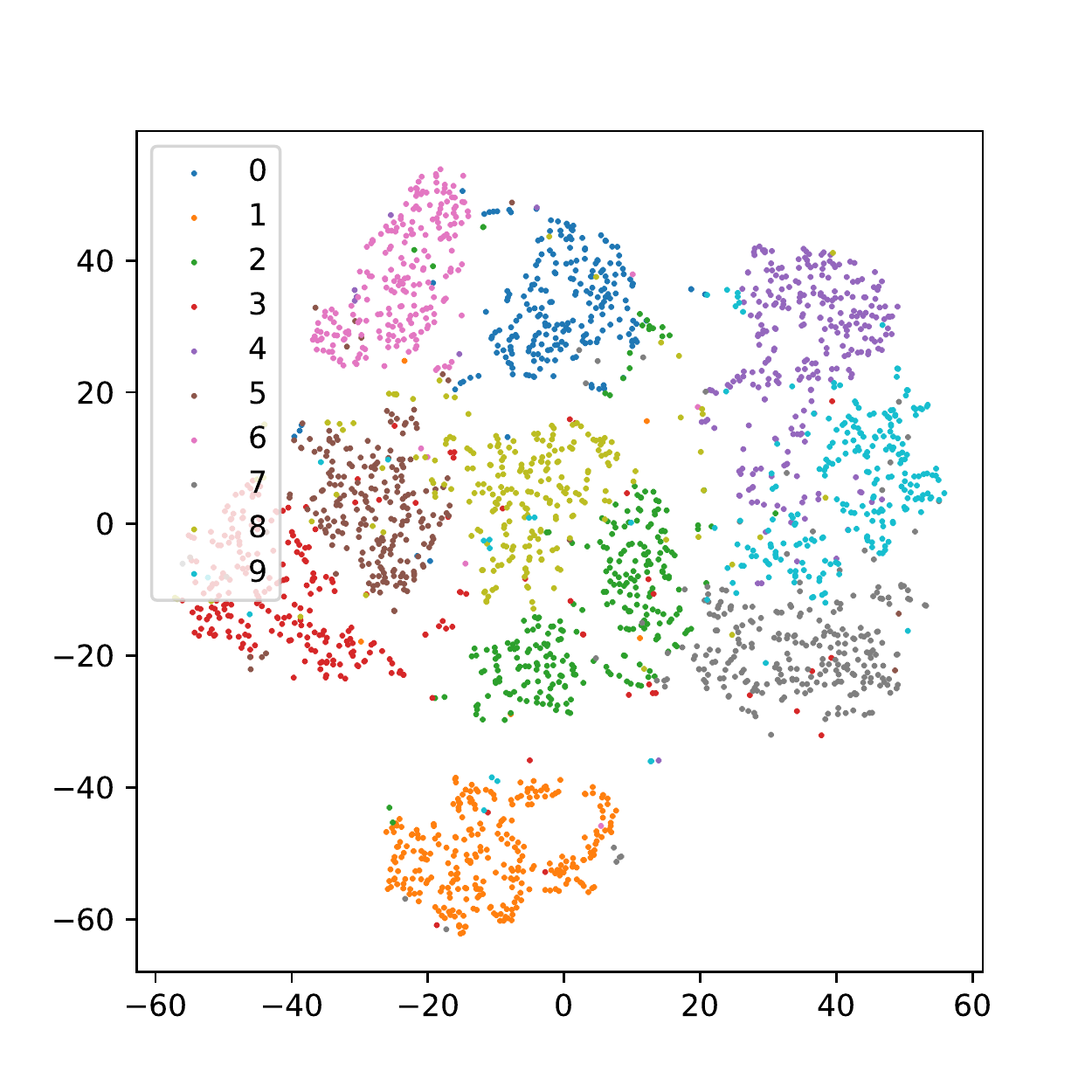} 
    
    &\includegraphics[scale=0.34]{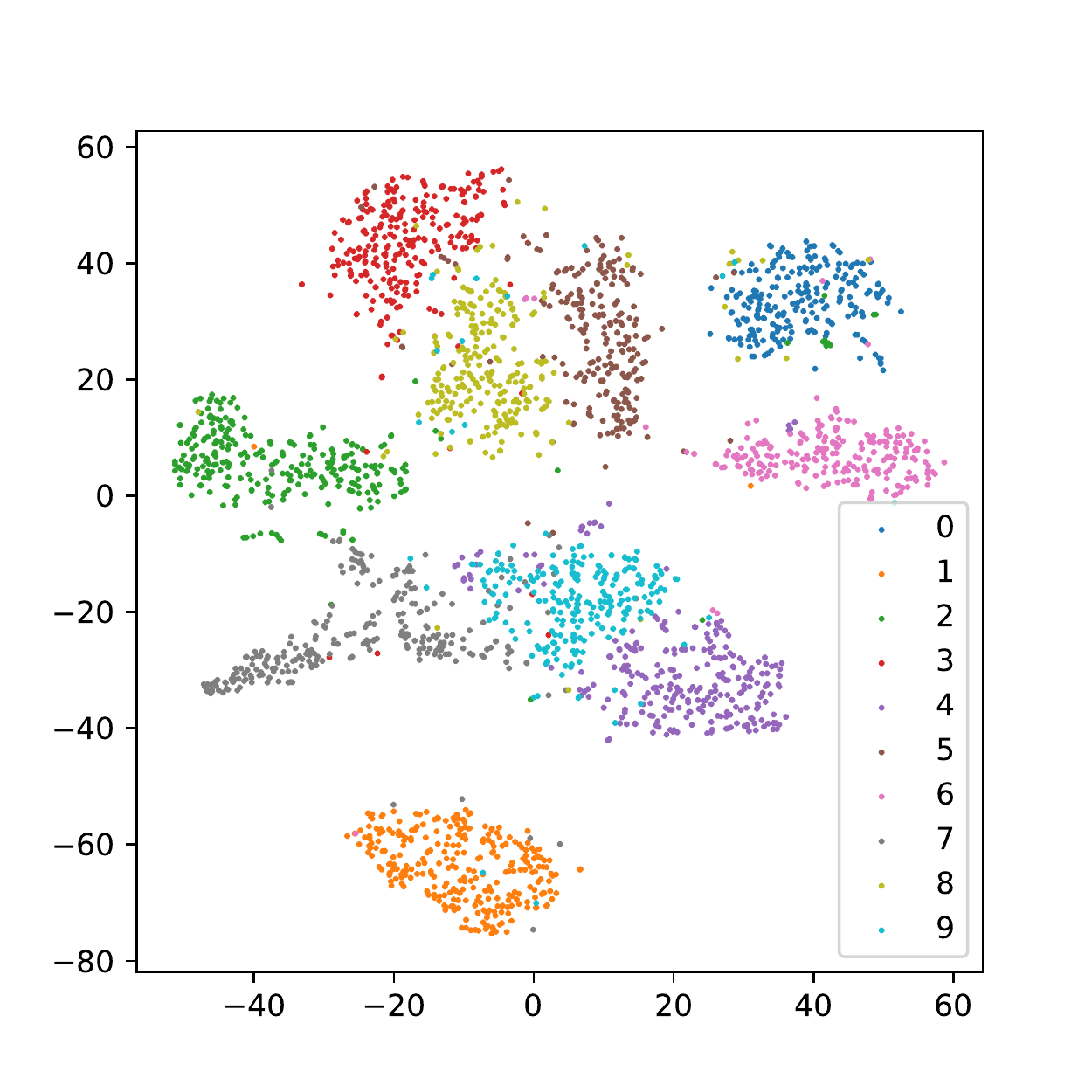} 
    
    \\
    PRAE & SWAE &WAE
    \\
 \includegraphics[scale=0.34]{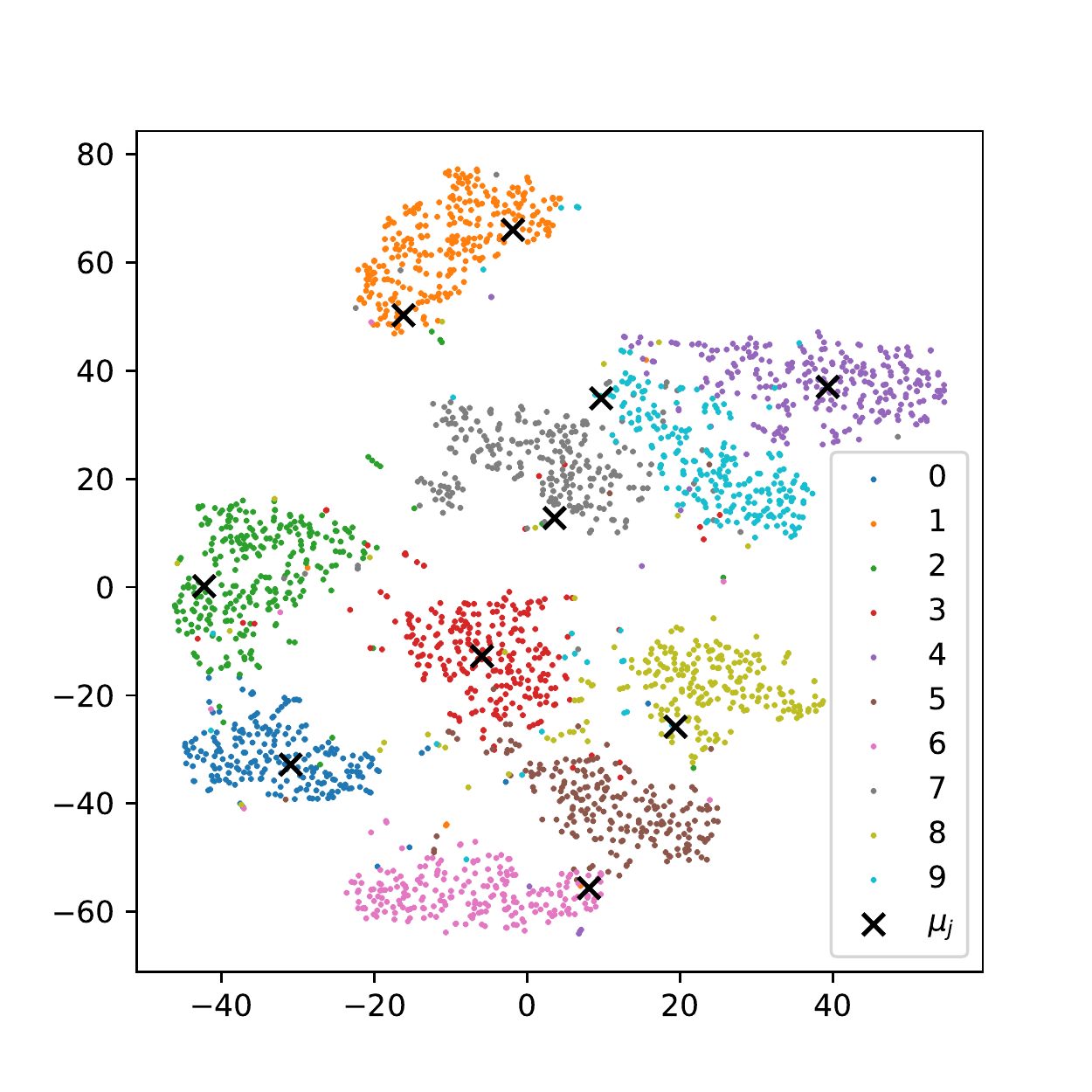} 
  &
    \includegraphics[scale=0.34]{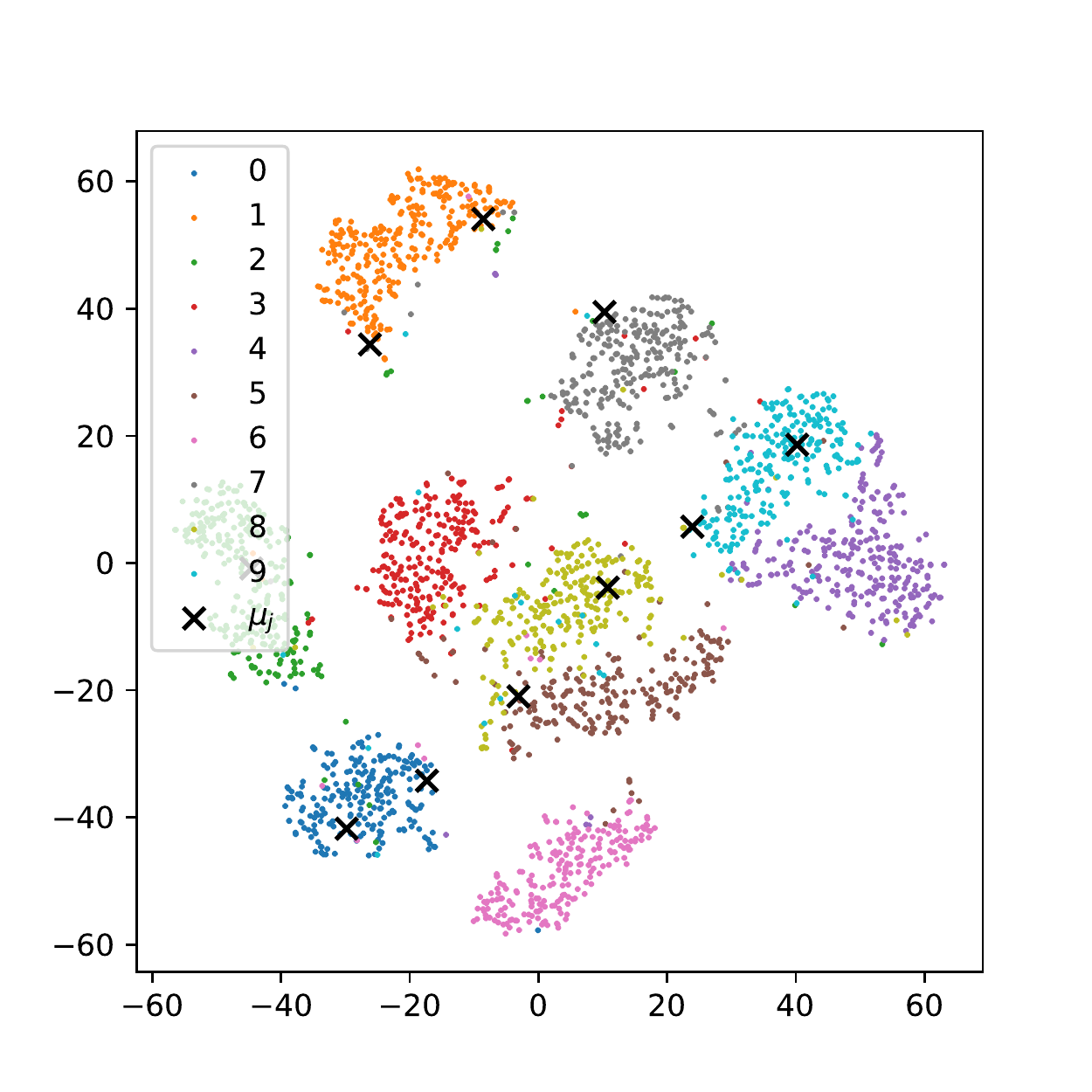} 
    &
    \includegraphics[scale=0.34]{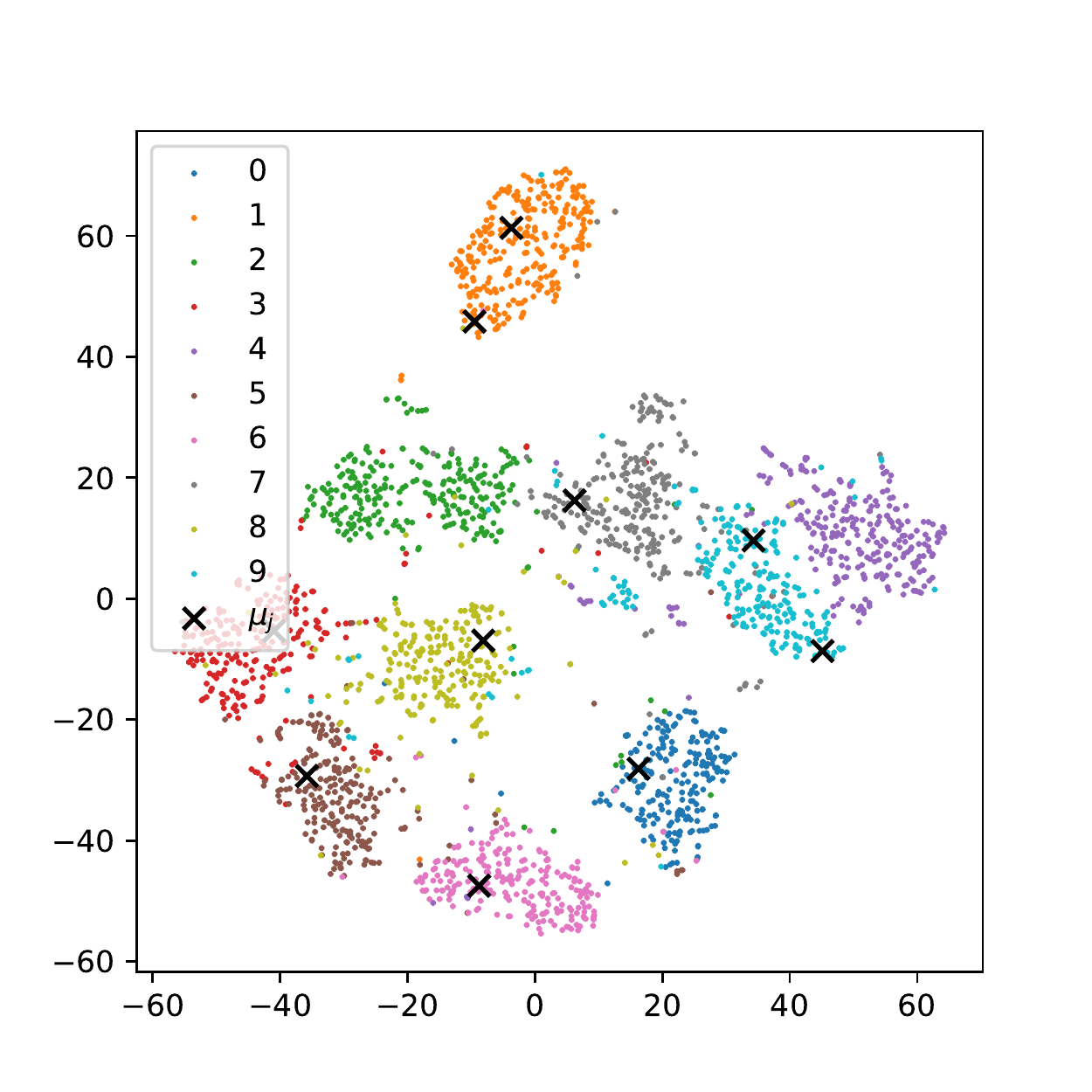}
    \\
    DRAE & m-DRAE & s-DRAE
  
   \\
   
 \includegraphics[scale=0.34]{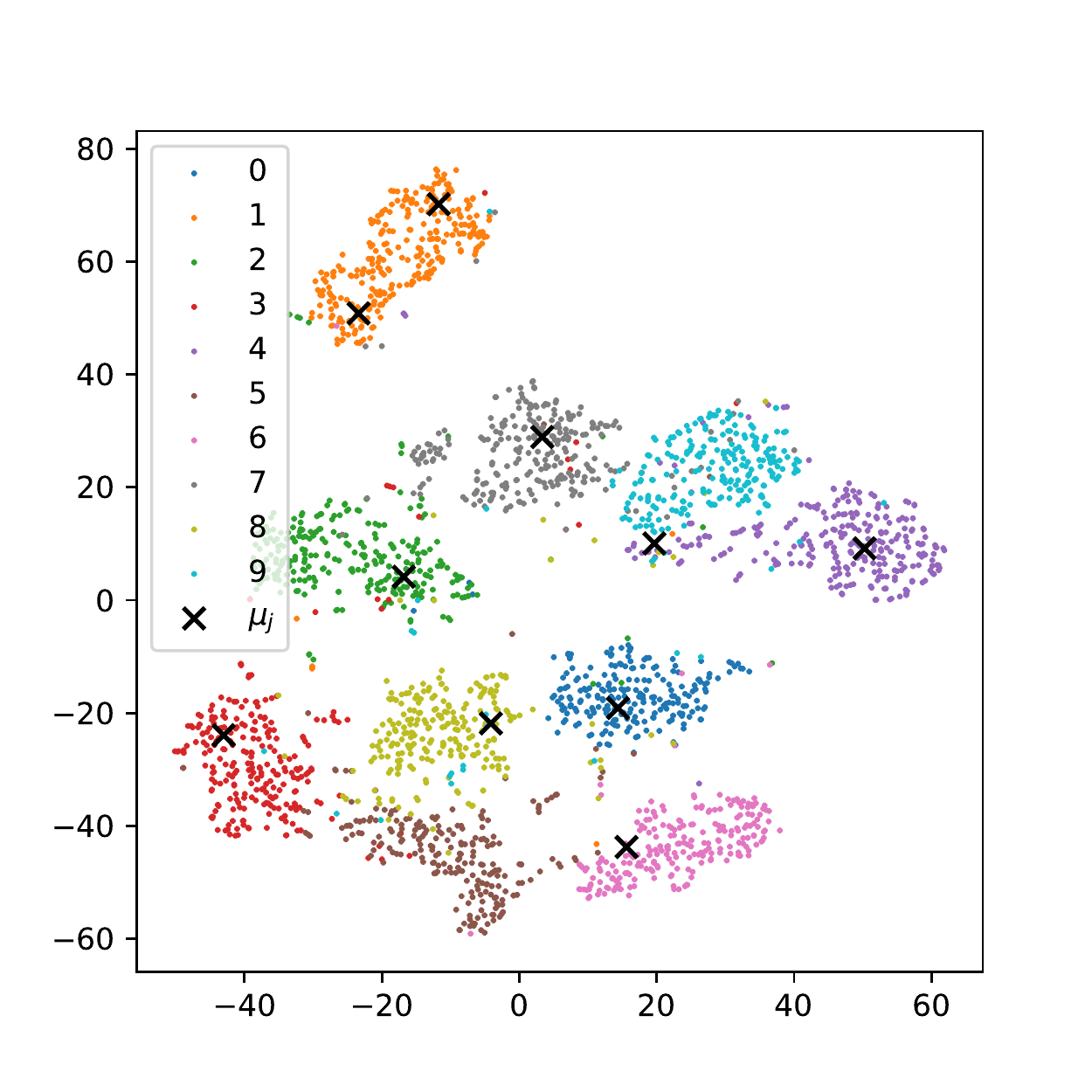} 
  &
    \includegraphics[scale=0.34]{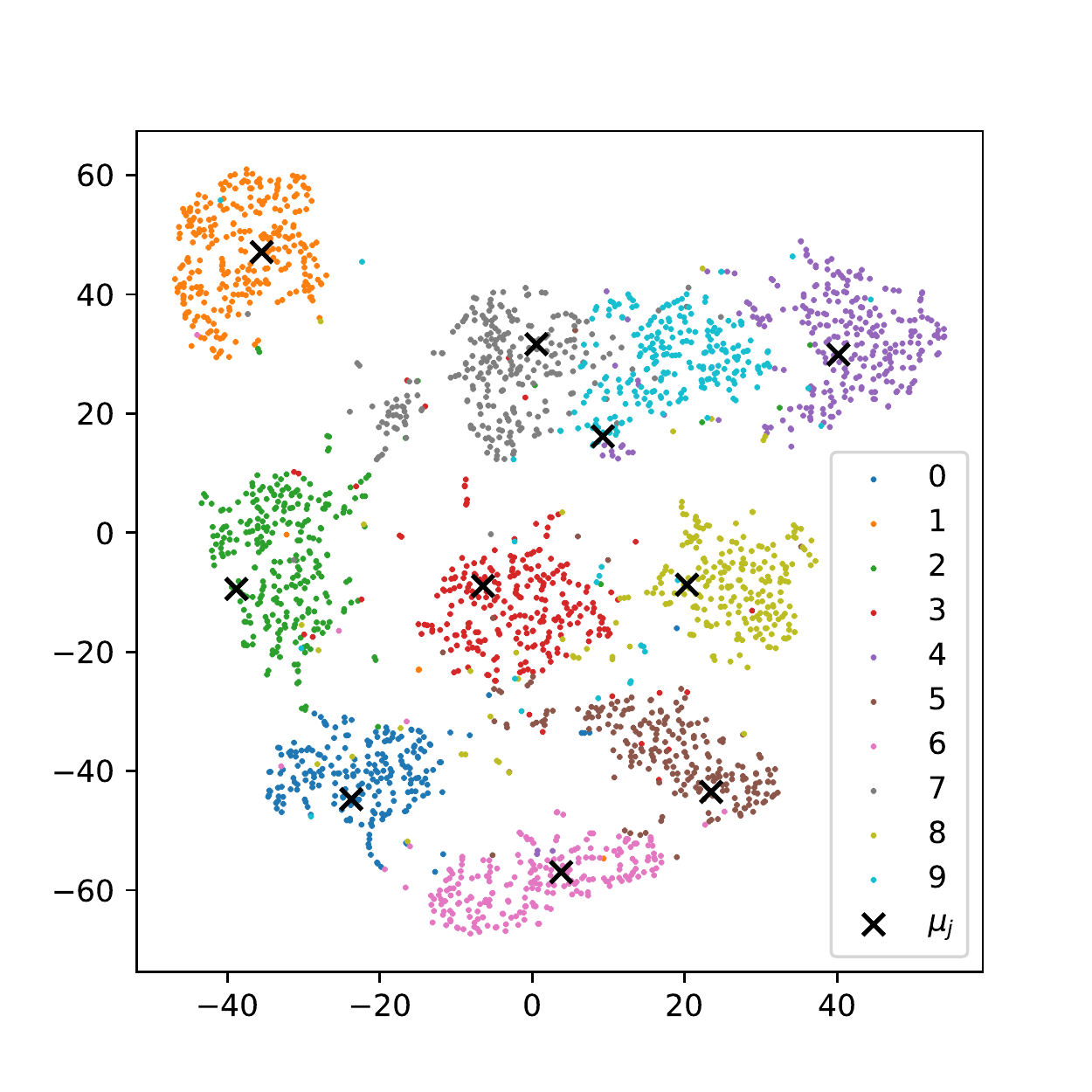} 
    &
    \includegraphics[scale=0.34]{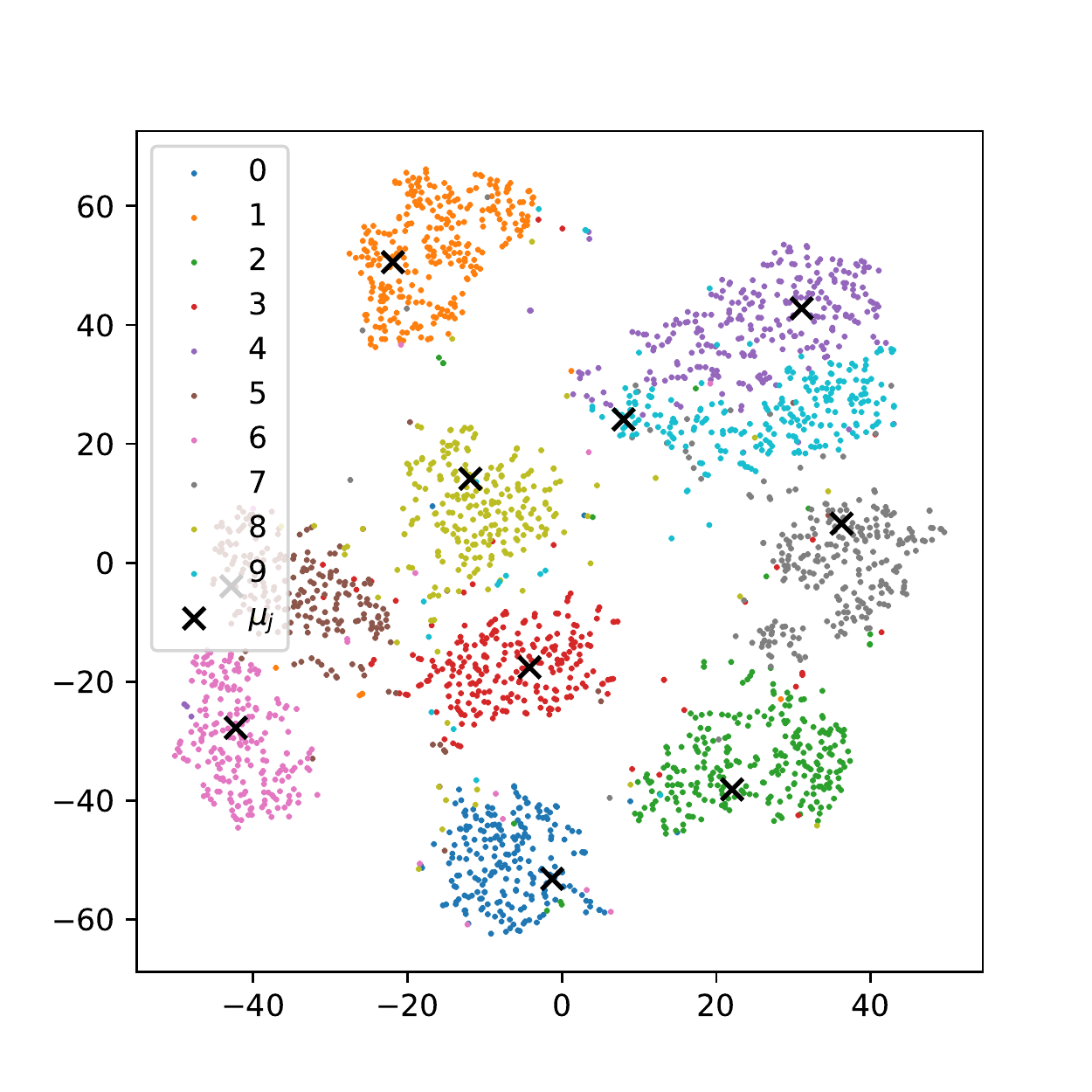}
    \\
    ps-DRAE &  ms-DRAE ($k=10$) &  ms-DRAE ($k=50$)
  \end{tabular}
     
\end{center}
  \caption{
  \footnotesize{ t-SNE on MNIST latent code, the $\mu_j$ are the means of components in the Gaussian mixture prior.
    }}
  \label{fig:TSNe}
  \vspace{-0.2 em}
\end{figure}
\subsection{Results on spherical deterministic relational autoencoder}
\label{subsec:more_result_sdrae}
\begin{figure}[!h]
\begin{center}
    
  \begin{tabular}{ccc}
 \includegraphics[scale=0.34]{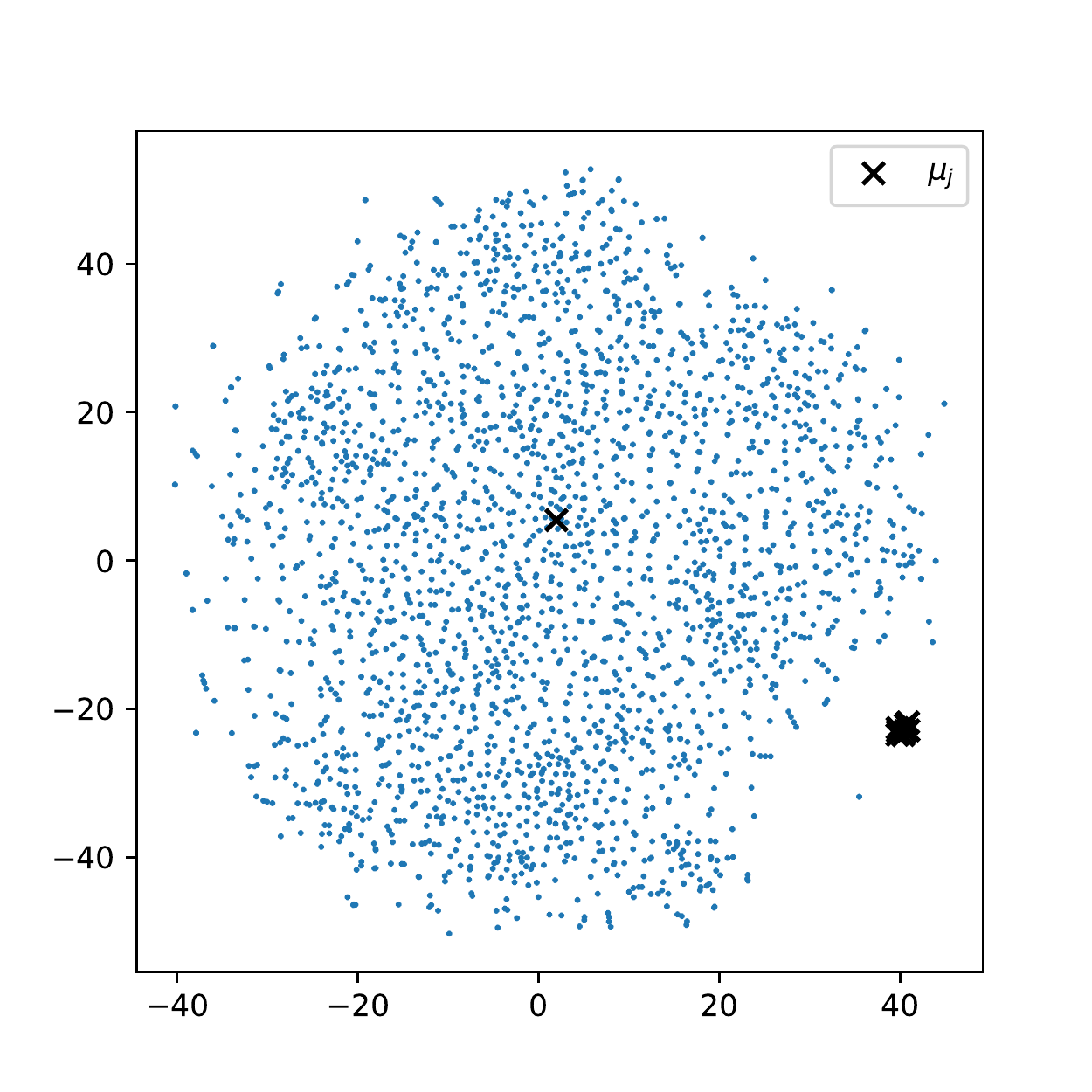} 
    
    &\includegraphics[scale=0.34]{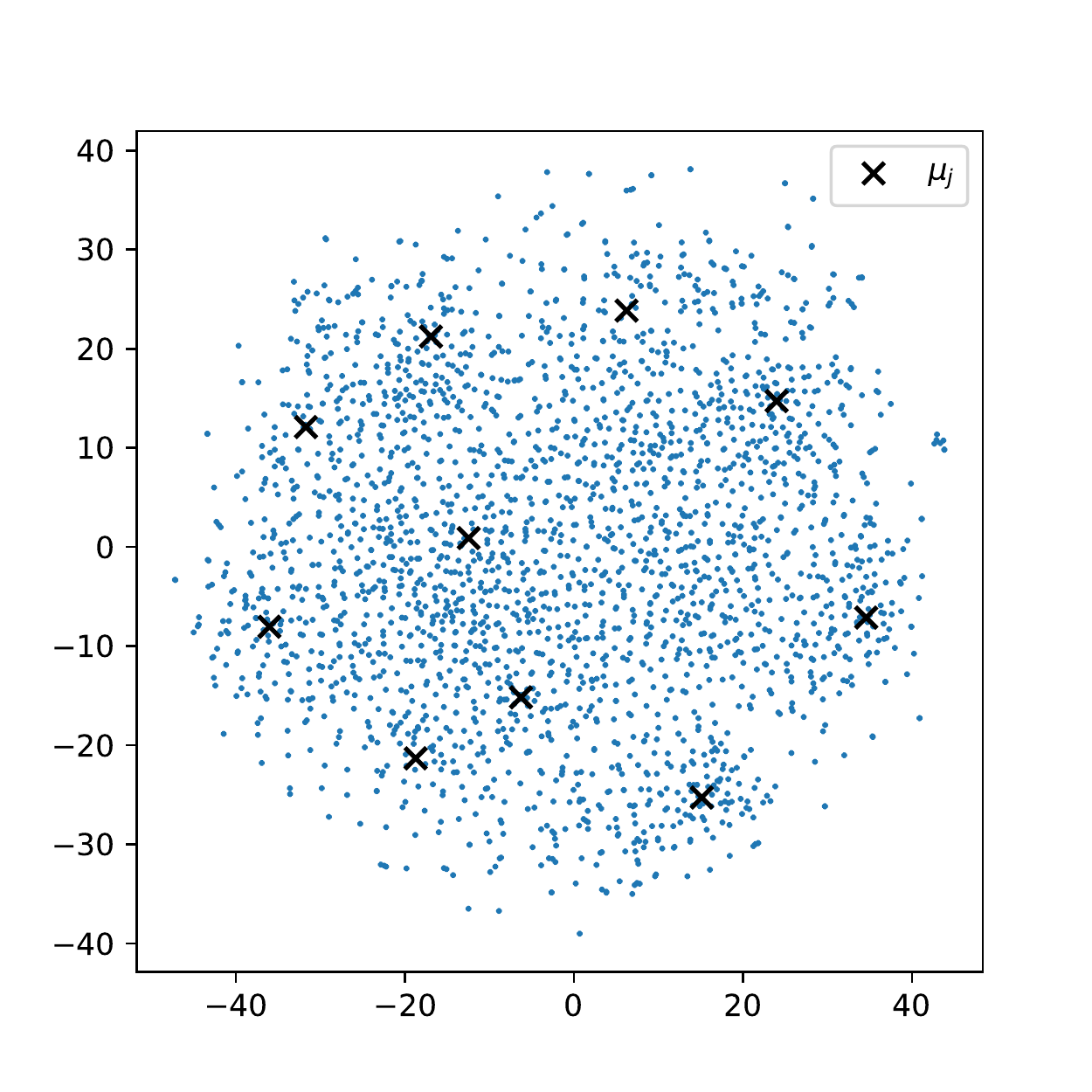} 
     &
    \includegraphics[scale=0.34]{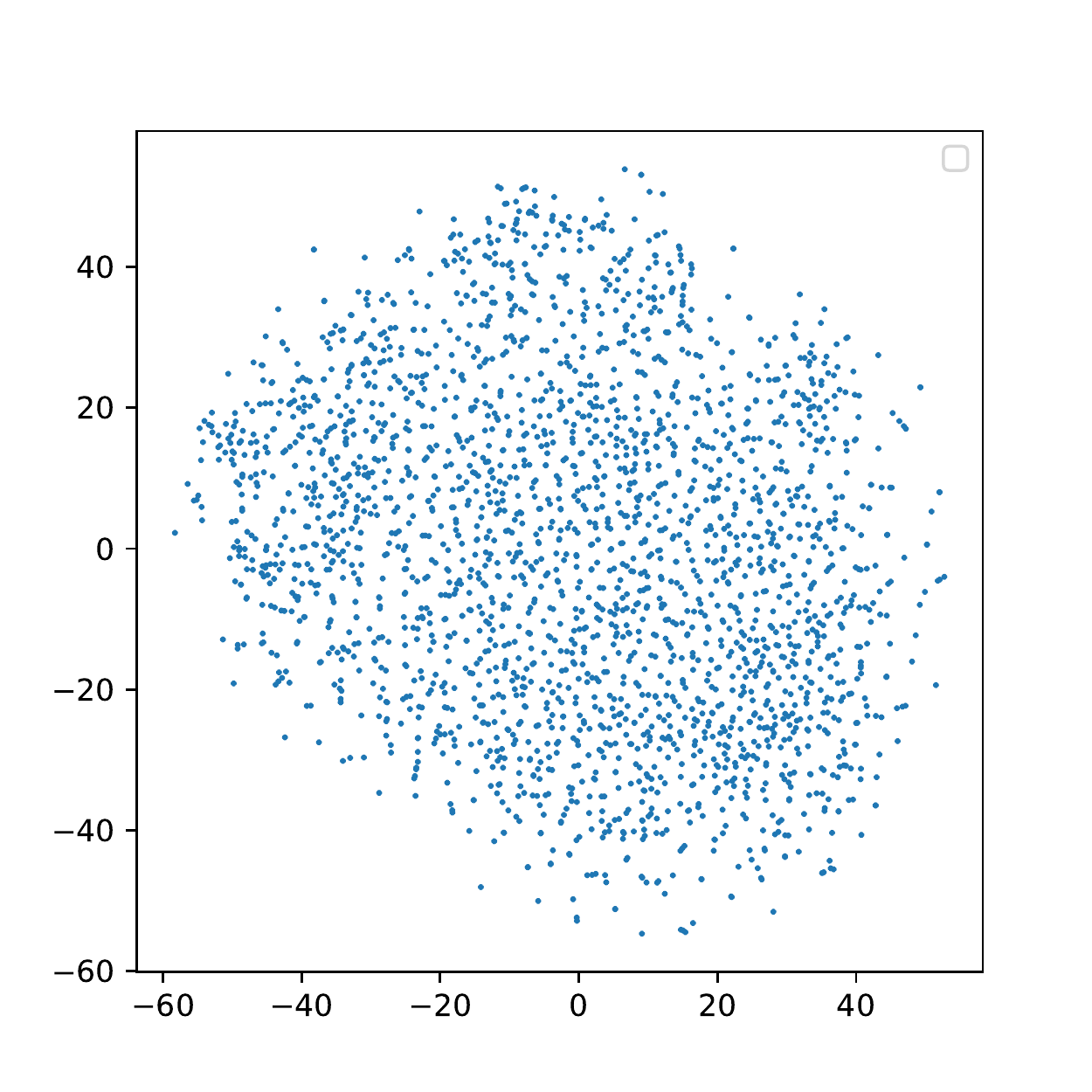} 
  
     \\
     GMVAE&PRAE&SWAE
    \\ 
    \includegraphics[scale=0.34]{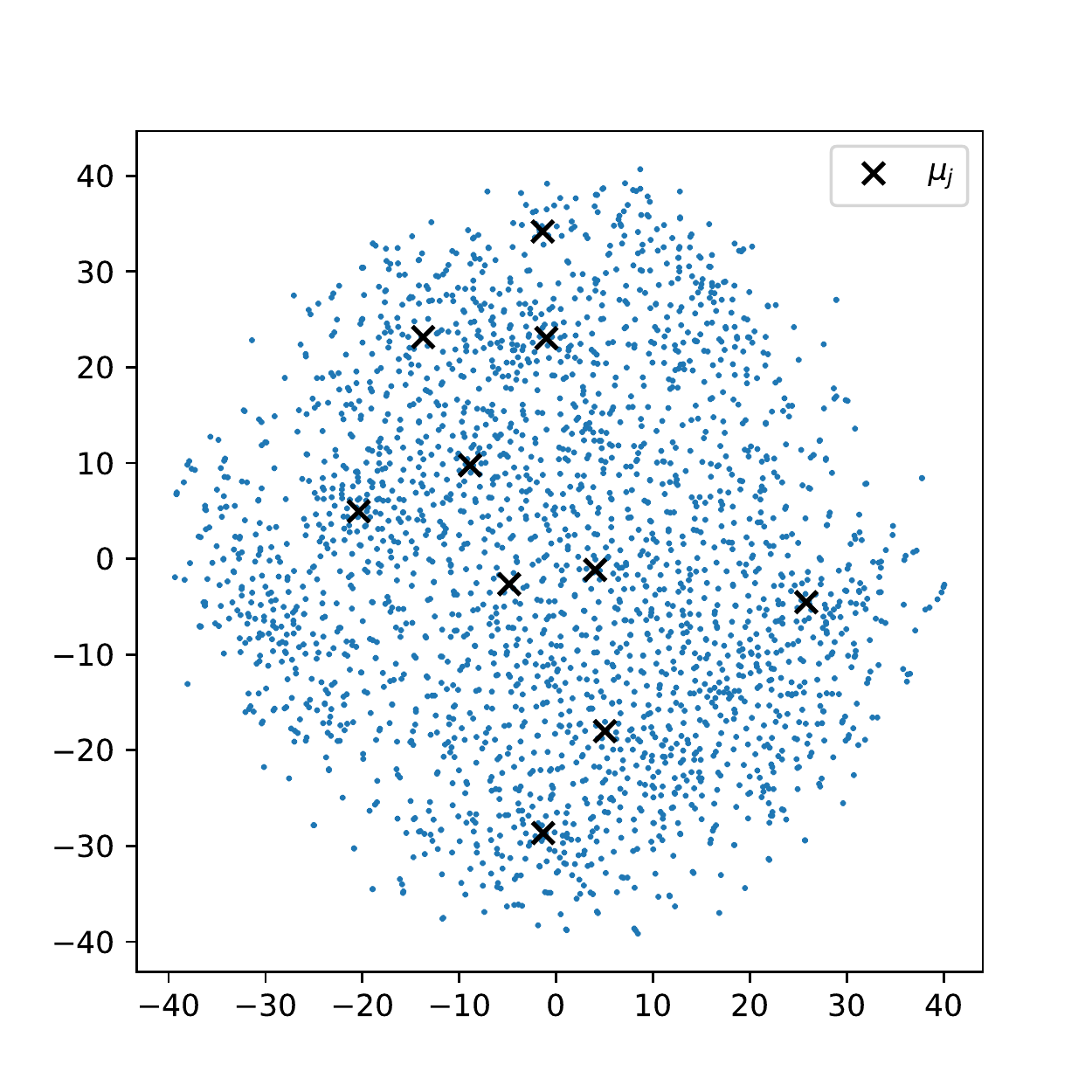} 
    &
    \includegraphics[scale=0.34]{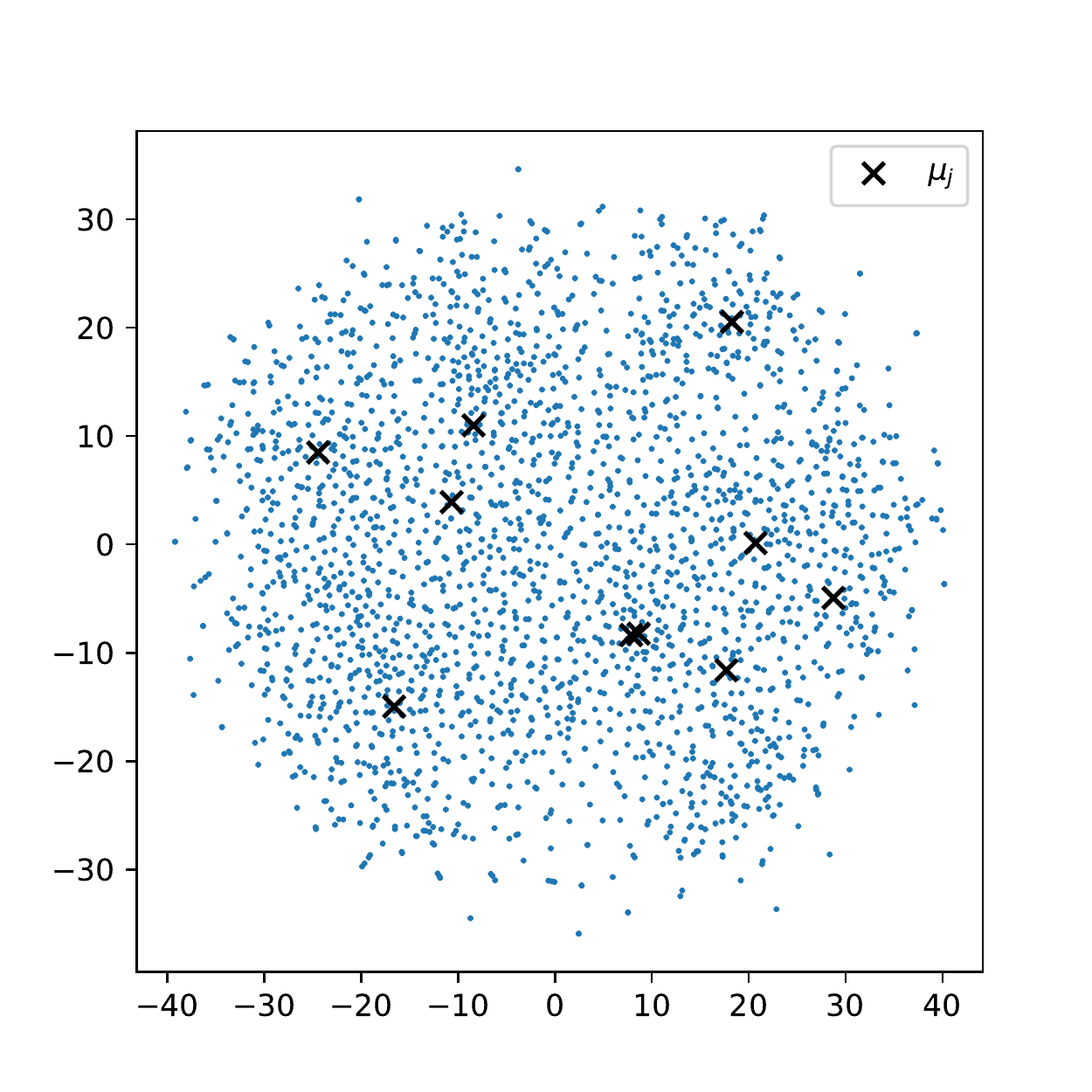} 
    &\includegraphics[scale=0.34]{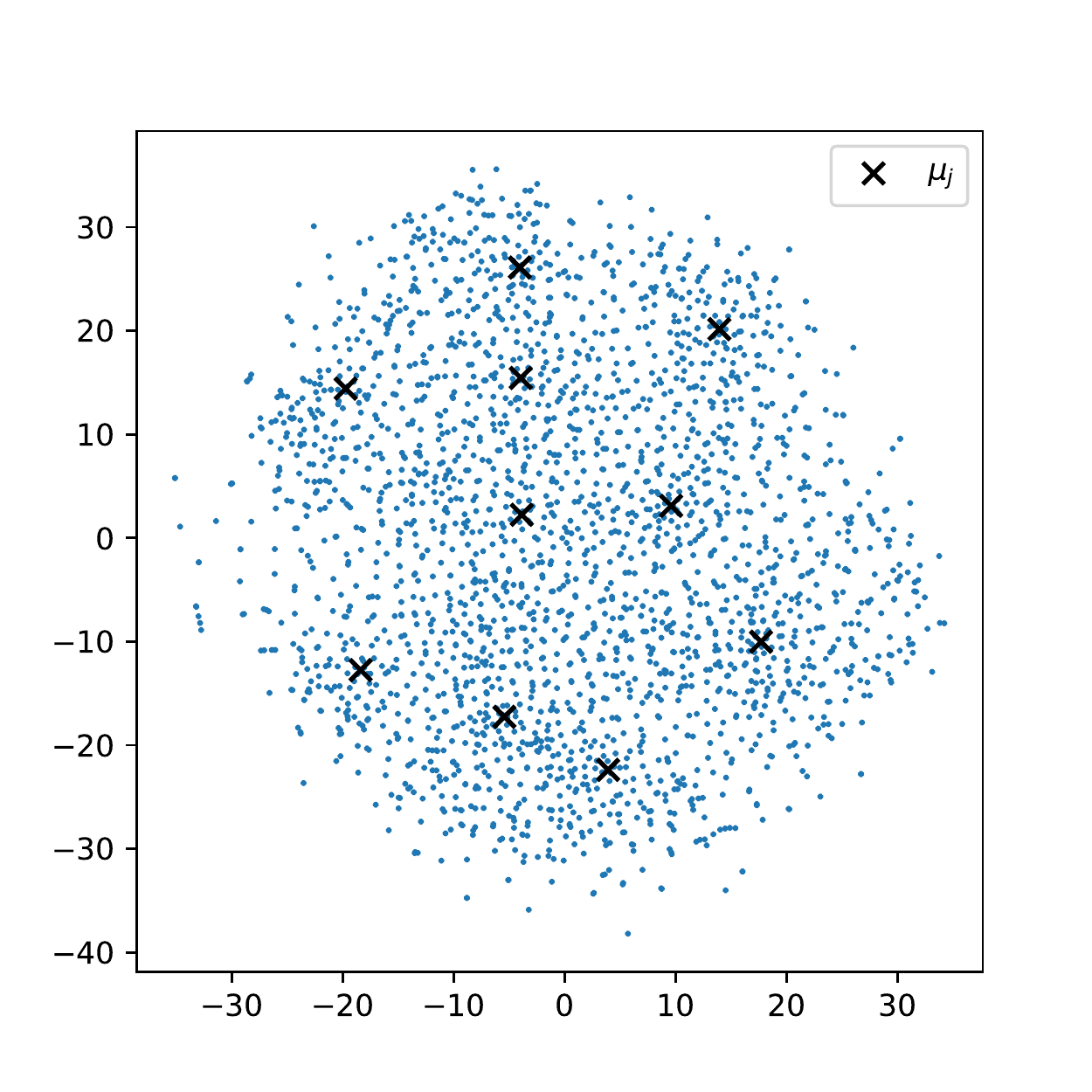}  
    \\
DRAE &m-DRAE &s-DRAE
\\ 
    \includegraphics[scale=0.34]{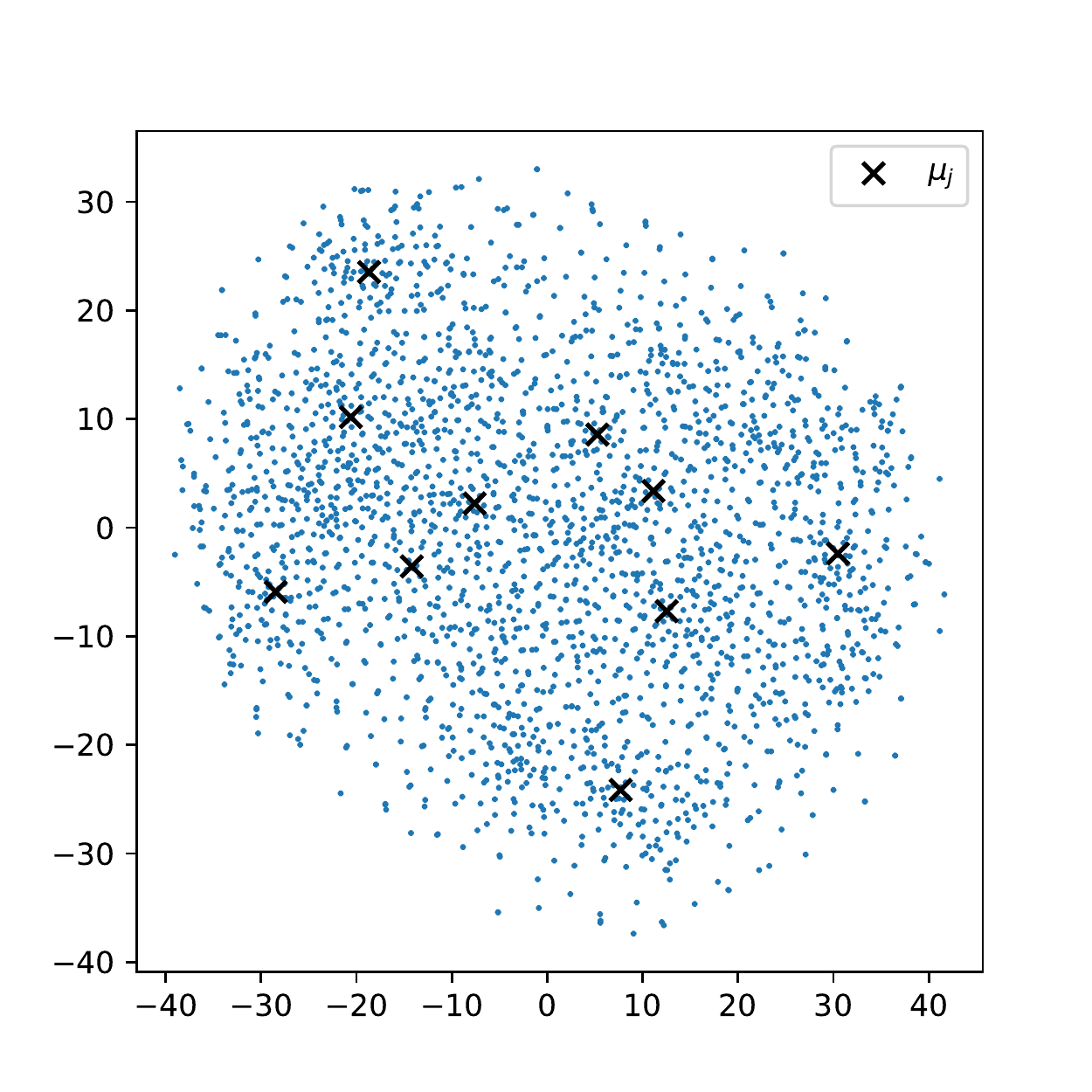} 
    &
    \includegraphics[scale=0.34]{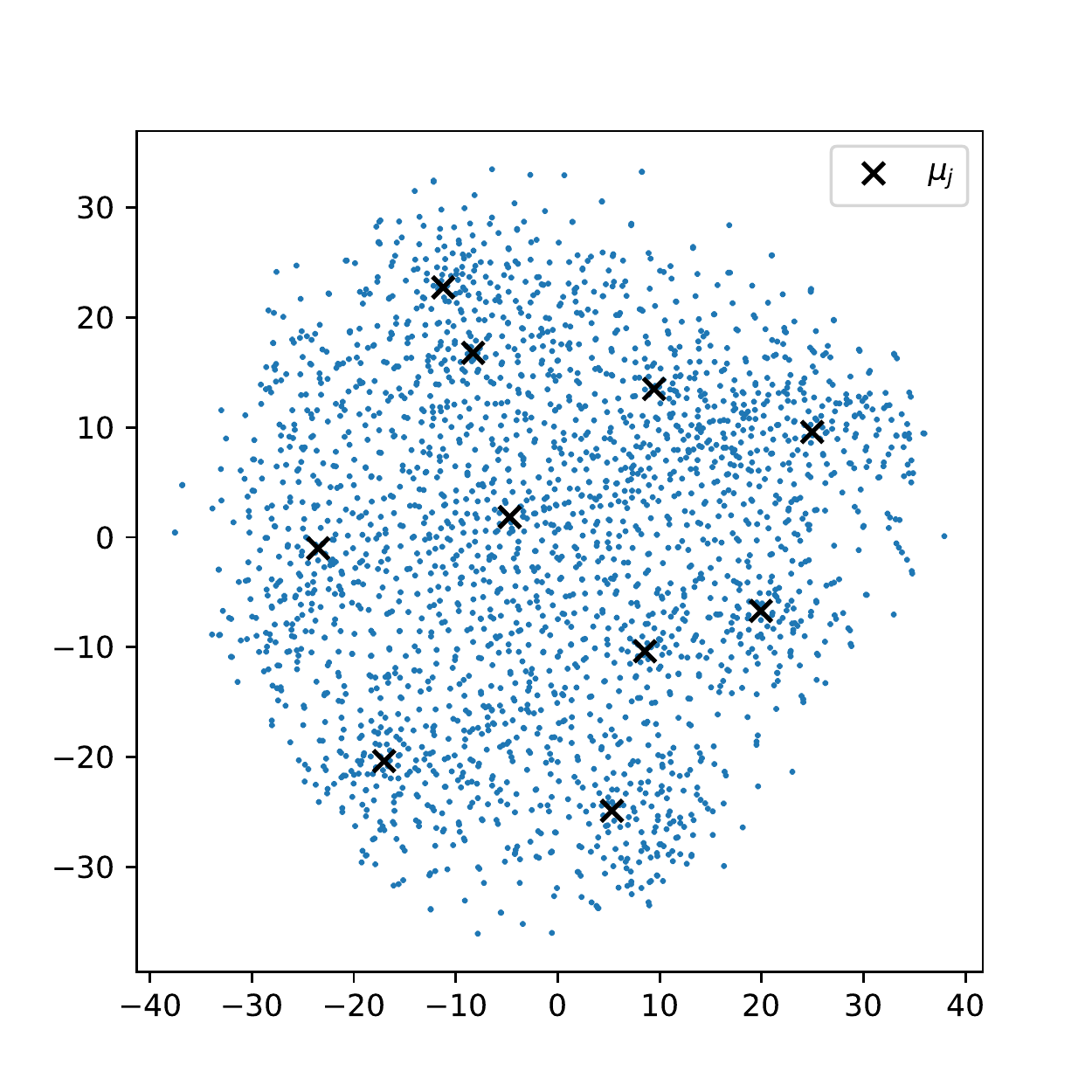} 
    &\includegraphics[scale=0.34]{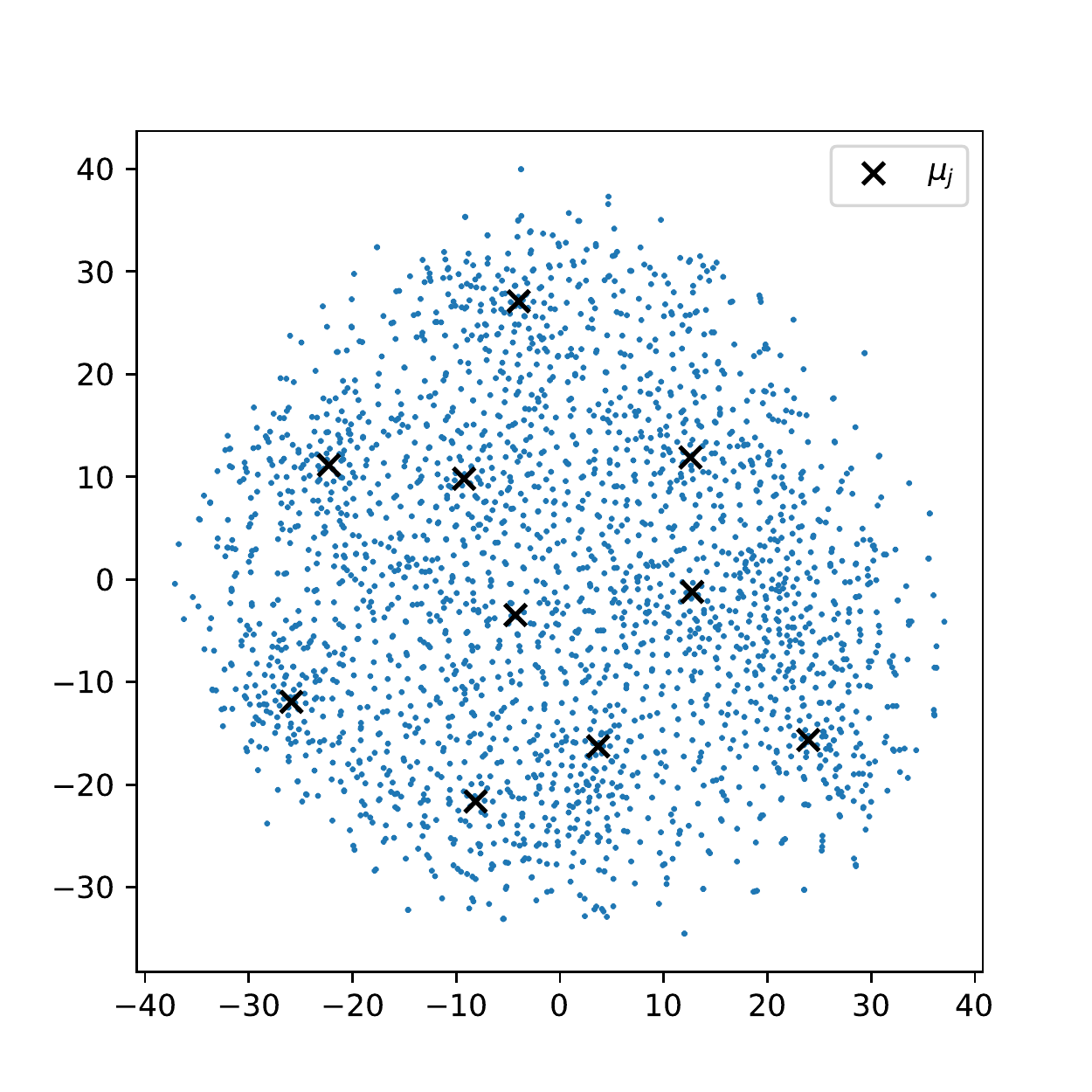}  
    \\
ps-DRAE &ms-DRAE ($k=10$) &ms-DRAE ($k=50$)
  \end{tabular}
  \end{center}
  \caption{
  \footnotesize{ t-SNE on CelebA latent code,  the $\mu_j$ are the means of components in the Gaussian mixture prior.
    }}
  \label{fig:Celebtsne}
  \vspace{-0.2 em}
\end{figure}

\textbf{Visualization of the latent space: } First, we plot the t-SNE visualization of the latent distribution of various types of autoencoders. For autoencoders that have a mixture of Gaussian distributions prior, the Gaussian means are also visualized. In Figure~\ref{fig:TSNe} with MNIST dataset, we find that Vampprior only learns collapsed Gaussian distributions that cannot cover the encoded distribution. Regarding GMVAE, its latent modes are overlapped, which might affect the generative diversity. For PRAE, a probabilistic relational regularized autoencoder, its Gaussian means seem to cover all the space but the clustering effect is not obvious. Therefore, the generated images from PRAE might be blurred. Three deterministic relational regularized autoencoders (DRAE, m-DRAE, s-DRAE) have diverse Gaussian components, which cover the latent space well, however, they still miss some small parts.  With the CelebA dataset in Figure~\ref{fig:Celebtsne} whose latent space dimension is much larger than that of MNIST (64 dimensions compared to 8 dimensions),  the differences between DRAE, m-DRAE, and s-DRAE can be seen clearly. In particular, s-DRAE gets a better visualized latent space and its prior can cover well that space while m-DRAE and DRAE seem to miss some parts of the space.

\begin{figure}[!h]
\begin{center}

  \begin{tabular}{ccc}
 \includegraphics[scale=0.28]{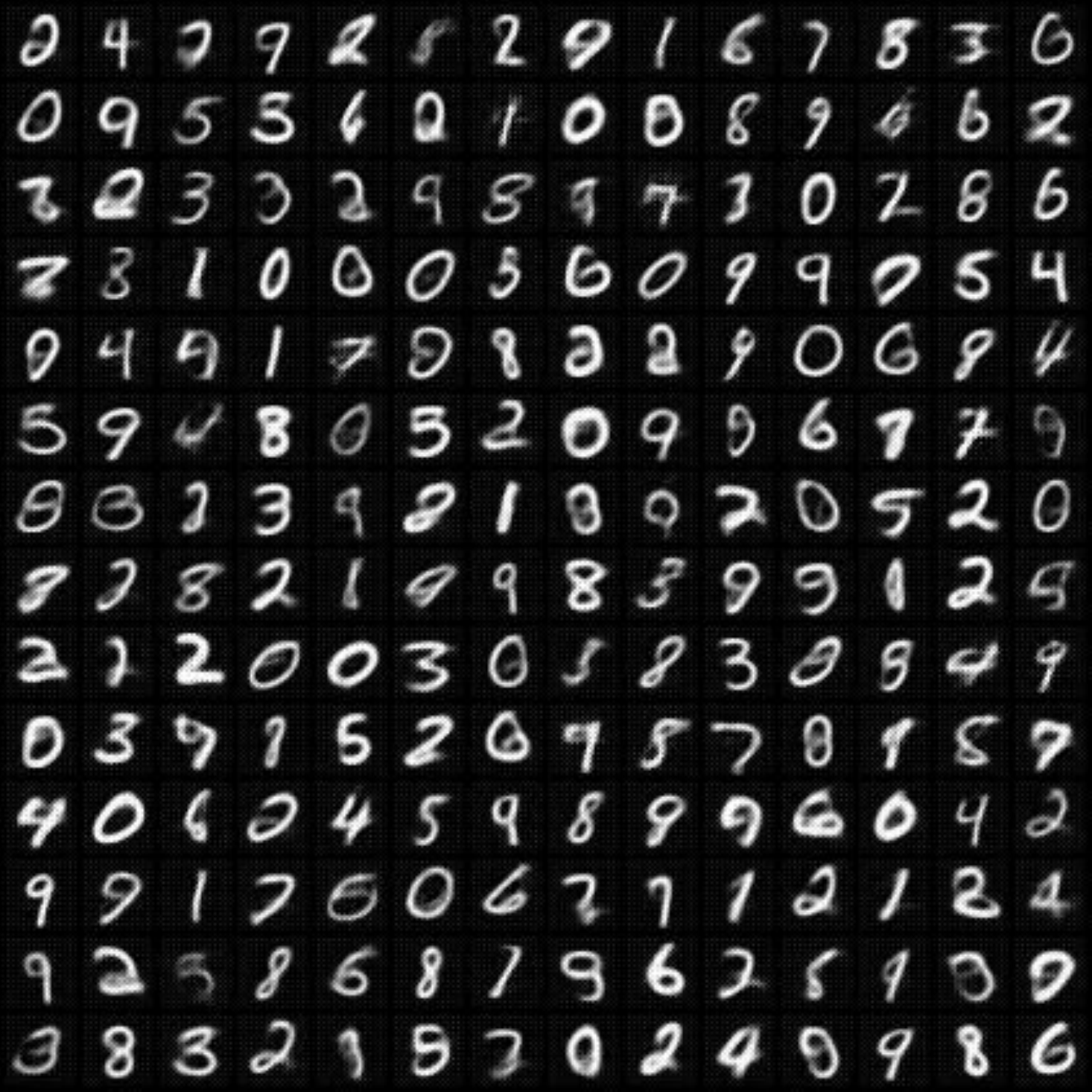} 
    
    &\includegraphics[scale=0.28]{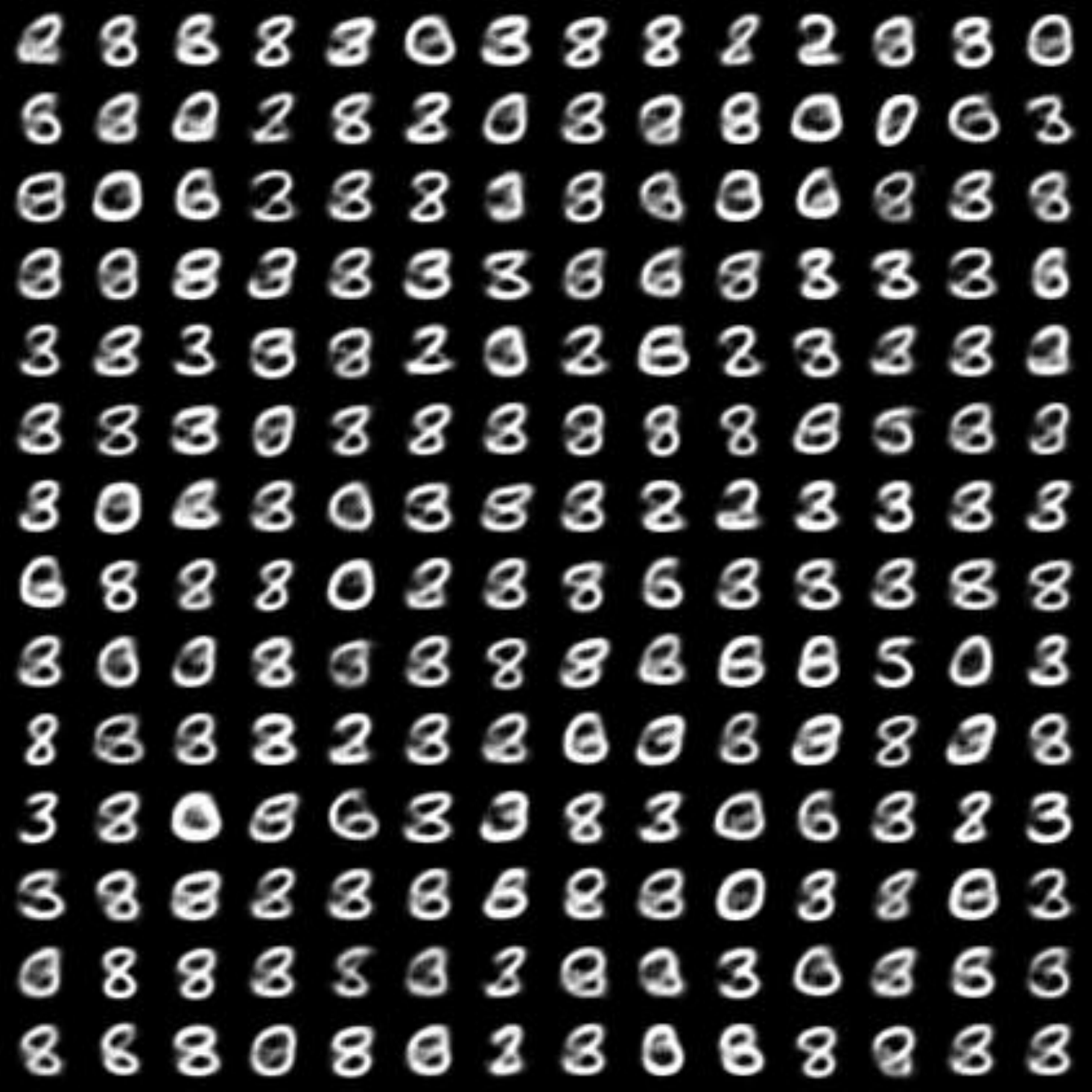} 
     &
 \includegraphics[scale=0.28]{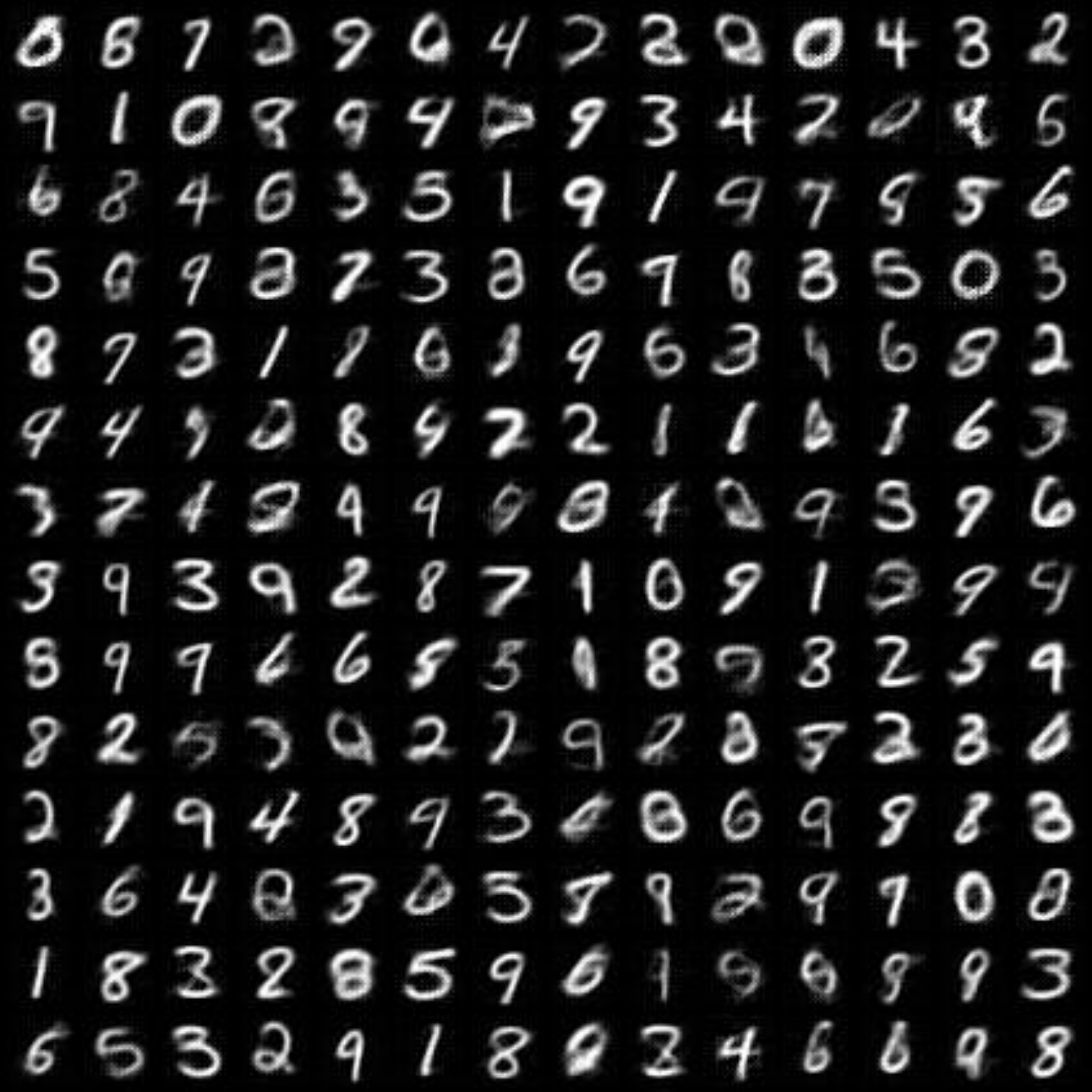} 
 \\
   VAE&Vampprior&GMVAE
  \\
    \includegraphics[scale=0.28]{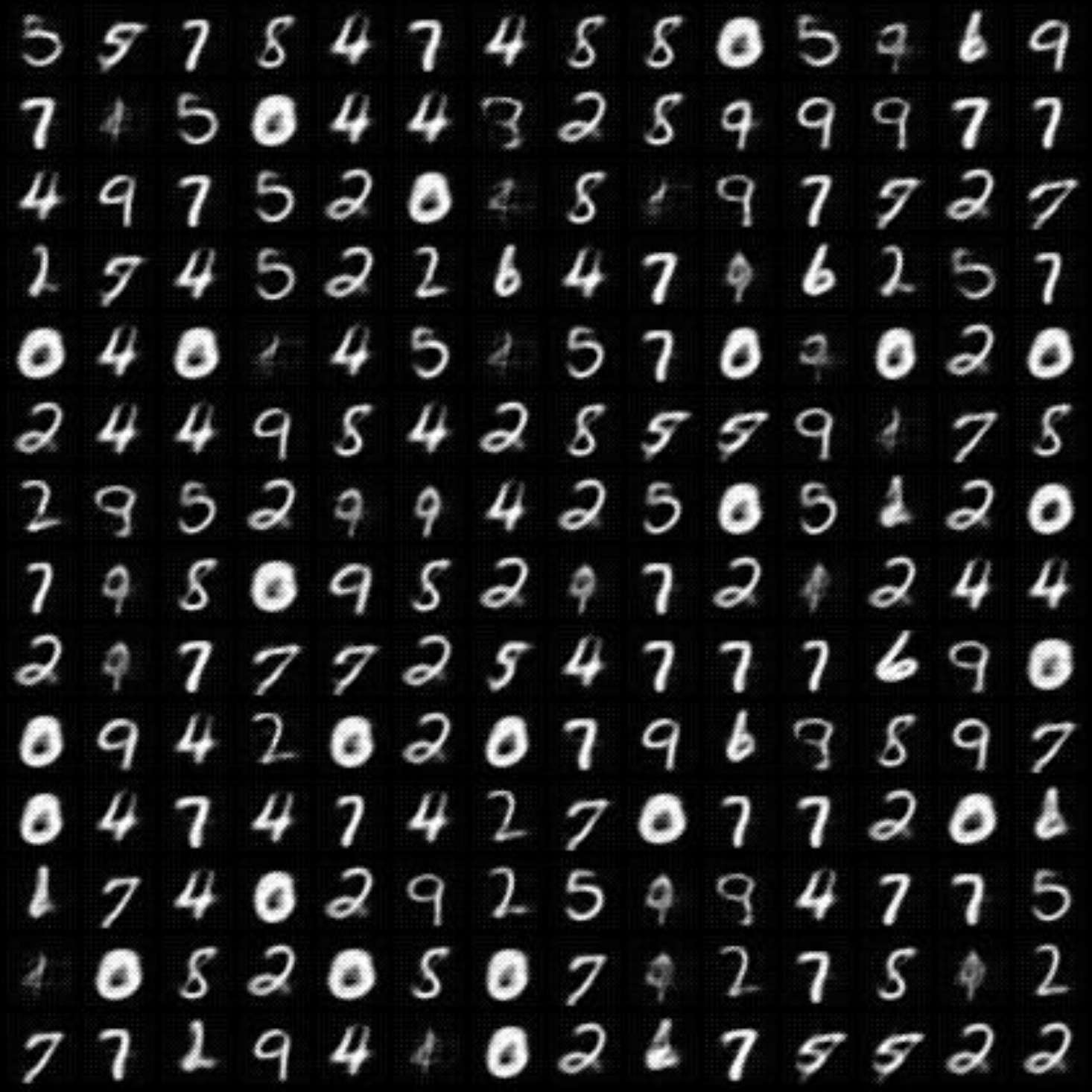} 
    &
    \includegraphics[scale=0.28]{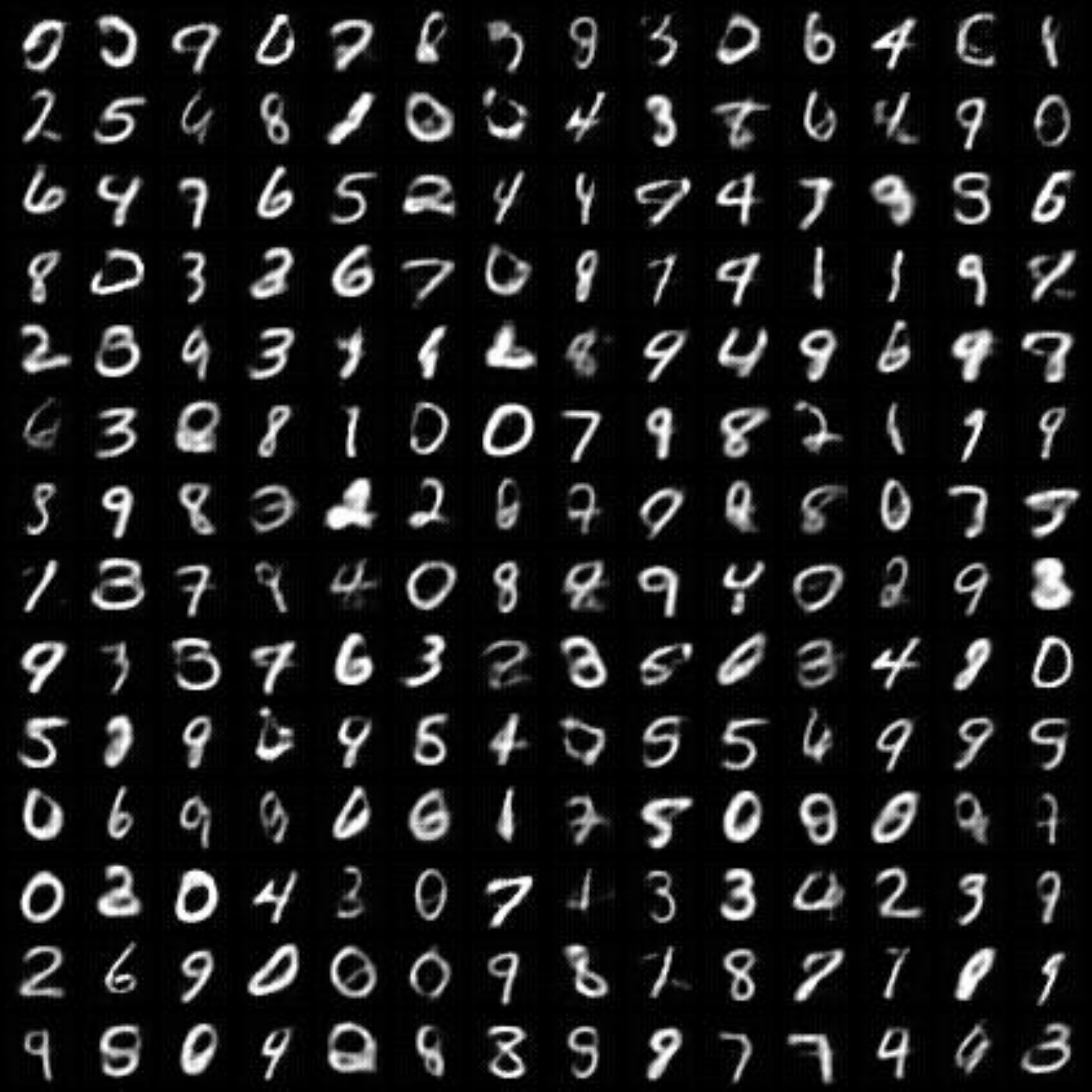} 
    
    &\includegraphics[scale=0.28]{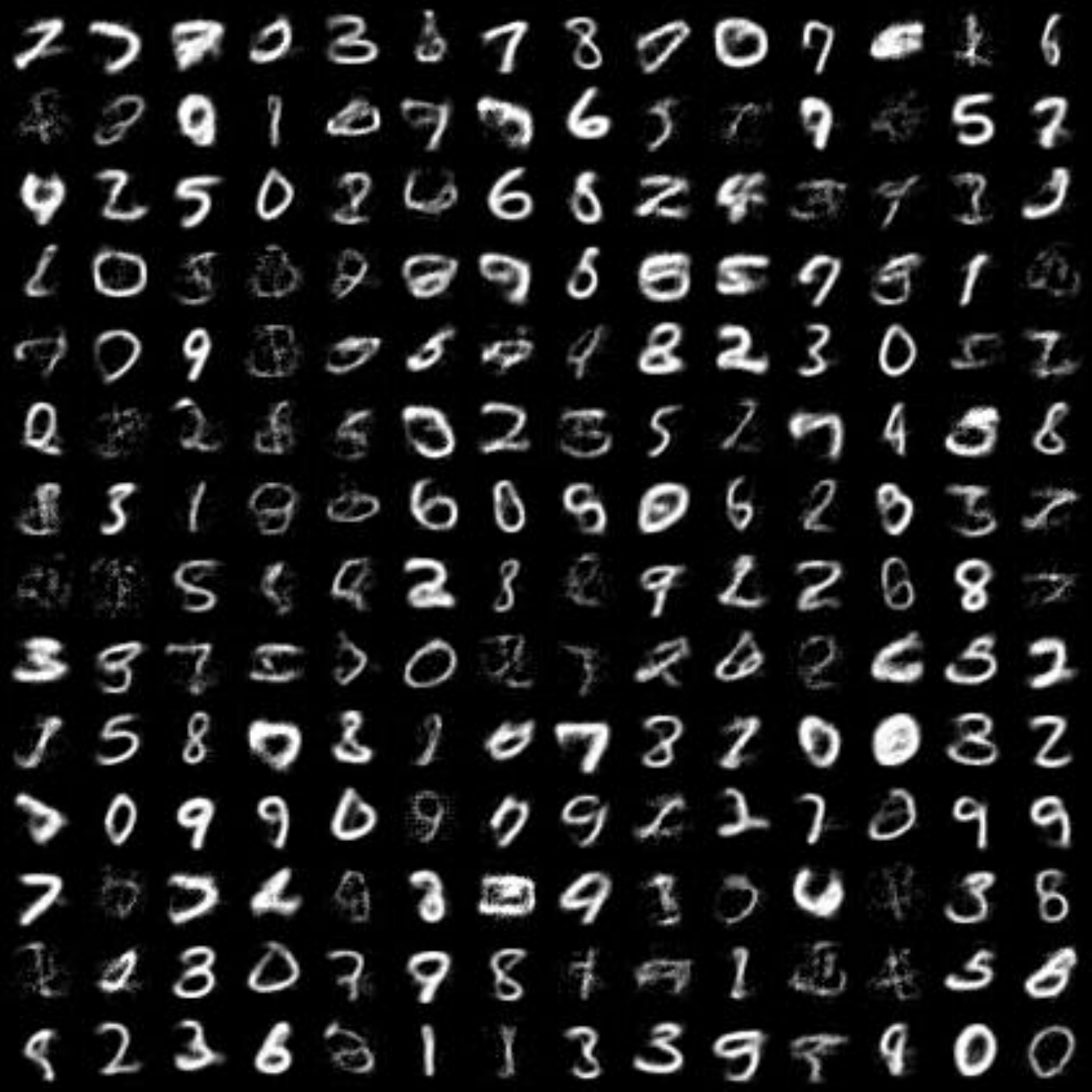}  
    \\
    PRAE &  SWAE &WAE
    \\
 \includegraphics[scale=0.28]{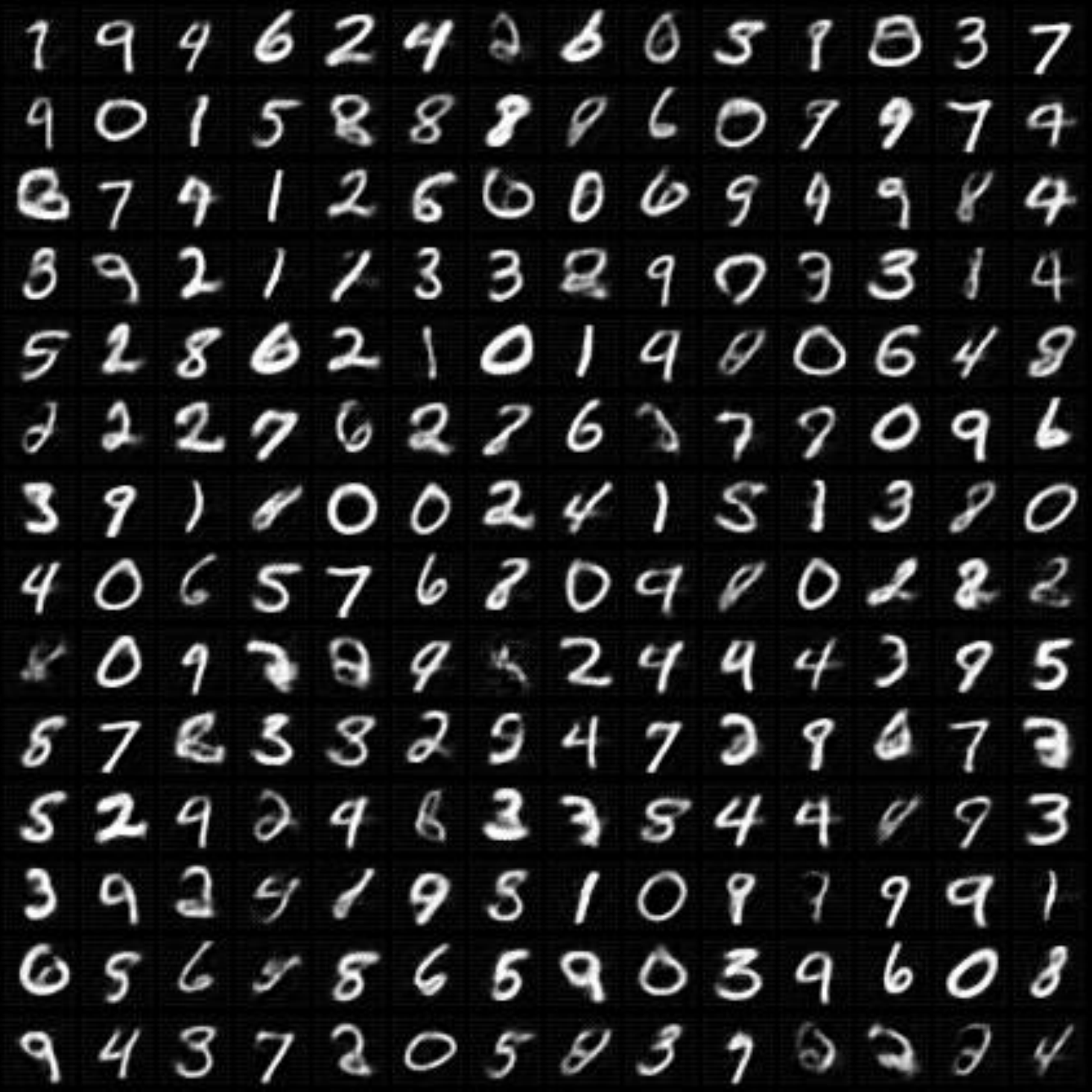} 
  &
    \includegraphics[scale=0.28]{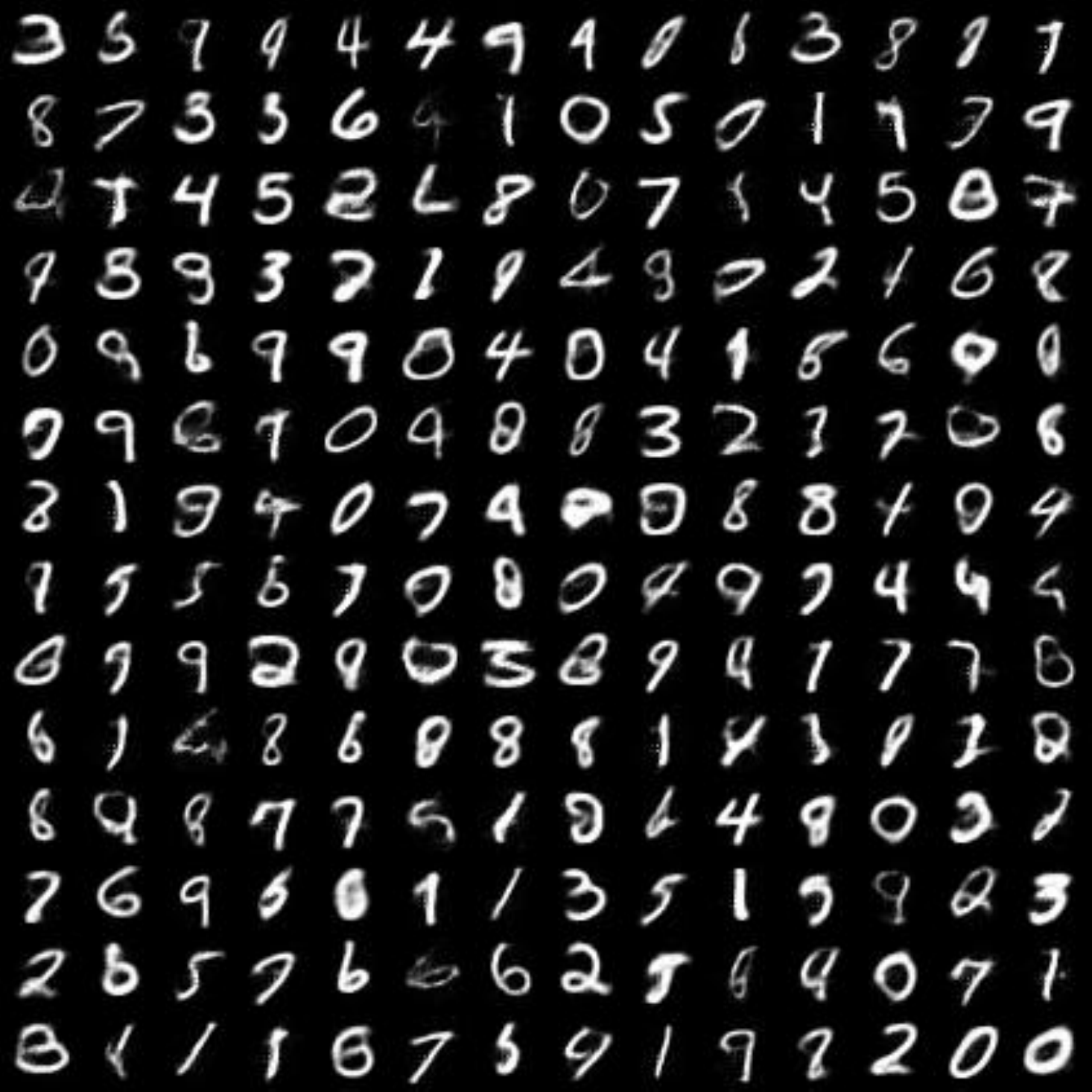} 
    &\includegraphics[scale=0.28]{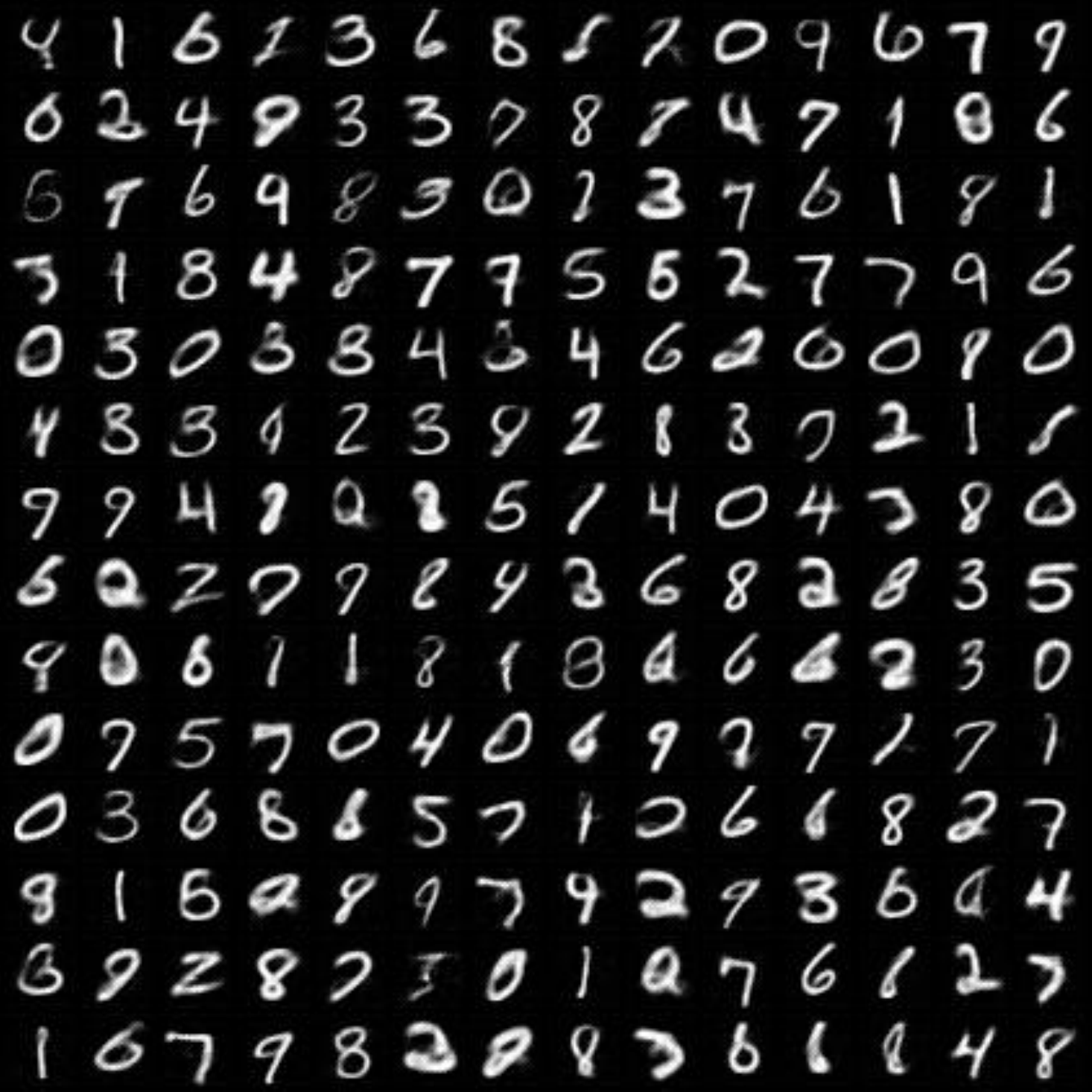}  
    \\
    DRAE & m-DRAE & s-DRAE
    \\
 \includegraphics[scale=0.28]{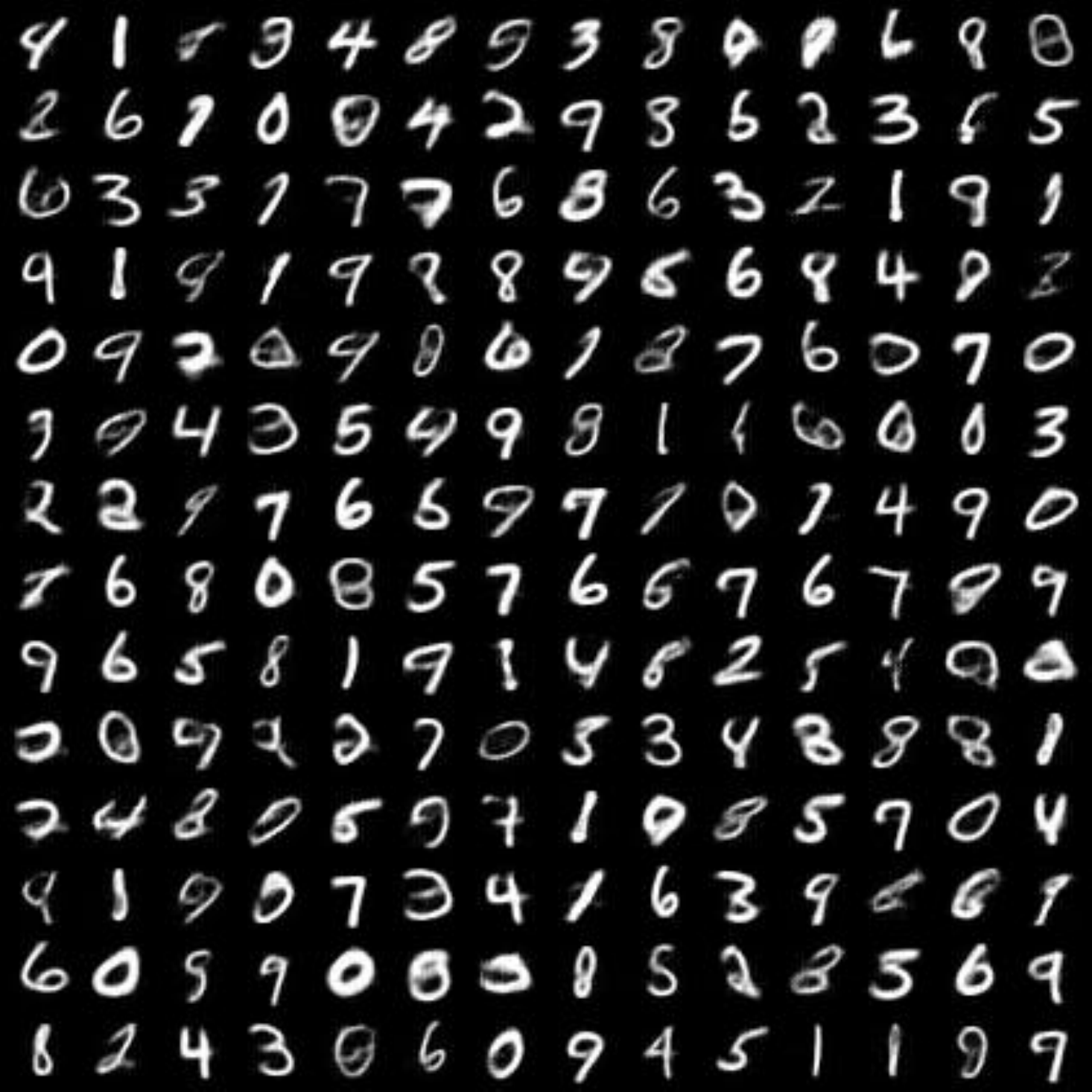} 
  &
    \includegraphics[scale=0.28]{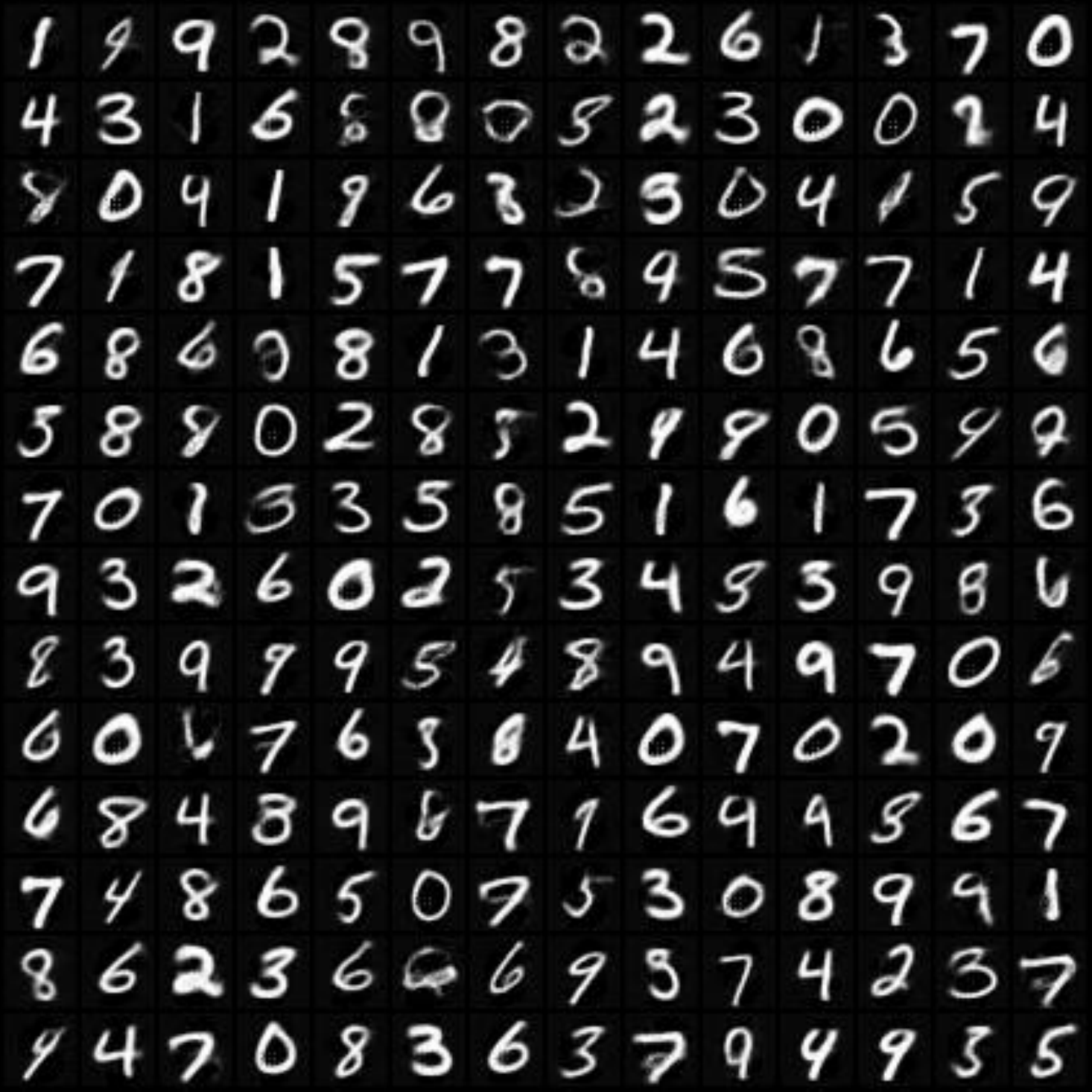} 
    &\includegraphics[scale=0.28]{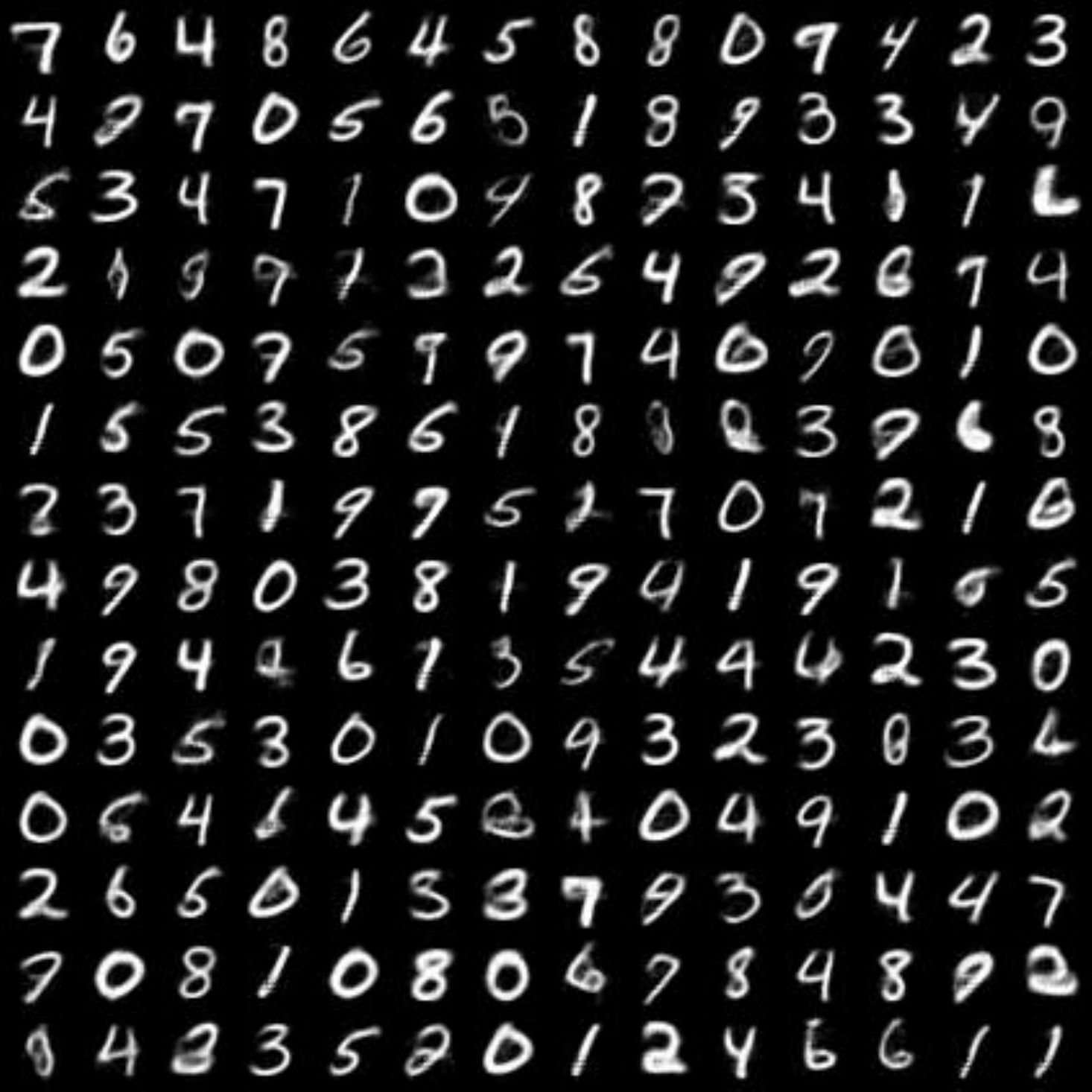}  
    \\
    ps-DRAE &  ms-DRAE ($k=10$) &  ms-DRAE ($k=50$)
  \end{tabular}
\end{center}
  \caption{
  \footnotesize{ Generated images on MNIST dataset. Here, $k$ denotes the number of components of mixture of vMF distributions in MSSFG.
    }}
  \label{fig:MNISTgen}
  \vspace{-0.2 em}
\end{figure}

\textbf{Synthesis images: } We first present randomly generated samples on MNIST dataset in Figure~\ref{fig:MNISTgen}. Based on these images, we can see that DRAE, m-DRAE and s-DRAE produce significantly more realistic images than other autoencoders. Because of the collapsing prior (see Figure~\ref{fig:TSNe}), Vampprior cannot generate diverse digits. Furthermore, by looking closely, images from s-DRAE are clear and can be recognized as digits while some images from DRAE and m-DRAE are blurred. 

\begin{figure}[!h]
\begin{center}

  \begin{tabular}{ccc}
 \includegraphics[scale=0.13]{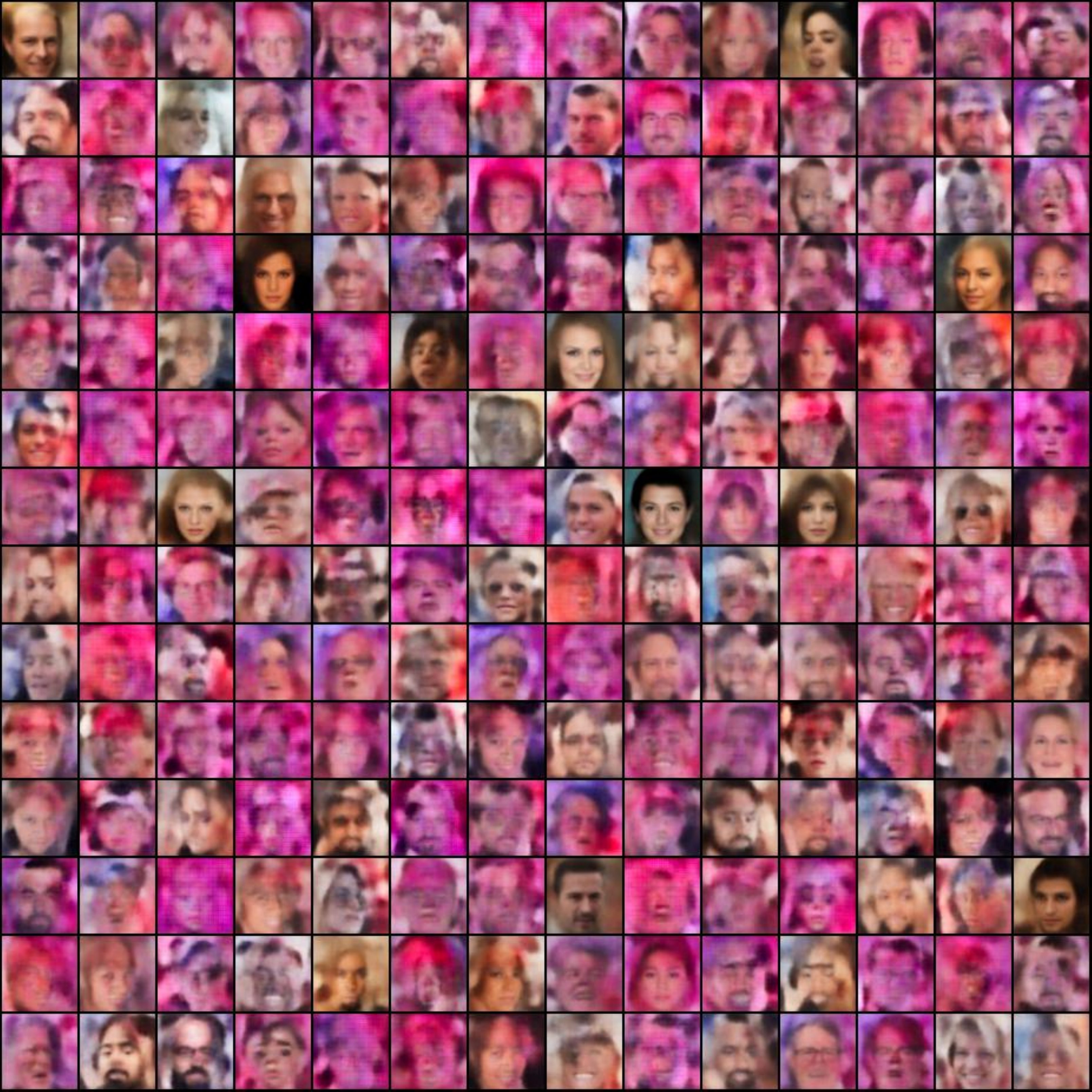} 
    
    &\includegraphics[scale=0.13]{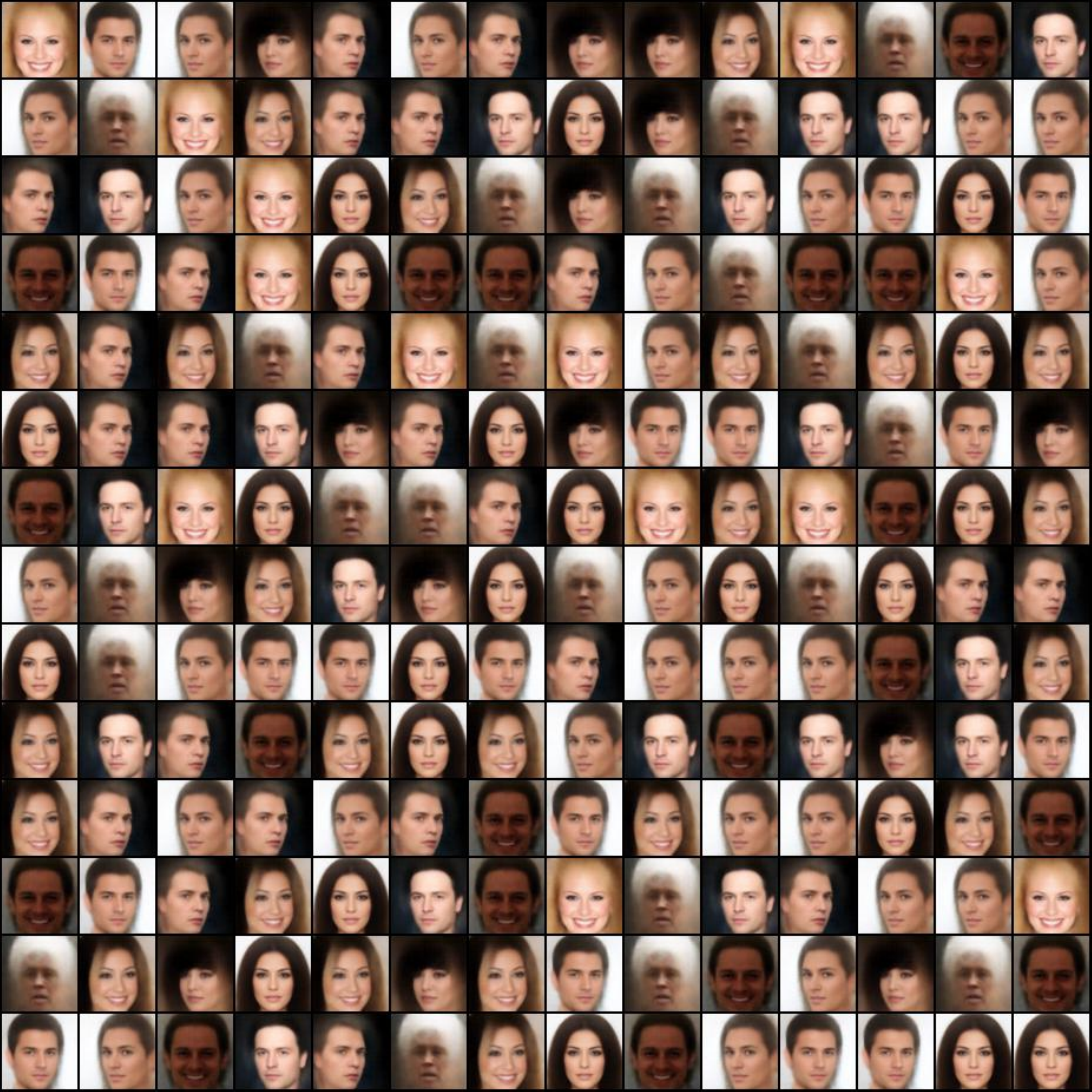} 
     &
    \includegraphics[scale=0.13]{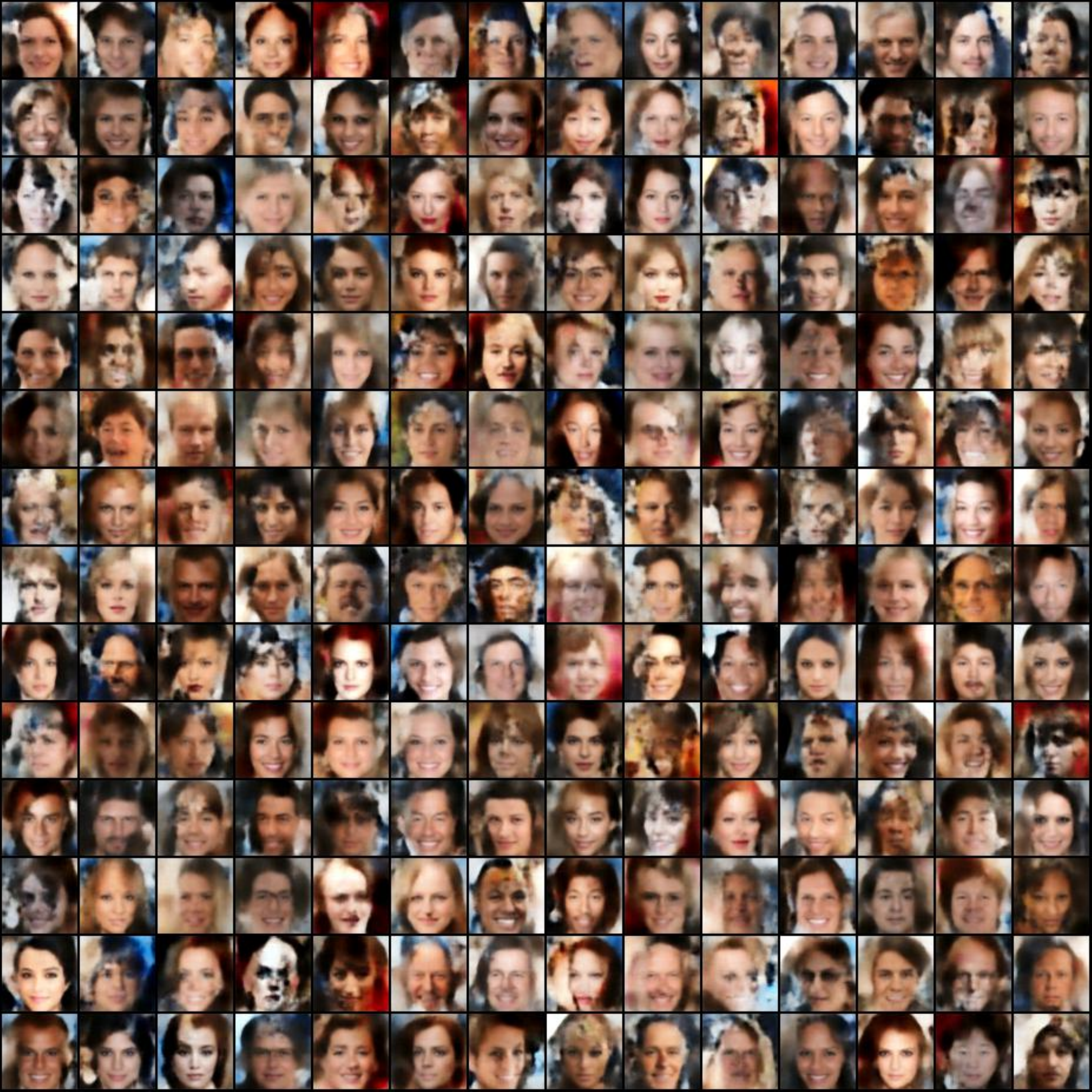} 
  
     \\
     GMVAE&PRAE&SWAE
    \\ 
    \includegraphics[scale=0.13]{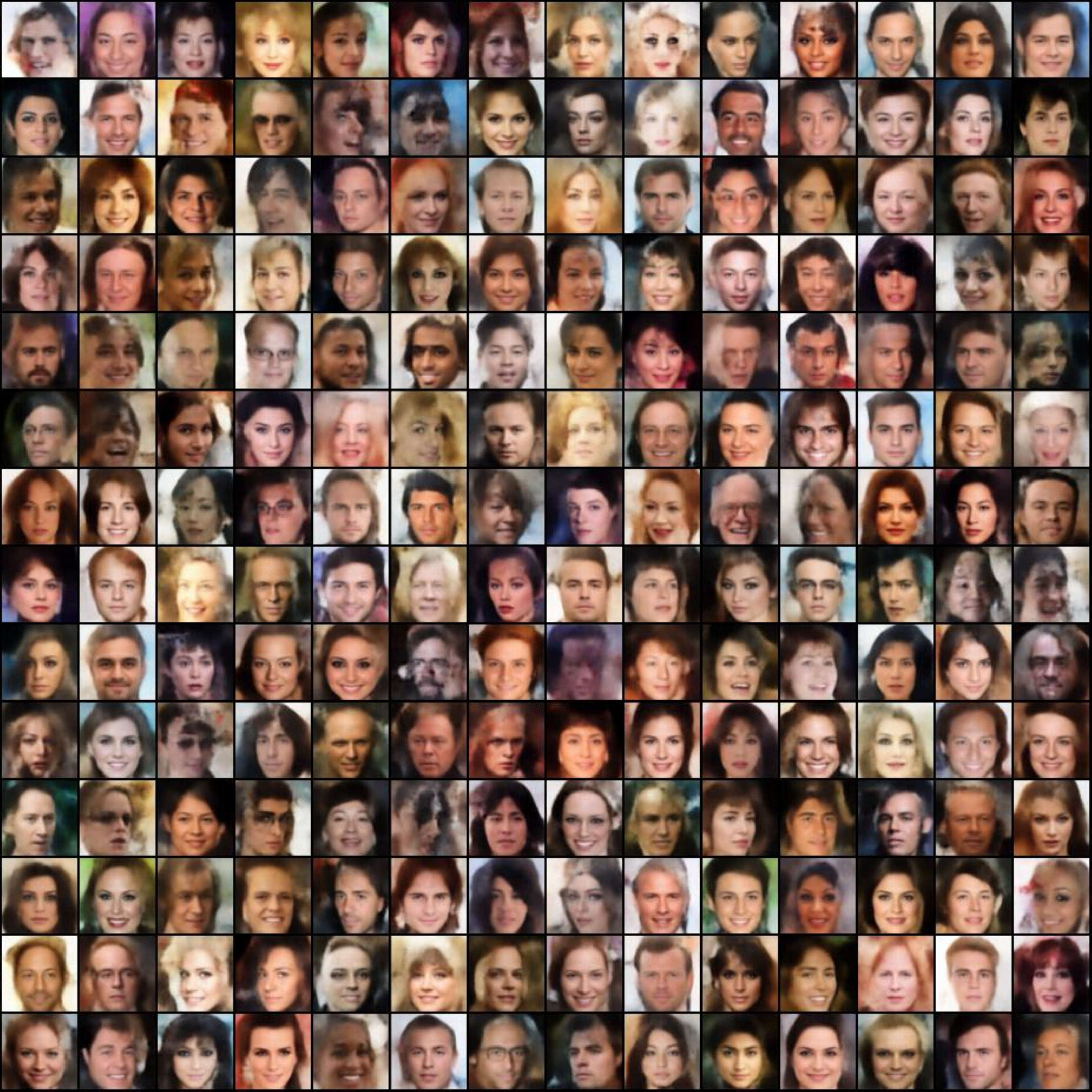} 
    &
    \includegraphics[scale=0.13]{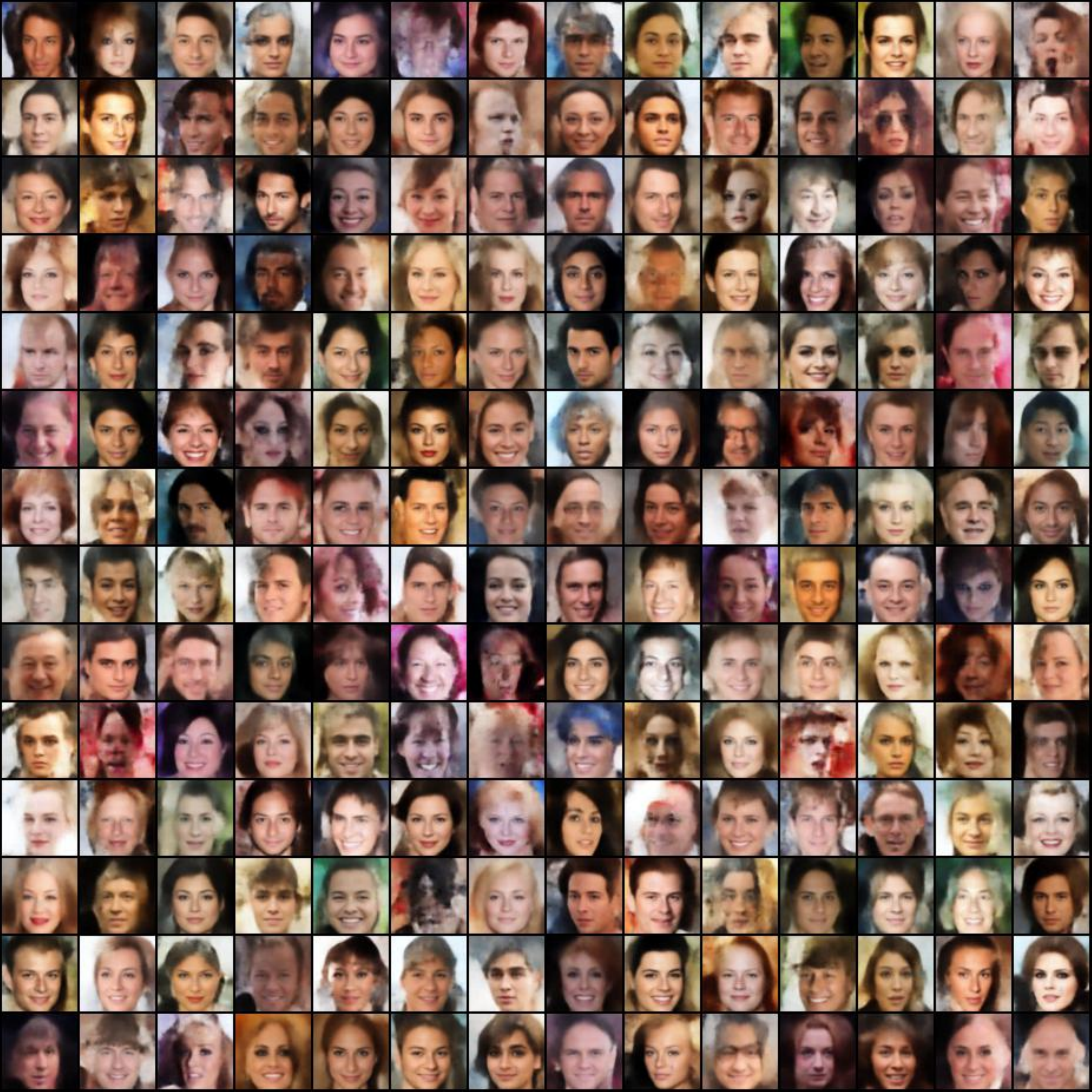} 
    &\includegraphics[scale=0.13]{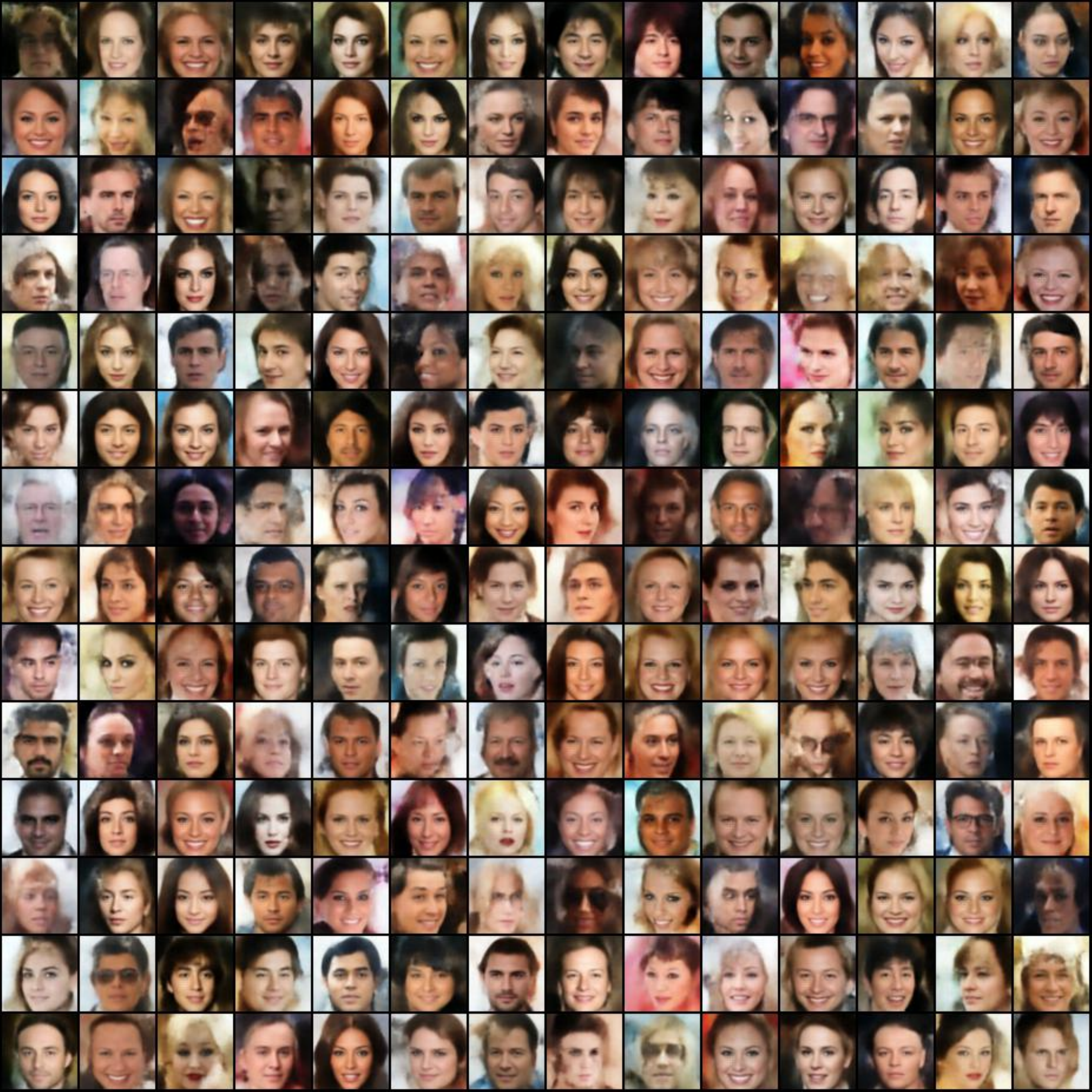}  
    \\
DRAE &m-DRAE &s-DRAE
\\ 
    \includegraphics[scale=0.13]{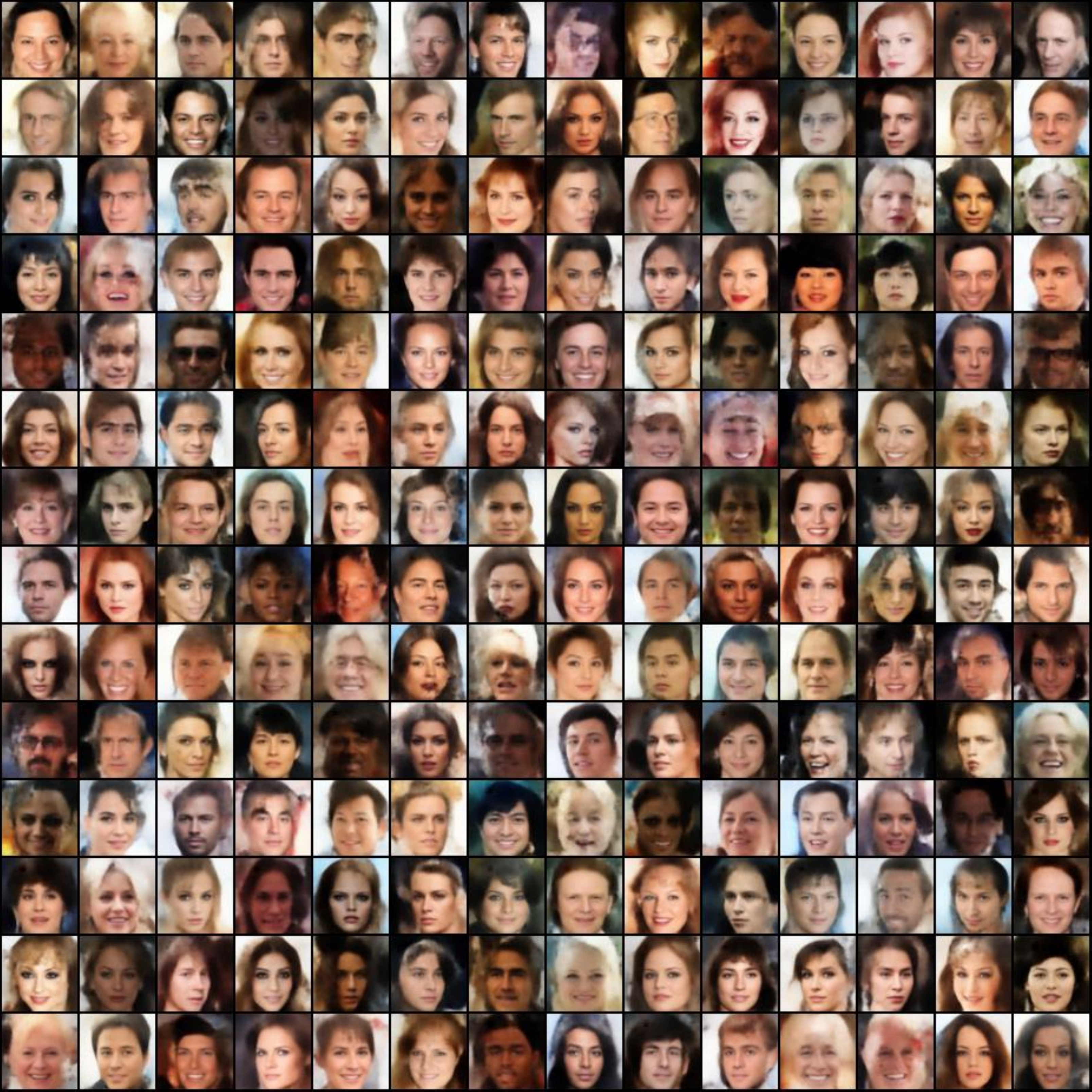} 
    &
    \includegraphics[scale=0.13]{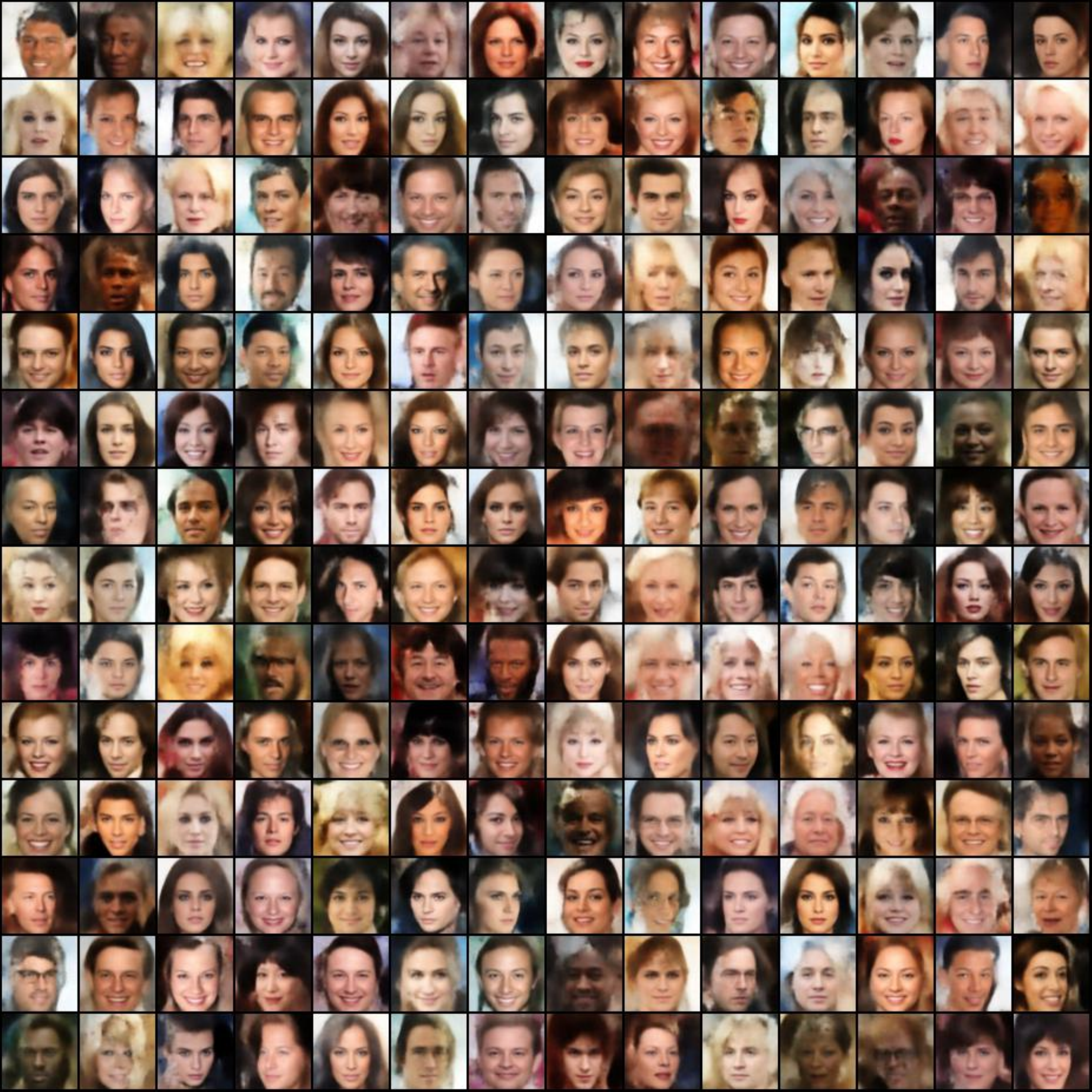} 
    &\includegraphics[scale=0.13]{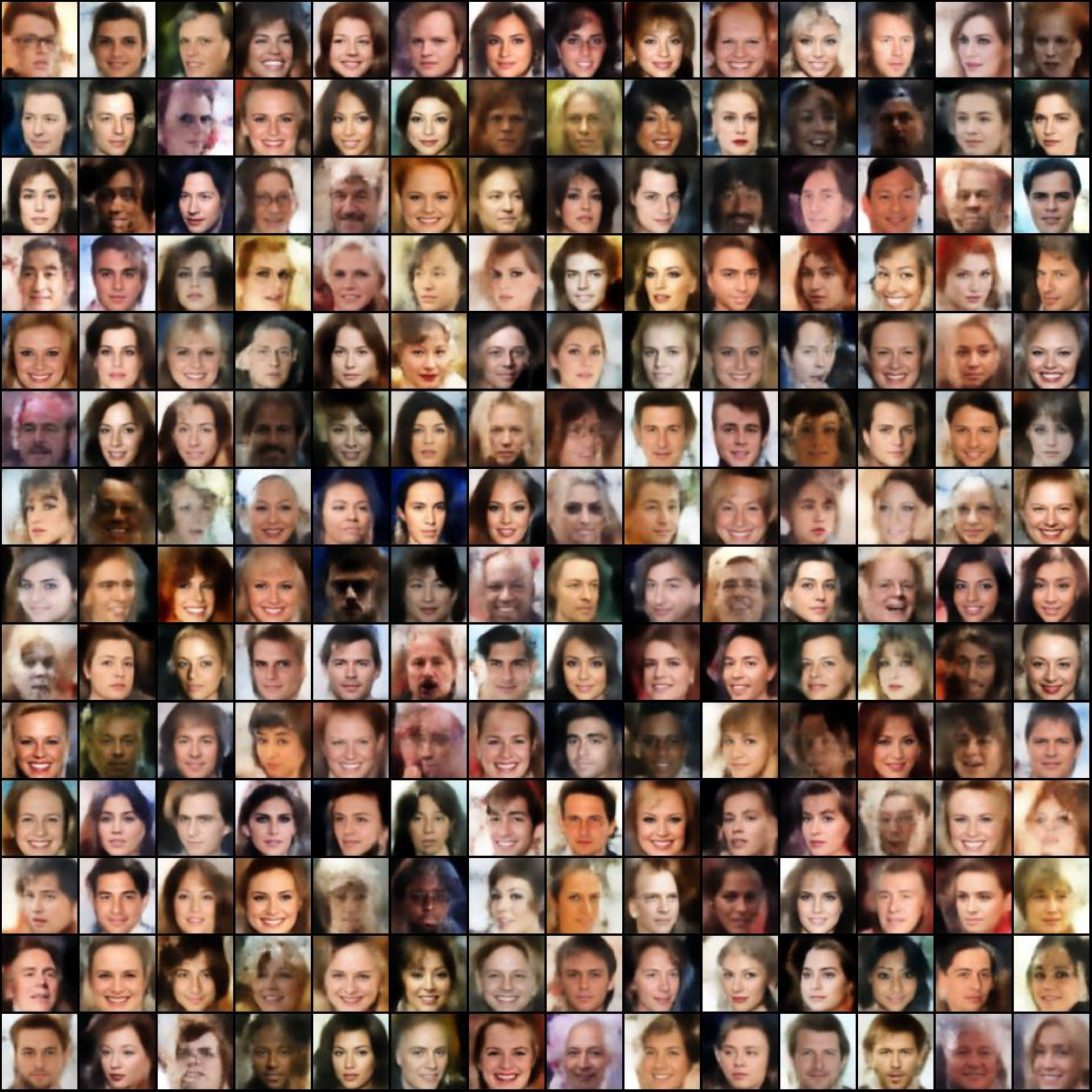}  
    \\
ps-DRAE &ms-DRAE ($k=10$)  &ms-DRAE ($k=50$) 
  \end{tabular}
      
\end{center}
  \caption{
  \footnotesize{ CelebA generated images.
    }}
  \label{fig:CelebAgen}
  \vspace{-0.2 em}
\end{figure}

Now, we move to the generated images with CelebA datasets in Figure~\ref{fig:CelebAgen}. We see that GMVAE and PRAE are not able to generate acceptable images. These two probabilistic autoencoders seem unstable to train on CelebA dataset. On the other hand, images obtained from DRAEs, i.e., DRAE, m-DRAE, and s-DRAE, are significantly better than other autoencoders. Nevertheless, just by only looking at the generated images, it is hard to know which autoencoder among these three autoencoders can produce better images. In order to evaluate their performances, we use FID scores and present them in Table~\ref{tab:FIDtable}. According to those  tables, s-DRAE is better than DRAE, m-DRAE and previous autoencoders on CelebA dataset.

\begin{figure}[!h]
\begin{center}

  \begin{tabular}{l c}
     Data &\includegraphics[scale=0.33]{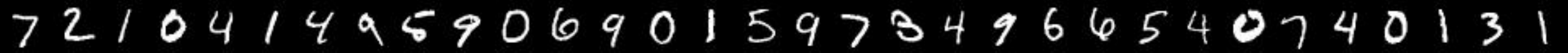}
     \\
     VAE  &\includegraphics[scale=0.33]{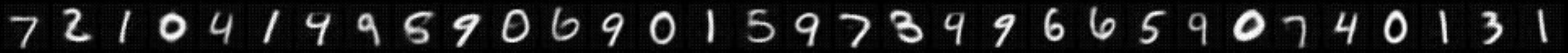}
     \\
     Vampprior &\includegraphics[scale=0.33]{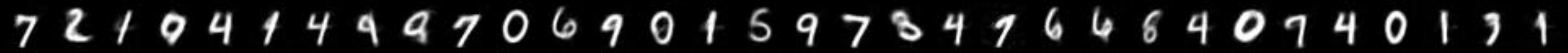}
     \\
    GMVAE &\includegraphics[scale=0.33]{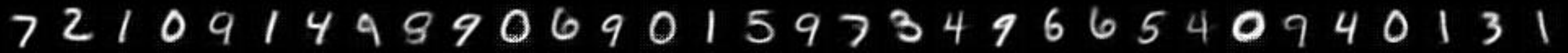}
     \\
     PRAE &\includegraphics[scale=0.33]{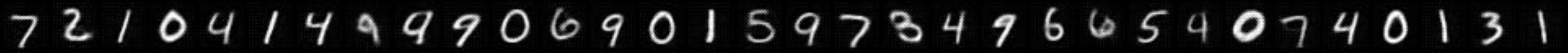}
     \\
     SWAE &\includegraphics[scale=0.33]{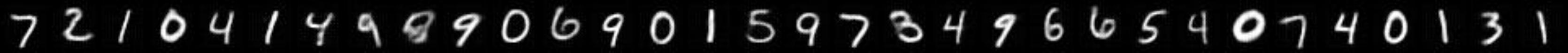}
     \\
     WAE  &\includegraphics[scale=0.33]{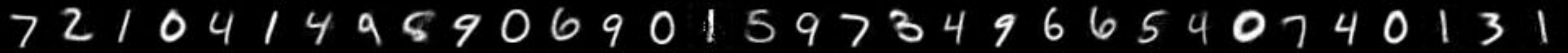}
     \\
     DRAE   &\includegraphics[scale=0.33]{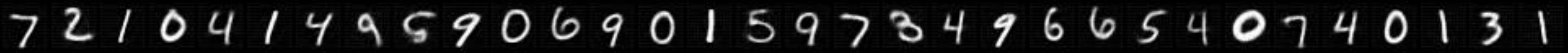}
     \\
     m-DRAE  &\includegraphics[scale=0.33]{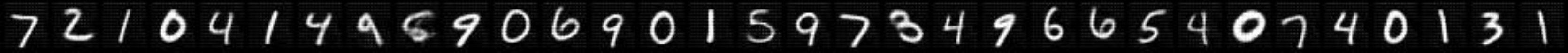}
     \\
     s-DRAE&\includegraphics[scale=0.33]{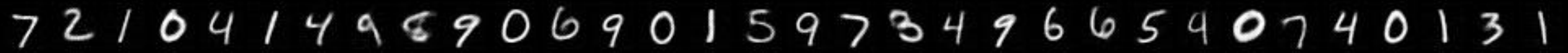}
      \\
     ps-DRAE   &\includegraphics[scale=0.33]{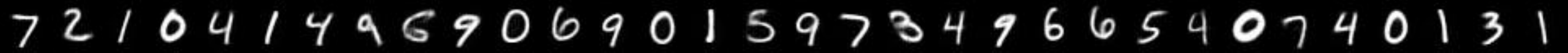}
     \\
     ms-DRAE (k=10)    &\includegraphics[scale=0.33]{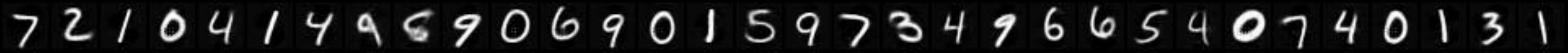}
     \\
     ms-DRAE (k=50)   &\includegraphics[scale=0.33]{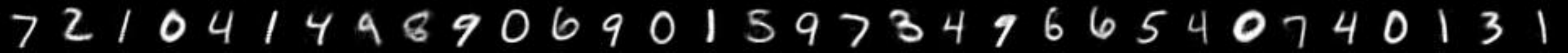}

  \end{tabular}
  \end{center}
  \caption{Reconstruction images on MNIST test set.
    }
  \label{fig:MNISTrecontruction}
\end{figure}

\begin{figure}[!h]
\begin{center}
  \begin{tabular}{l c}
     Data &\includegraphics[scale=0.15]{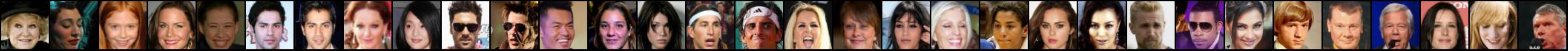}
     \\
     GMVAE  &\includegraphics[scale=0.15]{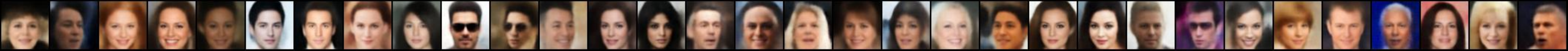}
     \\
     PRAE &\includegraphics[scale=0.15]{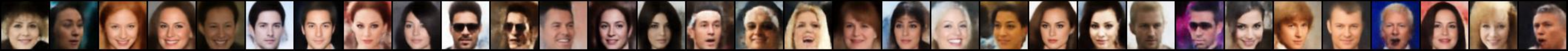}
     \\
    SWAE &\includegraphics[scale=0.15]{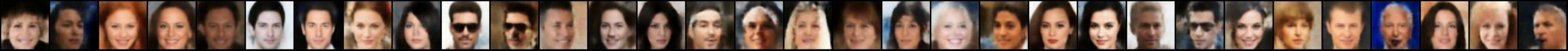}
     \\
     DRAE  &\includegraphics[scale=0.15]{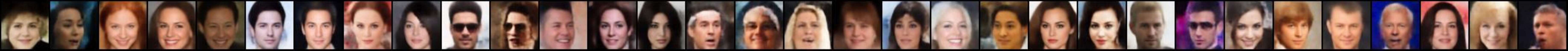}
     \\
     m-DRAE&\includegraphics[scale=0.15]{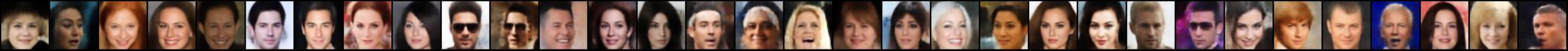}
     \\
     s-DRAE   &\includegraphics[scale=0.15]{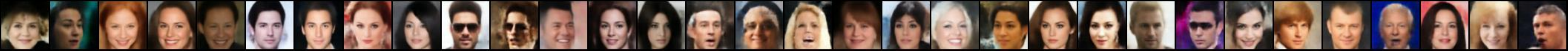}
     \\
     ps-DRAE  &\includegraphics[scale=0.15]{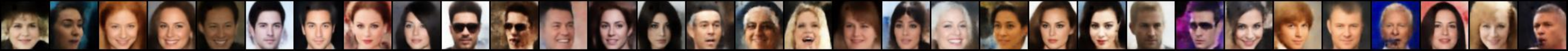}
     \\
     ms-DRAE (k=10)&\includegraphics[scale=0.15]{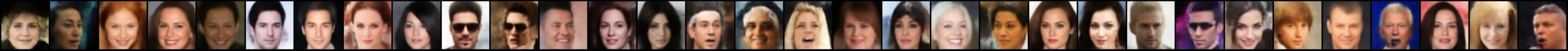}
     \\
     ms-DRAE (k=50)   &\includegraphics[scale=0.15]{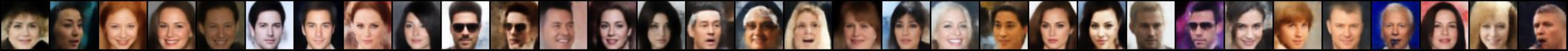}

  \end{tabular}
  \end{center}
  \caption{Reconstruction images on CelebA test set.
    }
  \label{fig:CelebArecontruction}
\end{figure}

\textbf{Reconstruction images: }Regarding the reconstruction ability, we show the reconstructed test-set images on MNIST dataset in Figure~\ref{fig:MNISTrecontruction}, and the reconstructed images test-set images on CelebA dataset in Figure~\ref{fig:CelebArecontruction}. These results show that relational regularizations in DRAE, m-DRAE, s-DRAE do not harm the reconstruction ability of the autoencoders and the reconstruction images from these models have at least the same quality as other considered autoencoders.

\begin{figure}[!h]
\begin{center}
  \begin{tabular}{cccc}
 \includegraphics[scale=0.3]{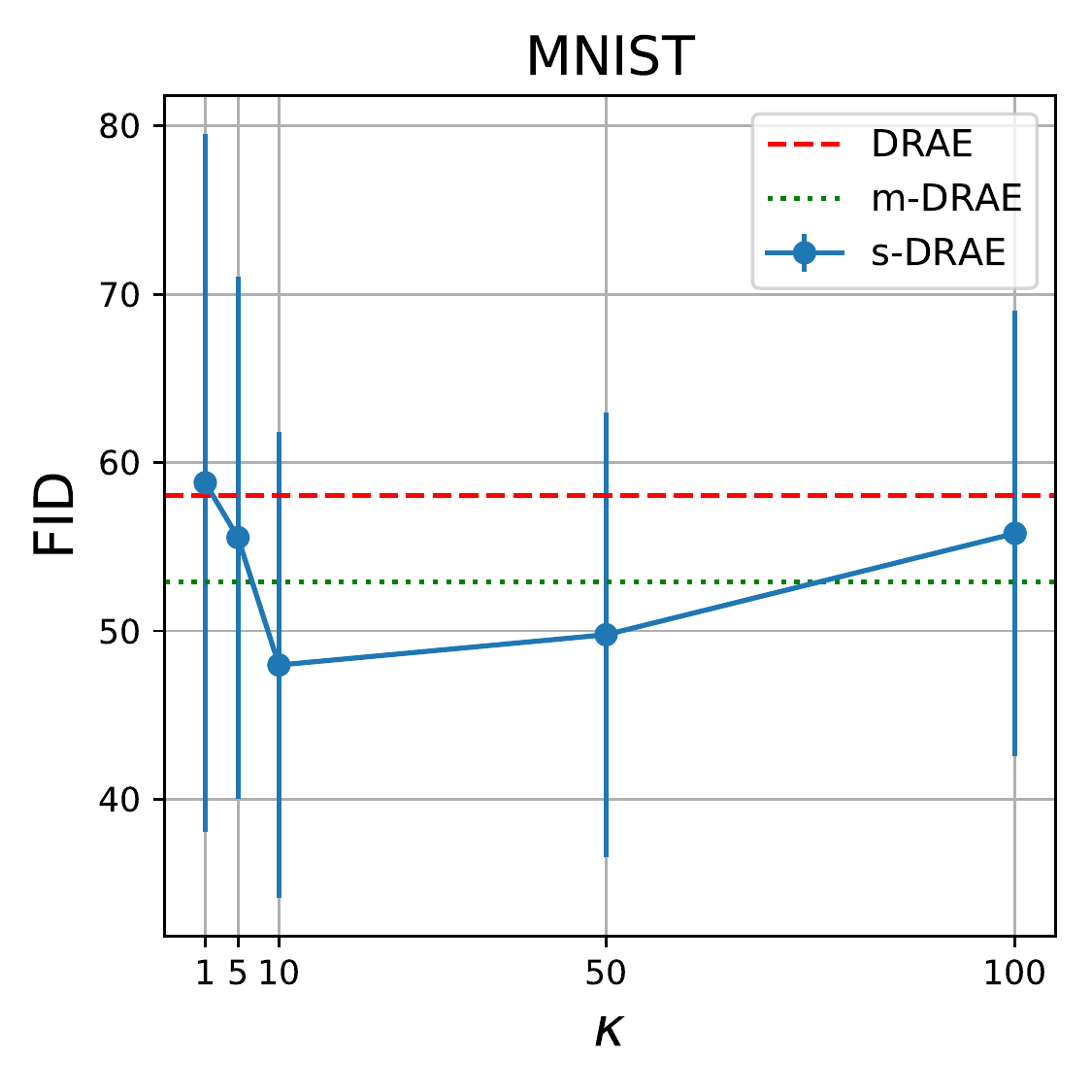} 
    
    &\includegraphics[scale=0.3]{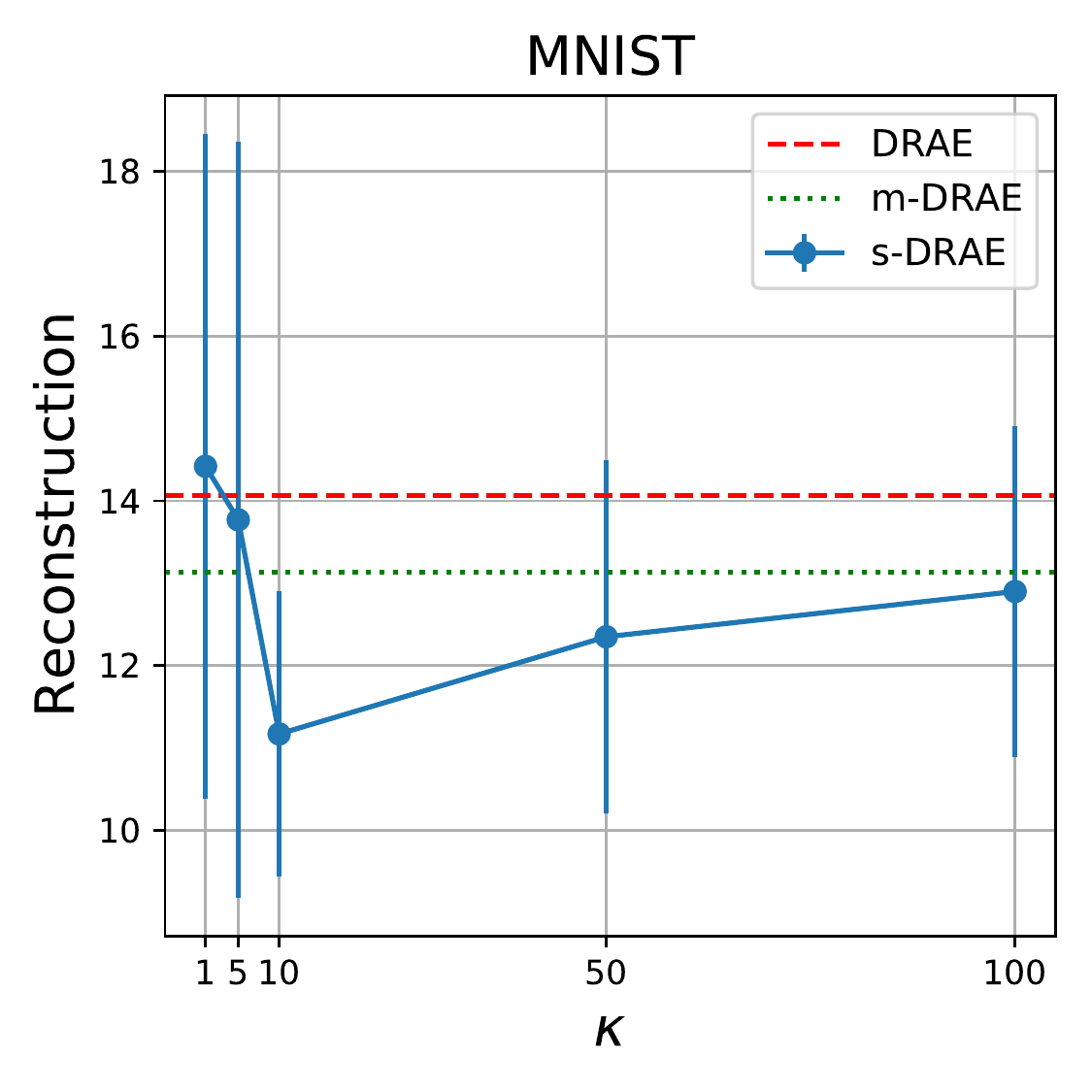} 
     &
    \includegraphics[scale=0.3]{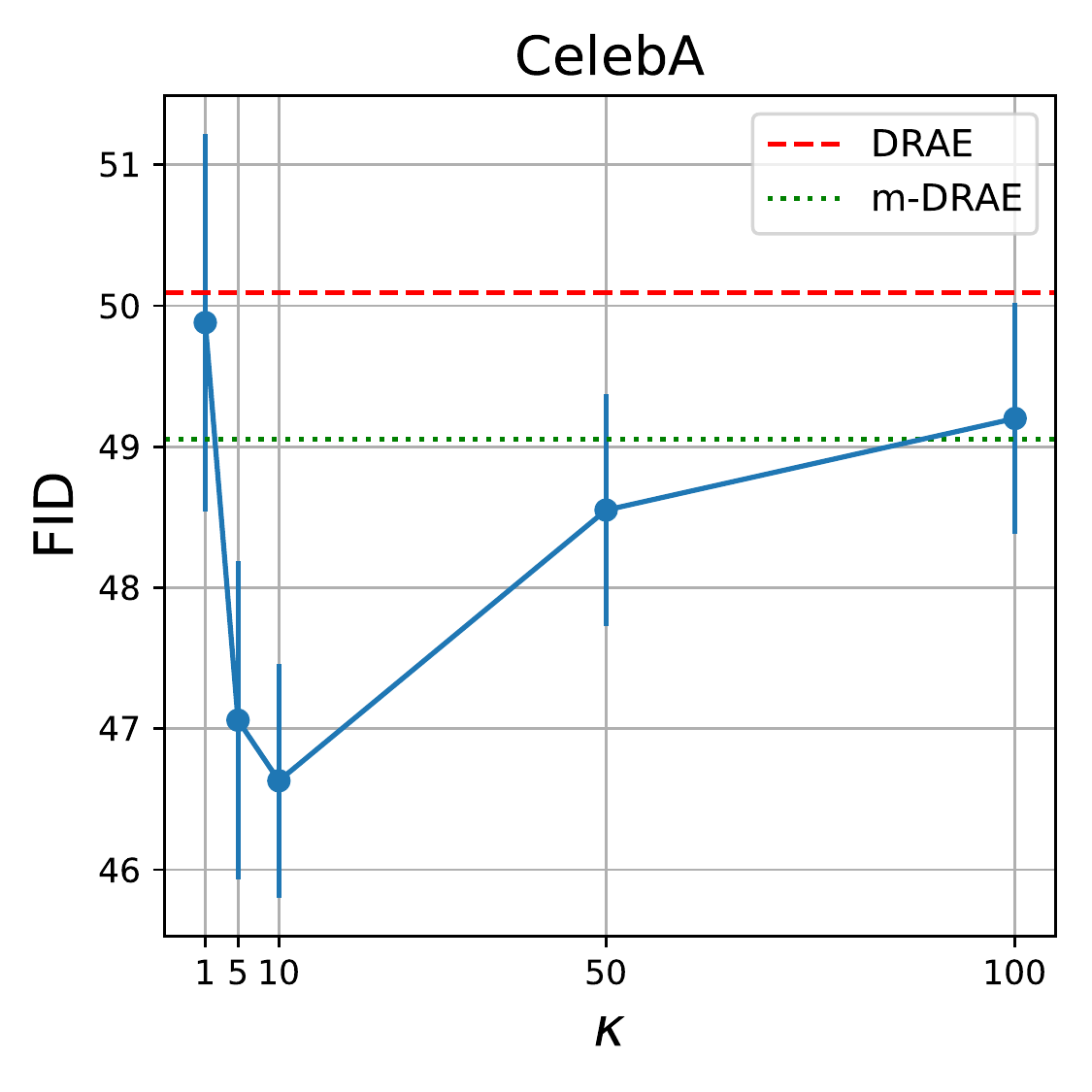} 
    &
    \includegraphics[scale=0.3]{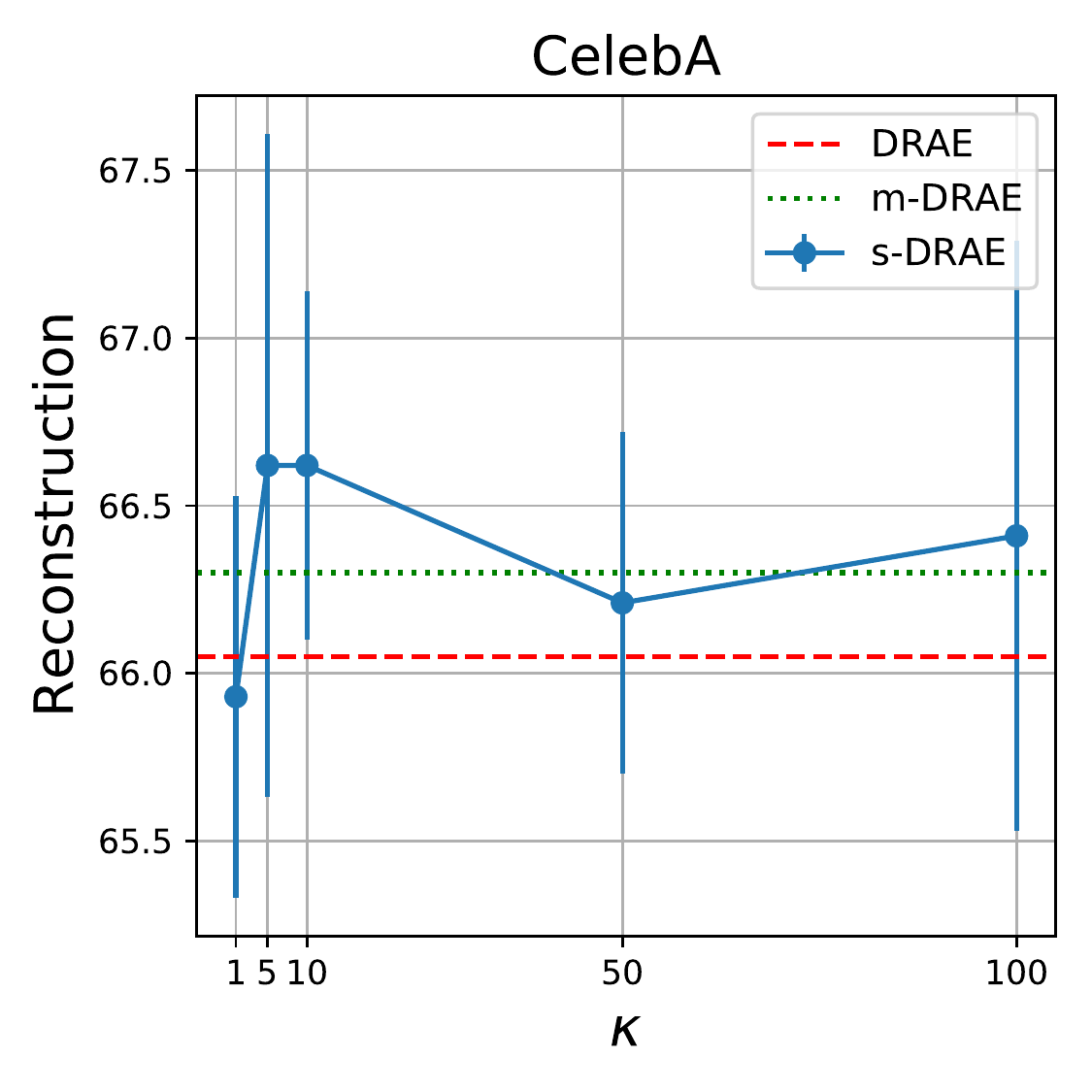} 
     
  \end{tabular}
  \end{center}
  \caption{
  \footnotesize{ The performance of s-DRAE ($L=50$) on MNIST and CelebA datasets when changing the value of $\kappa \in \{1,5,10,50,100 \}$ .
    }}
  \label{fig:kappa}
  \vspace{-0.5 em}
\end{figure}

\textbf{Sensitivity to the concentration parameter: } Here, we investigate the effect of $\kappa$ in training s-DRAE. We set the number of slices equal to $50$, and  $\kappa \in \{1,5,10,50,100\}$ and report reconstruction loss and FID score in Figure~\ref{fig:kappa}.  We find that the performance of s-DRAE is close to that of DRAE when $\kappa = 1$, namely, the reconstruction loss and FID score is nearly equal to the scores of DRAE . On the other extreme, when $\kappa = 100$, s-DRAE behaves like m-DRAE in both evaluation metrics. It confirms our claim that SSFG is a generalization of m-DRAE and DRAE.  About the best choice of $\kappa$, on MINIST, the best value of  $\kappa$ is 10 for both evaluation metrics. In detail, FID score and reconstruction loss decrease considerably when setting $\kappa$ from 1 to 10, then increase when $\kappa > 10$.  Meanwhile, on CelebA, the best value of $\kappa$ for reconstruction loss and FID score are different. With $\kappa = 10$, s-DRAE reaches the best FID score of about 46.6, while $\kappa = 1$ produces the lowest reconstruction loss. This partly explains why our s-DRAE gets the best FID score, not the best reconstruction loss among all tested autoencoders.  


    

\subsection{Results on mixture spherical deterministic relational autoencoder}
\label{subsec:exp_ms-DRAE}
As mentioned in Section~\ref{sub:extension_SSFG} and Appendix~\ref{Sec:MSFG}, there is an extension of SSFG that uses mixtures of vMF as the slicing distribution over directions and we denote the RAE using this new discrepancy as \emph{ms-DRAE}.  In practice, we use the uniform weight $\alpha_i=\frac{1}{k}$ for the vMF mixture and use the same value of $\kappa$ for every component. For the number of projections, we set $L=50$.

\begin{figure}[!h]
\begin{center}
  \begin{tabular}{cccc}
 \includegraphics[scale=0.3]{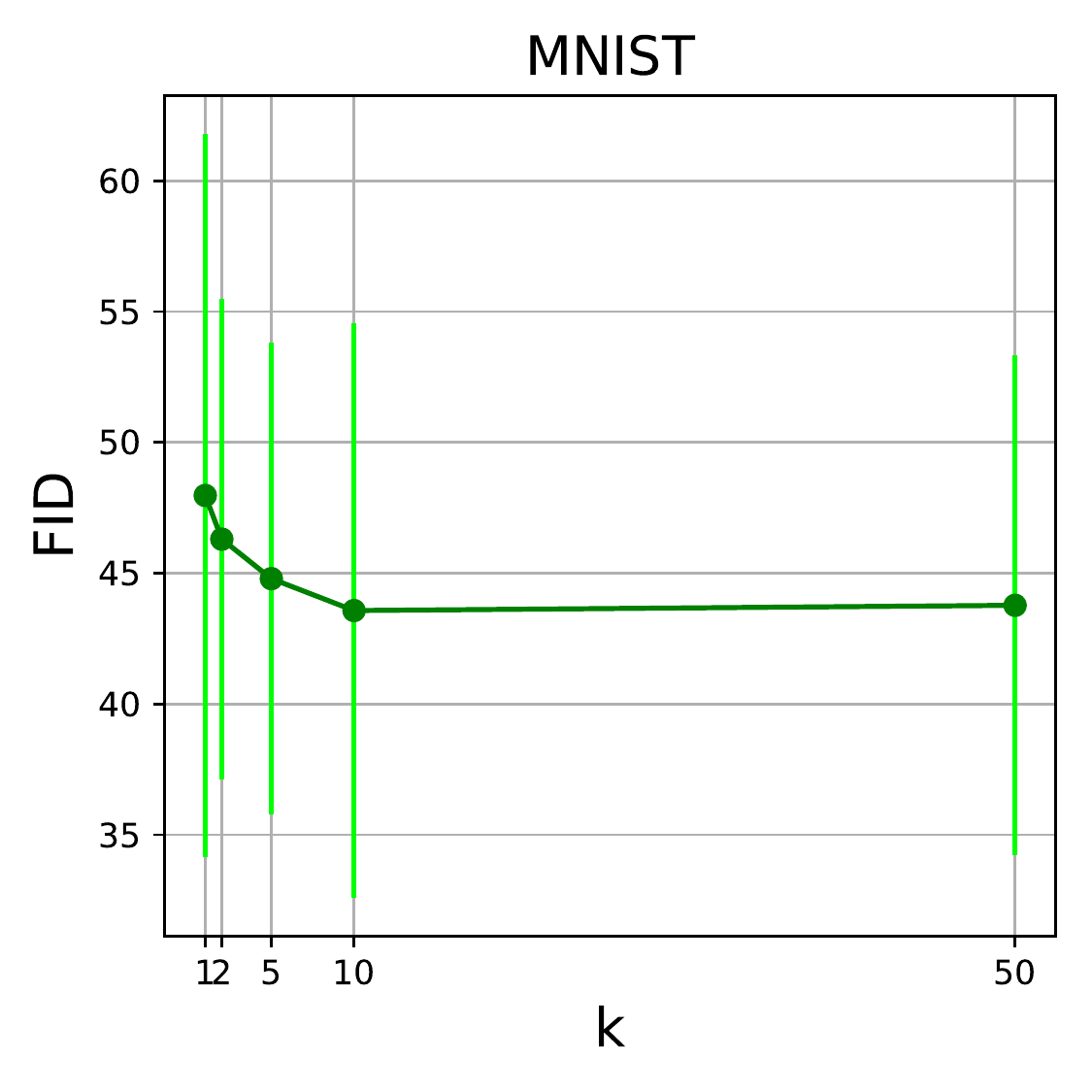} 
    
    &\includegraphics[scale=0.3]{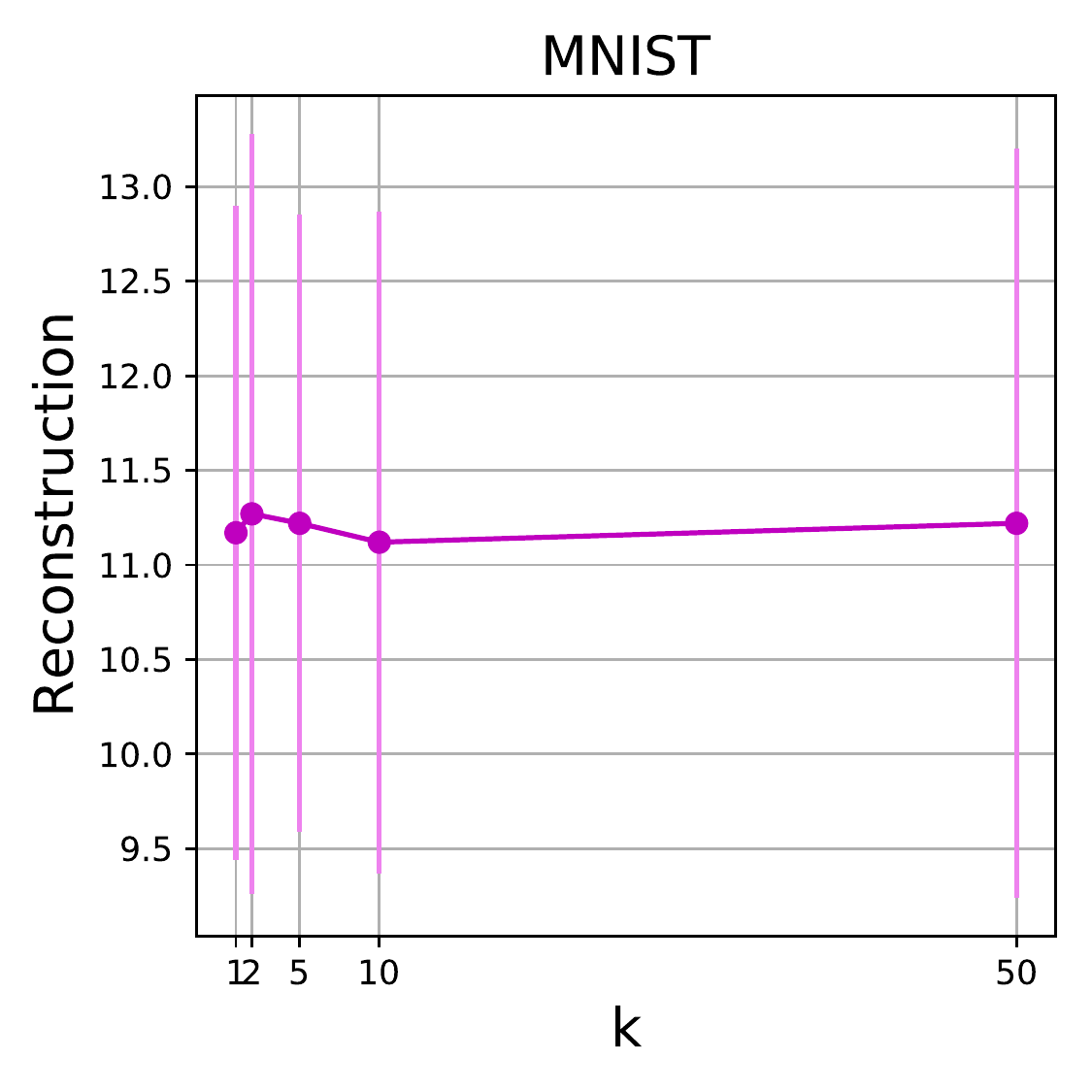} 
     &
    \includegraphics[scale=0.3]{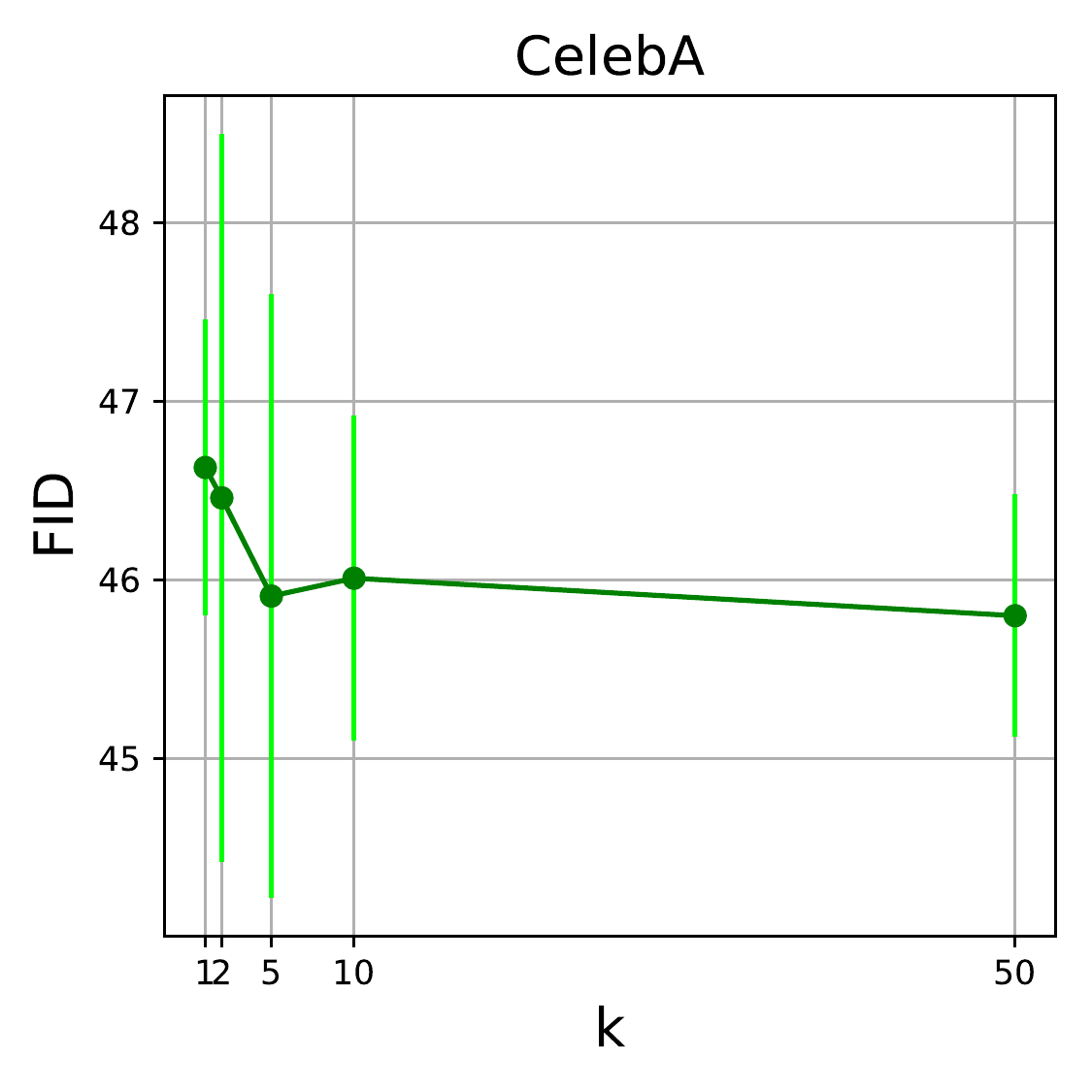} 
    &
    \includegraphics[scale=0.3]{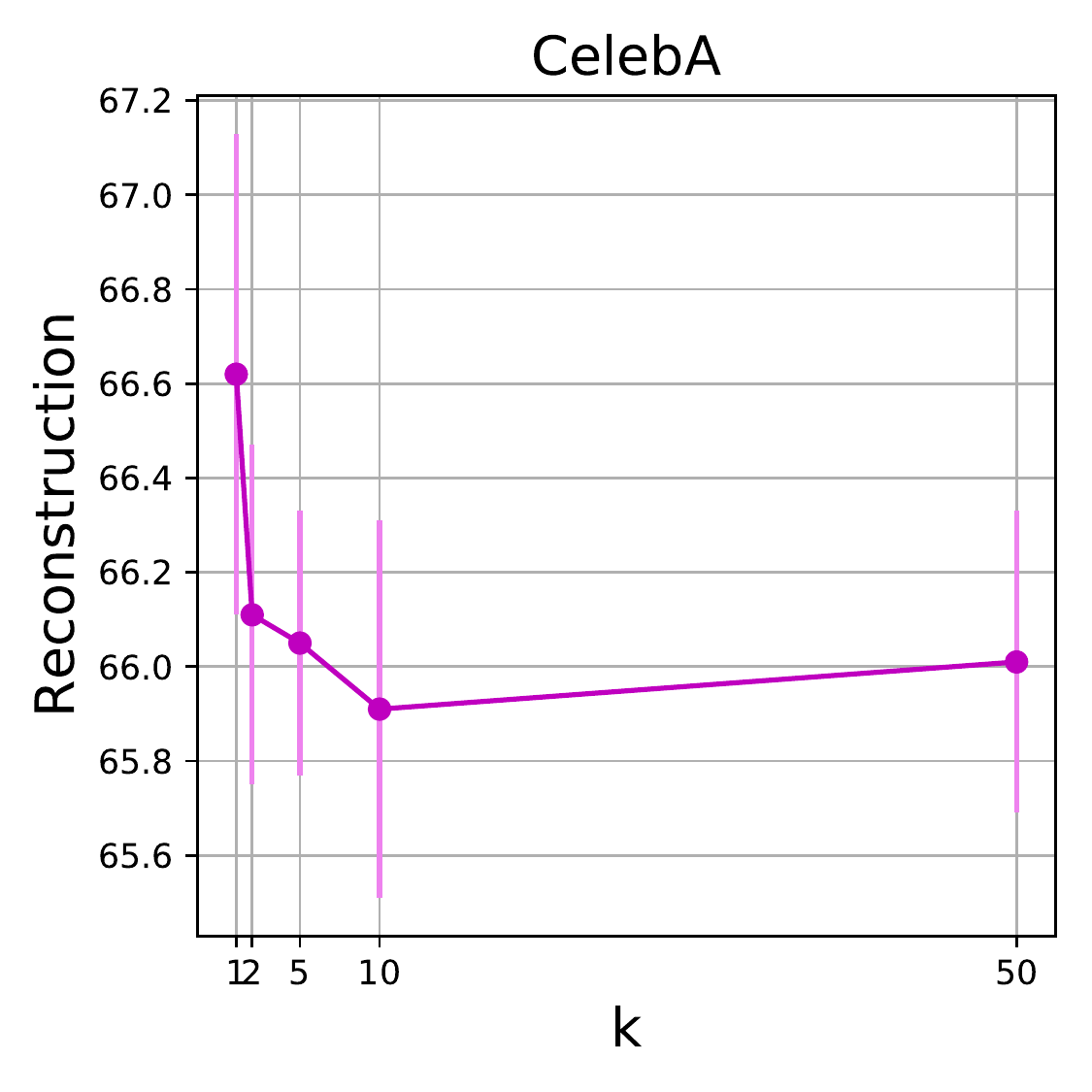} 
     \\
     
  \end{tabular}
  \end{center}
  \caption{
  \footnotesize{ The performance of ms-DRAE  when changing the number of vMF components ($k$).
    }}
  \label{fig:kmixture}
  \vspace{-0.2 em}
\end{figure}

\textbf{Visualization of the latent space: } In Figure ~\ref{fig:TSNe}, with MNIST dataset, the ms-DRAE  performs very well, with $k=10$ and $k=50$, ms-DRAE does not miss any data's mode, the clustering effect is also  very clear. With CelebA dataset in Figure ~\ref{fig:Celebtsne}, like s-DRAE, the latent space visualization of ms-DRAE with $k \in \{10,50\}$ is well covered by the mixtures of Gaussian prior.

\textbf{Synthesis images: } As shown in Figure ~\ref{fig:MNISTgen}, generated MNIST images from ms-DRAE are very realistic and easy to classify. With CelebA dataset, in Figure~\ref{fig:CelebAgen}, ms-DRAE can also produce the highest quality images among considered autoencoder. This quality is  quantified in Table~\ref{tab:FIDtable}, ms-DRAE gets the best FID score among all autoencoders.

\textbf{Reconstruction images: } According to the reconstructed test-set images on MNIST dataset in Figure~\ref{fig:MNISTrecontruction}, and the reconstructed images test-set images on CelebA dataset in Figure~\ref{fig:CelebArecontruction}, ms-DRAE provides reconstructed images that are at least comparable to other autoencoders.

\textbf{Increasing the number of vMF components: }We conduct experiment to see the effect of increasing the number of components, says $k$, and report results in  Figure \ref{fig:kmixture}.  For each value of $k \in \{1,2,5,10,50\}$ we search for its best value of the concentration parameter $\kappa \in \{1,5,10,50,100\}$. The figure shows that more vMF components enhance the quality of the generator on MNIST until $k = 10$, then flats after that. The reconstruction loss on MNIST changes slightly when $k$ increases, but it always between 11.0 and 11.5. Whereas on CelebA dataset, until $k = 5$, both FID and reconstruction loss go down sharply. When $k=10$, the FID score rises a little bit to about 46, while reconstruction loss continues to fall. In contrast, when $k=50$, FID score is reduced to about $45.8$ while reconstruction loss climbs to about 66. Overall, increasing the number of vMF components can affect positively the performance of the learned autoencoder.
\subsection{Results on power spherical deterministic relational autoencoder}
\label{subsec:exp_ps-DRAE}
In this appendix, we provide additional experiments with power spherical deterministic relational autoencoder (ps-DRAE).
\begin{figure}[!h]
 \begin{center}
  \begin{tabular}{cccc}
 \includegraphics[scale=0.3]{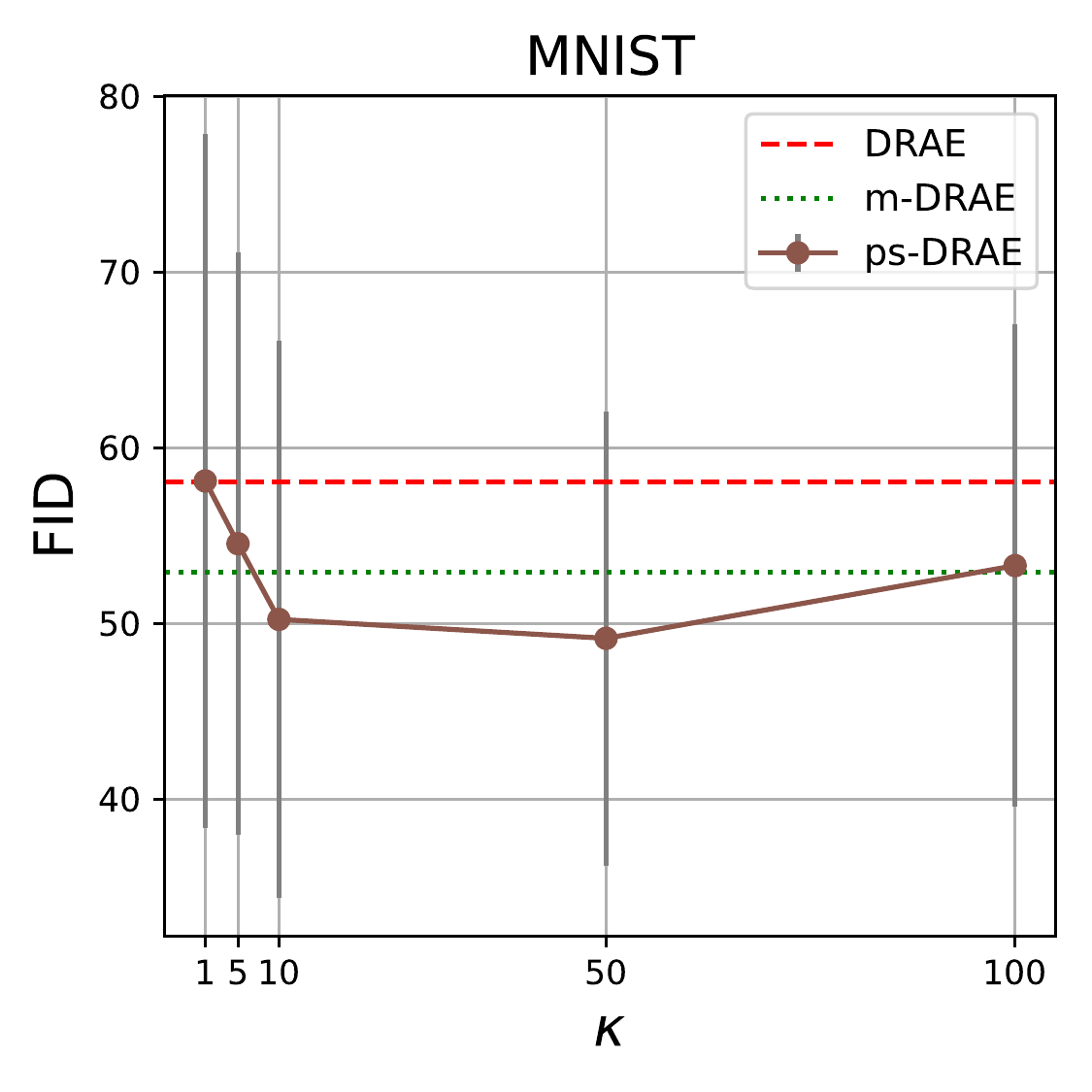} 
    
    &\includegraphics[scale=0.3]{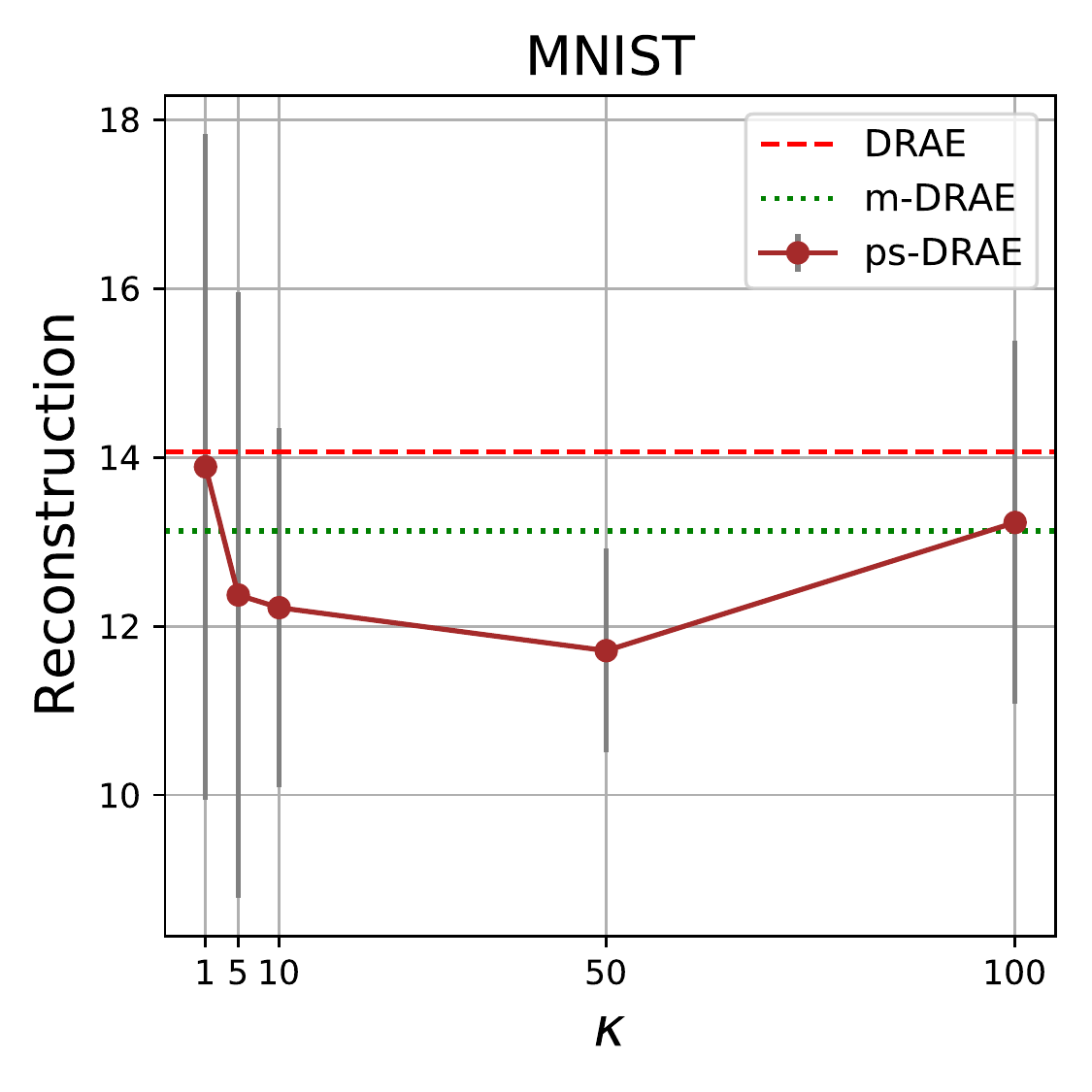} 
     &
    \includegraphics[scale=0.3]{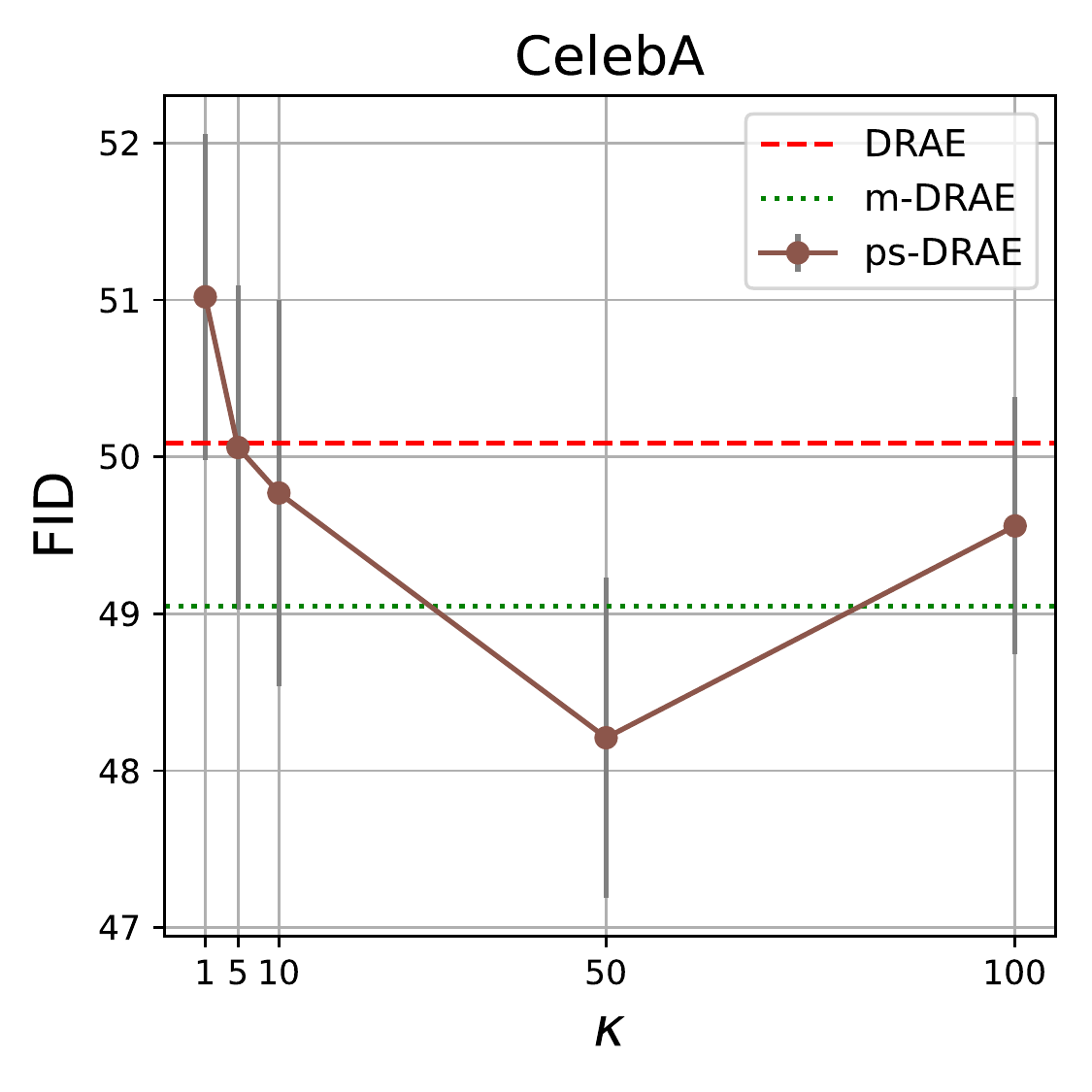} 
    &
    \includegraphics[scale=0.3]{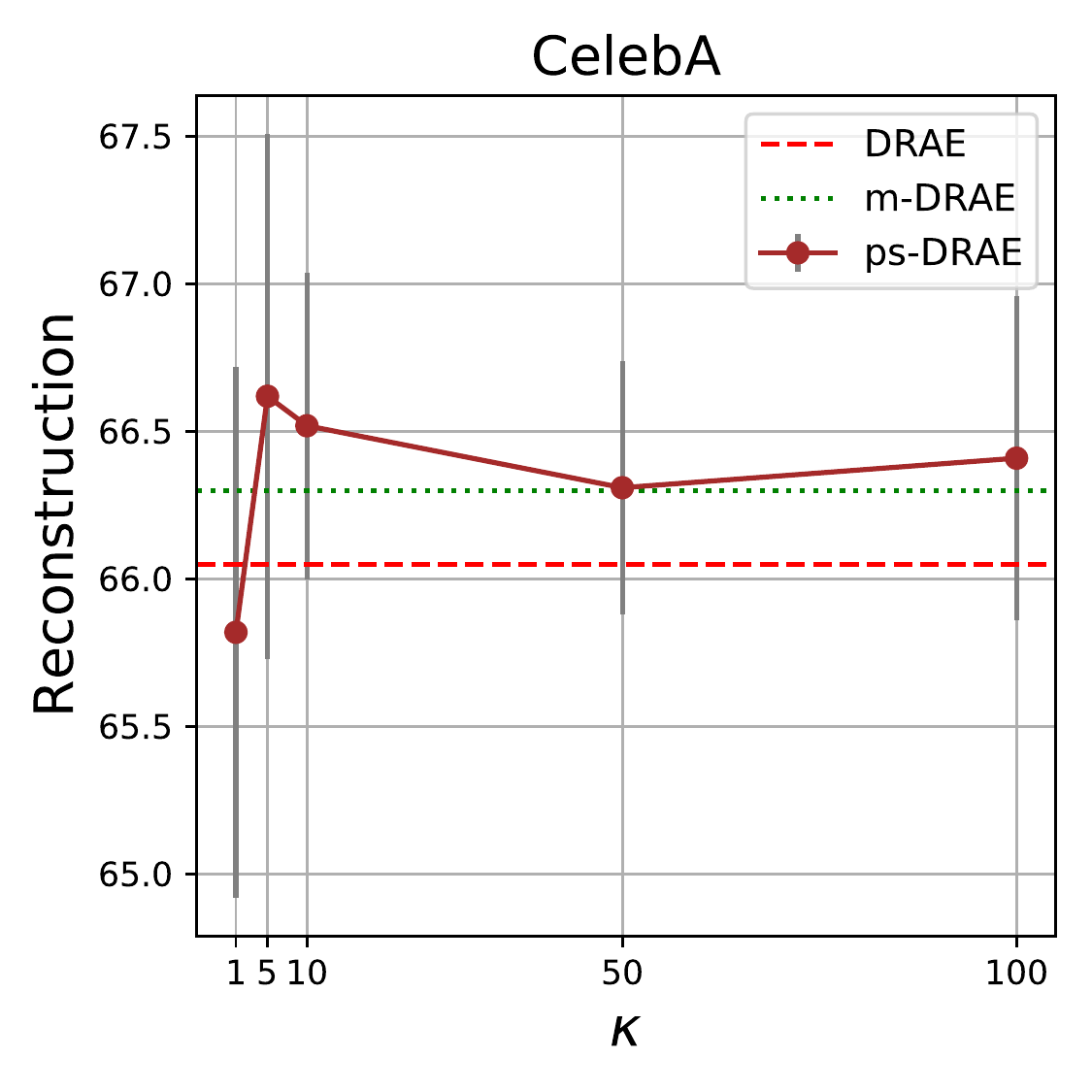} 
     
  \end{tabular}
  \end{center}
  \caption{
  \footnotesize{ The performance of ps-DRAE ($L=50$) on MNIST and CelebA datasets when changing the value of $\kappa \in \{1,5,10,50,100\}$.
    }}
  \label{fig:psdrae_kappa}
  \vspace{-0.5 em}
\end{figure}

\textbf{Visualization of the latent space: } In Figure~\ref{fig:TSNe}, with the MNIST dataset, the ps-DRAE  produces a comparable latent structure to s-DRAE, namely, its mixture Gaussian prior can capture quite well the latent code. With the CelebA dataset in Figure~\ref{fig:Celebtsne}, the latent space visualization of ps-DRAE is also well covered by the prior. 

\textbf{Synthesis images: } As shown in Figure~\ref{fig:MNISTgen}, generated MNIST images from ps-DRAE are comparable to s-DRAE, namely, the images are easy to classify into classes. With the CelebA dataset, in Figure~\ref{fig:CelebAgen}, ps-DRAE can also produce the good images of human faces. 

\textbf{Reconstruction images: } We find that ps-DRAE is also good at reconstructing images as s-DRAE. In Figures~\ref{fig:MNISTrecontruction}, and~\ref{fig:CelebArecontruction}, reconstructed images of ps-DRAE are also similar to the ground truth images.

\textbf{Sensitivity to the concentration parameter $\kappa$}: We find that the ps-DRAE has a similar effect as s-DRAE when we change the value of the concentration parameter $\kappa$. In particular, we plot the FID score and the reconstruction loss of ps-DRAE in Figure~\ref{fig:psdrae_kappa} for each value of $\kappa \in \{1,5,10,50,100\}$. On the MNIST dataset, when $\kappa=1$, both metrics of ps-DRAE are close to the values of DRAE. When $\kappa=100$, ps-DRAE behaves like m-DRAE in both FID score and reconstruction loss. When $\kappa \in \{5, 10, 50\}$, ps-DRAE has better FID score and reconstruction loss than both DRAE and m-DRAE. On the CelebA dataset, we also observe quite similar phenomenon with the FID score as that of the MNIST dataset. The reconstruction losses of ps-DRAE and those of m-DRAE and DRAE are very close regardless of the choice of $\kappa$.
         
        
         
   

\subsection{Ex-post density estimation autoencoders}
\label{subsec:pde}
In order to demonstrate the favorable performance of the SSFG, PSSFG and MSSFG over SFG, we adapt the DRAE framework to the ex-post density estimation procedure and test its generative quality.

Ex-post density estimation is a new procedure for training a generative autoencoder. It was proposed by~\cite{ghosh2019variational} and consists of two main steps. The first step is to train a regularized autoencoder by using the following objective:
\begin{align}
    \min_{\theta,\phi} \mathbb{E}_{p_d(x)} [\norm{x - G_\theta(E_\phi (x))}_2^2 + \lambda_1 \norm{E_\phi (x)}_2^2]+ \lambda_2 \norm{\theta}_2^2,
\end{align}
where $\lambda_1, \lambda_2$ are regularized positive parameters that will be chosen. After we find the optimal parameters $\theta^*$ and $\phi^*$ of the above objective function, the second step is to fit a density estimator $p_\psi(z)$ to the latent distribution $q_E (z) :=  \frac{1}{N} \sum_{i=1}^N \delta_{ E_{\phi^*}(x_i)}$. To generate samples from this model, we will first sample $z \sim p_\psi (z)$ and then get a new $x = G_{\theta^*}(z)$. 

In the experiments with ex-post density estimation, we compare SSFG and other relational discrepancies in the training procedure. More specifically, we learn an autoencoder with $\lambda_2=1$ and $\lambda_1 =0.1$ on the MNIST dataset (and $\lambda_2 = 1$ and $\lambda_1 = 1$ on the CelebA dataset) in the first step with 50 epochs. After this step, the trained autoencoder is shared among all methods. In the next  step, we again choose $p_\psi(z)$ as a mixture of Gaussian distributions and fit it to the latent distribution using relational discrepancies, e.g., SFG, max-SFG, SSFG, PSSFG, and MSSFG with again 50 epochs.

\textbf{Generative quality: }
We compute the FID score on MNIST dataset and CelebA dataset, then present them in Table \ref{tab:postfid}. According to this table, SSFG achieves a better FID score than SFG and max-SFG on this application. Compare to the traditional training of DRAE, the FID score on MNIST is significantly improved, it is much better than the score in Table \ref{tab:FIDtable}. For example, s-DRAE gets 47.97 while s-DRAE with ex-post density estimation training gets 37.42. On CelebA dataset, the new procedure performs worse than the traditional procedure in Table \ref{tab:FIDtable}, however, SSFG still gives the lowest FID score among all distances. Moreover, PSSFG also performs well in this task, it gives a comparable result to SSFG. About MSSFG, with more vMF components, the FID scores of MSSFG in both datasets decrease. So, MSSFG becomes the best choice of discrepancy for this task.

\begin{table}[!h]
\caption{FID table of ex-post density estimation autoencoders}
    \centering
    \begin{tabular}{lll}
    \toprule
         Method &  MNIST  &CelebA \\
         \midrule
         SFG&  41.85 $\pm$ 12.29 & 60.28 $\pm$ 2.56 \\
         max-SFG & 40.69 $\pm$ 5.96 & 60.06 $\pm$ 2.45\\
         SSFG& 37.42 $\pm$ 6.06 & 58.8 $\pm$ 1.97\\
         PSSFG & 38.05 $\pm$ 5.17 & 58.76 $\pm$ 1.88\\
         MSSFG (k=5) & 33.54 $\pm$ 7.12 & 58.21 $\pm$ 1.75 \\
         MSSFG (k=10) & 33.16 $\pm$ 6.96 & 57.48 $\pm$ 1.72\\
         MSSFG (k=50) & \textbf{33.11 $\pm$ 6.99} & \textbf{57.22 $\pm$ 1.7}\\
    \bottomrule
    \end{tabular}
    
    \label{tab:postfid}
\end{table}

\textbf{Visualization of the latent space: } Next, we show the t-SNE visualization of the autoencoder. On MNIST, the mixture of Gaussian prior learned with MSSFG (k=50) can cover  all the modes of the latent code. SSFG and PSSFG performs quite well, they only miss one mode. About SFG and Max-SFG, they both miss two modes of the latent code distribution. On CelebA, there is not too much difference, however, MSSFG seems to learn Gaussian means the best, those can cover almost all the latent space.

\begin{figure}[!h]
\begin{center}
    
  \begin{tabular}{ccc}
 \includegraphics[scale=0.34]{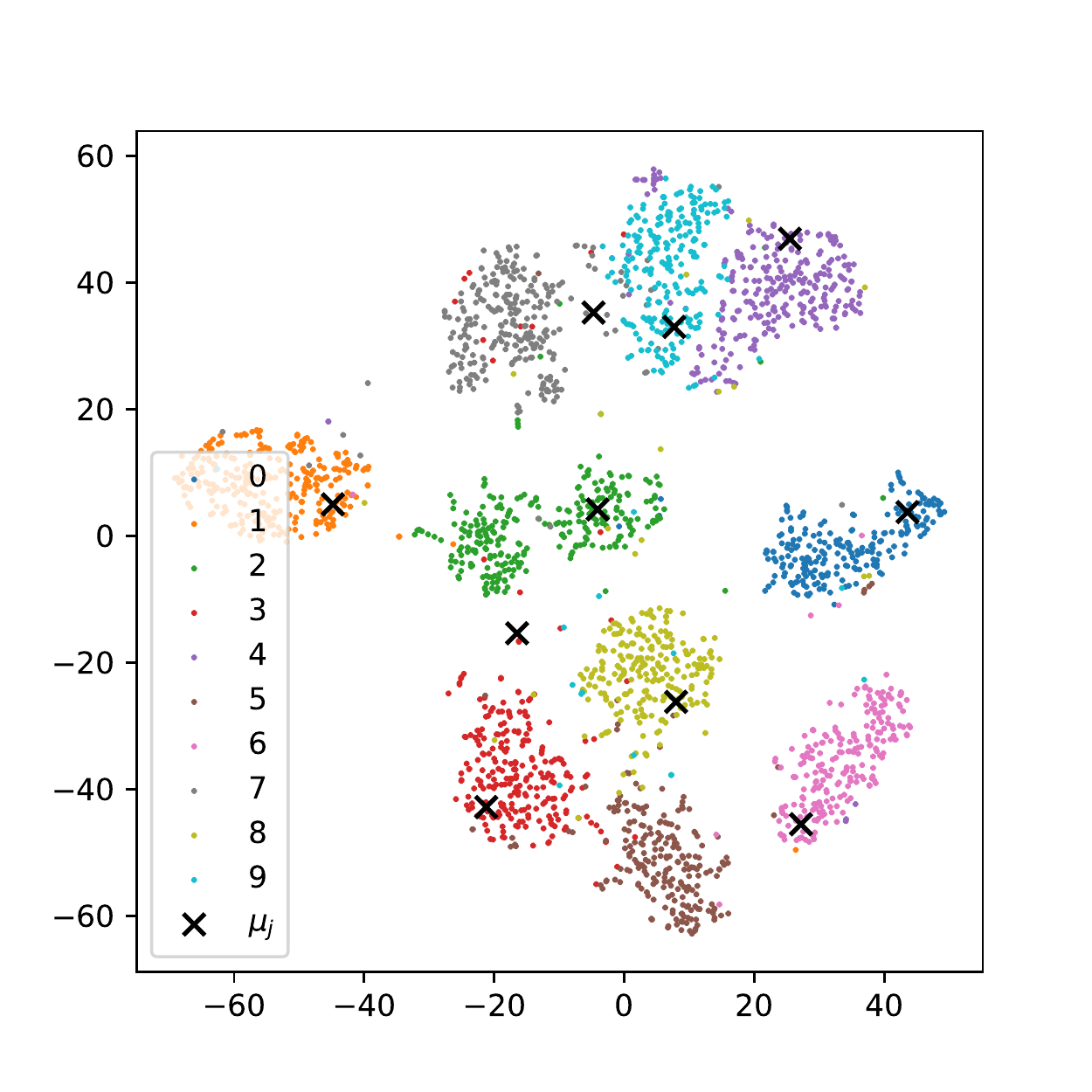} 
    
    &\includegraphics[scale=0.34]{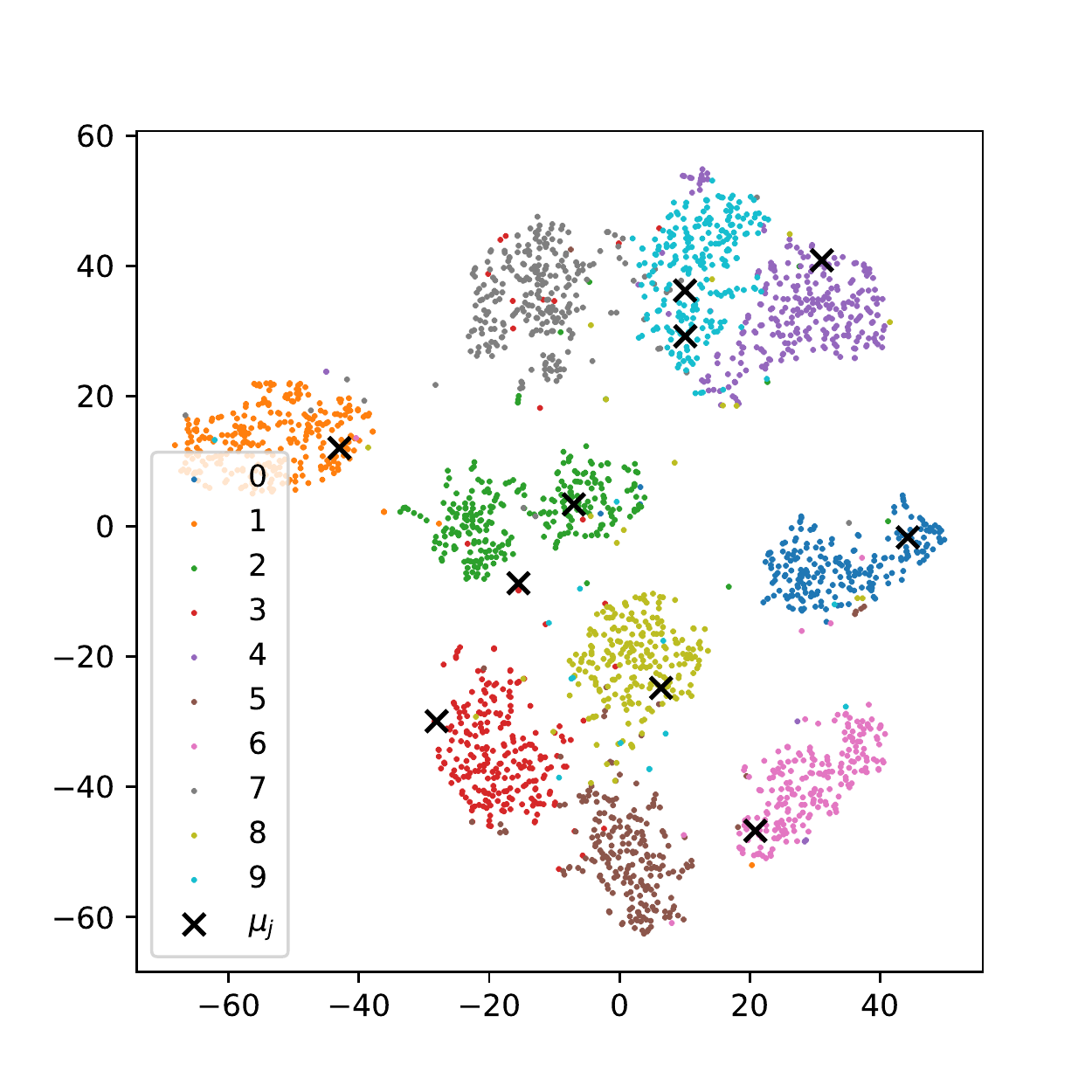} 
     &
    \includegraphics[scale=0.34]{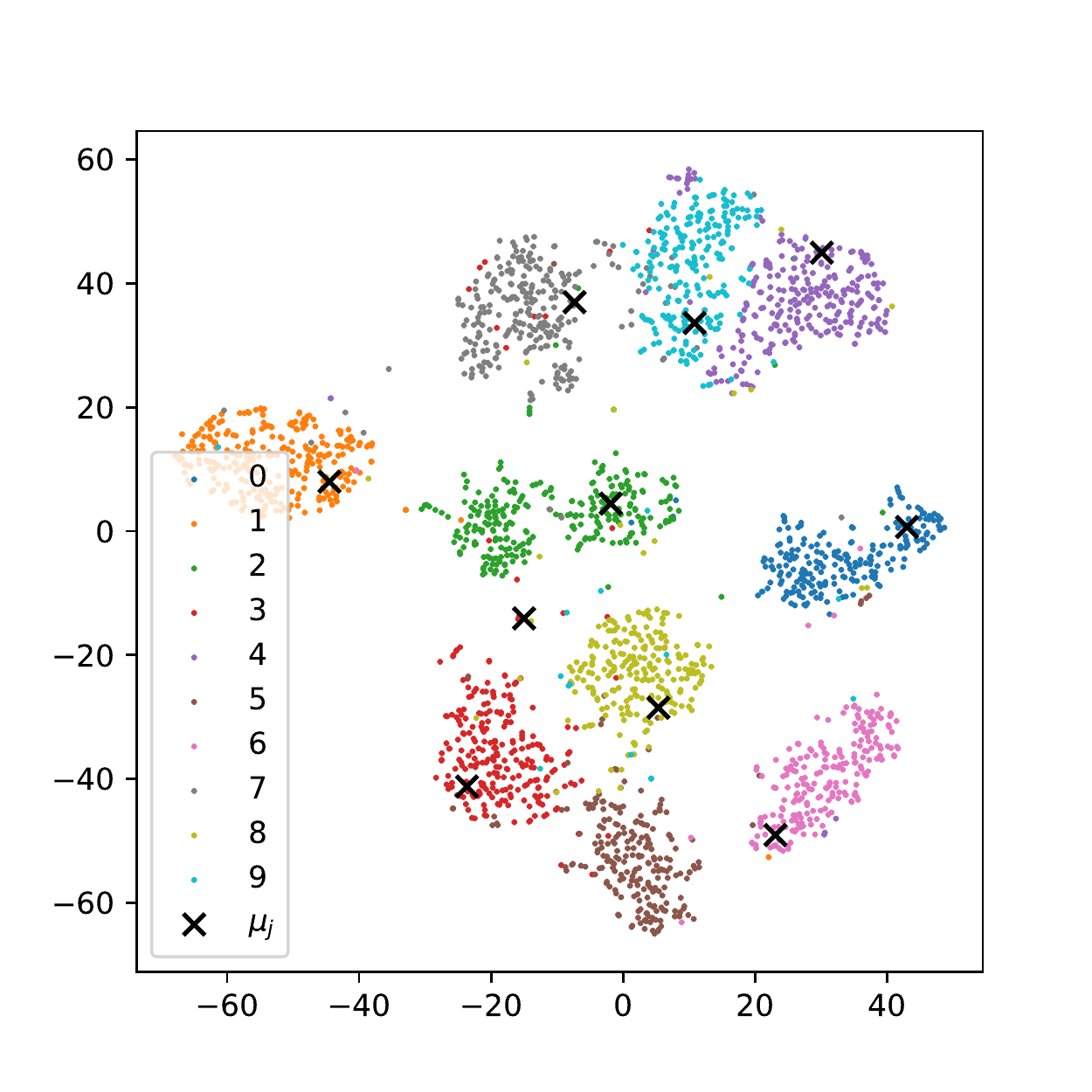} 
  
     \\
     SFG&Max-SFG&SSFG
    \\
    \includegraphics[scale=0.34]{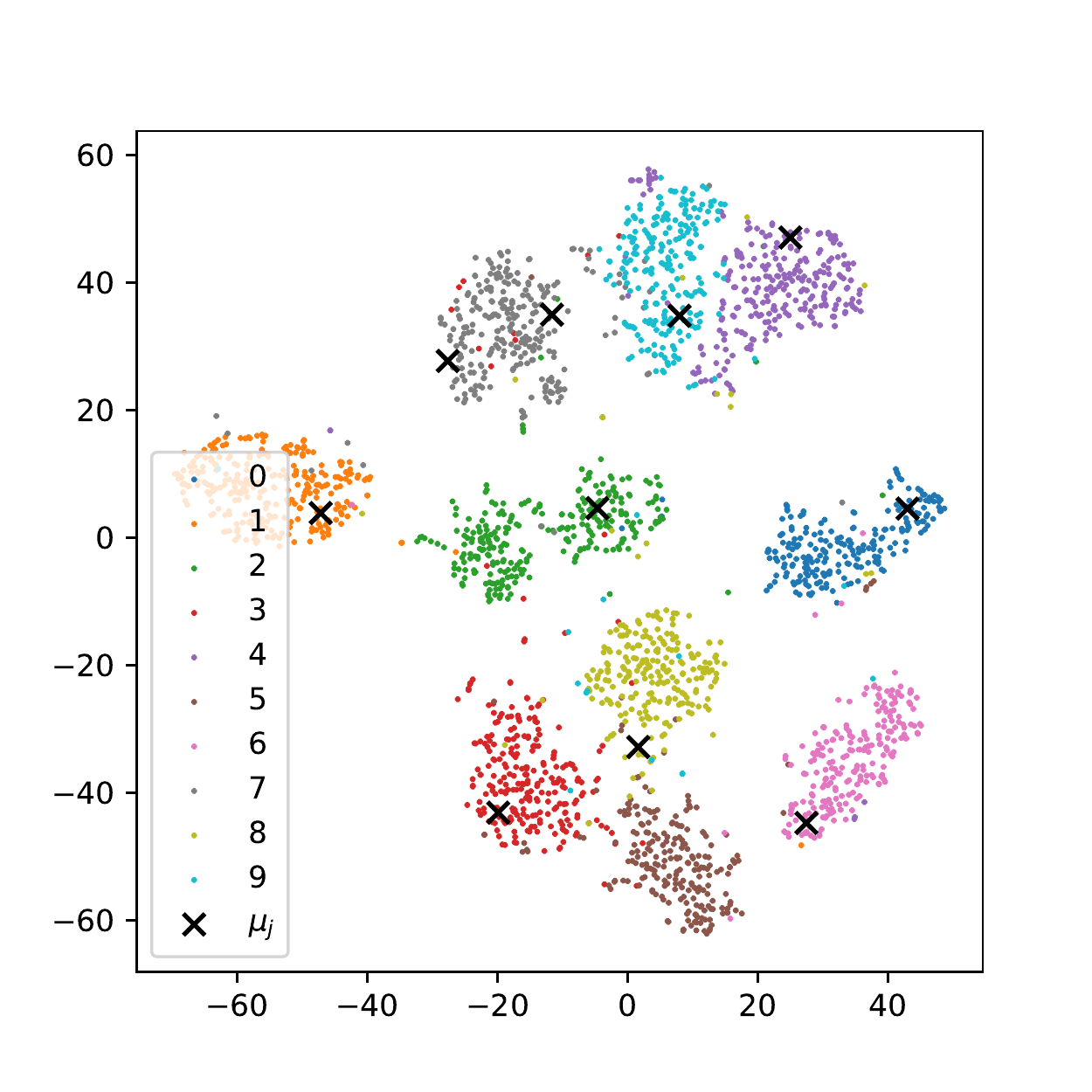} 
    
    &\includegraphics[scale=0.34]{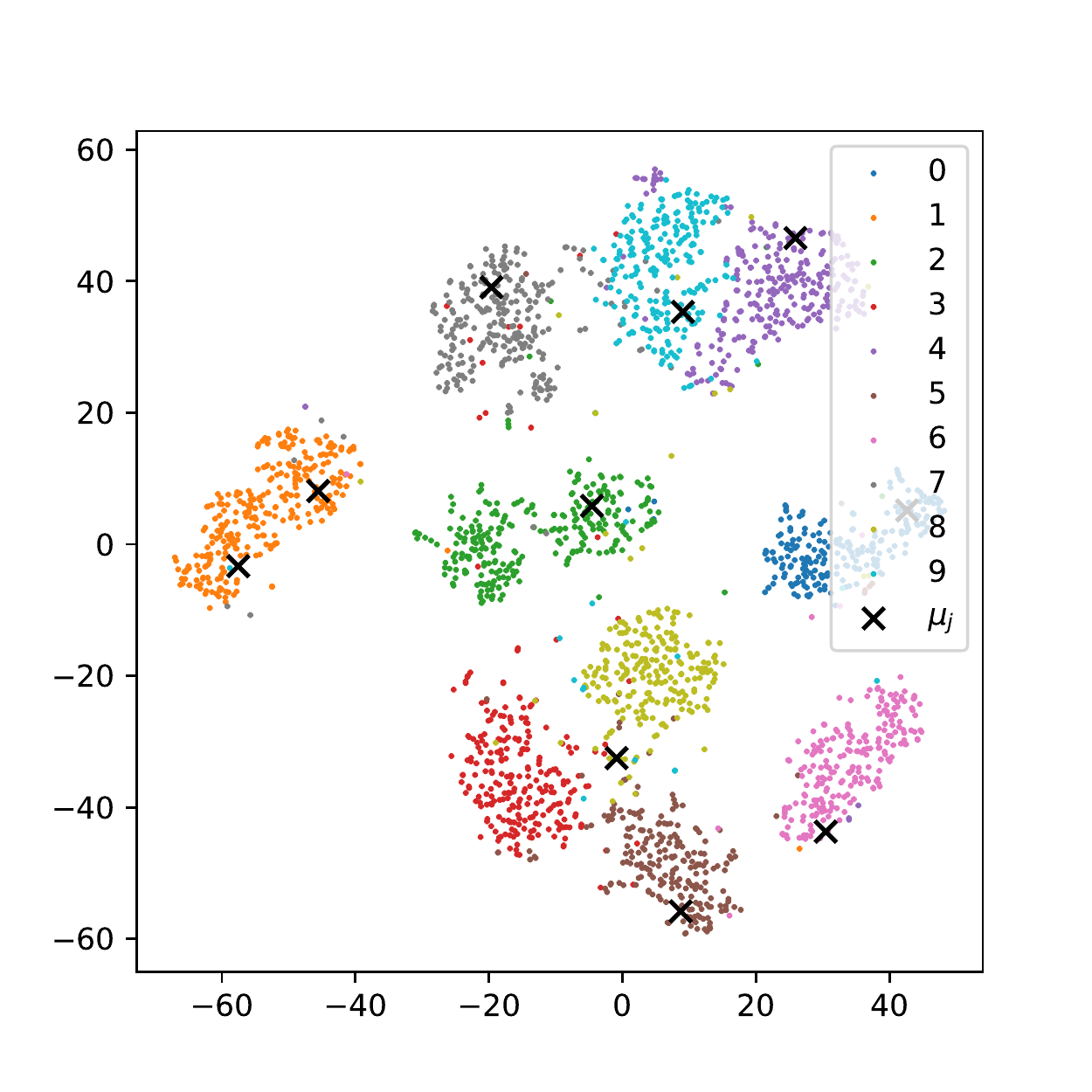} 
     &
    \includegraphics[scale=0.34]{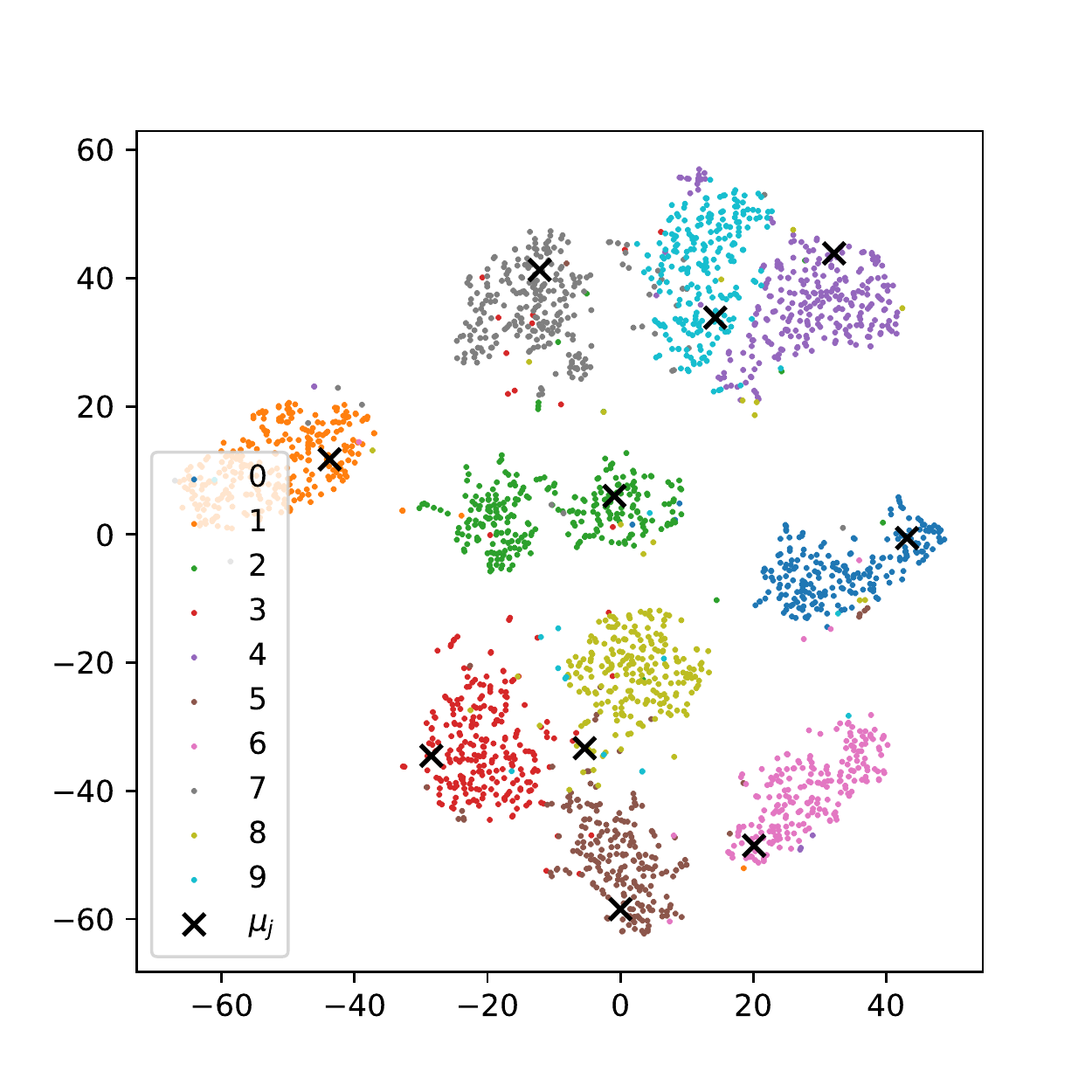} 
  
     \\
     PSSFG &MSSFG ($k=10$)&MSSFG ($k=50$)
  \end{tabular}
  \end{center}
  \caption{
  \footnotesize{  t-SNE on MNIST latent code with ex-post density estimation procedure,   the $\mu_j$ are the means of components in the Gaussian mixture prior.
    }}
  \label{fig:MNISTpdetsne}
  \vspace{-0.2 em}
\end{figure}

\begin{figure}[!h]
\begin{center}

  \begin{tabular}{ccc}
 \includegraphics[scale=0.34]{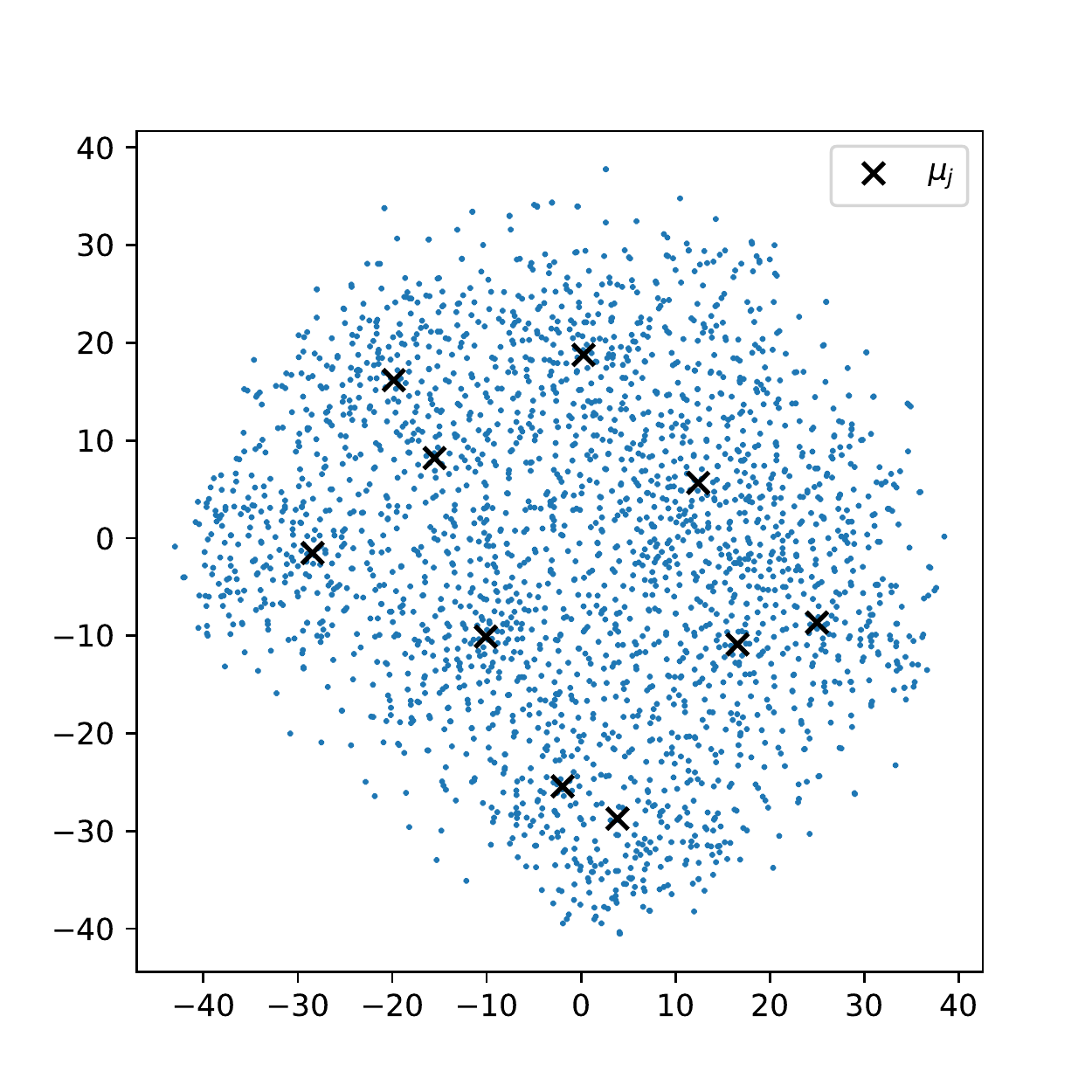} 
    
    &\includegraphics[scale=0.34]{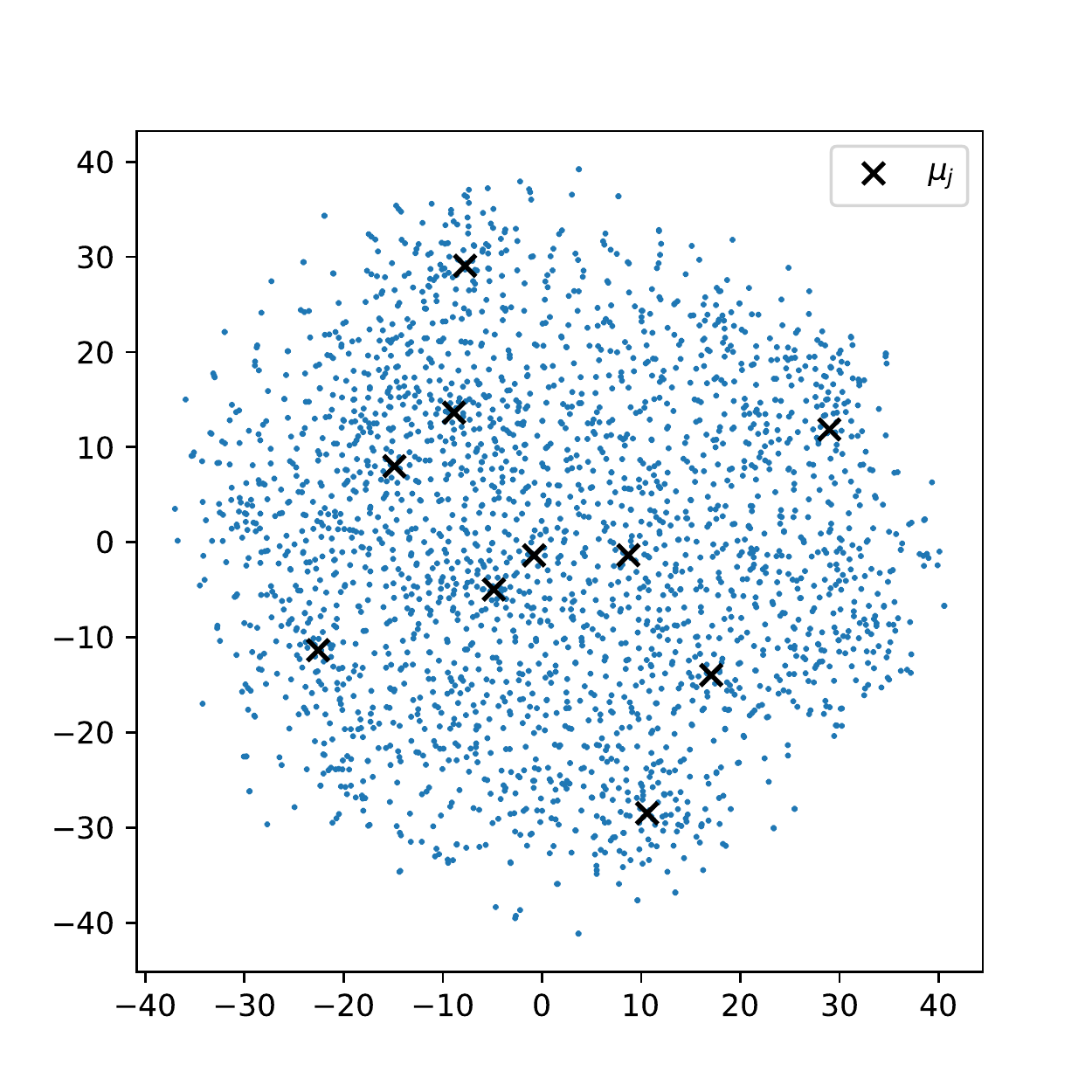} 
     &
    \includegraphics[scale=0.34]{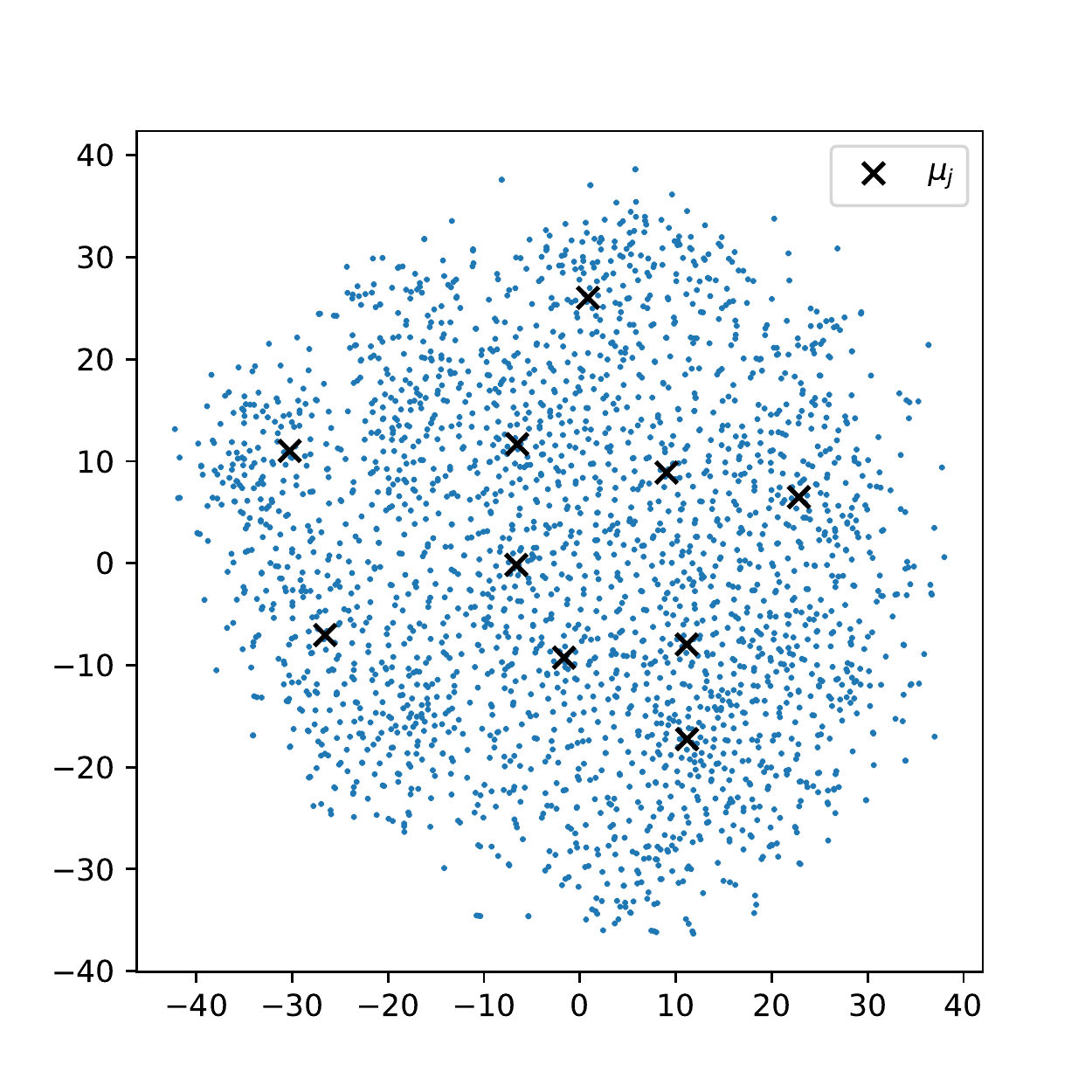} 
  
     \\
     SFG&Max-SFG&SSFG
     \\
      \includegraphics[scale=0.34]{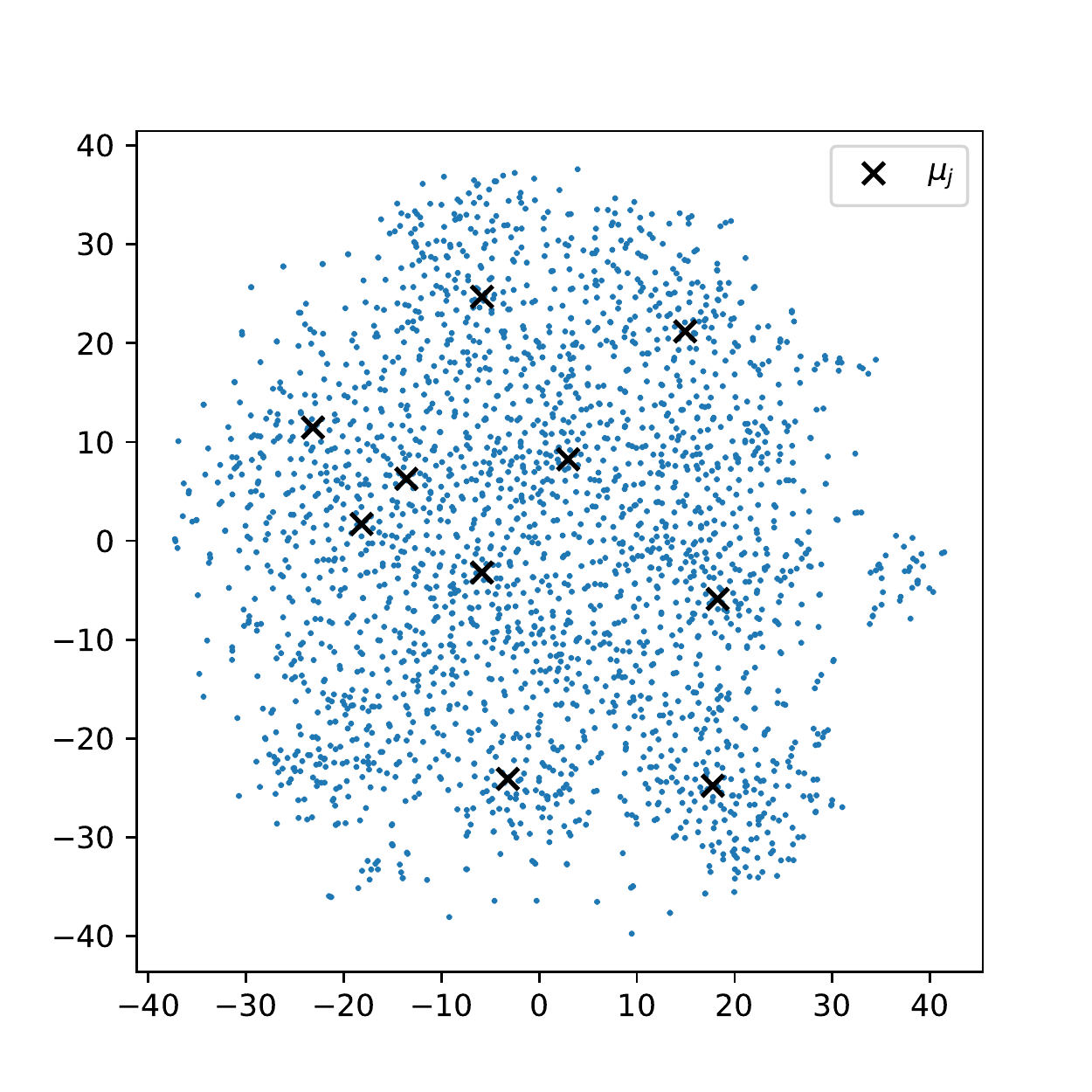} 
    
    &\includegraphics[scale=0.34]{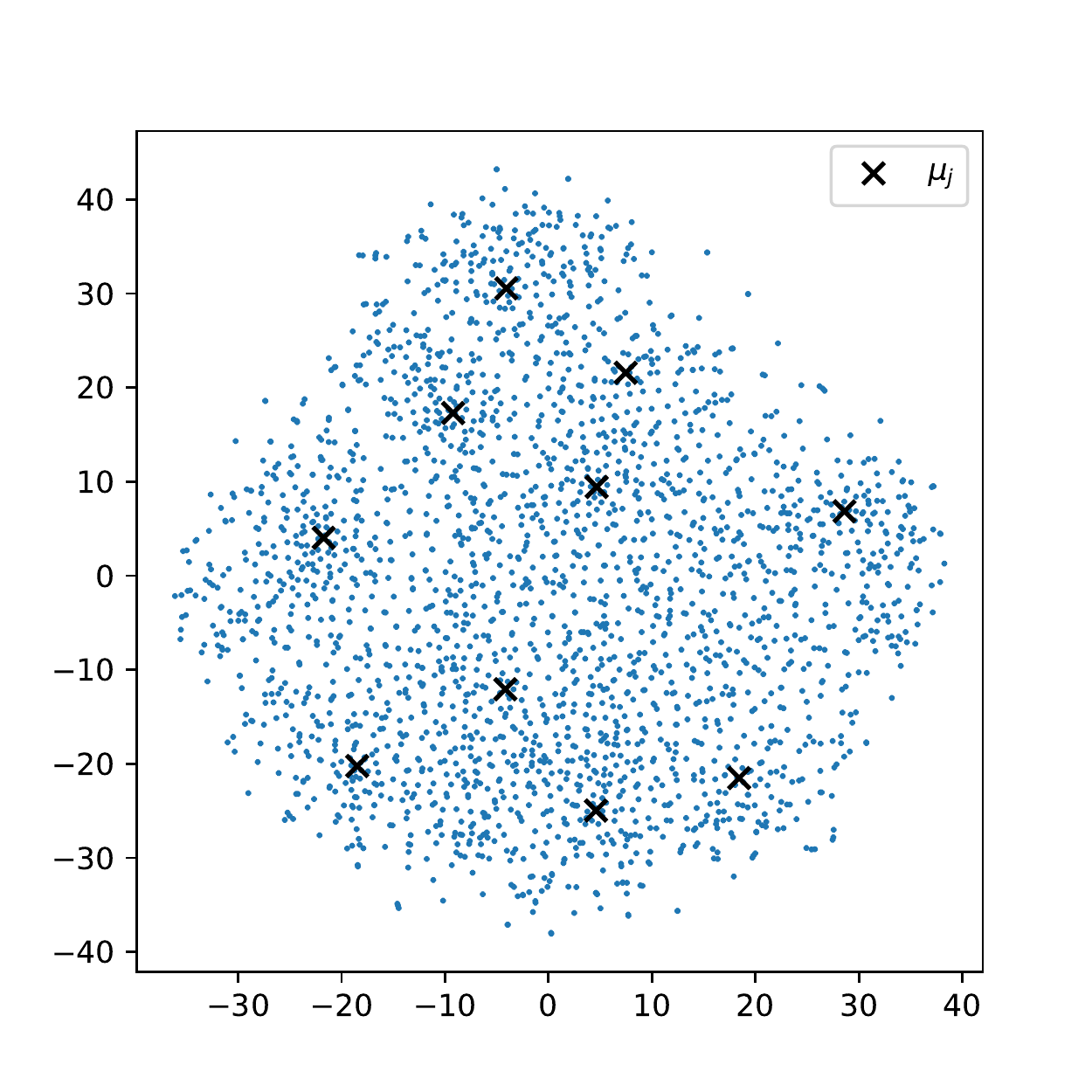} 
     &
    \includegraphics[scale=0.34]{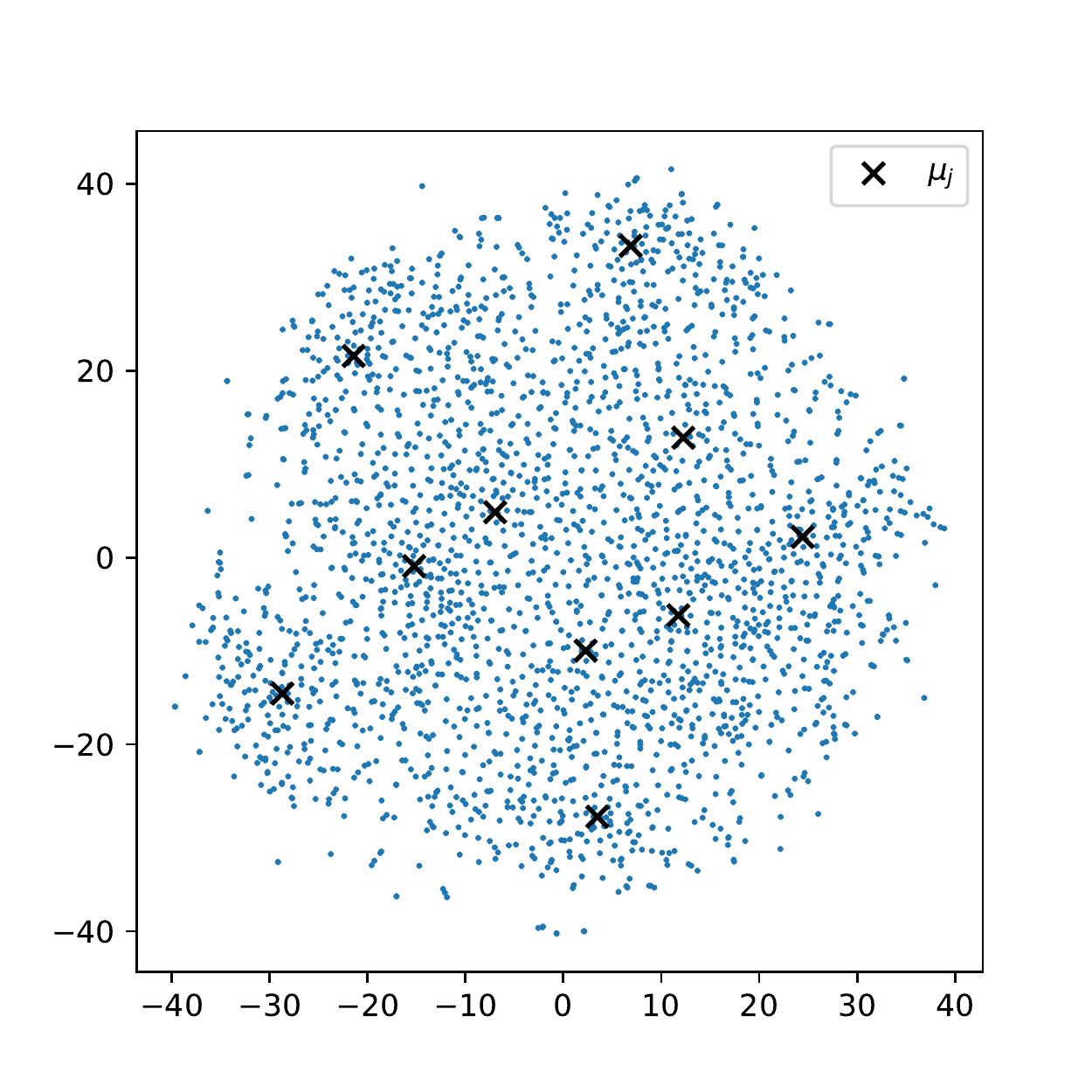} 
  
     \\
      PSSFG &MSSFG ($k=10$)&MSSFG ($k=50$)
    
  \end{tabular}
  \end{center}
  \caption{
  \footnotesize{ t-SNE on CelebA latent code with ex-post density estimation procedure, the $\mu_j$ are the means of components in the Gaussian mixture prior.
    }}
  \label{fig:CelebApdetsne}
  \vspace{-0.2 em}
\end{figure}

\begin{figure}[!h]
\begin{center}
  \begin{tabular}{ccc}
 \includegraphics[scale=0.28]{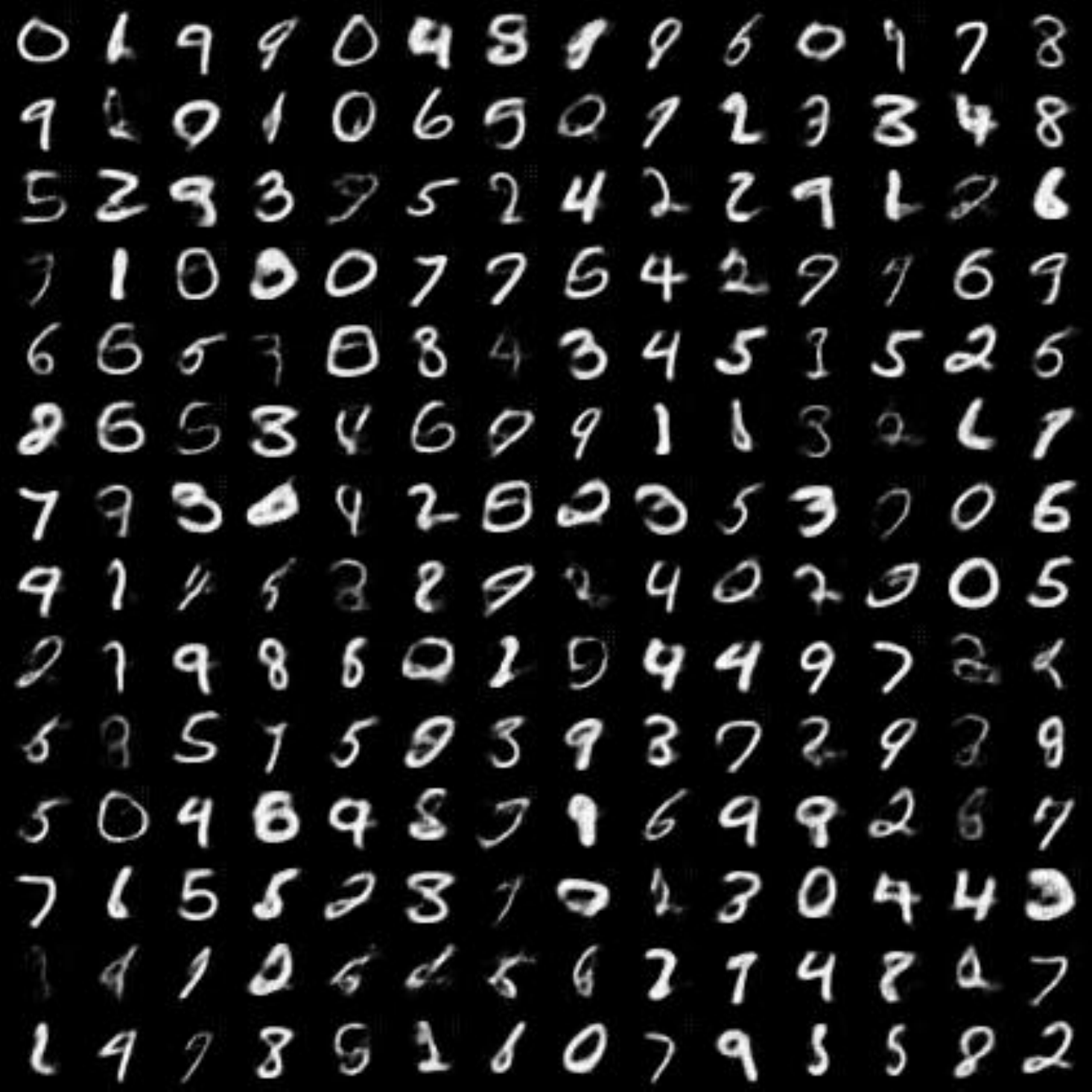} 
    &
    \includegraphics[scale=0.28]{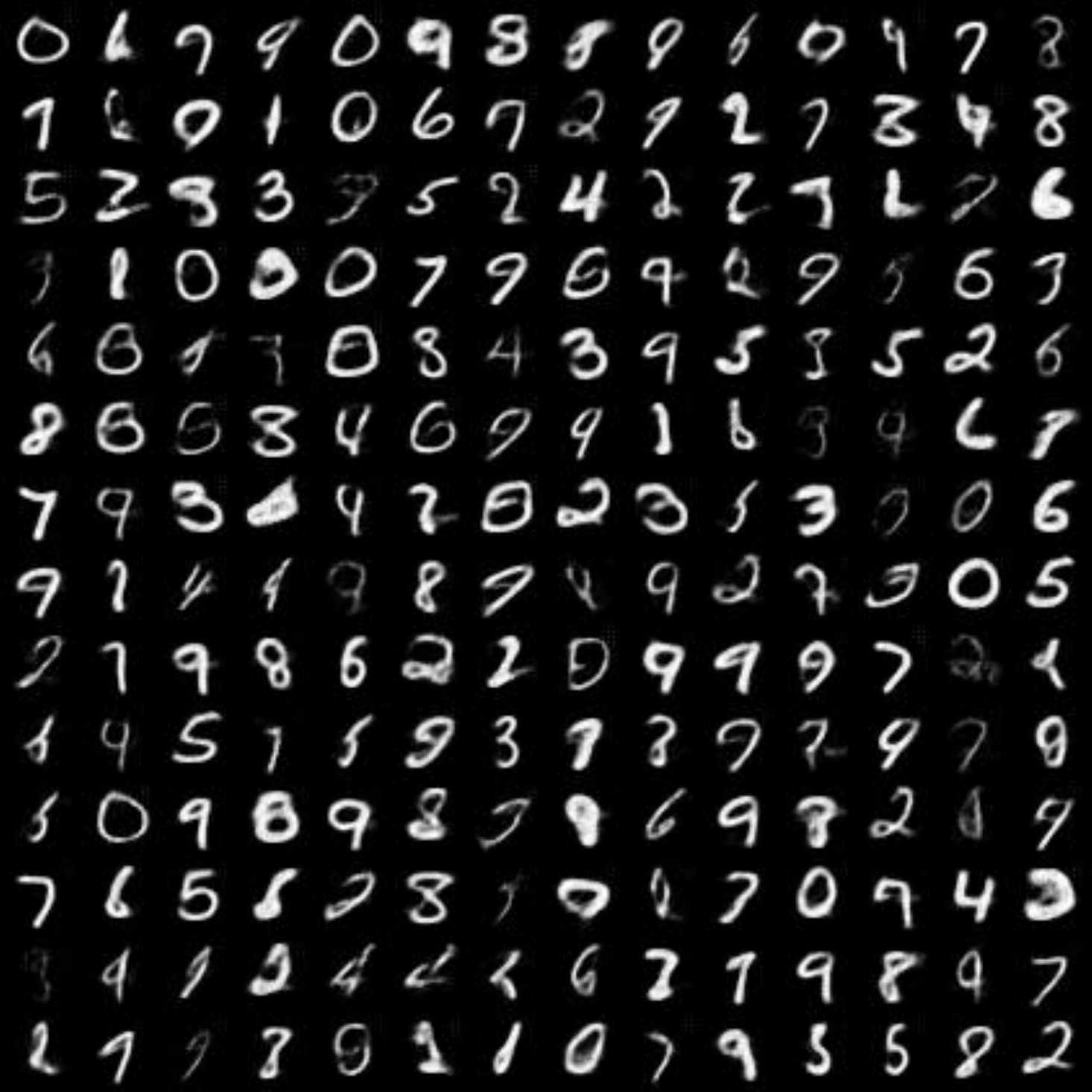} 
    &\includegraphics[scale=0.28]{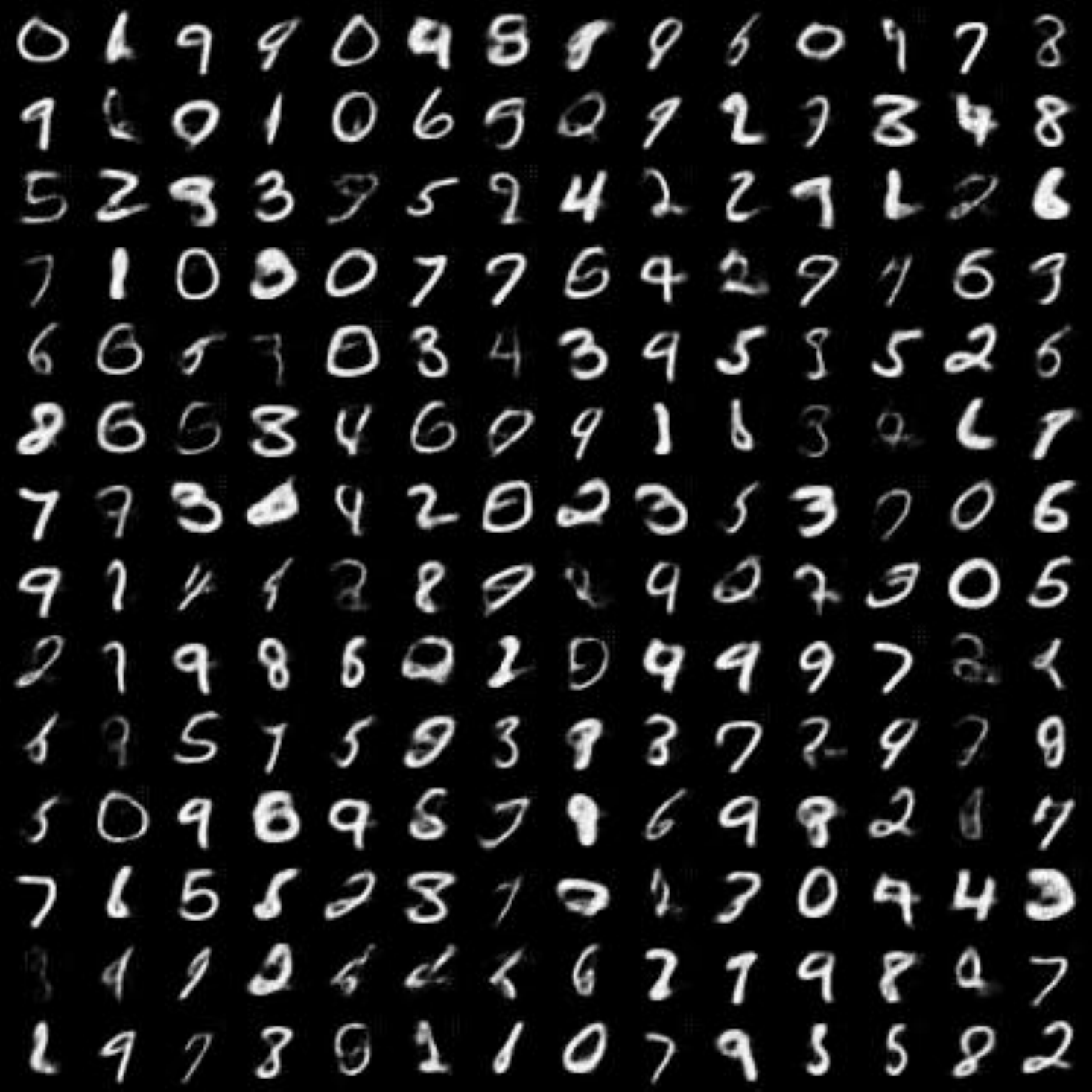} 
     
     \\
     SFG &max-SFG&  SSFG
     \\
     \includegraphics[scale=0.28]{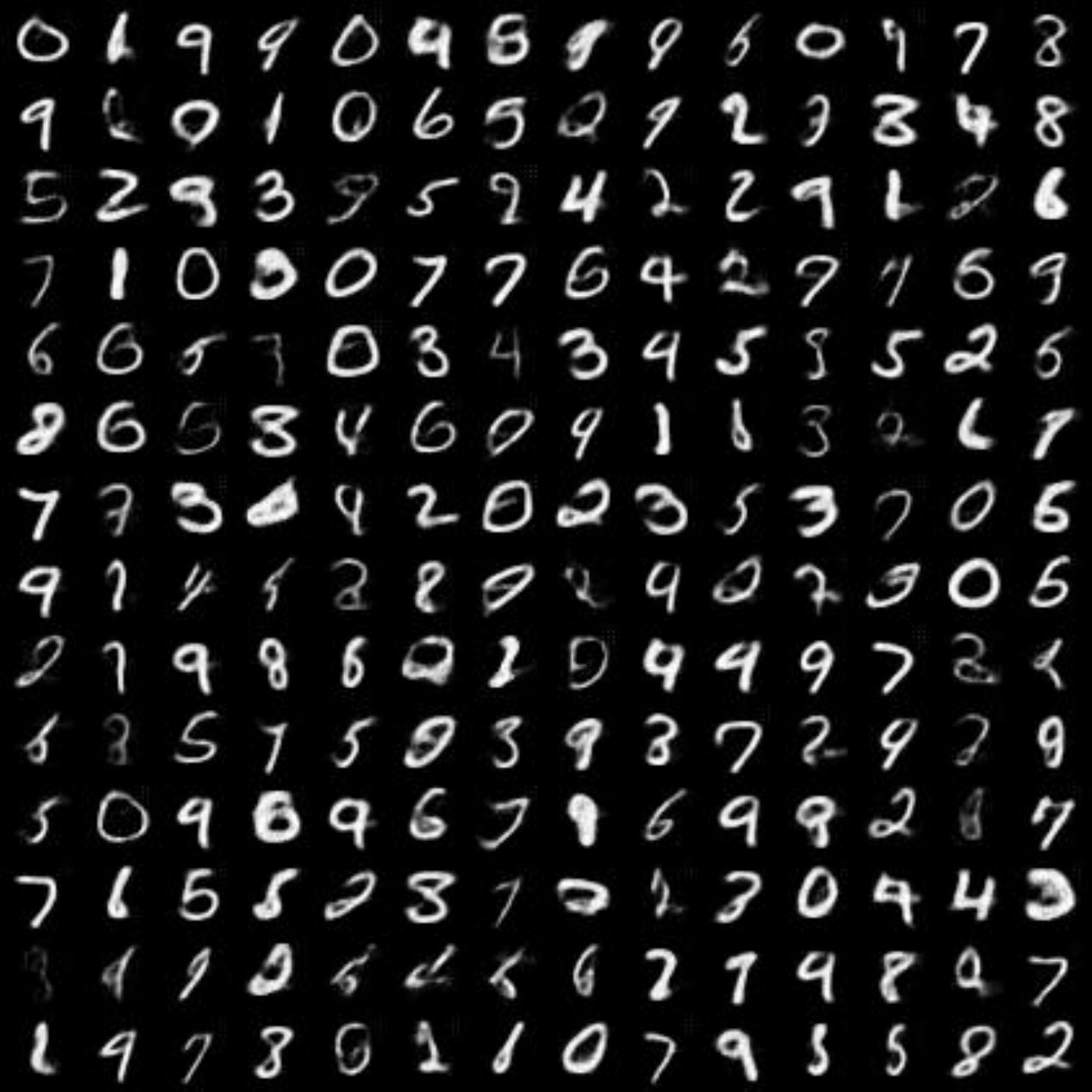} 
    &
    \includegraphics[scale=0.28]{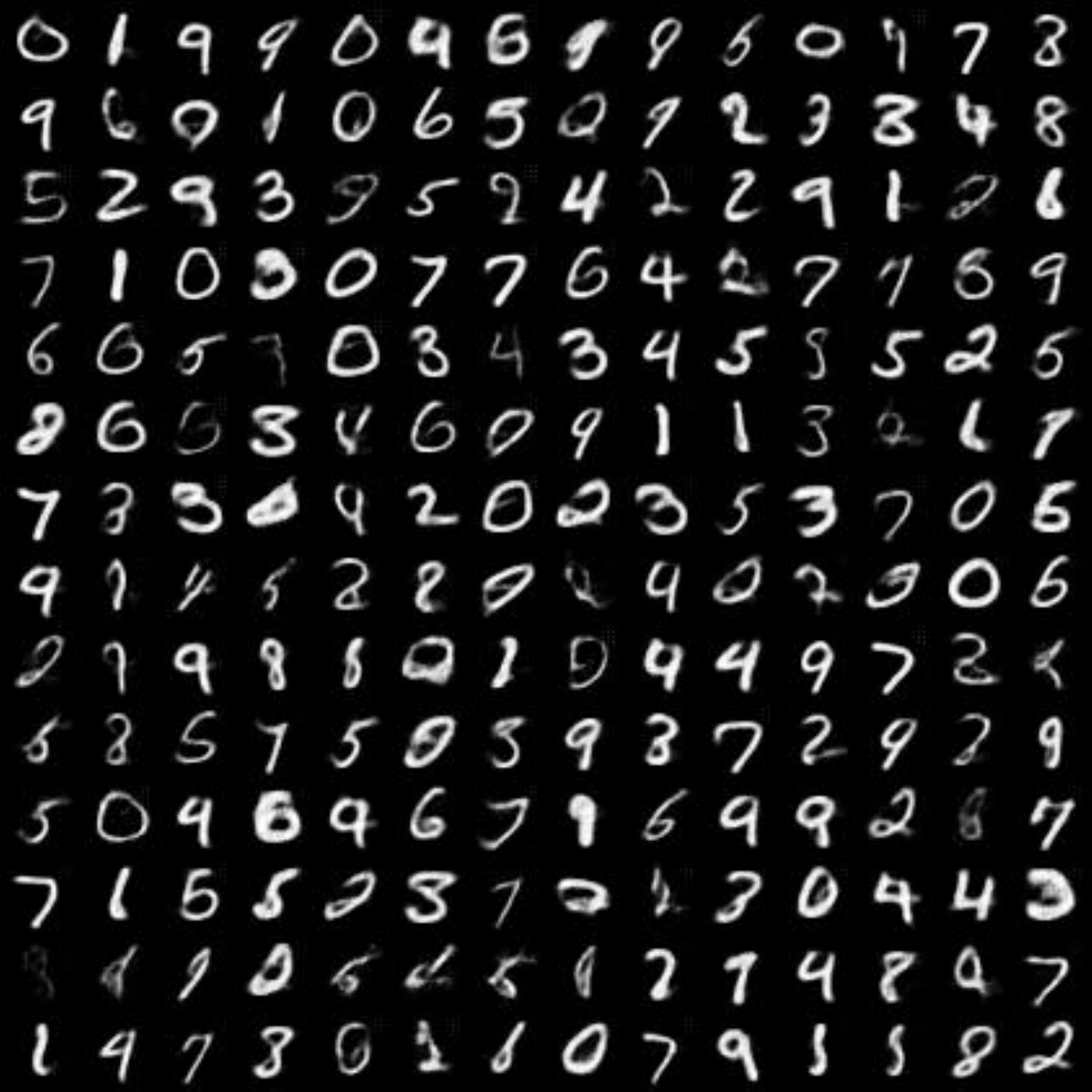} 
    &\includegraphics[scale=0.28]{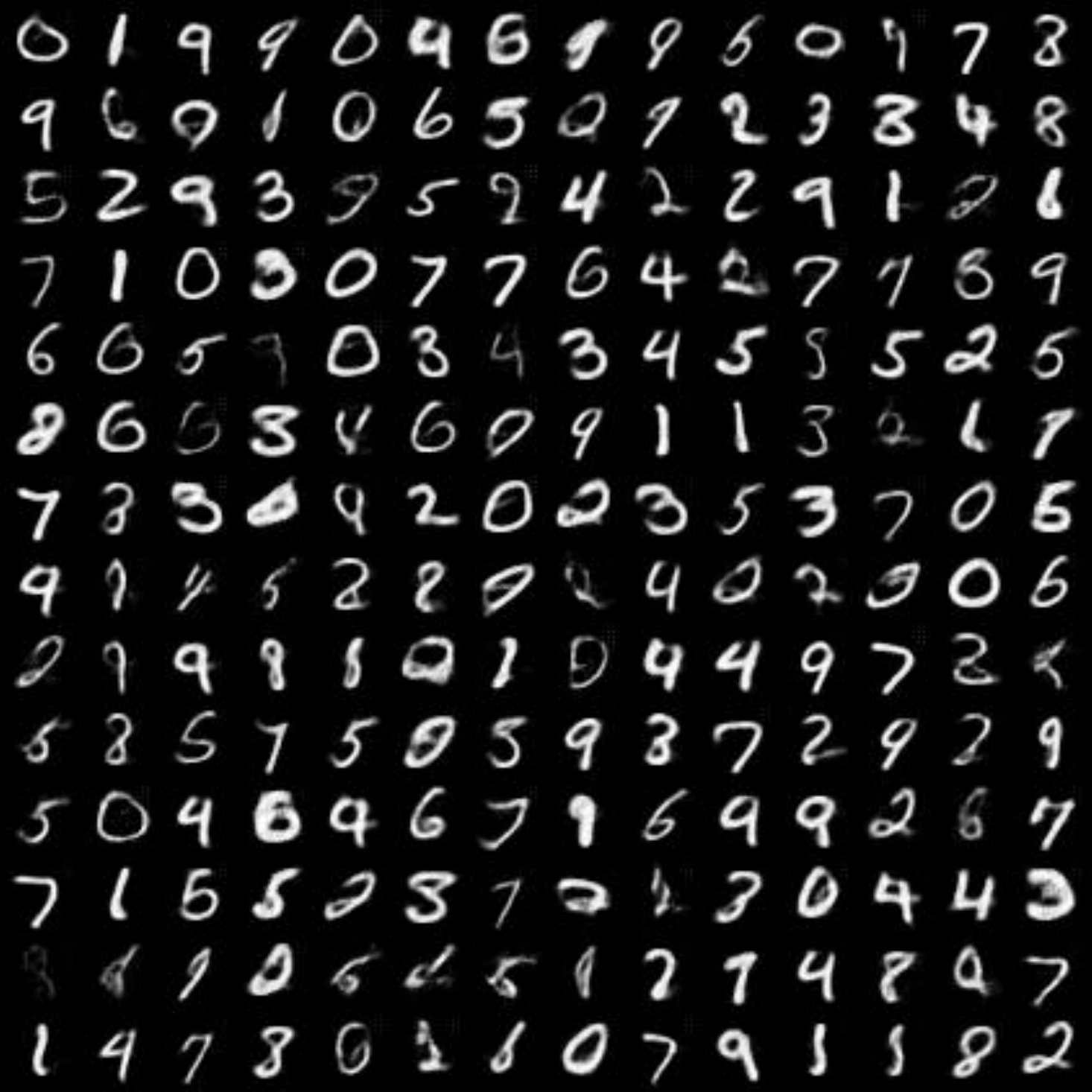} 
     
     \\
    PSSFG &MSSFG($k=10$)&  MSSFG($k=50$)

  \end{tabular}
  \end{center}
  \caption{
  \footnotesize{ MNIST ex-post density estimation generated images
    }}
  \label{fig:MNISTpdeAgen}
  \vspace{-0.2 em}
\end{figure}

\begin{figure}[!h]
\begin{center}
  \begin{tabular}{ccc}
 \includegraphics[scale=0.13]{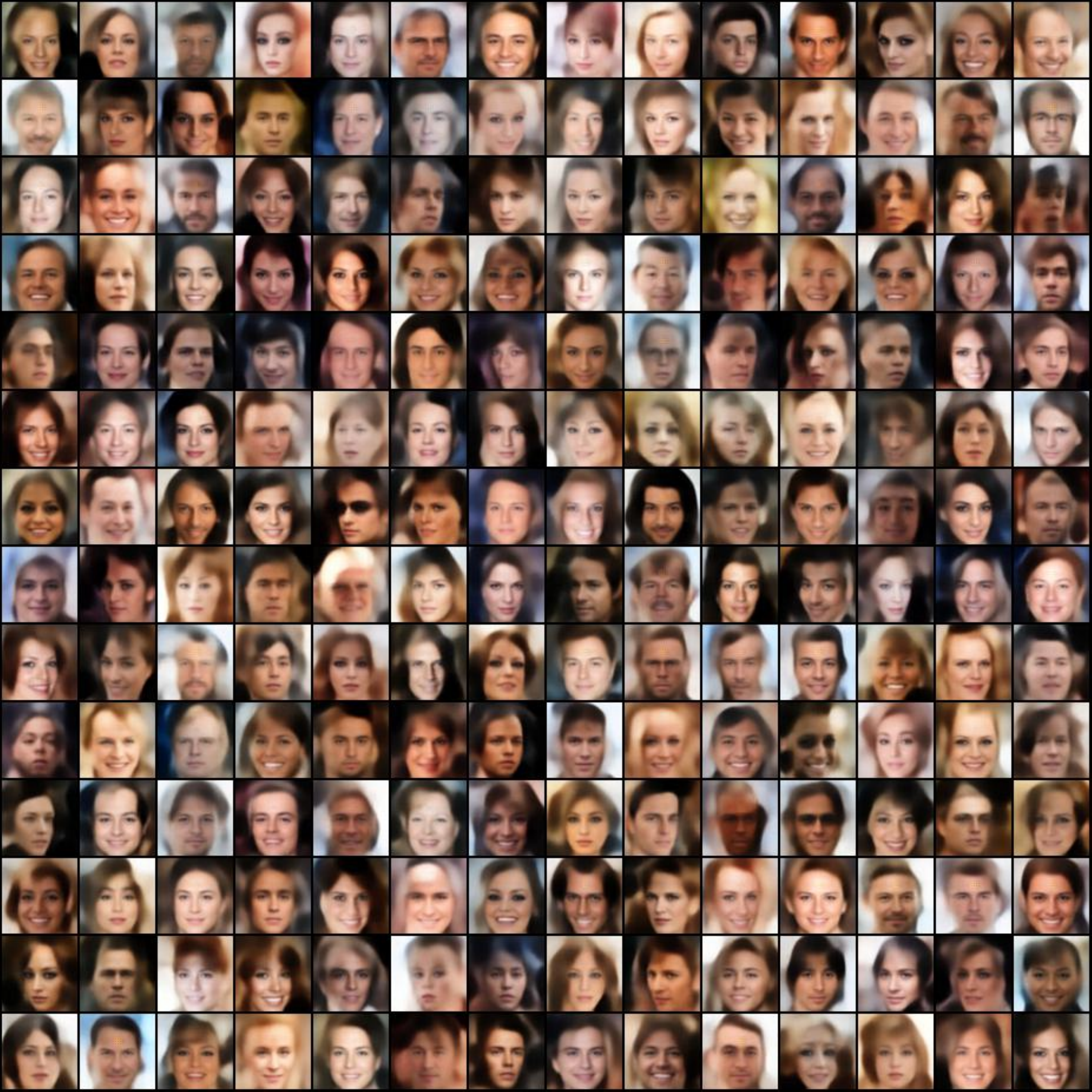} 
    &
    \includegraphics[scale=0.13]{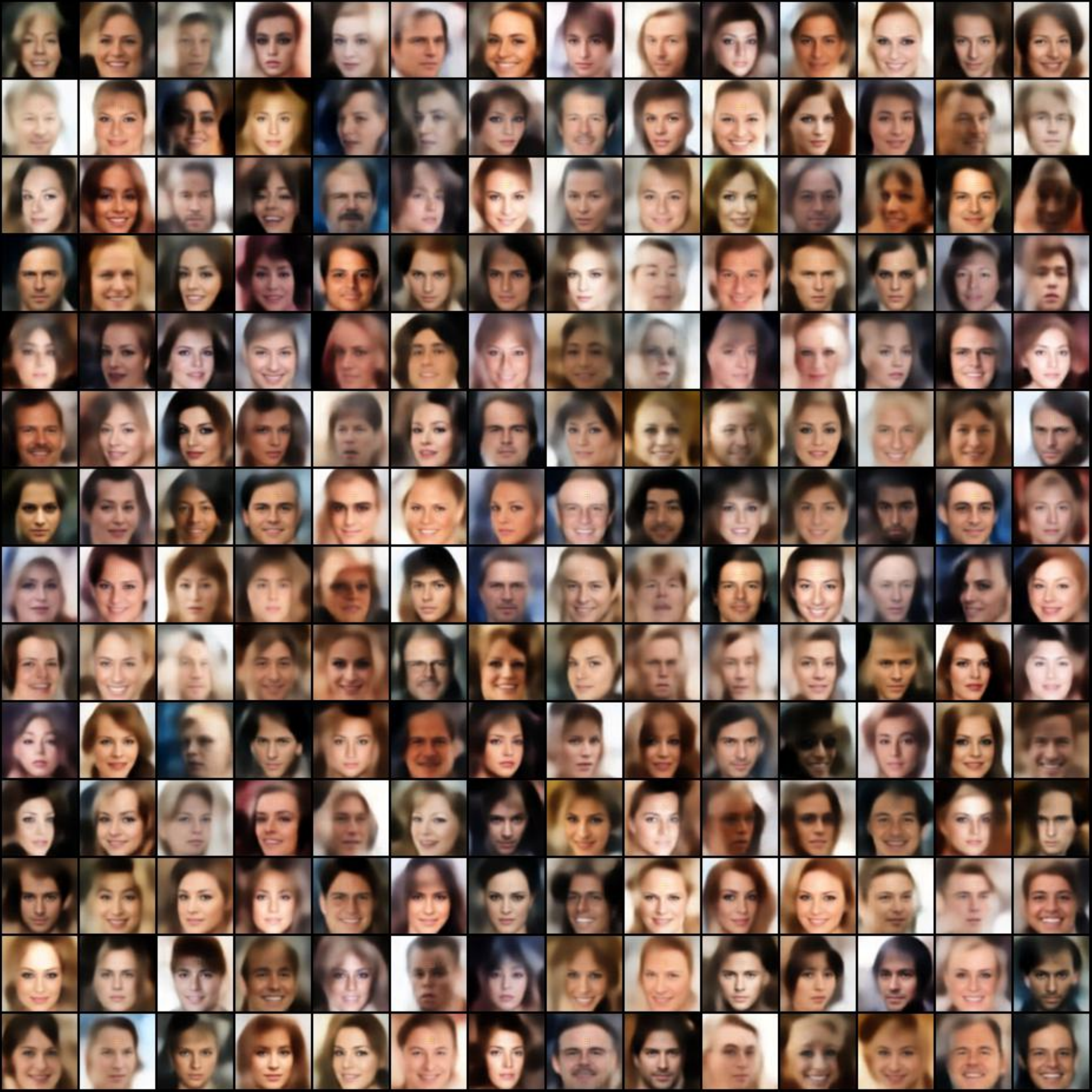} 
    &\includegraphics[scale=0.13]{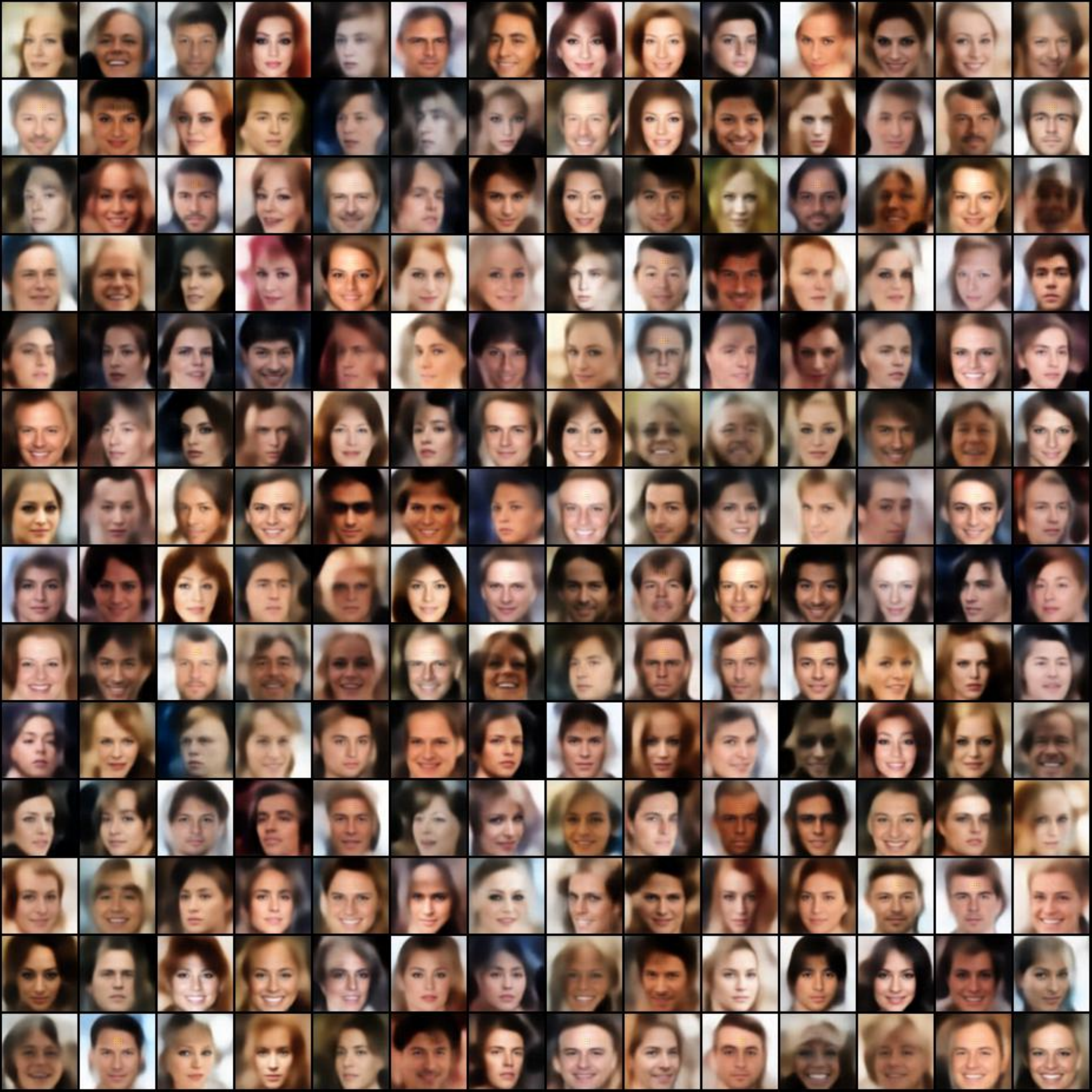} 
     
     \\
     SFG & max-SFG &  SSFG
    \\
    \includegraphics[scale=0.13]{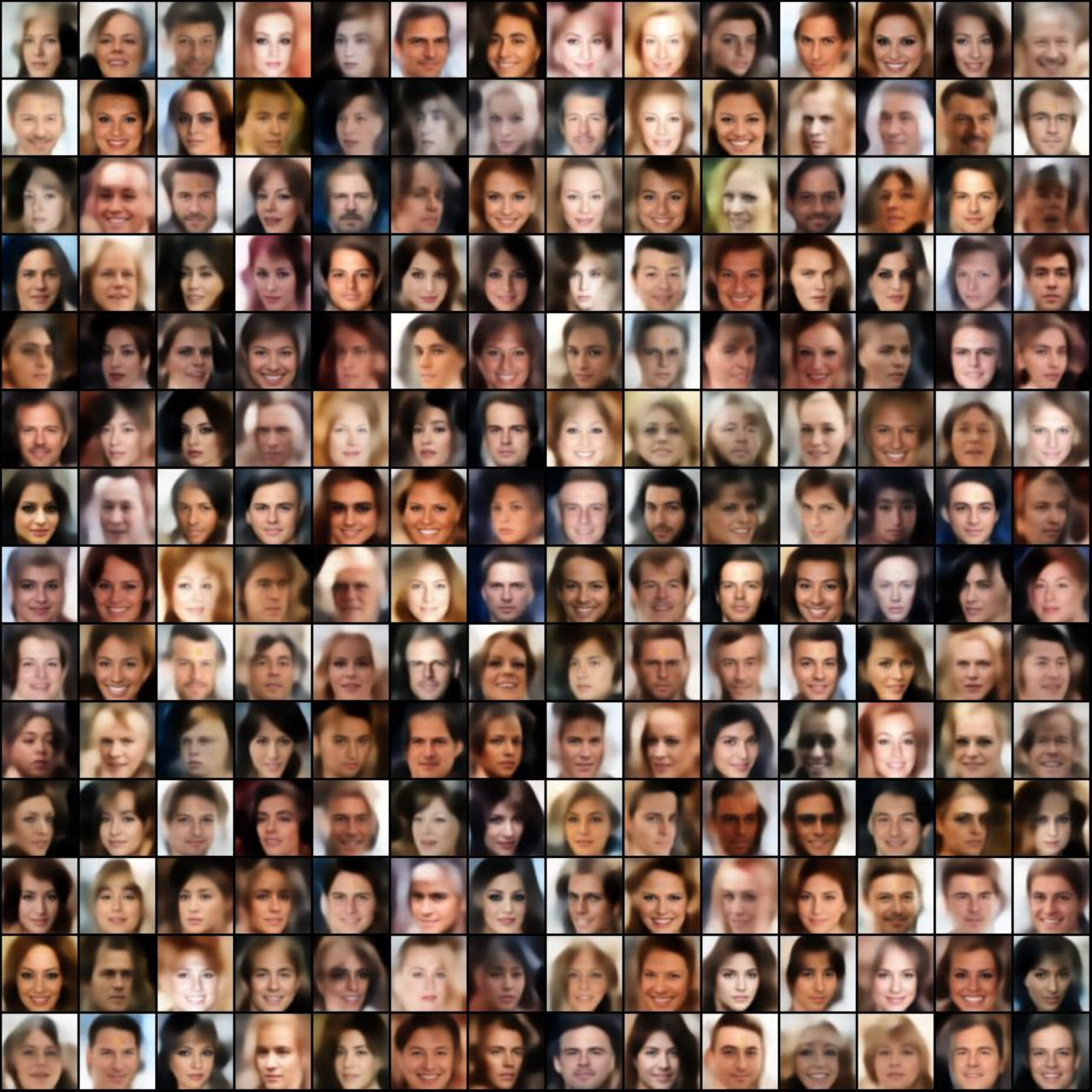} 
    &
    \includegraphics[scale=0.13]{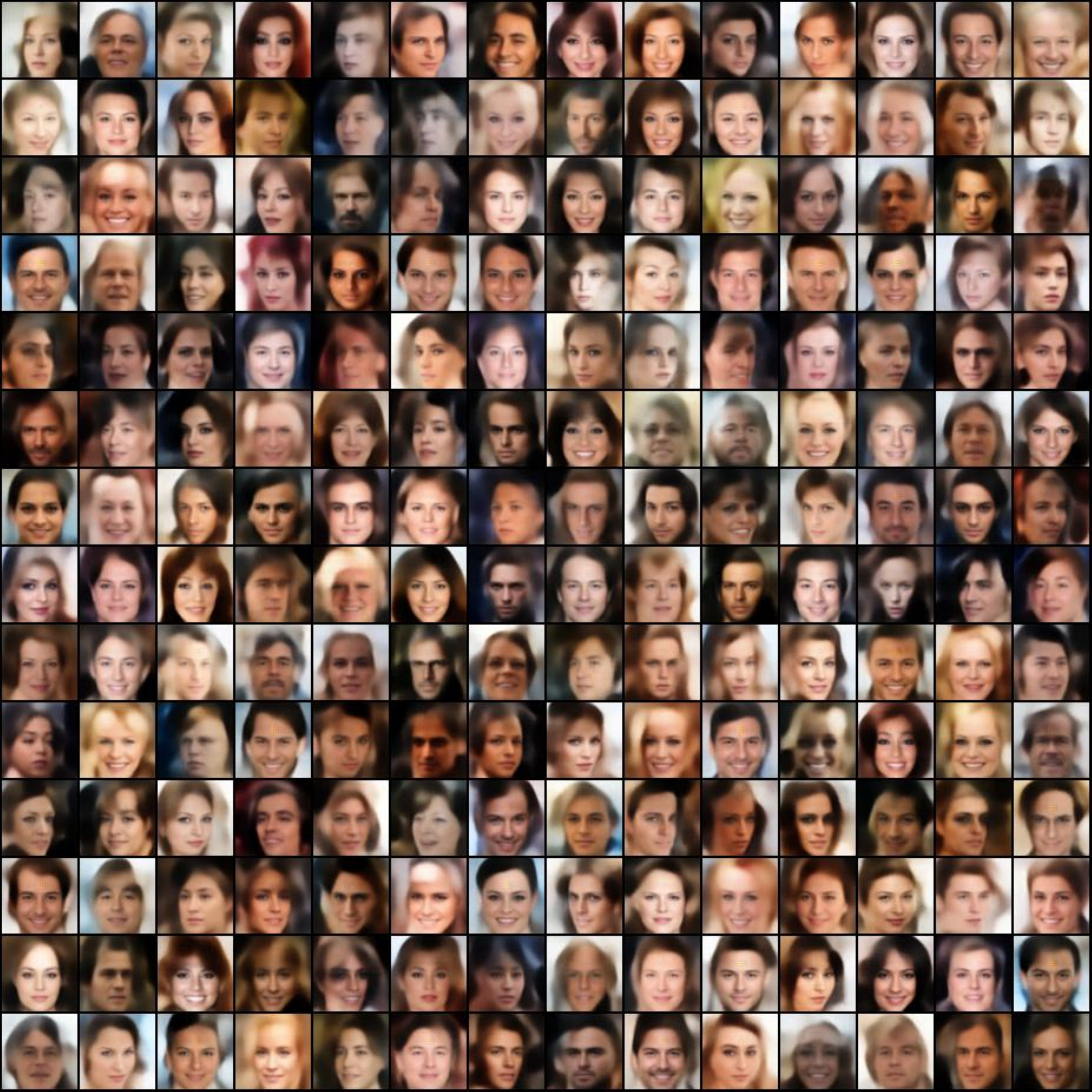} 
    &\includegraphics[scale=0.13]{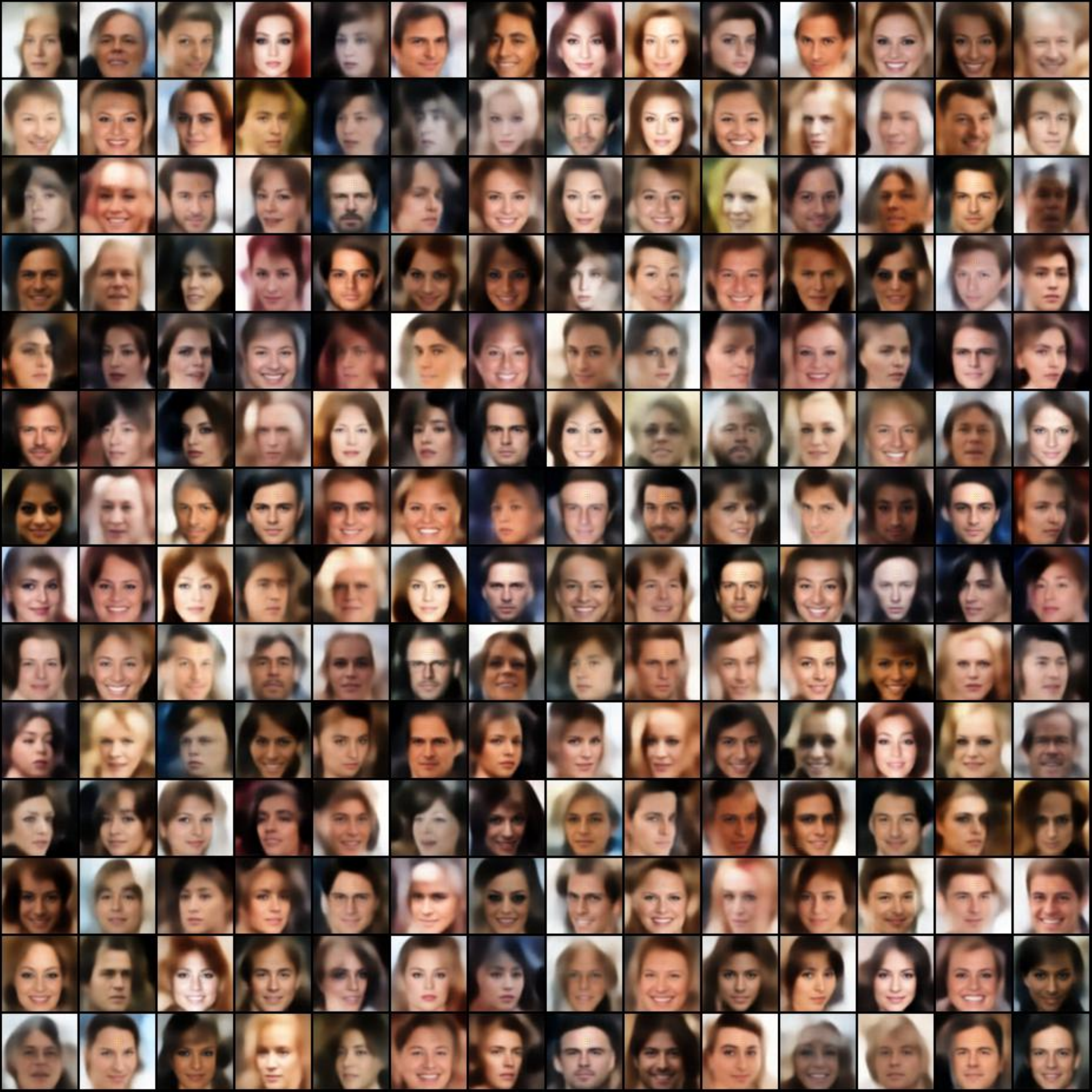} 
     
     \\
     PSSFG &MSSFG($k=10$)&  MSSFG($k=50$)
    
  \end{tabular}
  \end{center}
  \caption{
  \footnotesize{ CelebA ex-post density estimation generated images
    }}
  \label{fig:CelebApdegen}
  \vspace{-0.2 em}
\end{figure}

We show the generated images from trained models in Figure \ref{fig:MNISTpdeAgen} and Figure \ref{fig:CelebApdegen} (the reconstruction images are not shown because they are the same among all methods due to the shared first step in ex-post density estimation procedure). It is easy to see that the quality of SSFG's images is slightly better than SFG and Max-SFG.
\begin{figure}[!h]
\begin{center}

  \begin{tabular}{c}
    \includegraphics[scale=0.52]{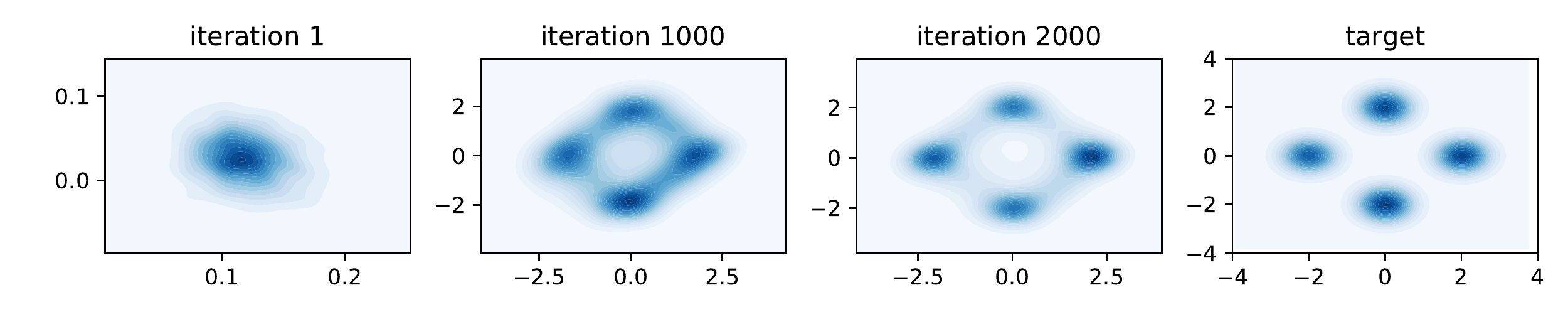} 
    \\
    SFG
    \\
    \includegraphics[scale=0.52]{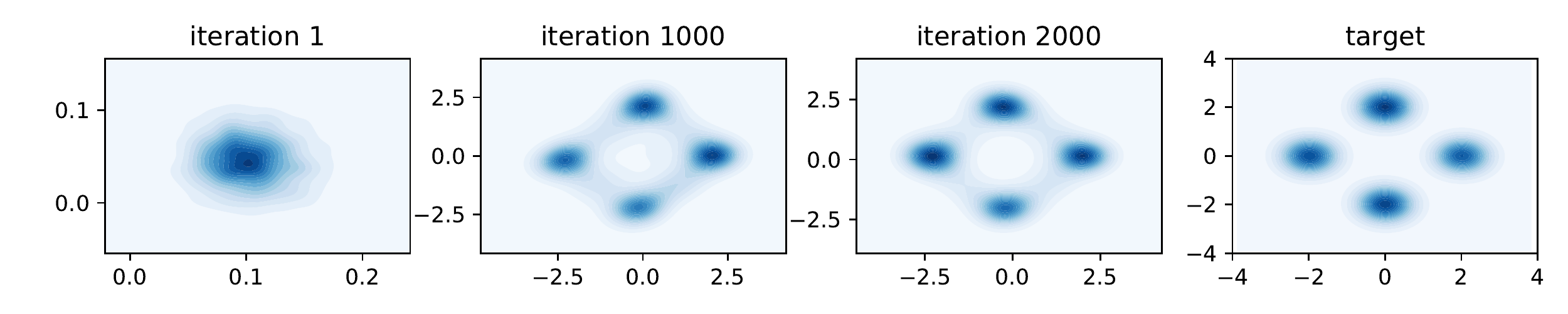} 
    \\
    max-SFG
    \\
     \includegraphics[scale=0.52]{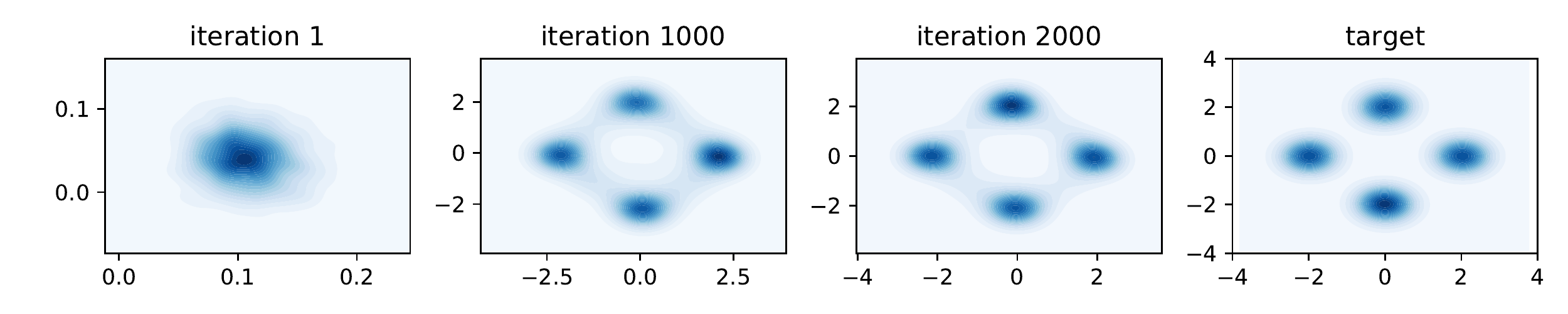} 
    \\
    SSFG
    \\
     \includegraphics[scale=0.52]{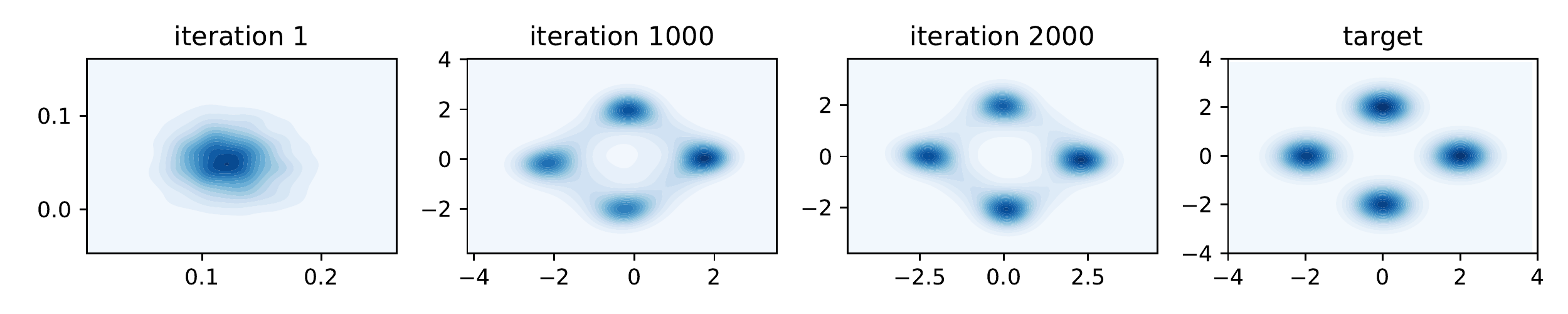} 
    \\
    PSSFG
    \\
     \includegraphics[scale=0.52]{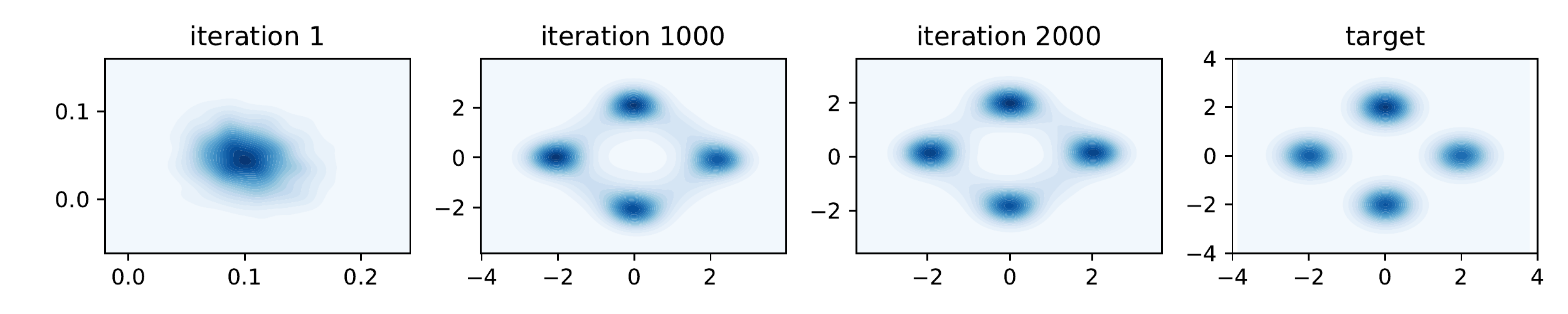} 
    \\
    MSSFG ($k=5$)
    \\
     \includegraphics[scale=0.52]{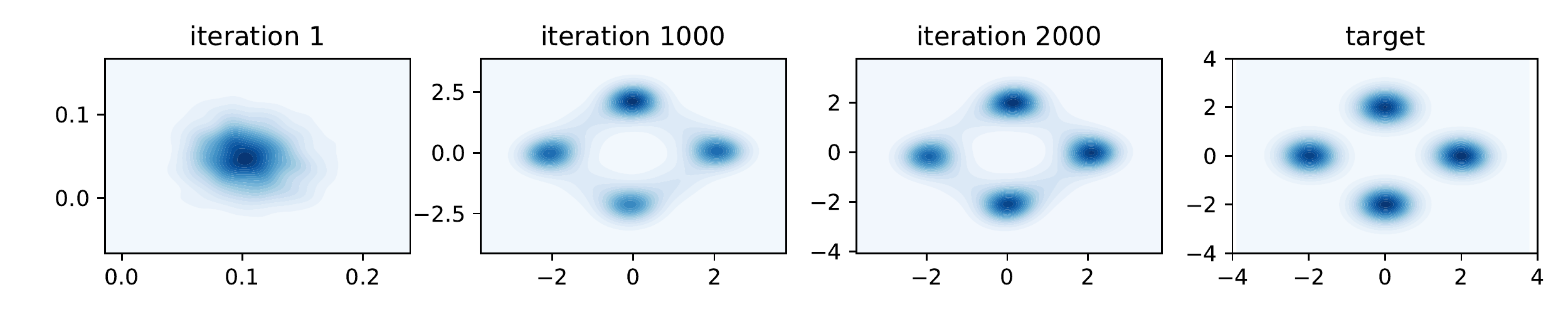} 
    \\
    MSSFG ($k=10$)
  \end{tabular}
\end{center}
  \caption{
  \footnotesize{ 4 Gaussian modes generation by variants of fused Gromov Wasserstein
    }}
  \label{fig:4guassians_fused}
  \vspace{-0.2 em}
\end{figure}
\subsection{GAN models}
\label{subsec:GAN}
Authors in \cite{bunne2019learning} have introduced a generative adversarial net \cite{goodfellow2014generative} variant that is based on Gromov Wasserstein, and it is adapted to used sliced Gromov Wasserstein in \cite{vayer2019sliced}. In this work, we replace SG by our spherical discrepancies and compare their performance in this application. 
\begin{align}
    G^* = \argmin_{G} D(p_{data}, G(Z)),
\end{align}
where $Z$ is a  low-dimensional random noise distribution (typically Gaussian), $G(Z)$ transforms  Z to a desired dimensional space.

In this toy experiment, we learn a $G$ function that is parametrized by a neural network to generate 4 Gaussian modes in 2-dimension. All settings are adapted from \url{https://github.com/bunnech/gw_gan}, and  we choose $\kappa = 1000$ for SSFG.

In Figure \ref{fig:4guassians_fused} , we show the learned distributions through iterations. The qualitative results show that max-SFG, SSFG, PSSFG and MSSFG helps the distribution converge faster to the target than SFG, the evidence is that the model distributions in the iteration 1000 of max-SFG and SSFG look more similar to 4 Gaussian modes. The difference between SSFG, max-SFG, PSSFG and MSSFG is not too clear, however, in the iteration 1000 we can see that MSSFG's distributions do not put too many samples on the intersection area between modes while  distributions of other discrepancies do.

\section{Experimental settings}
\label{Sec:exp_setting}
In this section, we provide detailed settings of experiments that we conduct in this paper.
\subsection{Neural Network architecture}
\textbf{For MNIST dataset:}

Encoder E: $x \in \mathbb{R}^{28\times 28} \to \text{Conv}_{128} \to \text{BatchNorm} \to \text{ReLU} \to   \text{Conv}_{256} \to \text{BatchNorm} \to \text{ReLU} \to   \text{Conv}_{512} \to \text{BatchNorm} \to \text{ReLU} \to   \text{Conv}_{1024} \to \text{BatchNorm} \to \text{ReLU} \to   \text{FC}_8 \to z \in \mathbb{R}^8$

Decoder G: $z \in \mathbb{R}^8 \to \text{FC}_{7x7x1024} \to \text{BatchNorm} \to \text{ReLU} \to   \text{FSConv}_{512}\to \text{BatchNorm} \to \text{ReLU} \to   \text{FSConv}_{256} \to \text{BatchNorm} \to \text{ReLU} \to   \text{FSConv}_{1} \to  x \in \mathbb{R}^{28\times 28}$

where $\text{Conv}_{k}$ denotes for a convolution with  $k$ $4 \times 4$ filters, $\text{FSConv}_{k}$ denotes  the fractional strided convolution with $k$ $4 \times 4$ filters and $\text{FC}_k$ is the fully connected layer mapping to $\mathbb{R}^k$. For VAE, PRAE, GMVAE and Vampprior, the encoder contains two FC layers for the mean and the logarithmic variance in the last layer.

\textbf{For CelebA dataset:}

Encoder E: $x \in \mathbb{R}^{64\times 64 \times 3} \to \text{Conv}_{128} \to \text{BatchNorm} \to \text{ReLU} \to   \text{Conv}_{256} \to \text{BatchNorm} \to \text{ReLU} \to   \text{Conv}_{512} \to \text{BatchNorm} \to \text{ReLU} \to   \text{Conv}_{1024} \to \text{BatchNorm} \to \text{ReLU} \to   \text{FC}_64 \to z \in \mathbb{R}^{64}$

Decoder G: $z \in \mathbb{R}^{64} \to \text{FC}_{8x8x1024} \to \text{BatchNorm} \to \text{ReLU} \to   \text{FSConv}_{512}\to \text{BatchNorm} \to \text{ReLU} \to   \text{FSConv}_{256} \to \text{BatchNorm} \to \text{ReLU} \to   \text{FSConv}_{3} \to  x \in \mathbb{R}^{64\times 64 \times 3}$

The Conv in CelebA also uses $4 \times 4$ filters. 

\subsection{Hyperparameter settings}
To train the autoencoder, we use Adam optimizer \cite{kingma2014adam} with learning rate equals 0.001, $\beta_1=0.5$, $\beta_2=0.999$. The number of epochs is 50 on MNIST and 40 on CelebA,  batch size is 100 for both datasets. The coefficient $\lambda$ (autoencoder regularization) is 1. The number of components of Gaussian mixture prior is set to 10. The fused parameter of fused Gromov Wasserstein $\beta=0.1$. The number of projections of sliced-discrepancies is 50 for fairness.

\textbf{For max-SFG, SSFG, PSSFG and MSSFG:} The maximum iteration of the optimization for max-direction (distribution over directions) is  10.

\subsection{Reconstruction and FID computation}
The reconstruction score is computed by the Mean square error between reconstructed images and original images in corresponding test set.

The FID score is computed between 10000 randomly generated images and all images from test set (for evaluation), and all images from validation set (for model selection). We use the implementation at \url{https://github.com/bioinf-jku/TTUR}.

\subsection{Tuning parameters}
\textbf{For SSFG, PSSFG and MSSFG}: We choose the $\kappa$ that has the lowest FID score on corresponding validation set.
\subsection{Code and computing system}

We use the code for SFG and autoencoder-baselines from \url{https://github.com/HongtengXu/Relational-AutoEncoders}, the code for GAN from \url{https://github.com/bunnech/gw_gan}, We run code on a single NVIDIA RTX 2080 Ti.

\end{document}

%% file: math_commands.tex
\usepackage{amsmath,amsfonts,bm}

\def\eqref#1{equation~\ref{#1}}

\def\1{\bm{1}}

\DeclareMathAlphabet{\mathsfit}{\encodingdefault}{\sfdefault}{m}{sl}
\SetMathAlphabet{\mathsfit}{bold}{\encodingdefault}{\sfdefault}{bx}{n}

\DeclareMathOperator*{\argmin}{arg\,min}

%% file: import.tex
\usepackage{microtype}
\usepackage{graphicx}
\usepackage{booktabs} 
\usepackage{amsmath}
\usepackage{amsfonts} 

\usepackage{epsf}
\usepackage{fancyhdr}
\usepackage{graphics}
\usepackage{graphicx}
\usepackage{psfrag}
\usepackage{microtype}

\usepackage{color}
\usepackage{amsthm}
\usepackage{amsfonts}
\usepackage{amsmath}
\usepackage{amssymb,bbm}
\usepackage{caption}
\usepackage{sidecap}
\sidecaptionvpos{figure}{c}

\usepackage{url}
\usepackage{hyperref}
\usepackage{algorithmic}
\usepackage{algorithm}

\newtheorem{assumption}{Assumption}

\long\def\comment#1{}
\comment{
\setlength{\topmargin}{0 in}
\setlength{\textwidth}{5.5 in}
\setlength{\textheight}{8 in}
\setlength{\oddsidemargin}{0.5 in}
}

\newtheorem{theorem}{Theorem}

\newtheorem{definition}{Definition}

\def\argmin{\textnormal{arg} \min}

\newcommand{\norm}[1]{\left\lVert#1\right\rVert}

\graphicspath{{./image/}}

\newcommand{\ba}{\begin{array}}
\newcommand{\ea}{\end{array}}

